\documentclass[12pt, oneside]{book}

\usepackage[english]{babel} 
\babelprovide[import]{hindi}
\babelprovide[import]{sanskrit}
\usepackage{url}

\usepackage[utf8]{inputenc}
\usepackage{graphicx}
\graphicspath{ {Images/} }
\usepackage{lipsum}
\usepackage[a4paper,width=150mm,top=25mm,bottom=25mm,bindingoffset=6mm]{geometry}

\usepackage{amsmath}

\usepackage{fancyhdr}
\usepackage{caption}
\usepackage{subcaption}

\newcommand{\institutionalreview}[1]{
\vspace{6pt}\noindent{\fontsize{10}{12.2}\selectfont\textbf{Institutional Review Board Statement:} {#1}\par}}

\newcommand{\authorcontributions}[1]{%
\vspace{6pt}\noindent{\fontsize{10}{12.2}\selectfont\textbf{Author Contributions:} {#1}\par}}

\newcommand{\funding}[1]{
\vspace{6pt}\noindent{\fontsize{10}{12.2}\selectfont\textbf{Funding:} {#1}\par}}

\newcommand{\informedconsent}[1]{
\vspace{6pt}\noindent{\fontsize{10}{12.2}\selectfont\textbf{Informed Consent Statement:} {#1}\par}}

\newcommand{\dataavailability}[1]{
\vspace{6pt}\noindent{\fontsize{10}{12.2}\selectfont\textbf{Data Availability Statement:} {#1}\par}}

\newcommand{\conflictsofinterest}[1]{%
\vspace{6pt}\noindent{\fontsize{10}{12.2}\selectfont\textbf{Conflicts of Interest:} {#1}\par}}

\newcommand{\abbreviations}[2]{\vspace{12pt}\noindent{\selectfont\textbf{#1}}{%
\par\vspace{3pt}\noindent{\fontsize{10}{12.2}\selectfont #2}\par}}

\newcommand{\acknowledgments}[1]{
\vspace{6pt}\noindent{\fontsize{10}{12.2}\selectfont\textbf{Acknowledgments:} {#1}\par}}

\usepackage[style=authoryear,sorting=anyt,mincitenames=1,maxcitenames=2,maxbibnames=4,uniquelist=false]{biblatex}
\usepackage{setspace}
\usepackage[table]{xcolor} 
\usepackage{multicol}
\usepackage{multirow} 
\usepackage{hyperref}
\hypersetup{
    colorlinks=true,
    linkcolor=black,
    filecolor=black,      
    urlcolor=black,
    citecolor=black,
    pdfpagemode=FullScreen,
    }

\usepackage{xurl}
\usepackage{setspace}
\usepackage{pdfpages}

\begin{document}

\begin{titlepage}

\thispagestyle{empty} 
\pagenumbering{gobble}

    \begin{center}
        
        \Huge
        \textbf{Enhancing Neural Machine Translation of Low-Resource Languages:}
        
        \vspace{0.5cm}
        
        \LARGE
        Corpus Development, Human Evaluation and Explainable AI Architectures
        \vspace{0.5cm}
        
        \Large
        \textbf{Séamus Lankford MSc, MBA, BEng}
        
        \vspace{0.3cm}
        
        Supervised by Prof. Andy Way\\
        and Dr Haithem Afli (Munster Technological University) \\
        
        \vfill
        
        \includegraphics[width=0.25\textwidth]{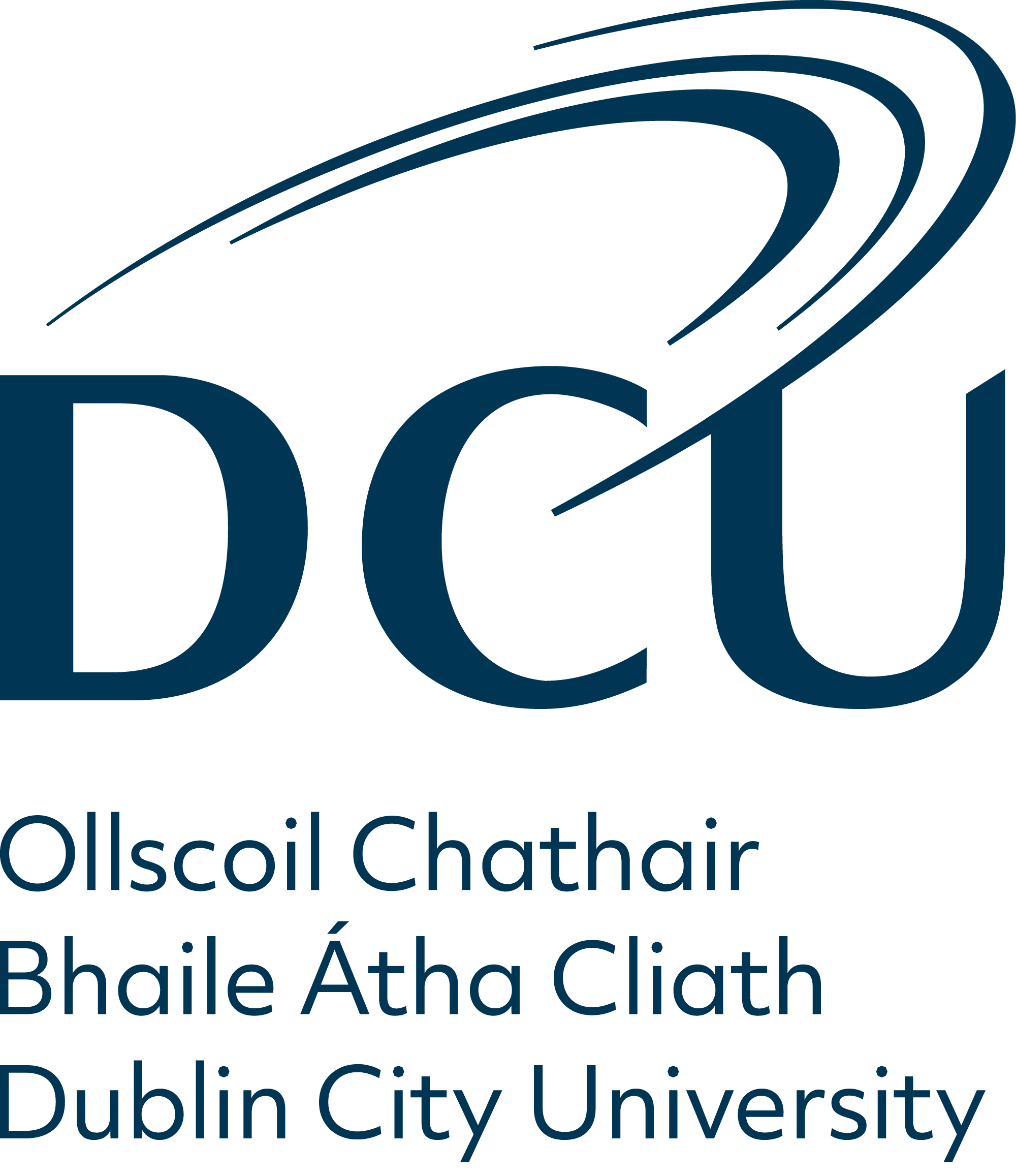}
        
        \vspace{0.5cm}
        
        \Large
   A Thesis Submitted for the Award of Doctor of Philosophy
        
        \vspace{0.5cm}
        
        \LARGE
	    \textsc{School of Computing\\
	    Dublin City University}
	    
	    \begin{flushright}
	
	    \Large
	    January 2024
	
	    \end{flushright}
        
    \end{center}
\end{titlepage}

\frontmatter 

\chapter*{Declaration}
\thispagestyle{plain}
\fancyfoot[C]{\thepage}
I hereby certify that this material, which I now submit for assessment on the programme of study leading to the award of Doctor of Philosophy (PhD) is entirely my own work, and that I have exercised reasonable care to ensure that the work is original, and does not to the best of my knowledge breach any law of copyright, and has not been taken from the work of others save and to the extent that such work has been cited and acknowledged within the text of my work.
\\
\\
Signed:   \\
ID No.: 20216607\\  
Date: January 2nd, 2024\\

\chapter*{Dedication}
\thispagestyle{plain}
\fancyfoot[C]{\thepage}
Letty agus Cormac atá caillte. I ndorn Dé go raibh a n-anamacha dílse.

\chapter*{Acknowledgements}
\thispagestyle{plain}
\fancyfoot[C]{\thepage}

I wish to thank my PhD supervisors, Professor Andy Way and Dr Haithem Afli for their unwavering support. Right throughout, Andy has been an exceptional mentor in my career development as a researcher. Despite a busy schedule, he was always available for our scheduled meetings and I was given a great deal of autonomy in how my research should be directed. Furthermore, Andy stepped in to steer me in the right direction whenever needed. Given his decades of experience, Andy helped in identifying research gaps and opportunities. It is a testament to Andy’s professionalism that he has retained his drive and interest in developing his students throughout this career and I consider myself fortunate that our paths have crossed. Likewise, I count myself lucky to have worked with Haithem who not only adopted the role of supervisor but also became an important ally in research projects not directly linked to my PhD. Haithem always exudes a positive attitude which helped greatly when the going got tough. The funding received from MTU was very much appreciated as was the support from Dr Teresa Lynn in the early days of my PhD journey. I must also thank my wife, Helen, who as always was steadfast in her support despite the countless hours I put in. My children, Darragh and Julie must also be acknowledged, not only for their support but also for their love of the Irish language which has been, in part, an inspiration for this work. Táim faoi chomaoin acu.

\chapter*{Motivation}
\thispagestyle{plain}
\fancyfoot[C]{\thepage}

The seeds of my interest in the fields of deep learning (DL) and natural language processing (NLP) were sown while taking an MSc in Artificial Intelligence during the Covid-19 pandemic. The decision to pursue such an MSc was motivated by the rise of artificial intelligence (AI) which can be used both as a force for positive and negative change. Motivated by how AI could be used to impact society positively, I signed up for the MSc programme which culminated with the publication of my mini-thesis: Automatic Neural Architecture using AutoML, Swarm Intelligence and Ensemble methods.

I grew up in a family where Irish was spoken with ease as a second language which created an awareness that language is a matter of inheritance, in the same way our place and family are. In my case, my second language was an inheritance from two generations previously, for I had the privilege of knowing and interacting with bilingual grandparents. They had acquired their early speech patterns in the 1890s, when vernacular in Ireland was still transitioning from Irish to English.  

Given my background as an Irish language speaker, it was clear the foundational principles which I learned throughout the MSc could be applied in the context of a PhD study focused on the neural machine translation (NMT) of low-resource languages. Initially, the research questions were not immediately clear. However, it soon emerged that enhancing NMT of low-resource languages should concentrate on three key pillars namely corpus development, improved human evaluation and the development of transparent and easily understood open-source NMT architectures. 

\chapter*{List of Publications}
\thispagestyle{plain}
\fancyfoot[C]{\thepage}
\begin{enumerate}

\item
Lankford, S., Afli, H. and Way, A., 2023. adaptMLLM: Fine-Tuning Multilingual Language Models on Low-Resource Languages with integrated LLM playgrounds. In: \textit{Information 14.12}, issn: 2078-2489, doi: 10.3390/info14120638, url: https://www.mdpi.com/2078-2489/14/12/638.

\item
Lankford, S., Afli, H. and Way, A., 2023. adaptNMT: an open-source, language-agnostic development environment for neural machine translation. In: \textit{Language Resources and Evaluation}, pp.1-26. url: https://doi.org/10.1007/s10579-023-09671-2

\item
Lankford, S., Afli, H. and Way, A., 2023. Design of an Open-Source Architecture for Neural Machine Translation. In: \textit{Proceedings of the 1st Workshop on Open Community-Driven Machine Translation}. Tampere, Finland: European Association for Machine Translation, pp. 15–20. \\ url: https://aclanthology.org/2023.crowdmt-1.2.

\item
Lankford, S., Afli, H. and Way, A., 2022. Human Evaluation of English-Irish Transformer-Based NMT. In: \textit{Information 13.7}, issn: 2078-2489. doi: 10.3390/info13070309. url: https://www.mdpi.com/2078-2489/13/7/309.

\item 
Lankford, S., Afli, H., Ní Loinsigh Ó., Way, A., 2022. “gaHealth: An English–Irish Bilingual Corpus of Health Data”. In: \textit{Proceedings of the Thirteenth Language Resources and Evaluation Conference}. Marseille, France: European Language Resources Association, pp. 6753–6758. \\ url: https://aclanthology.org/2022.lrec-1.727. 

\item 
Lankford, S., Afli, H., and Way, A. 2021. “Transformers for Low-Resource Languages: Is Féidir Linn!” In: \textit{Proceedings of Machine Translation Summit XVIII: Research Track}. Virtual: Association for Machine Translation in the Americas, pp. 48–60. url: https://aclanthology.org/2021.mtsummit-research.5.

\item 
Lankford, S., Afli, H., and Way, A. 2021. “Machine Translation in the Covid domain: an EN$\leftrightarrow$GA case study for LoResMT 2021”. In: \textit{Proceedings of the 4th Workshop on Technologies for MT of Low Resource Languages (LoResMT2021)}. Virtual: Association for Machine Translation in the Americas, pp. 144–150. url: https://aclanthology.org/2021.mtsummit-loresmt.15

\end{enumerate}
\begin{singlespace}

\pagestyle{plain}
\tableofcontents
\pagestyle{plain}

\fancyfoot[C]{\thepage}

\listoffigures
\thispagestyle{plain}
\fancyfoot[C]{\thepage}

\listoftables
\thispagestyle{plain}
\fancyfoot[C]{\thepage}

\end{singlespace}

\thispagestyle{plain}

\fancyfoot[C]{\thepage}
\clearpage

\thispagestyle{plain}
\begin{center}
    \Large
    \textbf{Enhancing Neural Machine Translation of Low-Resource Languages: Corpus Development, Human Evaluation and Explainable AI Architectures}
    
    \vspace{0.4cm}
    \large

    \vspace{0.4cm}
    \textbf{Séamus Lankford}
    
    \vspace{1.9cm}
    
    \textbf{Abstract}
\end{center}
\begin{singlespace}
In the current machine translation (MT) landscape, the Transformer architecture stands as the gold standard, especially for high-resource language pairs. This research delves into its efficacy for low-resource language pairs including both the English$\leftrightarrow$Irish and English$\leftrightarrow$Marathi language pairs. Notably, the study identifies the optimal hyperparameters and subword model type to significantly improve the translation quality of Transformer models for low-resource language pairs.

The scarcity of parallel datasets for low-resource languages can hinder MT development. To address this, we developed gaHealth, the first bilingual corpus of health data for the Irish language. Focusing on the health domain, models developed using this in-domain dataset exhibited very significant improvements in BLEU score when compared with models from the LoResMT2021 Shared Task. A subsequent human evaluation using the multidimensional quality metrics error taxonomy showcased the superior performance of the Transformer system in reducing both accuracy and fluency errors compared to an RNN-based counterpart. 

Furthermore, this thesis introduces adaptNMT and adaptMLLM, two open-source applications streamlined for the development, fine-tuning, and deployment of neural machine translation models. These tools considerably simplify the setup and evaluation process, making MT more accessible to both developers and translators. Notably, adaptNMT, grounded in the OpenNMT ecosystem, promotes eco-friendly natural language processing research by highlighting the environmental footprint of model development. Fine-tuning of MLLMs by adaptMLLM demonstrated advancements in translation performance for two low-resource language pairs: English$\leftrightarrow$Irish and English$\leftrightarrow$Marathi, compared to baselines from the LoResMT2021 Shared Task.
\end{singlespace}

\clearpage

\thispagestyle{plain}
\begin{center}
    \Large
    \textbf{
    Feabhas a chur ar Mheaisínaistriúchán Néarach Teangacha Ísealacmhainne: Forbairt Chorpais, Meastóireacht Dhaonna agus Ailtireacht IS Inmhínithe
}
    
    \vspace{0.4cm}
    \large

    \vspace{0.4cm}
    \textbf{Séamus Lankford}
    
    \vspace{1.9cm}
    
    \textbf{Achoimre}
\end{center}
\begin{singlespace}
Sa tírdhreach meaisínaistriúcháin (MA) reatha, seasann ailtireacht an Trasfhoirmeora mar an caighdeán is airde, go háirithe do phéirí teanga ardacmhainne. Déanann an taighde seo tochailt isteach ar a éifeachtúlacht le haghaidh péirí teanga nach bhfuil mórán acmhainní acu lena n-áirítear péirí teanga Béarla$\leftrightarrow$Gaeilge agus \sloppy{Béarla$\leftrightarrow$Maratais}. Go suntasach, aithníonn an staidéar na hipear-pharaiméadair agus an cineál múnla fho-fhocal chun feabhas suntasach a chur ar chaighdeán aistriúcháin mhúnlaí an Trasfhoirmeora le haghaidh péirí teanga ísealacmhainne.

Féadtar leis an easnamh de thacair shonraí chomhthreomhara constaic a chur roimh fhorbairt MA. Chun aghaidh a thabhairt air seo, d’fhorbraíomar gaHealth, an chéad chorpas sláinte dátheangach do theanga na Gaeilge. 
Ag díriú ar an bhfearann sláinte, léirigh múnlaí a forbraíodh ag baint úsáid as an tacar sonraí seo feabhsuithe suntasacha i scór BLEU nuair a chuirtear iad i gcomparáid le múnlaí ón Tasc Roinnte LoResMT2021.

Léirigh meastóireacht dhaonna ina dhiaidh sin a bhain úsáid as meádracha ilghnéitheacha cáilíochta earráide tascsanmaíochta ardfheidhmíocht chóras an \sloppy{Trasfhoirmeora} maidir le hearráidí cruinnis agus líofachta araon a laghdú i gcomparáid le macasamhail RNN-bunaithe.

Chomh maith leis sin, tugann an tráchtas seo adaptNMT agus adaptMLLM isteach, dhá fheidhm fhoinse-oscailte sruthaithe d’fhorbairt, mionchoigeartú agus úsáid mhúnlaí meaisínaistriúcháin néaraigh. Simplíonn na huirlisí an leagan amach agus an próiseas meastóireachta go mór, rud a dhéanann MA níos inrochtana le haghaidh forbróirí agus aistritheoirí araon.

Rud atá le tabhairt faoi ndeara ná go gcuireann adaptNMT, atá bunaithe i ngnáthóg an OpenNMT, taighde próiseála teanga nádúrtha éiceabháiche chun cinn trí achar timpeallachta an mhúnla fhorbartha a léiriú. Léirigh mionchoigeartú MMLManna ag adaptMLLM forbairtí i leith feidhmíochta aistriúcháin le haghaidh dhá phéire teanga ísealacmhainne: Béarla$\leftrightarrow$Gaeilge agus Béarla$\leftrightarrow$Maratais, i gcomparáid le bonnlínte ón Tasc Roinnte LoRESMT2021.
\end{singlespace}

\thispagestyle{plain}
\fancyfoot[C]{\thepage}

\mainmatter 
\fancyfoot[C]{\thepage}
\chapter{Introduction}
\thispagestyle{fancy}
The digital age has ushered in a plethora of innovations in machine learning (ML) and natural language processing (NLP), delivering solutions to difficult problems. Machine translation (MT) is among the prominent areas that have experienced a significant transformation over the years. While MT has made commendable strides in various language pairs, challenges remain in deploying these advanced models for low-resource languages. One such low-resource language pair that presents unique challenges is the English-to-Irish pair (EN$\leftrightarrow$GA).

The EU flags the importance of digital services being available in all languages to ensure a level playing field and full access to services for citizens, companies and governments (EU Digital Single Market 2018). This has been reinforced by a 2018 EU Parliament decision endorsing language equality in the digital age\footnote{\url{https://www.europarl.europa.eu/doceo/document/A-8-2018-0228_EN.html}} and is a particularly acute issue for Ireland in fostering and protecting the Irish language as the European Commission derogation of Irish as an official EU language expired at the end of 2021.

A major part of this research involves developing applications and methods to address the challenges of low-resource language technology while also addressing the problem of data scarcity affecting deep learning (DL) for digital engagement. Harnessing the full potential of neural machine translation (NMT) for low-resource languages like EN$\leftrightarrow$GA is multifaceted. This research helps deliver parity in support for Irish and other less-resourced languages. 

\section{Neural Networks: an overview}

\begin{figure}[h]
\centering 
\includegraphics[]{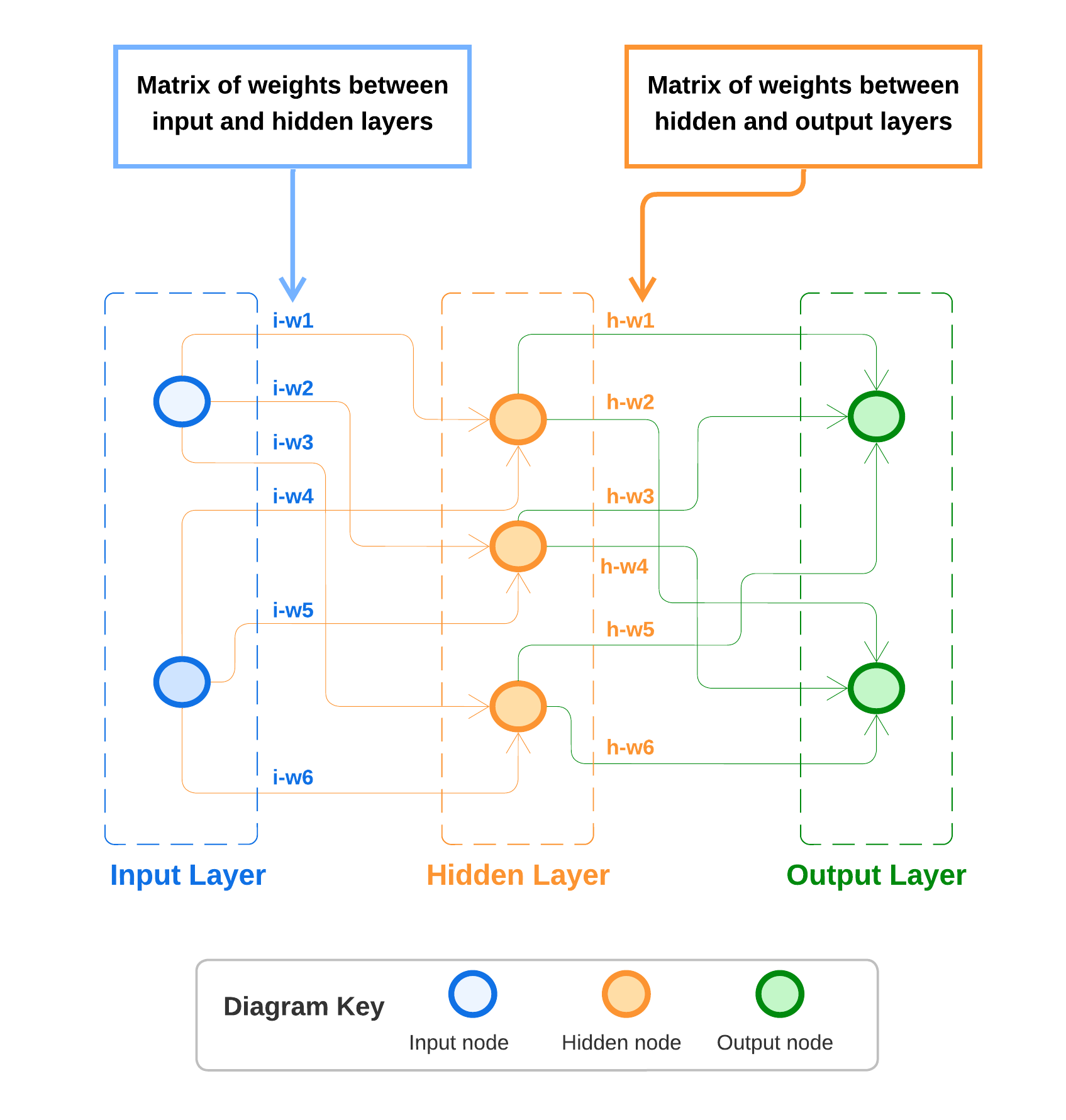}
\caption[Simplified feed-forward NN]{Simplified representation of a feed-forward NN with 3 layers. A matrix of weights, i-w1 to i-w6 connects the input and hidden layers whereas a separate matrix h-w1 to h-w6 connects the hidden and output layers.}
    \label{fig:nn}
\end{figure}

To grasp the concept of a neural network (NN), we can refer to the simplified diagram in Figure \ref{fig:nn}. While real-world NNs used for translating one language to another are considerably more complex, this basic representation helps offer a general insight into NMT. The input layer depicts the source language words, whereas the output layer showcases potential translations suggested by the NN. These interconnections have specific weights, which might be initially set at random.
 
When the network proposes a translation, its accuracy is gauged by comparing it to a known human expert translation. If there is a difference (or error) between the predicted translation and the reference, as perhaps assessed by BLEU~\parencite{papineni-etal-2002-bleu}, this error is fed back into the NN using back-propagation~\parencite{rumelhart1986learning}. This process adjusts the connection weights and retrains the network in a subsequent cycle, involving a complete forward and backward pass through the NN.

Ideally, the network's translation suggestions should come closer to the human reference with each iteration, indicating improved weights. If errors persist, the NN adjusts its weights, re-evaluates, and the cycle continues until no improvements in the BLEU score are observed. After repeatedly refining the network using a comprehensive set of examples from the training dataset, the final model is set. This optimised NN can then be introduced to new sentences it has not encountered before, marking the commencement of the actual translation process.

\section{Hyperparameters and Subword Models}

In the context of ML, \textit{hyperparameters} are distinct from the typical model \textit{parameters}. While model parameters are adjusted during training (e.g., weights and biases in a NN), hyperparameters are set before training begins. Common examples of hyperparameters include the learning rate, which governs how much model parameters are adjusted during training, batch size, which determines the number of data points considered in each iteration, and epochs, specifying how many times the algorithm will run through the entire training dataset. Regularisation parameters, which can help in adding penalties on the magnitude of parameters to prevent overfitting, are another set of important hyperparameters.

The significance of hyperparameters cannot be overstated. Their proper selection and tuning can dramatically influence the performance of the model. A suitable learning rate, for instance, can be the difference between a model that converges efficiently and one that does not converge at all. In scenarios with limited data, such as with low-resource languages, hyperparameter choices become even more critical. It is paramount to extract as much value as possible from the scarce data available, and well-tuned hyperparameters facilitate that.

Languages are diverse and rich, with some exhibiting significant morphological variations. This variability becomes a challenge when training language models, especially for languages that are under-represented in terms of digitally available data. Subword models offer a solution by breaking words into smaller, more manageable units. These units can range from individual characters to larger subword chunks. For example, the word ``unhappiness'' could be segmented into ``un-'', ``-happi-'', and ``-ness'' using a subword tokenization approach. 

For low-resource languages, the use of subword models brings multiple advantages. Firstly, it helps in managing the vocabulary size. Instead of handling every possible word form, the model deals with a more limited set of subword units which makes training of models faster. This approach also provides a mechanism for understanding and generating words the model has not explicitly seen during training. Instead of relegating unfamiliar words to a generic ``unknown'' category, the model can represent and process them using its known subword units. Furthermore, in the realm of transfer learning, where knowledge is transferred from data-rich languages to low-resource ones, subword models perform well. The shared subword units across languages make knowledge transfer more seamless.

Both hyperparameters and subword models play pivotal roles in language modelling, particularly for languages with limited data. While hyperparameters ensure the efficient and effective training of models, subword models offer a practical approach to tokenization that accounts for the morphological richness and data scarcity inherent in many of the world's languages.

\section{Neural Machine Translation}

NMT has achieved impressive performance for many language pairs, especially those with very large amounts of parallel data available (e.g., English to French). However, for low-resource languages, which are languages with limited available parallel data, there are challenges:

\begin{itemize}
  \item \textbf{Data Scarcity}: The primary challenge is the lack of substantial parallel corpora. Neural models, especially the DL models used for NMT, require large amounts of data to train effectively and generalise well to unseen inputs \parencite{koehn2017six, zoph-etal-2016-transfer, sutskever2014sequence}.
  \item \textbf{Noisy Training Data}: To overcome the data scarcity problem, sometimes researchers use web-crawled data or other less reliable sources to augment the training set. Such data can introduce noise due to incorrect alignments or low-quality translations \parencite{schwenk-etal-2021-ccmatrix, ott2019fairseq}.
  \item \textbf{Overfitting}: Due to the limited data, NMT models can overfit easily, which means they might memorise the training data rather than generalising from it. As a result, their performance on unseen data might be suboptimal \parencite{vaswani2017attention, 10.5555/1162264}.
  \item \textbf{Morphological Complexity}: Many low-resource languages are morphological rich, meaning they have a large number of word forms due to inflection, derivation, or compounding. This complexity can pose challenges for NMT models which might not see enough examples of each morphological variant in the training data \parencite{ataman-federico-2018-compositional,sennrich-etal-2016-neural}.
  \item \textbf{Limited Pre-trained Models or Embeddings}: For popular languages, there are often pre-trained models or word embeddings available that can be fine-tuned for specific tasks. For many low-resource languages, these might not exist, forcing researchers to start from scratch \parencite{10.1613/jair.1.11640}.
  \item \textbf{Domain Mismatch}: The available data for low-resource languages might be restricted to specific domains (e.g., religious texts or legal documents). Training on such a restricted domain can limit the extent to which a model can generalise to other domains.
  \item \textbf{Evaluation Challenges}: Even if a model is trained for a low-resource language, evaluating its performance can be problematic due to the lack of standard benchmarks, reference translations, or even native speakers who can assess the translations' quality.
  \item \textbf{Cultural and Contextual Nuances}: All languages convey cultural and contextual nuances. With limited data, the model might miss these subtleties, leading to translations that might be technically correct but culturally or contextually inappropriate or odd.
\end{itemize}

To address these challenges, researchers are exploring various methods, such as using transfer learning, building multilingual models, applying data augmentation techniques and developing new corpora. Translating low-resource languages with high accuracy remains an active area of research. A core objective of our NMT work is to address some of the challenges outlined above by posing and answering the research questions in the next section. An in-depth discussion of NMT architecture and the mathematical first principles governing its operation are covered in Section \ref{nmt-detail}.

\section{Research Questions}

Having introduced the concept of NMT in the context of low-resource languages, we will now discuss the specific research questions this PhD thesis seeks to address. 

\begin{itemize}
    \item RQ$_1$ - How can hyperparameters and subword models be optimised for \sloppy{Transformer-based} MT in low-resource language settings?
    \item RQ$_2$ - What is the benefit and impact of using in-domain datasets in improving the performance of MT models for low-resource language pairs, and how can these datasets be effectively developed and utilised?   
    \item RQ$_3$ - How do NMT systems, specifically Transformer and recurrent neural network (RNN) based models, differ in terms of accuracy and fluency errors when evaluated using a human evaluation technique such as the Multidimensional Quality Metrics (MQM) error taxonomy?
    \item RQ$_4$ - How can the process of NMT development, evaluation, and deployment be streamlined for both developers and translators, while also considering environmental sustainability?
\end{itemize}

\subsubsection{RQ$_1$ How can hyperparameters and subword models be optimised for Transformer-based MT in low-resource language settings?}\label{rw_1}

The effectiveness of fine-tuning the Transformer model for the low-resource EN$\leftrightarrow$GA language pair was investigated in the paper submitted to the MT Summit conference in 2021. Hyperparameter optimisation (HPO) and the impact of different subword models on translation performance were evaluated. The observations made as part of this study were used to choose the hyperparameters for developing an EN$\rightarrow$GA model entered into the LoResMT2021 Shared Task\footnote{\url{https://machinetranslate.org/loresmt2021}}~\parencite{lankford2021machine}. The focus of the shared task was to develop MT models for translating Covid-related data. The experimental findings from this shared task were the basis of a separate paper. This paper highlighted the role of careful selection of Transformer hyperparameters and demonstrated that choosing a 16k BPE SentencePiece submodel yielded high-performing translation models in a low-resource setting \parencite{kudo2018sentencepiece}. 

The research question was also explored as part of the 
human evaluation of EN$\leftrightarrow$GA Transformer-Based NMT where the impact of modifying hyperparameter settings and regularisation techniques on the performance of NMT models was evaluated~\parencite{lankford2022human}.

The theme is further addressed in Chapters 7 and 8 which cover the research papers on open-source architectures for fine-tuning NMT models~\parencite{lankfordlrev} and MLLM models~\parencite{lankford2023adaptLLM}.

\subsubsection{RQ$_2$ - What is the impact of using small in-domain datasets in improving the performance of MT models for low-resource language pairs, and how can these datasets be effectively developed and utilised?}\label{rw_2}

The fine-tuning experiments conducted as part of the winning LoResMT2021 Shared Task entry demonstrate that augmenting in-domain data, even by relatively modest amounts, for EN$\leftrightarrow$GA translations in the Covid domain leads to much better performance when compared with other techniques.  

Given the success of the initial approach of using augmented data for fine-tuning Covid models, a separate study was conducted that resulted in the development of an in-domain health dataset. The development of the gaHealth corpus\footnote{\url{https://github.com/seamusl/gaHealth}} showcases the process and benefits of focusing on specific domains for low-resource languages~\parencite{lankford2022lrec}. 

To build a bilingual corpus of health data, we selected multiple sources of professionally translated documents from within the Irish government, all of which are publicly available. In particular, the bilingual strategy statements and annual reports of the Irish Department of Health since 2010 were chosen. 

Furthermore, a dataset of Covid-related data, developed for a previous study~\parencite{lankford2021machine}, was incorporated into a larger health dataset. This amalgamated corpus, gaHealth, consists of 16,201 lines of parallel text.  

The main contribution of this work is to present an ongoing translation project that aims to build a parallel corpus, gaHealth, of health data for the Irish language by fully utilising freely available parallel documents. 

Due to the issues encountered during the conversion of PDF documents, we developed guidelines for researchers developing low-resource corpora to aid in the conversion process~\parencite{lankford2022lrec}. In addition to developing the gaHealth corpus, we trained and evaluated translation models for in-domain health data.  

There is no such corpus available according to the best of our knowledge, so gaHealth has become a useful resource in the NLP community, especially for those working with the Irish language domain. Addressing this research question has helped to increase the resources available for a low-resource language. The outcome of the work was published at the LREC 2022 conference in Marseille, France. 

\subsubsection{RQ$_3$ - How do MT systems, specifically Transformer and RNN-based models, differ in terms of accuracy and fluency errors when evaluated using a human evaluation technique such as the Multidimensional Quality Metrics (MQM) error taxonomy?}\label{rw_3}

In addressing this research question, a quantitative fine-grained manual evaluation was conducted which compared the performance of MT systems. Using the MQM~\parencite{lommel2014multidimensional} error taxonomy, a human evaluation of the error types generated by an RNN-based system and a Transformer-based system was conducted. Furthermore, Scalar Quality Metrics (SQM)~\parencite{freitag-etal-2021-experts} were also used so a combined approach helped end users to understand translation quality from a human, rather than a solely automatic perspective. 

An integral part of the human evaluation was to highlight the linguistic weaknesses associated with certain MT architectures which were discovered as part of the linguistic observations. Our findings show the best-performing Transformer system significantly reduces both accuracy and fluency errors when compared with an RNN-based model. 

\subsubsection{RQ$_4$ - How can the process of NMT development, evaluation, and deployment be streamlined for both developers and translators, while also considering environmental sustainability?}\label{rw_4}

To address this research question, two separate open-source tools were developed and journal papers describing them in detail were written \parencite{lankfordlrev,lankford2023adaptLLM}. The manner in which the features of each tool addressed this research question is outlined below.

\begin{enumerate}
    \item Streamlining for both developers and translators:

    \begin{itemize}
    \item The applications are designed for both technical and non-technical users in the field of MT.
    \item The setup of the development environment and the creation of training, validation, and test splits are simplified.
    \item HPO is made user-friendly through an intuitive interface.
    \end{itemize}
    \item Evaluation and Deployment:
    \begin{itemize}   
    \item Models can be evaluated using BLEU, TER~\parencite{snover2006study} and ChrF~\parencite{popovic2015chrf}.
    \item Deployment is facilitated as a translation service within the application itself.
    \end{itemize}
    
    \item Environmental Sustainability:
    
    \begin{itemize}
    \item The application contains a green report that flags power consumption and kgCO\textsubscript2 emissions generated during model development. The inclusion of this feature creates an awareness of the environmental cost amongst developers and translators. This helps in addressing the environmental sustainability aspect of the research question.   
    \end{itemize}

\end{enumerate}
\section{Thesis Outline}

\begin{figure}[h!]
\centering 
\includegraphics[width=16cm]{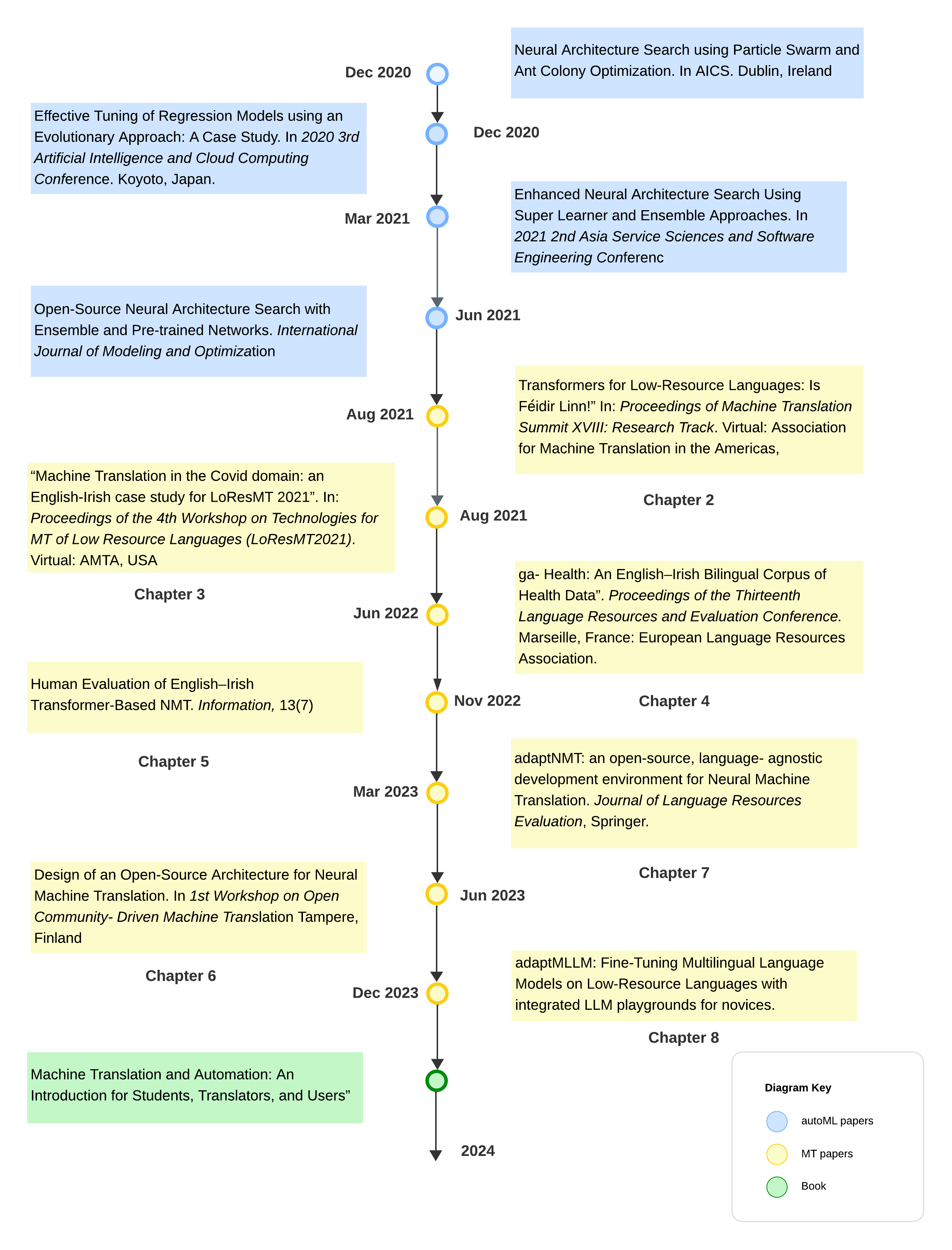}
\caption[Timeline of publications and mapping to thesis chapters.]{Timeline of publications and mapping to thesis chapters. The illustration highlights the initial set of papers on automated machine learning (AutoML), followed by the MT papers and culminating in a book on the topic of MT and automation~\parencite{moorkens-lankford-way}. }
    \label{fig:timeline}
\end{figure}

The outline of the thesis provides an overview of the papers, the principal motivation for the work and the key research contributions. The adopted approach dedicates a chapter to each published paper which focuses on the outlined research questions. At the start of each chapter, a context section outlines the motivation for the research work and illustrates the flow in how the research questions were addressed. The timeline of the paper publications and their mapping to chapters within the thesis is illustrated in Figure \ref{fig:timeline}.

\textbf{Chapter 2: Transformers for Low-Resource Languages:} This chapter examines the effectiveness of the Transformer model, particularly in MT scenarios where limited training data is available, such as with the EN$\leftrightarrow$GA language pair. By testing model configuration parameters, the study found significant performance improvements. Notably, the choice of subword model was crucial, with SentencePiece models using both unigram and BPE methods being examined. Overall, this paper demonstrates that Transformers can be efficient for a low-resource language pair.

\textbf{Chapter 3: Machine Translation in the Covid domain:} an EN$\rightarrow$GA case study for LoResMT2021 Shared Task. Focused on the specific challenge of translating EN$\rightarrow$GA Covid-related data, this chapter describes the paper which explored different domain adaptation techniques. The study also introduced a unique EN$\leftrightarrow$GA dataset concerning health and education in the Covid context. Remarkably, just by adding 5k lines to an 8k in-domain baseline dataset, the BLEU score improved by a massive 27 points, with the Transformer architecture delivering the best performance.

\textbf{Chapter 4: gaHealth: An EN$\leftrightarrow$GA Bilingual Corpus of Health Data:} Highlighting the challenges and potential of MT for low-resource languages, this chapter discusses the motivation for gaHealth, a bilingual corpus geared towards health data in the EN$\leftrightarrow$GA language pair. The research showcased how using a specific in-domain dataset, in this case for health, could significantly improve translation performance. By leveraging the gaHealth corpus, BLEU scores improved by a substantial 22.2 points when stacked against models from the LoResMT2021 Shared Task. The paper also shared linguistic guidelines for developing such datasets.

\textbf{Chapter 5: Human Evaluation of EN$\leftrightarrow$GA Transformer-Based NMT:} The previous work focused on automatic evaluation whereas the ultimate validation of the research effort can only be determined by a human analysis of the MT output. Therefore the chapter focuses on human evaluation of Transformer-based NMT for the EN$\leftrightarrow$GA pair. The paper found that Transformer-optimised models significantly outperformed RNN models. A fine-grained manual evaluation using the MQM error taxonomy identified that the Transformer models considerably cut down on both accuracy and fluency errors. The linguistic observations noted by our translators also highlight the typical errors which occur in EN$\leftrightarrow$GA MT. Suggestions on how these errors could potentially be mitigated are outlined in the future work section of Chapter 9.

\textbf{Chapter 6: Design of an Open-Source Architecture for Neural Machine Translation:} The paper presented at the inaugural CrowdMT workshop in Tampere is discussed in this chapter. As a workshop paper, the research objective was to present a high-level view of the adaptNMT architecture. A more in-depth analysis of the tool is provided in the LREV journal paper which is covered in Chapter 7.  

\textbf{Chapter 7: adaptNMT: an open-source, language-agnostic development environment for Neural Machine Translation:} This chapter includes the paper which introduced adaptNMT, an integrated open-source application designed for both newcomers and experts in MT. It simplifies various processes, from setting up the environment to model training, while also enabling hyperparameter customisation through a user-friendly interface. The application integrates graphing for training progress visualisation and uses SentencePiece for subword segmentation. One useful feature is the green report that brings to light the environmental cost associated with model development in terms of power consumption and kgCO\textsubscript2 emissions, thus encouraging researchers to choose an alternative greener infrastructure. 

\textbf{Chapter 8: adaptMLLM: Fine-Tuning Multilingual Language Models on Low-Resource Languages with integrated LLM playgrounds:} Continuing the theme of open-source tools, this chapter describes the paper on adaptMLLM which describes the tool designed to fine-tune MLLMs. This application is user-friendly, streamlining the process of setting up the development environment and customising hyperparameters. Tested on two low-resource language pairs, EN$\leftrightarrow$GA and English-Marathi (EN$\leftrightarrow$MR), it demonstrated significant performance improvements over the bespoke NMT system which won the LoResMT2021 Shared Task. 

\textbf{Chapter 9: Conclusion:} The concluding chapter revisits the research questions and highlights the research contributions before finally discussing the lessons learnt as well as avenues for future research. In the future work section, roadmaps are laid out for future development work on the adaptNMT and adaptMLLM applications. While the performance of MT systems developed through adaptMLLM is higher, it also demands an infrastructure with a higher specification. Therefore, it is anticipated there will be separate roles for both applications going forward.

\section{Research Contributions}

The transformative potential of the Transformer model in MT is undeniable, as its adoption has heralded state-of-the-art (SOTA) results for many language pairs. However, as illustrated in our paper \textit{``Transformers for Low-Resource Languages - is féidir linn!''},\footnote{The title of the paper was inspired by Obama's 2011 presidential visit to Ireland where he adapted his campaign slogan \textit{``Yes we can!''} to its Gaelic equivalent: \textit{``Is Féidir Linn!''}.} when faced with limited training data, even advanced NMT models like the Transformer may fall short of expectations. This limitation underscores the necessity to fine-tune models and optimise hyperparameters for better translation performance, particularly for low-resource language pairs like EN$\leftrightarrow$GA. The pivotal role of subword models, especially in determining MT translation performance in low-resource scenarios was emphasised. In addressing RQ1, the Transformer architecture was optimised yielding impressive gains in BLEU scores and other performance metrics, thus bridging the gap posed by insufficient amounts of training data.  

The relevance and potential of domain-specific translation cannot be overstated, especially in the current global landscape marked by unpredictable challenges. The recent Covid-19 pandemic highlighted the urgent need for accurate translation within the health domain, especially for languages that lack large resources. In the paper, \textit{``MT in the Covid domain: an EN$\leftrightarrow$GA case study for LoResMT2021 Shared Task''}, cf. Chapter 3, the efficacy of domain adaptation techniques was explored to address RQ2. Through this research, the significance of developing specialised datasets for domain-specific translation was highlighted. Augmenting in-domain data with a modest amount of domain-specific data yielded MT models which obtained 1st place in the LoResMT2021 Shared Task. Such findings emphasise the potential that lies in domain-focused MT models and drive advancements in domain-specific translation.

The idea of domain specificity was further developed by creating a corpus specifically for the health sector. As highlighted in the paper \textit{``gaHealth: An English–Irish Bilingual Corpus of Health Data''}, cf. Chapter 4, there is a marked absence of parallel datasets tailored for low-resource languages. While the rush to build larger, more encompassing datasets is understandable, the merits of focused, in-domain datasets have often been overshadowed. The development of an in-domain corpus and subsequent results from training an MT system with the gaHealth corpus demonstrate the tangible benefits of deploying in-domain datasets for translation tasks. Moreover, in addressing RQ2, this study has provided a valuable resource for the wider NLP community.

In addressing the research question posed by RQ3, a detailed human evaluation was carried out to validate the translation outputs of RNN and Transformer models for a low-resource language pair. The approach taken and the findings are described in the paper, `\textit{`Human Evaluation of English–Irish Transformer-Based NMT''}, cf. Chapter 5. We utilised the MQM error framework to investigate the types of errors produced by both RNN and Transformer-based systems. Native Irish speakers were collaborated with for this assessment, and one of the research contributions was to establish how a human evaluation could be conducted with limited resources. As such, we adopted a simplified MQM error taxonomy, pairing it with an SQM. By engaging two native speakers and analysing 25 reference translations, we were able to validate the outputs from both the RNN and Transformer models. Our results indicate the top-performing Transformer system notably minimises errors in both accuracy and fluency when compared with its RNN counterpart. A significant takeaway from our study is the linguistic insights shared by our translators, which are elaborated on in the related paper (cf. Section \ref{lo}).  

These practices were combined into a new framework which was subsequently applied when evaluating the output from MLLMs in our later work. Furthermore, in a separate contribution, this human evaluation component was integrated into the adaptNMT and adaptMLLM applications effectively bridging RQ3 with RQ4.

The research question identified by RQ4 was addressed in the subsequent journal paper, \textit{``adaptNMT: an open-source, language-agnostic development environment for Neural Machine Translation''}, cf. Chapter 7. The application introduces an open-source application designed to simplify the NMT development process. adaptNMT is a representation of the increasing demand for tools that are intuitive, powerful, and user-friendly. One of the key features of this tool is its commitment to sustainable NLP research, evident from its green report, which calculates the environmental costs of model training. By highlighting the environmental cost of training models, users are encouraged to think about the carbon footprint of their research activities and are consequently incentivised to use a greener infrastructure or pre-trained models, where possible.  Furthermore, adaptNMT's role in the broader NLP community is underlined by its open-source nature, inviting collaborative enhancement from researchers worldwide.

Expanding on the foundation set by adaptNMT, RQ4 was further developed in the paper \textit{``adaptMLLM: Fine-Tuning Multilingual Language Models on Low-Resource Languages with integrated LLM playgrounds''}, cf. Chapter 8. Recognising the transformative potential of MLLMs and LLMs, the paper showcases the very significant gains in translation quality achieved by fine-tuning MLLMs and LLMs for low-resource language pairs. The adaptMLLM tool democratises cutting-edge NLP technologies, demonstrating that SOTA results can be achieved without the backing of colossal research infrastructures. Moreover, its open-source nature mirrors the ethos of adaptNMT, advocating for community-driven advancements in the field.

\section{Publications}

My PhD research has culminated in the publication of eleven peer-reviewed papers, and presentations both virtual and physical at conferences in Dublin, Kyoto, Macau, Orlando and Tampere. Best presentation awards were received at the ACM conferences in Kyoto (2020)\footnote{\url{http://www.adip.org/2020.html}} and Macau (2021).\footnote{\url{http://www.aibc.org/2021.html}} Furthermore, I developed a translation system for LoResMT2021 which obtained first place in the EN$\rightarrow$GA shared task~\parencite{ojha2021findings}.

All of the recent work, comprising seven publications, focuses on advancements in NMT and the development of user-friendly tools. These MT-related papers have been included as individual chapters in this thesis. The full set of peer-reviewed MT publications is outlined below.

\begin{enumerate}

\item
Lankford, S., Afli, H. and Way, A., 2023. adaptMLLM: Fine-Tuning Multilingual Language Models on Low-Resource Languages with integrated LLM playgrounds. In: \textit{Information 14.12}, issn: 2078-2489, doi: 10.3390/info14120638, url: https://www.mdpi.com/2078-2489/14/12/638.

\item
Lankford, S., Afli, H. and Way, A., 2023. adaptNMT: an open-source, language-agnostic development environment for neural machine translation. In: \textit{Language Resources and Evaluation}, pp.1-26. \\ url: https://doi.org/10.1007/s10579-023-09671-2

\item
Lankford, S., Afli, H. and Way, A., 2023. Design of an Open-Source Architecture for Neural Machine Translation. In: \textit{Proceedings of the 1st Workshop on Open Community-Driven Machine Translation}. Tampere, Finland: European Association for Machine Translation, pp. 15–20. \\ url: https://aclanthology.org/2023.crowdmt-1.2.

\item
Lankford, S., Afli, H. and Way, A., 2022. Human Evaluation of English-Irish Transformer-Based NMT. In: \textit{Information 13.7}, issn: 2078-2489. doi: 10.3390/info13070309. url: https://www.mdpi.com/2078-2489/13/7/309.

\item 
Lankford, S., Afli, H., Ní Loinsigh Ó., Way, A., 2022. “gaHealth: An English–Irish Bilingual Corpus of Health Data”. In: \textit{Proceedings of the Thirteenth Language Resources and Evaluation Conference}. Marseille, France: European Language Resources Association, pp. 6753–6758. url: https://aclanthology. org/2022.lrec-1.727.

\item 
Lankford, S., Afli, H., and Way, A. 2021. “Transformers for Low-Resource Languages: Is Féidir Linn!” In: \textit{Proceedings of Machine Translation Summit XVIII: Research Track}. Virtual: Association for Machine Translation in the Americas, pp. 48–60. url: https://aclanthology.org/2021.mtsummit-research.5.

\item 
Lankford, S., Afli, H., and Way, A. 2021. “Machine Translation in the Covid domain: an EN$\leftrightarrow$GA case study for LoResMT 2021”. In: \textit{Proceedings of the 4th Workshop on Technologies for MT of Low Resource Languages (LoResMT2021)}. Virtual: Association for Machine Translation in the Americas, pp. 144–150. url: https://aclanthology.org/2021.mtsummit-loresmt.15

\end{enumerate}


\chapter{Transformers for Low-Resource Languages}
\section{Context}

The motivation for this research stems from the recognition that Transformer models are now SOTA in the MT arena, offering remarkable performance on well-resourced language pairs. However, such models often struggle when dealing with low-resource language pairs due to the scarcity of training data. This limitation has led to relatively few experiments and advancements in using Transformers for such language pairs. As a result, my motivation was to address RQ1 by exploring and optimising Transformer models specifically for translating low-resource languages such as the EN$\leftrightarrow$GA language pair.

The key motivations and objectives of this research can be summarised as follows:
\begin{itemize}
   
\item \textbf{Addressing Low-Resource Challenges}: The research seeks to tackle the challenges posed by low-resource language pairs, where conventional NMT models tend to perform poorly due to insufficient training data. The motivation is to bridge the performance gap and enhance translation quality for underrepresented language pairs.

\item \textbf{Hyperparameter Optimisation}: The study focuses on hyperparameter optimisation for Transformer models. This is motivated by the understanding that the performance of MT models is heavily influenced by the choices made in configuring the model's hyperparameters. The goal is to identify the most effective set of hyperparameters for EN$\leftrightarrow$GA translation. 

\item \textbf{Subword Model Selection}: Another key motivation is to determine the optimal subword model for low-resource translation. This choice is critical as it impacts the model's ability to handle the linguistic nuances of the target language. The research evaluates different subword models, including SentencePiece models with unigram and BPE approaches.

\item \textbf{Model Architecture Exploration}: Various aspects of the Transformer architecture were investigated to enhance translation performance. This included modifying the number of layers, testing regularisation techniques, and evaluating the ideal number of attention heads. The goal is to identify the best architectural configurations for the given language pair.

\item \textbf{Benchmarking and Comparison}: To assess the effectiveness of the optimised Transformer models, the research benchmarks them against existing translation systems, including Google Translate. This comparison provides empirical evidence of the improvements achieved by their proposed approach.

\item \textbf{Reducing Post-Editing Effort}: By improving translation quality, the research aims to reduce the post-editing effort required for low-resource language pairs. This has practical implications for making MT more accessible and cost-effective in scenarios where human expertise is required for quality control.
\end{itemize}

In summary, the motivation for this paper is to address the challenges faced by low-resource language pairs in MT by optimising Transformer models. The study explores hyperparameter selection, subword modelling, and architectural modifications to achieve substantial performance improvements, ultimately making MT more effective and accessible for EN$\leftrightarrow$GA translation.

\clearpage

   \begin{center}
       \vspace*{1cm}

       \textbf{Transformers for Low-Resource Languages: Is Féidir Linn!}
            
       \vspace{1.5cm}

       \textbf{Séamus Lankford \\ Haithem Afli \\ Andy Way}

       \vfill

Proceedings of Machine Translation Summit XVIII \\ 
Research Track. Virtual: Association for Machine Translation in the Americas \
August 16 - 20, 2021 \\ 
Florida, USA \\
       \vspace{0.8cm}
                 
       ADAPT Centre\\
       Dublin City University\\
       Ireland\\

        \vspace{0.5cm}
    \url{https://aclanthology.org/2021.mtsummit-research.5.pdf}            
   \end{center}

\clearpage

\section{Abstract} 
The Transformer model is state-of-the-art in MT. However, in general, neural translation models often underperform on language pairs with insufficient training data. As a consequence, relatively few experiments have been carried out using this architecture on low-resource language pairs. In this study, hyperparameter optimisation of Transformer models in translating the low-resource English-Irish language pair is evaluated. We demonstrate that choosing appropriate hyperparameters leads to considerable performance improvements. Most importantly, the correct choice of subword model is shown to be the biggest driver of translation performance. SentencePiece models using both unigram and BPE approaches were appraised. Variations on model architectures included modifying the number of layers,  testing various regularisation techniques and evaluating the optimal number of heads for attention. A generic 55k DGT corpus and an in-domain 88k public admin corpus were used for evaluation. A Transformer-optimised model demonstrated a BLEU score improvement of 7.8 points when compared with a baseline RNN model.  Improvements were observed across a range of metrics, including TER, indicating a substantially reduced post-editing effort for Transformer-optimised models with 16k BPE subword models. Benchmarked against Google Translate, our translation engines demonstrated significant improvements. The question of whether or not Transformers can be used effectively in a low-resource setting of English-Irish translation has been addressed. Is féidir linn - yes we can.

\section{Introduction}

The advent of neural machine translation (NMT) has heralded an era of high-quality translations. However, these improvements have not been manifested in the translation of all languages. Large datasets are a prerequisite for high-quality NMT. This works well in the context of well-resourced languages where there is an abundance of data. In the context of low-resource languages which suffer from a sparsity of data, alternative approaches must be adopted. 

An important part of this research involves developing applications and models to address the challenges of low-resource language technology. Such technology incorporates methods to address the data scarcity affecting deep learning for digital engagement of low-resource languages.  

It has been shown that an out-of-the-box NMT system, trained on English-Irish (EN$\leftrightarrow$GA) data, achieves a lower translation quality compared with using a tailored SMT system (Dowling et al, 2018). It is in this context that further research is required in the development of NMT for low-resource languages and the Irish language in particular.

Most research on choosing subword models has focused on high-resource languages~\parencite{ding2019call, gowda2020finding}. In the context of developing models for EN$\leftrightarrow$GA translation, there are no clear recommendations on the choice of subword model types. One of the objectives of this study is to identify which type of subword model performs best in this low-resource scenario.

\section{Background}
   
Native speakers of low-resource languages are often excluded from useful content since, more often than not, online content is not available to them in their language of choice. Such a digital divide and the resulting social exclusion experienced by second-language speakers, such as refugees living in developed countries, has been well documented in the research literature ~\parencite{macfarlane2008responses, alam2015digital}.

Research on machine translation (MT) in low-resource scenarios directly addresses this challenge of exclusion via pivot languages ~\parencite{liu2018pivot}, and indirectly, via domain adaptation of models ~\parencite{ghifary2016deep}. Breakthrough performance improvements in the area of MT have been achieved through research efforts focusing on NMT \parencite{bahdanau2014neural, cho2014properties}. Consequently, state-of-the-art (SOTA) performance has been attained on multiple language pairs~\parencite{bojar-etal-2017-findings, bojar-etal-2018-findings}.

\subsection{Irish Language}
The Irish language is a primary example of such a low-resource language that will benefit from this research. NMT involving Transformer model development will improve the performance in specific domains of low-resource languages. Such research will address the end of the Irish language derogation in the European Commission in 2021\footnote{\url{http://amtaweb.org/wp-content/uploads/2020/11/MT-in-EU-Overview-with-Voiceover-Andy-Way-KEYNOTE-K1.pdf}} helping to deliver parity in support for Irish in online digital engagement.
\subsection{Hyperparameter Optimisation}

Hyperparameters are employed to customise machine learning models such as translation models. It has been shown that machine learning performance may be improved through HPO rather than just using default settings~\parencite{sanders2017informing}.

The principle methods of HPO are grid search \parencite{montgomery2001design} and random search \parencite{JMLR:v13:bergstra12a}]. Grid search is an exhaustive technique which evaluates all hyperparameter permutations. However, as the number of features grows, the amount of data permutations grows exponentially making optimisation expensive in the context of developing long-running translation models. 
 
An effective, and less computationally intensive, alternative is to use random search which samples random configurations. 

\subsubsection{Recurrent Neural Networks}

Recurrent neural networks (RNNs) are often used for the tasks of natural language processing, speech recognition and MT. RNN models enable previous outputs to be used as inputs while having hidden states. In the context of MT, such neural networks were ideal due to their ability to process inputs of any length. Furthermore, the model sizes do not necessarily increase with the size of its input. Commonly used variants of RNN include bidirectional (BRNN) and deep (DRNN) architectures. However, the problem of vanishing gradients coupled with the development of attention-based algorithms often leads to Transformer models performing better than RNNs.

\subsubsection{Transformer}
The greatest improvements have been demonstrated when either the RNN or the CNN architecture is abandoned completely and replaced with an attention mechanism creating a much simpler and faster architecture known as Transformer \parencite{vaswani2017attention}. 
Transformer models use attention to focus on previously generated tokens. The approach allows models to develop a long memory which is particularly useful in the domain of language translation. Performance improvements to both RNN and CNN approaches may be achieved through the introduction of such attention layers in the translation architecture.

Experiments in MT tasks show such models are better in quality due to greater parallelisation while requiring significantly less time to train. 

\subsection{Subword Models}

Translation, by its nature, requires an open vocabulary and the use of subword models aims to address the fixed vocabulary problem associated with NMT. Rare and unknown words are encoded as sequences of subword units. By adapting the original Byte Pair Encoding (BPE) algorithm~\parencite{gage1994new}, the use of BPE submodels can improve translation performance~\parencite{sennrich2015neural,kudo2018subword}. 

Designed for NMT, SentencePiece, is a language-independent subword tokenizer that provides an open-source C++ and a Python implementation for subword units. An attractive feature of the tokenizer is that SentencePiece trains subword models directly from raw sentences~\parencite{kudo2018sentencepiece}.

\subsubsection{Byte Pair Encoding compared with Unigram}
BPE and unigram language models are similar in that both encode text using fewer bits but each uses a different data compression principle (dictionary vs. entropy). In principle, we would expect the same benefits with the unigram language model as with BPE. However, unigram models are often more flexible since they are probabilistic models that output multiple segmentations with their probabilities.

\begin{figure}
    \centering
    \includegraphics[width=12cm]{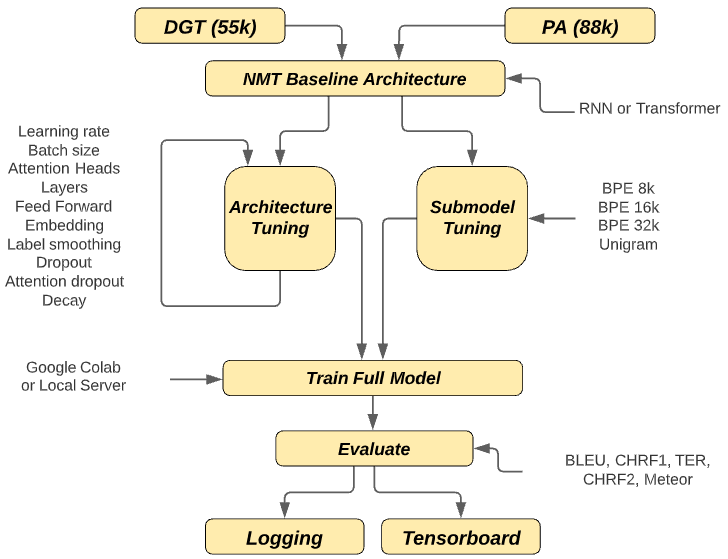}
    \caption{Proposed approach of Transformers for low-resource languages}
    \label{fig:approach_mtsummit}
\end{figure}

\begin{center}
\begin{table}
\center
\begin{tabular}{ll}
\hline
\textbf{Hyperparameter} & \textbf{Values}                \\ \hline
Learning rate            & 0.1, 0.01, 0.001, \textbf{2}            \\ \hline
Batch size               & 1024, \textbf{2048},  4096, 8192       \\ \hline
Attention heads          & \textbf{2}, 4, \textbf{8}                     \\ \hline
Number of layers         & 5, \textbf{6}                           \\ \hline
Feed-forward dimension   & \textbf{2048}                           \\ \hline
Embedding dimension      & 128, \textbf{256}, 512                  \\ \hline
Label smoothing          & \textbf{0.1}, 0.3                       \\ \hline
Dropout                  & 0.1, \textbf{0.3}                       \\ \hline
Attention dropout        & \textbf{0.1}                            \\ \hline
Average Decay            & 0, \textbf{0.0001}                      \\ \hline
\end{tabular}
\caption[Hyperparameter optimisation for Transformer models]{Hyperparameter optimisation for Transformer models. Optimal hyperparameters are highlighted in bold. The highest performing model trained on the 55k DGT corpus uses 2 attention heads whereas the best model trained with the larger 88k PA dataset uses 8 attention heads.}
\label{tab:hpo-table-trans}
\end{table}
\end{center}

\section{Proposed Approach}

HPO of RNN models in low-resource settings has previously demonstrated considerable performance improvements. The extent to which such optimisation techniques may be applied to Transformer models in similar low-resource scenarios is evaluated as part of this study. Evaluations included modifying the number of attention heads, the number of layers and experimenting with regularisation techniques such as dropout and label smoothing. Most importantly, the choice of subword model type and the vocabulary size are evaluated. 

To test the effectiveness of our approaches, optimisation was carried out on two EN$\leftrightarrow$GA parallel datasets: a general corpus of 52k lines from the Directorate General for Translation (DGT) and an in-domain corpus of 88k lines of Public Administration (PA) data. With DGT, the test set used 1.3k lines and the development set comprised 2.6k lines. In the case of the PA dataset, there were 1.5k lines of test data and 3k lines of validation.  All experiments involved concatenating source and target corpora to create a shared vocabulary and a shared SentencePiece subword model.  The impact of using separate source and target subword models was not explored.

The approach adopted is illustrated in Figure \ref{fig:approach_mtsummit}. Two baseline architectures, RNN and Transformer, are evaluated. On evaluating the hyperparameter choices for Transformer models, the values outlined in Table \ref{tab:hpo-table-trans} were tested using a random search approach. A range of values for each hyperparameter was tested using short cycles of 5k training steps. Once an optimal value, within the sampled range was identified, it was locked in for tests on subsequent hyperparameters.

\subsection{Architecture Tuning}

Given the long training times associated with NMT, it is difficult and costly to tune systems using a conventional grid search approach. Therefore a random search approach was adopted in the HPO of our transformer models. 

With low-resource datasets, the use of smaller and fewer layers has previously been shown to improve performance \parencite{araabi2020optimizing}. Performance of low-resource NMT has also been demonstrated to improve in cases where shallow Transformer models are adopted ~\parencite{van2020optimal}. Guided by these findings, configurations were tested which varied the number of neurons in each layer and modified the number of layers used in the Transformer architecture.

The impact of regularisation, by applying varying degrees of dropout to Transformer models, was evaluated. Configurations using smaller (0.1) and larger values (0.3) were applied to the output of each feed-forward layer.

\subsection{Subword Models}

It has become standard practice to incorporate word segmentation approaches, such as Byte-Pair-Encoding (BPE) when developing NMT models. Previous work shows that subword models may be particularly beneficial for low-resource languages since rare words are often a problem. Reducing the number of BPE merge operations resulted in substantial improvements of 5 BLEU points (Sennrich and Zhang 2019) when tested on RNN models.

In the context of EN$\leftrightarrow$GA translation, there is no clear agreement as to what constituted the best approach. Consequently, as part of this study, subword regularisation techniques, involving BPE and unigram models were evaluated to determine the optimal hyperparameters for maximising translation performance. BPE models with varying vocabulary sizes of 4k, 8k, 16k and 32k were tested.

\section{Empirical Evaluation}

\subsection{Experimental Setup}
\subsubsection{Datasets}
The performance of the Transformer and RNN approaches is evaluated on EN$\leftrightarrow$GA parallel datasets. Two datasets were used in the evaluation of our models namely the publicly available DGT dataset which may be broadly categorised as generic and an in-domain dataset which focuses on public administration data.

The DGT, and its Joint Research Centre, have made available all Translation Memory (TM; i.e. sentences and their professionally produced translations) which cover all official European Union languages~\parencite{steinberger2013dgt}. 

Data provided by the Department of Tourism, Culture, Arts, Gaeltacht, Sport and Media in Ireland formed the majority of the data in the public administration dataset. This includes staff notices, annual reports, website content, press releases and official correspondence. 

Parallel texts from the Digital Corpus of the European Parliament (DCEP) and the DGT are included in the training data. Crawled data, from sites of a similar domain are included. Furthermore, a parallel corpus collected from Conradh na Gaeilge (CnaG), an Irish language organisation that promotes the Irish language, was included. The dataset was compiled as part of a previous study which carried out a preliminary comparison of SMT and NMT models for the Irish language ~\parencite{dowling2018smt}. 

\subsubsection{Infrastructure}
Models were developed using a lab of machines each of which has an AMD Ryzen 7 2700X processor, 16GB memory, a 256GB SSD and an NVIDIA GeForce GTX 1080 Ti. Rapid prototype development was enabled through a Google Colab Pro subscription using NVIDIA Tesla P100 PCIe 16GB graphic cards and up to 27GB of memory when available~\parencite{Bisong2019}.

Our MT models were trained using the Pytorch implementation of OpenNMT 2.0, an open-source toolkit for NMT \parencite{klein2017opennmt}. 

\subsubsection{Metrics}

As part of this study, several automated metrics were used to determine the translation quality. All models were trained and evaluated on both the DGT and PA datasets using the BLEU~\parencite{papineni-etal-2002-bleu}, TER \parencite{snover2006study} and ChrF \parencite{popovic2015chrf} evaluation metrics. Case-insensitive BLEU scores, at the corpus level,  are reported. Model training was stopped once an early stopping criteria of no improvement in validation accuracy for 4 consecutive iterations was recorded.

\subsection{Results}

\subsubsection{Performance of subword models}
The impact on translation accuracy when choosing a subword model is highlighted in Tables \ref{tab:dgtvanilla-table-trans} - \ref{tab:patrans-table-trans}.  In training both RNN and Transformer architectures, incorporating any submodel type led to improvements in model accuracy. This finding is evident when training either the smaller generic DGT dataset or the larger in-domain PA dataset. 

Using an RNN architecture on DGT, as illustrated in Table \ref{tab:dgtvanilla-table-trans}, the best-performing model with a 32k unigram submodel, achieved a BLEU score 7.4\% higher than the baseline.  With the PA dataset using an RNN, as shown in Table \ref{tab:pavanilla-table}, the model with the best BLEU, TER and ChrF3 scores again used a unigram submodel.

\begin{table}
\centering
\begin{tabular}{lllllllll}
\hline
\textbf{Architecture} &
  \textbf{BLEU} $\uparrow$ &
  \textbf{TER} $\downarrow$ &
  \textbf{ChrF3} $\uparrow$ &
  \textbf{Steps} &
  \textbf{\begin{tabular}[c]{@{}l@{}}Runtime \\ (hours)\end{tabular}} &
  \textbf{kgCO\textsubscript2} \\ \hline
dgt-rnn-base    & 52.7       & 0.42  & 0.71 & 75k  & 4.47 & 0 \\
dgt-rnn-bpe8k   & 54.6       & 0.40 & 0.73 & 85k  & 5.07 & 0 \\
dgt-rnn-bpe16k  & 55.6 & 0.39 & 0.74 & 100k & 5.58 & 0 \\
dgt-rnn-bpe32k  & 55.3       & 0.39 & 0.74 & 95k  & 4.67 & 0 \\ 
dgt-rnn-unigram & \textbf{55.6} & \textbf{0.39} & \textbf{0.74} & 105k & 5.07 & 0 \\ \hline
\end{tabular}
\caption{RNN performance on DGT dataset of 52k lines}
\label{tab:dgtvanilla-table-trans}
\end{table}

\begin{table}
\centering
\begin{tabular}{lllllllll}
\hline
\textbf{Architecture} &
  \textbf{BLEU} $\uparrow$ &
  \textbf{TER} $\downarrow$ &
  \textbf{ChrF3} $\uparrow$ &
  \textbf{Steps} &
  \textbf{\begin{tabular}[c]{@{}l@{}}Runtime \\ (hours)\end{tabular}} &
  \textbf{kgCO\textsubscript2} \\ \hline
pa-rnn-base     & 40.4 & 0.47 & 0.63 & 60k  & 2.13 & 0 \\
pa-rnn-bpe8k   & 41.5 & 0.46 & 0.64 & 110k  & 4.16 & 0 \\
pa-rnn-bpe16k  & 41.5 & 0.46 & 0.64 & 105k & 3.78 & 0 \\
pa-rnn-bpe32k  & 41.9 & 0.47 & 0.64& 100k & 2.88 & 0 \\
pa-rnn-unigram & \textbf{41.9} & \textbf{0.46} & \textbf{0.64} & 95k & 2.75 & 0 \\ \hline
\end{tabular}
\caption{RNN performance on PA dataset of 88k lines}
\label{tab:pavanilla-table}
\end{table}

There are small improvements in BLEU scores when the RNN baseline is compared with models using a BPE submodel of either 8k, 16k or 32k words, as illustrated in Tables \ref{tab:dgtvanilla-table-trans} and \ref{tab:pavanilla-table}. The maximum BLEU score improvement of 1.5 points (2.5\%) is quite modest in the case of the public admin corpus.  However, there are larger gains with the DGT corpus. A baseline RNN model, trained on DGT, achieved a BLEU score of 52.7 whereas the highest-performing BPE variant,  using a 16k vocab, recorded an improvement of nearly 3 points with a score of 55.6. 
\begin{table}
\centering
\begin{tabular}{lllllllll}
\hline
\textbf{Architecture} &
  \textbf{BLEU} $\uparrow$ &
  \textbf{TER} $\downarrow$ &
  \textbf{ChrF3} $\uparrow$ &
  \textbf{Steps} &
  \textbf{\begin{tabular}[c]{@{}l@{}}Runtime \\ (hours)\end{tabular}} &
  \textbf{kgCO\textsubscript2} \\ \hline
dgt-trans-base      & 53.4 & 0.41 & 0.72 & 55k  & 14.43 & 0.81 \\
dgt-trans-bpe8k     & 59.5 & 0.34 & 0.77 & 200k & 24.48 & 1.38 \\
dgt-trans-bpe16k    & \textbf{60.5} & \textbf{0.33} & \textbf{0.78} & 180k & 26.90 & 1.52 \\
dgt-trans-bpe32k    & 59.3 & 0.35 & 0.77 & 100k & 18.03 & 1.02 \\ 
dgt-trans-unigram   & 59.3 & 0.35  & 0.77 & 125k & 21.95 & 1.24 \\ \hline
\end{tabular}
\caption[Transformer performance on 52k DGT dataset]{Transformer performance on 52k DGT dataset. The highest-performing model uses 2 attention heads. All other models use 8 attention heads.}
\label{tab:trans-table-trans}
\end{table}

\begin{table}
\centering
\begin{tabular}{lllllllll}
\hline
\textbf{Architecture} &
  \textbf{BLEU} $\uparrow$ &
  \textbf{TER} $\downarrow$ &
  \textbf{ChrF3} $\uparrow$ &
  \textbf{Steps} &
  \textbf{\begin{tabular}[c]{@{}l@{}}Runtime \\ (hours)\end{tabular}} &
  \textbf{kgCO\textsubscript2} \\ \hline
pa-trans-base    & 44.1 & 0.44 & 0.66 & 20k  & 5.97  & 0.34 \\
pa-trans-bpe8k   & 46.6 & \textbf{0.40} & 0.68 & 160k & 20.1  & 1.13 \\
pa-trans-bpe16k  & \textbf{47.1} & 0.41 & \textbf{0.68} & 100k & 14.22 & 0.80  \\
pa-trans-bpe32k  & 46.8 & 0.41 & 0.68 & 70k  & 12.7  & 0.72 \\ 
pa-trans-unigram & 46.6 & 0.42 & 0.68 & 75k  & 13.34 & 0.75 \\ \hline
\end{tabular}
\caption[Transformer performance on 88k PA dataset]{Transformer performance on 88k PA dataset. All models use 8 attention heads.}
\label{tab:patrans-table-trans}
\end{table}

In the context of Transformer architectures, highlighted in Table \ref{tab:trans-table-trans} and Table \ref{tab:patrans-table-trans}, the use of subword models delivers significant performance improvements for both the DGT and public admin corpora. The performance gains for Transformer models are far greater than RNN models. Baseline DGT Transformer models achieve a BLEU score of 53.4 while a Transformer model, with a 16k BPE submodel, has a score of 60.5 representing a BLEU score improvement of 13\% at 7.1 BLEU points. 

For translating into a morphologically rich language, such as Irish, the ChrF metric has proven successful in showing a strong correlation with human translation~\parencite{stanojevic2015results}. In the context of our experiments, it worked well in highlighting the performance differences between RNN and Transformer architectures. 

\begin{figure}
    \centering
    \includegraphics[width=12cm]{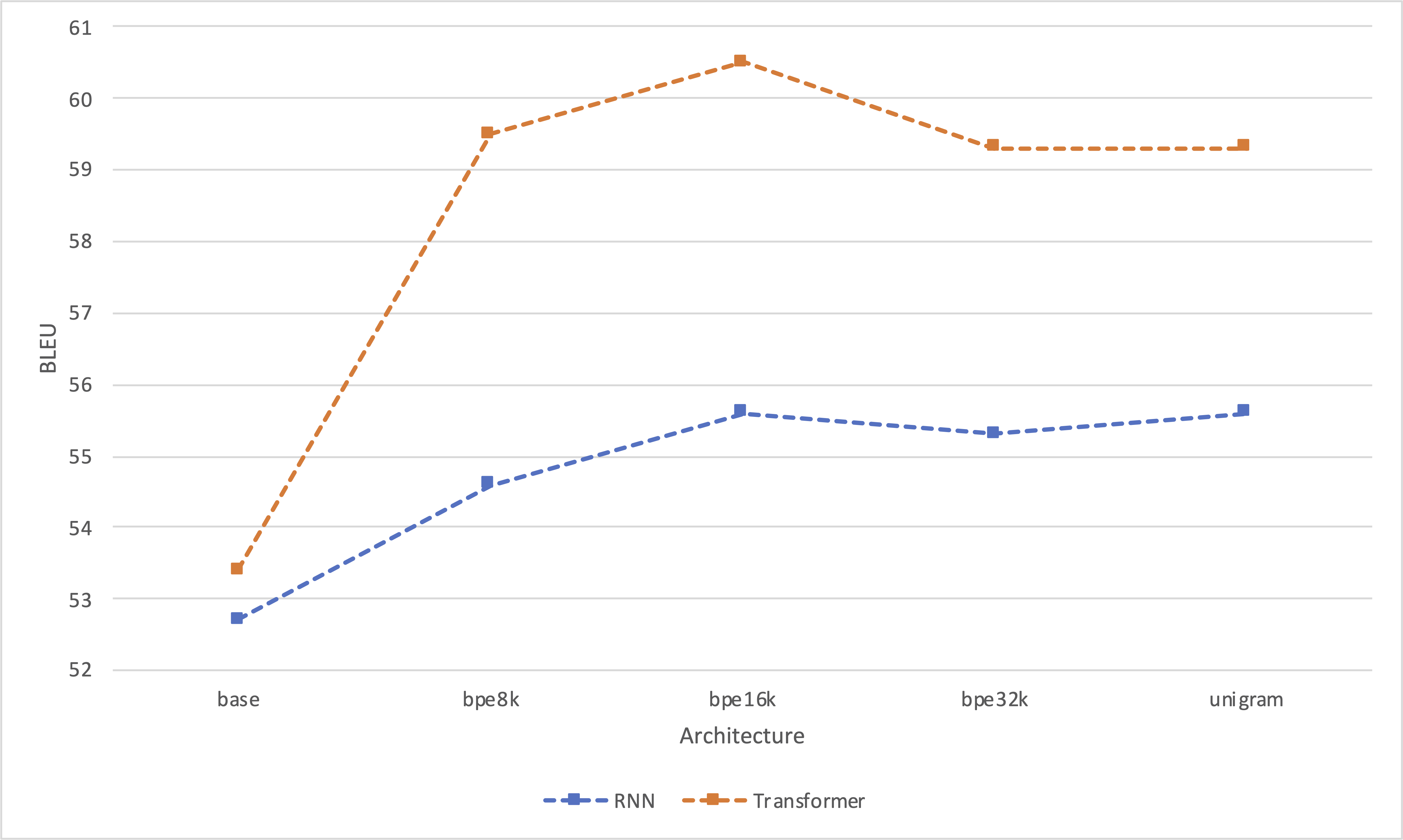}
    \caption{BLEU performance for all model architectures}
    \label{fig:dgt-trans}
\end{figure}

\subsubsection{Transformer performance compared with RNN}

\begin{figure}[h]
    \centering
    \includegraphics[width=12cm]{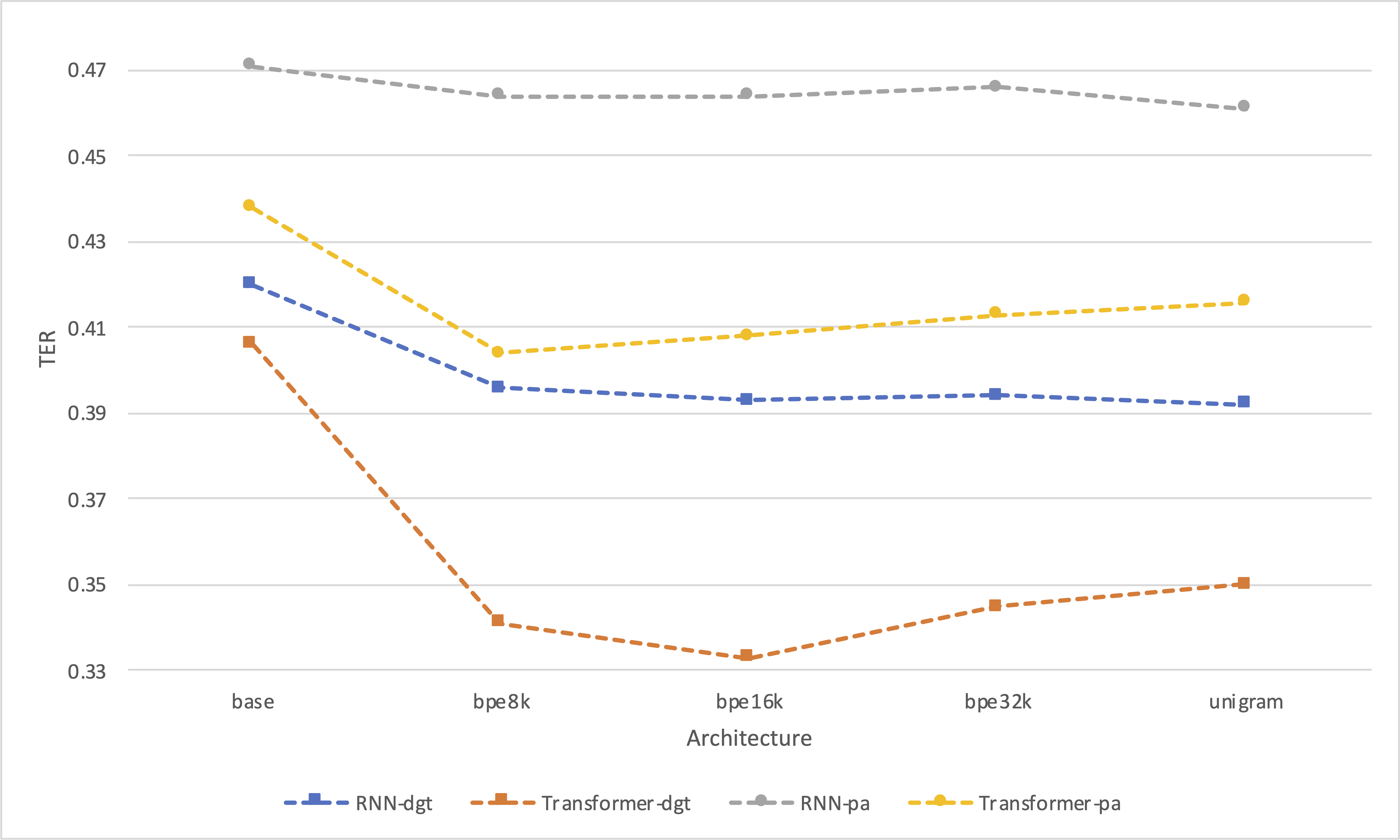}
    \caption{TER performance for all model architectures}
    \label{fig:pacompare-trans}
\end{figure}

The performance of RNN models is contrasted with the Transformer approach in Figure \ref{fig:dgt-trans} and Figure \ref{fig:pacompare-trans}. Transformer models, as anticipated, outperform all their RNN counterparts. It is interesting to note the impact of choosing the optimal vocabulary size for BPE submodels. Both datasets demonstrate that choosing a BPE vocabulary of 16k words yields the highest performance. 

Furthermore, the TER scores highlighted in Figure \ref{fig:pacompare-trans} reinforce the findings that using 16k BPE submodels on Transformer architectures leads to better translation performance. The TER score for the DGT Transformer 16k BPE model is significantly better (0.33) when compared with the baseline performance (0.41).

\begin{figure}[h]
\centering
    \includegraphics[width=7.25cm]{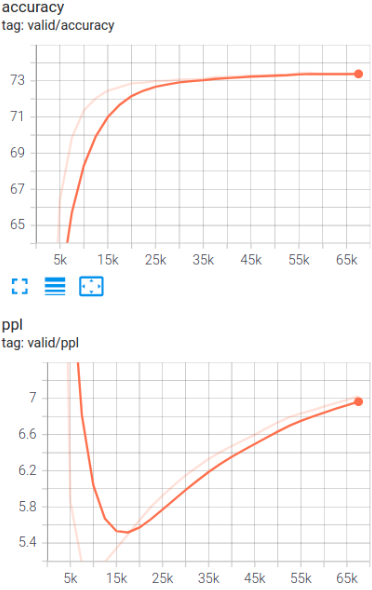}
    \caption{Training DGT Transformer baseline}
    \label{fig:train-basetrans}
\end{figure}

\begin{figure}[h]
\centering
    \includegraphics[width=7.25cm]{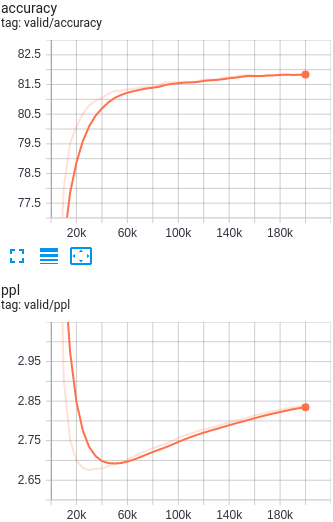}
    \caption{Training DGT Transformer 16k BPE}
    \label{fig:train-bpe-trans}
\end{figure}

\section{Environmental Impact}
Motivated by the findings of Stochastic Parrots \parencite{bender2021dangers}, energy consumption during model development was tracked. Prototype model development used Colab Pro, which as part of Google Cloud is carbon neutral \parencite{lacoste2019quantifying}. However, longer running Transformer experiments were conducted on local servers using 324 gCO\textsubscript2 per kWh\footnote{https://www.seai.ie/publications/Energy-in-Ireland-2020.pdf} \parencite{sei2020}. The net result was just under 10 kgCO\textsubscript2 created for a full run of model development. Models developed during this study will be reused for ensemble experiments in future work.

\section{Discussion}
Validation accuracy, and model perplexity, in developing the baseline and optimal models for the DGT corpus are illustrated in Figure \ref{fig:train-basetrans} and Figure \ref{fig:train-bpe-trans}. Rapid convergence was observed while training the baseline model such that little accuracy improvement occurred after 20k steps. Including a subword model led to much slower convergence and there were only marginal gains after 60k steps. Furthermore, it is observed that training the DGT model, with a 16k BPE submodel, boosted validation accuracy by over 8\% compared with its baseline.

With regard to the key metric of perplexity, it is shown to rise after training for 15k steps in the baseline models. PPL was observed to rise at later stages, typically after 40k steps in models developed using subword models. Perplexity (PPL), shows how many different, equally probable words can be produced during translation. As a metric for translation performance, it is important to keep low scores so the number of alternative translations is reduced. Therefore, for future model development, it may be worthwhile to set PPL as an early stopping hyperparameter.

On examining the PPL graphs of Figure \ref{fig:train-basetrans} and Figure \ref{fig:train-bpe-trans}, it is clear that a lower global minimum is achieved when the Transformer approach is used with a 16k BPE submodel. The PPL global minimum (2.7) is over 50\% lower than the corresponding PPL for the Transformer base model (5.5). Such a finding illustrates that choosing an optimal submodel delivers significant performance gains.

Translation engine performance was benchmarked against Google Translate's \footnote{https://translate.google.com/} EN$\leftrightarrow$GA translation service which is freely available on the internet. Four random samples were selected from the English source test file and are presented in Table~\ref{tab:translations-trans}. Translation of these samples was carried out on the optimal DGT Transformer model and using Google Translate. Case insensitive, sentence level BLEU scores were recorded and are presented in Table~\ref{tab:trans-google-trans}. The results are encouraging and indicate well-performing translation models on the DGT dataset.

The optimal hyperparameters selected in this discovery process are identified in bold in Table~\ref{tab:hpo-table-trans}. A higher initial learning rate of 2 coupled with an average decay of 0.0001 led to longer training times but more accurate models. Despite setting an early stopping hyperparameter, many of the Transformer builds continued for the full cycle of 200k steps over periods of 20+ hours. 

Training transformer models with a reduced number of attention heads led to a marginal improvement in translation accuracy with a smaller corpus. Our best-performing model on a 55k DGT corpus, with 2 heads and a 16k BPE submodel, achieved a BLEU score of 60.5 and a TER score of 0.33. By comparison, using 8 heads with the same architecture and dataset yielded 60.3 for the BLEU and 0.34 for the TER. In the case of a larger 88k PA corpus, all transformer models using 8 heads performed better than equivalent models using just 2 heads.

\begin{table}[h]

\begin{tabular}{p{6.9cm}p{6.9cm}}\hline
\textbf{Source Language (English)} & \textbf{Reference Human Translation (Irish)} \\  \hline
A clear harmonised procedure, including the necessary criteria for disease–free status, should be established for that purpose. & Ba cheart nós imeachta comhchuibhithe soiléir, lena n-áirítear na critéir is gá do stádas saor ó ghalar, a bhunú chun na críche sin. \\ \hline
the mark is applied anew, as appropriate. & déanfar an mharcáil arís, mar is iomchuí.\\\hline
If the court decides that a review is justified on any of the grounds set out in paragraph 1, the judgment given in the European Small Claims Procedure shall be null and void. &  Má chinneann an chúirt go bhfuil bonn cirt le hathbhreithniú de bharr aon cheann de na forais a leagtar amach i mír 1, beidh an breithiúnas a tugadh sa Nós Imeachta Eorpach um Éilimh Bheaga ar neamhní go hiomlán. \\ \hline
households where pet animals are kept; & teaghlaigh ina gcoimeádtar peataí; \\ \hline

\end{tabular}
\caption{Samples of human reference translations}
\label{tab:translations-trans}
\end{table}

\begin{table}[h!]
\begin{tabular}{p{5cm}p{1.75cm}p{5cm}p{1.75cm}}\hline
\textbf{Transformer (16 kBPE)} & \textbf{BLEU} $\uparrow$ & \textbf{Google Translate} & \textbf{BLEU} $\uparrow$\\ \hline

Ba cheart nós imeachta soiléir comhchuibhithe, lena n-áirítear na critéir is gá maidir le stádas saor ó ghalair, a bhunú chun na críche sin.& 61.6 & Ba cheart nós imeachta comhchuibhithe soiléir, lena n-áirítear na critéir riachtanacha maidir le stádas saor ó ghalair, a bhunú chun na críche sin. & 70.2 \\ \hline 
go gcuirtear an marc i bhfeidhme, de réir mar is iomchuí. & 21.4 & cuirtear an marc i bhfeidhm as an nua, de réir mar is cuí. & 6.6 \\ \hline
Má chinneann an chúirt go bhfuil bonn cirt le hathbhreithniú ar aon cheann de na forais a leagtar amach i mír 1, beidh an breithiúnas a thugtar sa Nós Imeachta Eorpach um Éilimh Bheaga ar neamhní. & 77.3 & Má chinneann an chúirt go bhfuil údar le hathbhreithniú ar aon cheann de na forais atá leagtha amach i mír 1, beidh an breithiúnas a thugtar  sa Nós Imeachta Eorpach um Éilimh Bheaga ar neamhní & 59.1 \\ \hline
teaghlaigh ina gcoimeádtar peataí; & 100 & teaghlaigh ina gcoinnítear peataí; & 30.2 \\ \hline
\end{tabular}
\caption[Transformer model compared with Google Translate]{Transformer model compared with Google Translate using random samples from the DGT corpus. Full evaluation of Google Translate on the DGT test set, with 1.3k lines, generated a BLEU score of 46.3 and a TER score of 0.44. Comparative scores on the test set using our Transformer model, with 2 attention heads and 16k BPE submodel realised 60.5 for BLEU and 0.33 for TER.}
\label{tab:trans-google-trans}
\end{table}

Standard Transformer hyperparameters for batch size (2048) and the number of encoder and decoder layers (6) were both observed to perform well on the DGT and PA corpora. Reducing hidden neurons to 256 and increasing regularisation dropout to 0.3 improved translation performance and were chosen when building all Transformer models.

\section{Conclusion}

In our paper, we demonstrated that a random search approach to hyperparameter optimisation leads to the development of high-performing translation models. 

We have shown that choosing subword models, in our low-resource scenarios, is an important driver for the performance of MT engines. Moreover, the choice of vocabulary size leads to varying degrees of performance. Within the context of low-resource EN$\rightarrow$GA translation, we achieved optimal performance, on a 55k generic corpus and an 88k in-domain corpus, when a Transformer architecture with a 16k BPE submodel was used. 
The importance of selecting hyperparameters in training low-resource Transformer models was also demonstrated. By reducing the number of hidden layer neurons and increasing dropout, our models performed significantly better than baseline models and Google Translate.

Performance improvement of our optimised Transformer models, with subword segmentation, was observed across all key indicators namely a higher validation accuracy, a PPL achieved at a lower global minimum, a lower post-editing effort and a higher translation accuracy.  

\chapter{MT in the Covid domain: Shared Task for LoResMT2021}
\section{Context}

\begin{figure}[ht!]
\centering
  \includegraphics{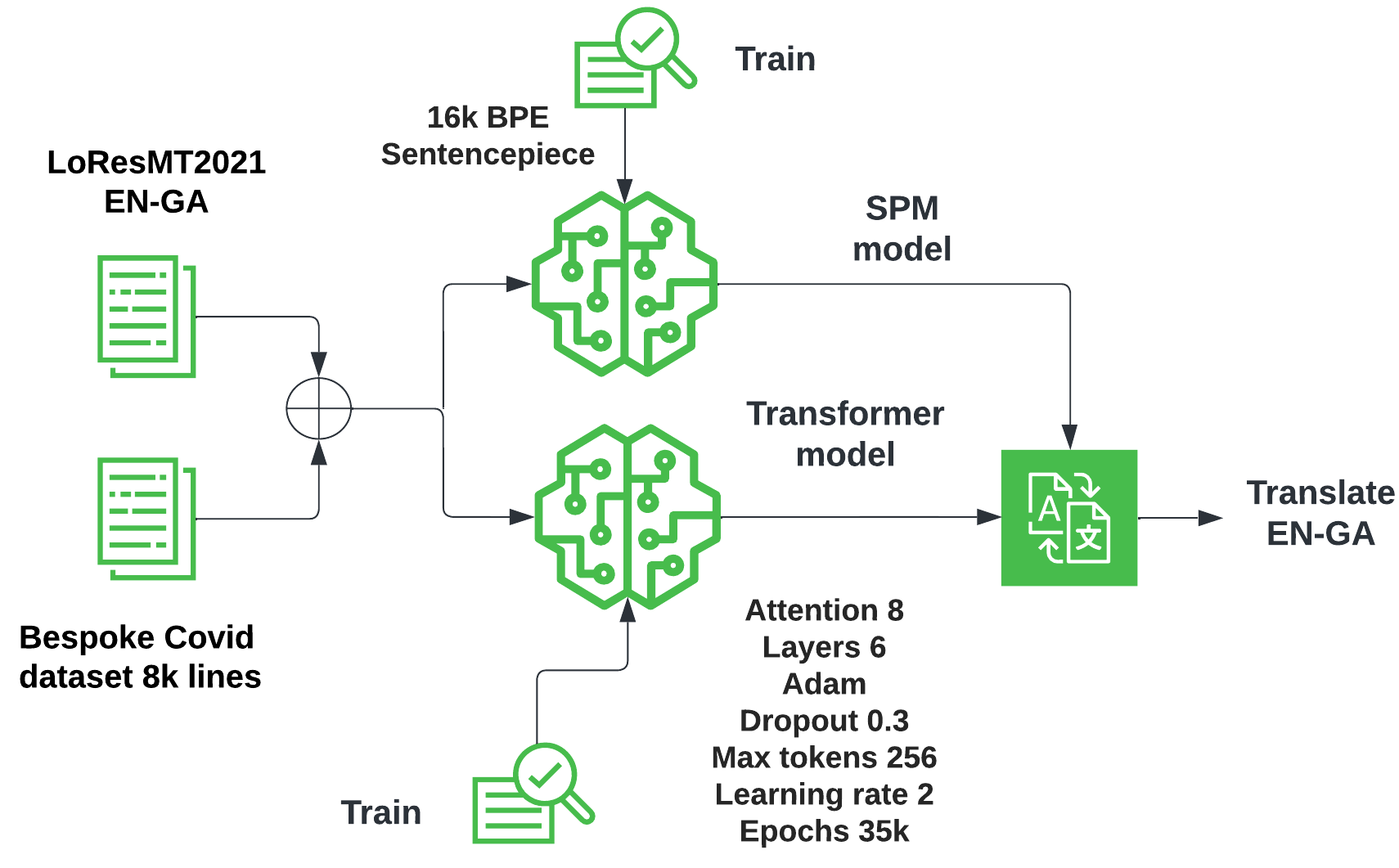}
  \caption{Proposed approach for LoResMT2021 Shared Task}
 \label{fig:adapnmt_approach}
\end{figure}

In refining the Transformer model with a custom dataset, we investigated the research objectives outlined in RQ1 and RQ2. A significant contribution of this study was showcasing that hyperparameters specified in \textit{``Transformers for Low-Resource Languages - is féidir linn!''} could be employed to excel in a shared task targeting low-resource languages. Therefore, the hyperparameters identified in the HPO experiments of Table \ref{tab:hpo-table-trans} were employed. Another key contribution was to demonstrate the notable improvement in translation quality when the shared task dataset was augmented by a compact bespoke in-domain dataset.  

The approach I adopted in developing the model for the LoResMT2021 Shared Task is illustrated in Figure \ref{fig:adapnmt_approach}. As the winning entry in the EN$\rightarrow$GA direction, the system achieved 36.0 BLEU, 0.531 TER and 0.6 ChrF. By combining the baseline LoResMT2021 5k line dataset with a bespoke 8k line Covid dataset, an amalgamated Covid-specific corpus of 13k lines was created. Using this dataset, a Transformer model was trained from scratch. Our system won by a wide margin with the second-placed team achieving a 25.8 BLEU.

\clearpage

   \begin{center}
       \vspace*{1cm}

       \textbf{Machine Translation in the Covid domain: an English-Irish case study - LoResMT2021}
            
       \vspace{1.5cm}

       \textbf{Séamus Lankford \\ Haithem Afli \\ Andy Way}

       \vfill

Proceedings of the 4th Workshop on Technologies for \\ 
MT of Low-Resource Languages (LoResMT2021)  \\ 
Virtual: Association for Machine Translation in the Americas \\
August 16 - 20, 2021 \\ 
Florida, USA \\
1st place in EN$\rightarrow$GA Shared Task
       \vspace{0.8cm}
                 
       ADAPT Centre\\
       Dublin City University\\
       Ireland\\

        \vspace{0.5cm}
    \url{https://aclanthology.org/2021.mtsummit-loresmt.15.pdf} \end{center}

\clearpage

\section{Abstract}
Translation models for the specific domain of translating Covid data from English to Irish were developed. Domain adaptation techniques, using a Covid-adapted generic 55k corpus from the Directorate General of Translation, were applied. Fine-tuning, mixed fine-tuning and combined dataset approaches were compared with models trained on an extended in-domain dataset.  As part of this study, an English-Irish dataset of Covid related data, from the Health and Education domains, was developed. The highest-performing model used a Transformer architecture trained with an extended in-domain Covid dataset. In the context of this study, we have demonstrated that extending an 8k in-domain baseline dataset by just 5k lines improved the BLEU score by 27 points. 

\section{Introduction}
Neural machine translation (NMT) has routinely outperformed statistical machine translation (SMT) when large parallel datasets are available~\parencite{crego2016systran, wu2016google}. Furthermore, Transformer-based approaches have demonstrated impressive results in moderate low-resource scenarios~\parencite{lankford2021transformer}. 
NMT involving Transformer model development will improve the performance in specific domains of low-resource languages~\parencite{araabi2020optimizing}. However, the benefits of NMT are less clear when using very low-resource machine translation (MT) on in-domain datasets of less than 10k lines. 

The Irish language is a primary example of a low-resource language that will benefit from such research. This paper reports the results of the MT system developed for the English–to-Irish (EN$\rightarrow$GA) Shared Task at LoResMT2021 \parencite{ojha2021findings}. Relevant work is presented in the background section followed by an overview of the proposed approach. The empirical findings are outlined in the results section. Finally, the key findings are presented and discussed. 

\section{Background}
   
\subsection{Transformer}

A novel architecture called Transformer was introduced in the paper ‘Attention Is All You Need’ \parencite{vaswani2017attention}. Transformer is an architecture for transforming one sequence into another with the help of an encoder and decoder without relying on recurrent neural networks (RNNs).

Transformer models use attention to focus on previously generated tokens. This approach allows models to develop a long memory which is particularly useful in the domain of language translation. 

\subsection{Domain Adaptation}

Domain adaptation is a proven approach to addressing the paucity of data in low-resource settings. Fine-tuning an out-of-domain model by further training with in-domain data is effective in improving the performance of translation models (Freitag and Al-Onaizan, 2016; Sennrich et al., 2016). With this approach, an NMT model is initially trained using a large out-of-domain corpus. Once fully converged, the out-of-domain model is further trained by fine-tuning its parameters with a low-resource in-domain corpus. 

A modification to this approach is known as mixed fine-tuning \parencite{chu2017empirical}. With this technique, an NMT model is trained on out-of-domain data until fully converged. This serves as a base model which is further trained using the combined in-domain and out-of-domain datasets. 

\section{Proposed Approach}

\begin{figure} [h]
    \centering
    \includegraphics[width=15.45cm]{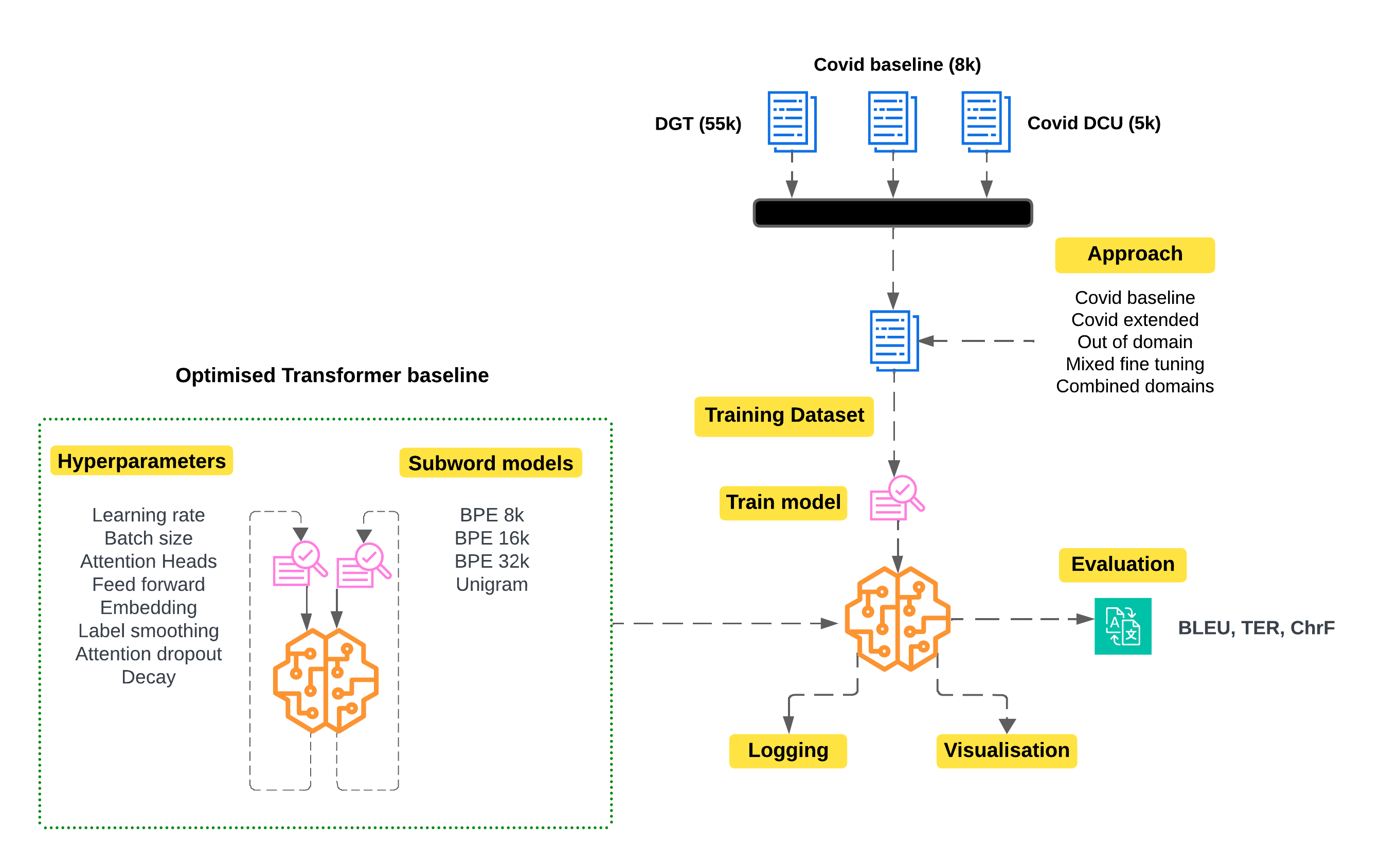}
    \caption[Proposed approach of MT in Covid domain]{Proposed approach of MT in the Covid domain. Optimal hyperparameters are applied to Transformer models which are trained using one of several possible approaches. The training dataset composition is determined by the chosen approach. Models are subsequently evaluated using a suite of metrics.}
    \label{fig:approach_covid}
\end{figure}

Hyperparameter optimisation of RNN models in low-resource settings has previously demonstrated considerable performance improvements \parencite{sennrich2019revisiting}. The extent to which such optimisation techniques may be applied to Transformer models in similar low-resource scenarios was evaluated in a previous study~\parencite{lankford2021transformer}. Evaluations included modifying the number of attention heads, the number of layers and experimenting with regularisation techniques such as dropout and label smoothing. Most importantly, the choice of subword model type and the vocabulary size were evaluated.

To test the effectiveness of our approaches, models were trained using three EN$\rightarrow$GA parallel datasets: a general corpus of 52k lines from the Directorate General for Translation (DGT) and two in-domain corpora of Covid data (8k and 5k lines). All experiments involved concatenating source and target corpora to create a shared vocabulary and a shared SentencePiece \parencite{kudo2018sentencepiece} subword model. The impact of using separate source and target subword models was not explored.

The approach adopted is illustrated in Figure \ref{fig:approach_covid} and the datasets used in evaluating this approach are outlined in Table \ref{tab:approach}. All models were developed using a Transformer architecture. 

\begin{table} [h]
\centering
\begin{tabular}{llllll}
\hline
\textbf{Approach} &
 \textbf{Source} &
 \textbf{Lines} & \\ \\ \hline
Covid baseline & Baseline & 8k &  \\
Covid extended & Baseline + Covid\_DCU & 13k & \\
Out-of-domain & DGT & 52k &  \\
Fine-tuned & Baseline + Covid\_DCU + DGT & 65k & \\
Mixed fine-tuned & Baseline + Covid\_DCU + DGT & 65k & \\
Combined domains & Baseline + Covid\_DCU + DGT & 65k & \\ \hline
\end{tabular}
\caption{Datasets used in proposed approach for MT in Covid domain}
\label{tab:approach}
\end{table}

\subsection{Architecture Tuning}

Long training times associated with NMT make it costly to tune systems using conventional grid search approaches. A previous study identified the hyperparameters required for optimal performance ~\parencite{lankford2021transformer}. Reducing the number of hidden layer neurons and increasing dropout led to significantly better performance. Furthermore, within the context of low-resource EN$\rightarrow$GA translation, using a 16k BPE submodel resulted in the highest-performing models. The Transformer hyperparameters, chosen in line with these findings, are outlined in Table \ref{tab:hpo-table-trans-covid}.
\begin{center}
\begin{table}
\center
\begin{tabular}{ll}
\hline
\textbf{Hyperparameter} & \textbf{Values}                \\ \hline
Learning rate            & 0.1, 0.01, 0.001, \textbf{2}            \\ \hline
Batch size               & 1024, \textbf{2048},  4096, 8192       \\ \hline
Attention heads          & \textbf{2}, 4, \textbf{8}                     \\ \hline
Number of layers         & 5, \textbf{6}                           \\ \hline
Feed-forward dimension   & \textbf{2048}                           \\ \hline
Embedding dimension      & 128, \textbf{256}, 512                  \\ \hline
Label smoothing          & \textbf{0.1}, 0.3                       \\ \hline
Dropout                  & 0.1, \textbf{0.3}                       \\ \hline
Attention dropout        & \textbf{0.1}                            \\ \hline
Average Decay            & 0, \textbf{0.0001}                      \\ \hline
\end{tabular}
\caption[Hyperparameter optimisation for Transformer models]{Hyperparameter optimisation for Transformer models. Optimal parameters are highlighted in bold. The highest-performing model trained on the 55k DGT corpus uses 2 attention heads ~\parencite{lankford2021transformer}.}
\label{tab:hpo-table-trans-covid}
\end{table}
\end{center}

\section{Empirical Evaluation}

\subsection{Experimental Setup}
\subsubsection{Datasets}
The performance of the Transformer approach is evaluated on EN$\rightarrow$GA parallel datasets in the Covid domain. Three datasets were used in the evaluation of our models. These consisted of a baseline Covid dataset (8k) provided by MT Summit 2021~\parencite{ojha2021findings}, an in-domain Covid dataset (5k) developed at DCU and a publicly available out-of-domain dataset (52k) provided by DGT~\parencite{steinberger2013dgt}. 

\subsubsection{Infrastructure}
Models were developed using a lab of machines each of which has an AMD Ryzen 7 2700X processor, 16GB memory, a 256GB SSD and an NVIDIA GeForce GTX 1080 Ti. Rapid prototype development was enabled through a Google Colab Pro subscription using NVIDIA Tesla P100 PCIe 16GB graphic cards and up to 27GB of memory when available~\parencite{Bisong2019}.

Our MT models were trained using the Pytorch implementation of OpenNMT 2.0, an open-source toolkit for NMT~\parencite{klein2017opennmt}. 

\subsubsection{Metrics}

Automated metrics were used to determine the translation quality. All models were trained and evaluated using the BLEU~\parencite{papineni-etal-2002-bleu}, TER \parencite{snover2006study} and ChrF~\parencite{popovic2015chrf} evaluation metrics. Case-insensitive BLEU scores, at the corpus level, are reported. Model training was stopped once an early stopping criteria of no improvement in validation accuracy for 4 consecutive iterations was recorded.

\subsection{Results}

Experimental results achieved using a Transformer architecture, with either 2 or 8 attention heads, are summarised in Table \ref{tab:trans-2heads} and in Table \ref{tab:trans-8heads}. Clearly, in the context of our low-resource experiments, it can be seen there is little performance difference using Transformer architectures with a differing number of attention heads. The largest difference occurs when using a fine-tuned approach (2.1 BLEU points). However, the difference between a 2-head and an 8-head approach is less than 1 BLEU point for all other models. The highest-performing approach uses the extended Covid dataset (13k) which is a combination of the MT summit Covid baseline and a custom DCU Covid dataset. This Transformer model, with 2 heads, performs well across all key translation metrics (BLEU: 36.0, TER: 0.63 and ChrF3: 0.32).

\begin{figure}
     \begin{subfigure}[b]{0.5\textwidth}
         \centering
         \includegraphics[width=\textwidth]{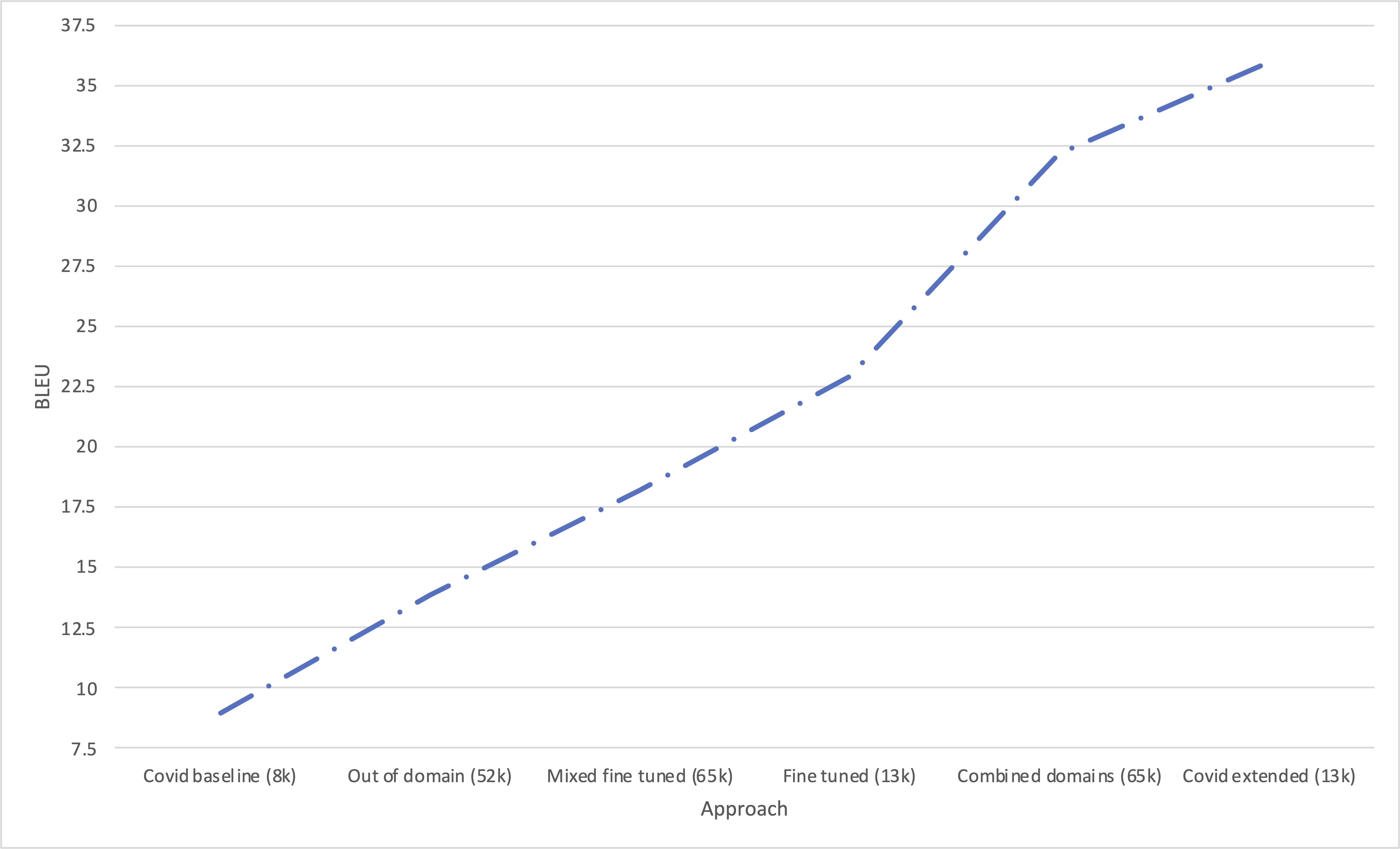}
         \caption{BLEU}
         \label{fig:bleu}
     \end{subfigure}
     \begin{subfigure}[b]{0.5\textwidth}
         \centering
         \includegraphics[width=\textwidth]{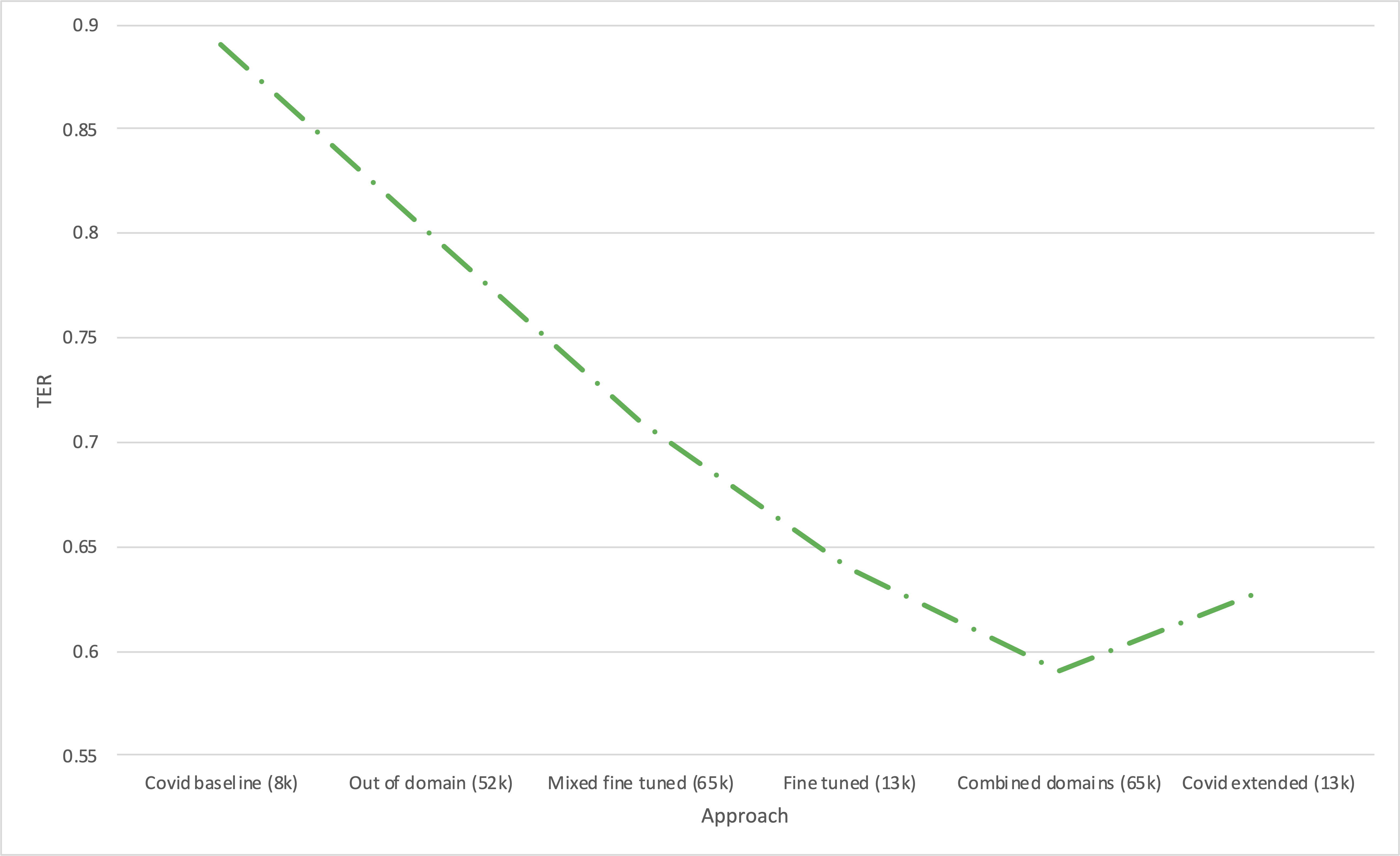}
         \caption{TER}
         \label{fig:ter}
     \end{subfigure}
        \caption{Translation performance using Transformers with 2 heads}
        \label{fig:transper}
\end{figure}

\begin{table}[]
\begin{tabular}{lcccccc}
\hline
\textbf{System} & \multicolumn{1}{l}{\textbf{Heads}} & \multicolumn{1}{l}{\textbf{Lines}} & \multicolumn{1}{l}{\textbf{Steps}} & \multicolumn{1}{l}{\textbf{BLEU} $\uparrow$} & \multicolumn{1}{l}{\textbf{TER} $\downarrow$ } & \multicolumn{1}{l}{\textbf{ChrF3} $\uparrow$} \\ \hline
Covid baseline & 2 & 8k & 35k & 9.0 & 0.89 & 0.32 \\ 
Covid extended & 2 & 13k & 35k & 36.0 & 0.63 & 0.54 \\ 
Out-of-domain & 2 & 52k & 200k & 13.9 & 0.80 & 0.41 \\ 
Fine-tuned & 2 & 65k & 35k & 22.9 & 0.64 & 0.42 \\ 
Mixed fine-tuned & 2 & 65k & 35k & 18.2 & 0.71 & 0.42 \\ 
Combined domains & 2 & 65k & 35k & 32.2 & 0.59 & 0.55 \\ \hline
\end{tabular}
\caption{Comparison of Transformer performance with 2 attention heads}
\label{tab:trans-2heads}
\end{table}

\begin{table}[]
\begin{tabular}{lcccccc}
\hline
\textbf{System} & \multicolumn{1}{l}{\textbf{Heads}} & \multicolumn{1}{l}{\textbf{Lines}} & \multicolumn{1}{l}{\textbf{Steps}} & \multicolumn{1}{l}{\textbf{BLEU} $\uparrow$ } & \multicolumn{1}{l}{\textbf{TER} $\downarrow$ } & \multicolumn{1}{l}{\textbf{ChrF3} $\uparrow$ } \\ \hline
Covid baseline & 8 & 8k & 35k & 9.6 & 0.91 & 0.33 \\ 
Covid extended & 8 & 13k & 35k & 35.7 & 0.61 & 0.55 \\ 
Out-of-domain & 8 & 52k & 200k & 13.0 & 0.80 & 0.40 \\ 
Fine-tuned & 8 & 65k & 35k & 25.0 & 0.63 & 0.43 \\ 
Mixed fine-tuned & 8 & 65k & 35k & 18.0 & 0.71 & 0.42 \\ 
Combined domains & 8 & 65k & 35k & 32.8 & 0.59 & 0.57 \\ \hline
\end{tabular}
\caption{Comparison of Transformer performance with 8 attention heads}
\label{tab:trans-8heads}
\end{table}

The worst-performing model uses the Covid baseline which is not surprising given that only 8k lines are available. The performance of the higher resourced models (out-of-domain, fine-tuned, mixed fine-tuned and combined domains) all lag that of the Covid extended model. In particular, the out-of-domain model, using the DGT dataset, performs very poorly with a BLEU score of just 13.9 on a Transformer model with 2 heads.

The BLEU and TER scores for all approaches are illustrated in Figure \ref{fig:bleu} and Figure \ref{fig:ter}. As expected, there is a high level of inverse correlation between BLEU and TER. Well-performing models, with high BLEU scores, also required little post-editing effort as indicated by their lower TER scores. 

\section{Discussion}

Standard Transformer parameters identified in a previous study were observed to perform well~\parencite{lankford2021transformer}. Reducing hidden neurons to 256 and increasing regularisation dropout to 0.3 improved translation performance and these hyperparameters were chosen when building all Transformer models. Furthermore, a batch size of 2048 and using 6 layers for the encoder and decoder were chosen throughout. 

The results demonstrate that translation performance for specific domains is driven by the amount of data which is available for that specific domain. It is noteworthy that an in-domain dataset of 13k lines (Covid extended), trained for just 35k steps outperformed by 22.1 BLEU points the corresponding out-of-domain 52k dataset (DGT) which was trained for 200k steps. 

\section{Conclusion and Future Work}

In our paper, we demonstrated that a high-performing in-domain translation model can be built with a dataset of 13k lines. Developing a small in-domain dataset, of just 5k lines, improved the BLEU score by 27 points when models were trained with the combined Covid baseline and custom Covid dataset.     

Following on from our previous work, careful selection of Transformer hyperparameters, and using a 16k BPE SentencePiece submodel, enabled rapid development of high-performing translation models in a low-resource setting. 

Within the context of our research in low-resource EN$\rightarrow$GA translation, we have shown that augmenting in-domain data, by a small amount, performed better than approaches which incorporate fine-tuning, mixed fine-tuning or the combination of domains.    

As part of our future work, we plan to develop EN$\leftrightarrow$GA MT models trained on a dataset derived from the health domain (cf. Chapter 4). Domain adaptation, through fine-tuning such models with the Covid extended dataset may further improve Covid MT performance.

\chapter{\textbf{{\em gaHealth}: EN$\leftrightarrow$GA Bilingual Corpus of Health Data}}
\section{Context}

While MT has made significant strides in high-resource language pairs, there is a noticeable scarcity of parallel data for low-resource languages, particularly in specialised domains like health. An example of a high-resource language pair that has near-perfect MT in a specific domain can be seen in the English$\leftrightarrow$French pair. Weather forecasting in Canada, which is provided both in English and French, has led to one of the best MT systems ever produced, namely the METEO system \parencite{thouin-1981-meteo}. Having been perfected over decades, the only errors made by the system are due to erroneous human input.

With regard to low-resource languages, the existing efforts often prioritise quantity over domain specificity. The focus of this research is to address, in part, this imbalance and the key motivations for this research are driven by RQ2. A separate contribution from the research was the development of a set of guidelines for the pre-processing, alignment and validation of a corpus for a low-resource language pair. An important motivating factor for answering this research question is that key information first comes out in ``major'' languages before it does in less well-resourced languages. Consequently, people who would prefer to access information in their preferred language are forced to operate in a language they otherwise would not choose. In particular, this paper deals with several aspects of the research question which are highlighted below:

\begin{itemize}

\item \textbf{Addressing Low-Resource Language Challenges}: The scarcity of parallel data for low-resource languages hinders the development of effective MT models. This scarcity is particularly acute in specialised domains like health. The study aims to alleviate this issue by creating a focused, in-domain dataset.

\item \textbf{Importance of In-Domain Data}: The benefits of using in-domain data for training MT models are highlighted. In contrast to generic datasets, in-domain datasets are tailored to a specific domain (in this case, health), which can lead to more accurate and contextually relevant translations for that domain.

\item \textbf{Development of gaHealth Corpus}: The research introduces the gaHealth corpus, which is the first bilingual corpus of health data for the Irish language. This corpus is designed to be a valuable resource for the NLP community, especially those working within the Irish language domain. The importance of open access is recognised and the gaHealth corpus has been shared online, inviting further research and contributions from the wider community.

\item \textbf{Empirical Demonstration of Benefits}: The study provides empirical evidence to support the use of in-domain data, showing significant improvements in translation performance. The comparison with top-performing models from a shared task (LoResMT2021) highlights the substantial gains achieved through the use of the gaHealth corpus.

\item \textbf{Guidelines for Corpus Development}: Recognising the challenges encountered during the conversion of PDF documents, guidelines were established to facilitate the conversion process. This contribution not only aids in the development of the gaHealth corpus but also provides valuable insights for similar projects.

\item \textbf{Future Directions}: The research outlines a clear roadmap for future work. This includes expanding the gaHealth corpus as more Irish language documents become available, refining the models, and conducting a deep linguistic investigation to understand the nuances of model performance. Additionally, it is planned to extend the effort to other key domains such as Education and Finance.

\end{itemize}

In summary, the research addresses a critical gap in the availability of specialised, in-domain data for low-resource languages, particularly in the context of health-related content for the Irish language. By creating the gaHealth corpus and demonstrating its effectiveness in improving translation models, the study provides a valuable resource for the NLP community and lays the foundation for future advancements in this domain. More importantly, it has contributed to high-quality MT systems which can assist Irish-language speakers in accessing health information in their preferred language.  

\clearpage

   \begin{center}
       \vspace*{1cm}

       \textbf{\textbf{{\em gaHealth}: An EN$\leftrightarrow$GA Bilingual Corpus of Health Data}}
            
       \vspace{1.5cm}

       \textbf{Séamus Lankford \\ Haithem Afli \\ Órla Ní Loinsigh \\ Andy Way}

       \vfill

Proceedings of the Thirteenth Language Resources and Evaluation Conference\\ 
European Language Resources Association \\
20-25 June 2022 \\
Marseille, France \\

       \vspace{0.8cm}
                 
       ADAPT Centre\\
       Dublin City University\\
       Ireland\\

        \vspace{0.5cm}

    \url{https://aclanthology.org/2022.lrec-1.727.pdf}
    
\end{center}

\clearpage

\section{Abstract}

Machine translation is a mature technology for many high-resource language pairs. However, in the context of low-resource languages, there is a paucity of parallel data datasets available for developing translation models. Furthermore, the development of datasets for low-resource languages often focuses on simply creating the largest possible dataset for generic translation. The benefits and development of smaller in-domain datasets can easily be overlooked. To assess the merits of using in-domain data, a dataset for the specific domain of health was developed for the low-resource English-to-Irish language pair. Our study outlines the process used in developing the corpus and empirically demonstrates the benefits of using an in-domain dataset for the health domain. In the context of translating health-related data, models developed using the {\em gaHealth} corpus demonstrated a maximum BLEU score improvement of 22.2 points (40\%) when compared with top-performing models from the LoResMT2021 Shared Task. Furthermore, we define linguistic guidelines for developing {\em gaHealth}, the first bilingual corpus of health data for the Irish language, which we hope will be of use to other creators of low-resource data sets. {\em gaHealth} is now freely available online and is ready to be explored for further research.

\section{Introduction}
\label{sec:intro}

Improvements in performance in natural language processing (NLP) tasks are typically to be seen when deep learning models are used. However, deep learning requires large amounts of data for model training. Consequently, the availability of large amounts of textual data has become fundamental to the success of NLP applications, such as language modelling \parencite{buck2014n} and neural machine translation (NMT) \parencite{bahdanau2014neural,sennrich2016edinburgh}.

A popular method of developing such corpora for machine translation (MT) tasks is to crawl and parse bilingual web pages to general parallel corpora. However, given the nature of low-resource languages, there are often insufficient websites available in both the languages of study. Accordingly, a lack of web content typically hinders the development of NLP applications for low-resource languages.
 
The motivation underpinning our present work comes from the challenges we faced in developing high-performing MT models in low-resource settings \parencite{afli2017sentiment}. In this work, we developed the first bilingual corpus of health data for the English$\leftrightarrow$Irish (EN$\leftrightarrow$GA) language pair. A procedure was created to extract, clean and select appropriate sentences to build a bilingual corpus. In addition, we built a high-performing MT model for translating in-domain health data.

\section{Related work}
\label{sec:relWork}

\subsection{Transformer}

Transformer \parencite{vaswani2017attention} is an architecture for transforming an input sequence into an output sequence via an encoder and decoder without relying on recurrent neural networks. Transformer models use attention to focus on previously generated tokens. This approach allows models to develop a long memory which is particularly useful in the domain of language translation. 

\subsection{Transformer Hyperparameter Optimisation}

Hyperparameter optimisation of Transformer models in translating the low-resource English-Irish (EN$\rightarrow$GA) language pair has been evaluated in previous studies \parencite{lankford2021transformer}. 
Carefully selecting the appropriate subword model has been shown to be an important driver of translation performance. A Transformer architecture, using a 16k BPE SentencePiece subword model, demonstrated optimal performance. 

\subsection{Neural MT}
Using large bilingual corpora, NMT approaches require the training of neural networks to learn a statistical model for MT. The technique has demonstrated state-of-the-art translation performance on many benchmarks. However, one of the key factors in enabling the development of high-performing NMT models is the availability of large amounts of parallel data \parencite{koehn2017six,sennrich2019revisiting}.

\section{Proposed Approach}
\label{sec:meth}
\subsection{Sources for {\em gaHealth} Development}
\label{subsec:signCorp}
To build a bilingual corpus of health data, we selected multiple sources of professionally translated documents from within the Irish government, all of which are publicly available. In particular, the bilingual strategy statements and annual reports of the Irish Department of Health since 2010 were chosen. 

Furthermore, a dataset of Covid-related data, developed for a previous study~\parencite{lankford2021machine}, was incorporated into a larger health dataset. Given the pace at which Covid-19 data was being published, translated Irish website content often lagged behind the English-language counterpart. Website snapshots taken by the WayBack Machine\footnote{\url{https://archive.org/web/}} \parencite{arora2016using} proved particularly useful in creating good parallel data from unaligned parallel websites. Extracts from the corpus are illustrated in Table \ref{tab:corpus-samples}.

\begin{table}[]
\begin{tabular}{p{1.5cm}p{12cm}}
\hline
\textbf{Type} &
  \textbf{Sentence} \\ \hline \\
EN-1 &
  The Programme for Government makes strong commitments to strengthen community-based care, including primary care and social care, and sees this as fundamental to advancing Sláintecare reforms. \\ 
GA-1 &
Tá gealltanais láidre sa Chlár Rialtais maidir le cúram pobalbhunaithe a neartú, cúram príomhúil agus cúram sóisialta san áireamh, agus breathnaítear air sin mar chuid bhunriachtanach d’athchóirithe Sláintecare a chur ar aghaidh. \\ 
EN-2 &
The Hepatitis C Compensation Tribunal (Amendment) Act 2006 established a statutory scheme to address insurance difficulties experienced by persons infected with Hepatitis C and HIV through the administration within the State of blood and blood products. \\
GA-2 &
Faoin Acht um Binse Cúitimh i ndáil le Heipitíteas C (Leasú) 2006, bunaíodh scéim reachtúil chun dul i ngleic le deacrachtaí árachais a bhí ag daoine a bhí ionfhabhtaithe le Heipitíteas C agus VEID trí fhuil agus táirgí fola a tugadh dóibh laistigh den Stát. \\ 
EN-3 &
To provide virological evidence on the presence and extent of undetected community transmission of covid-19 and monitor positivity rates among individuals presenting ill or acute respiratory tract infections to primary care. \\
GA-3 &
Fianaise víreolaíoch a sholáthar maidir le láithreacht agus méid an tarchuir pobail anaithnid de covid-19 agus monatóireacht a dhéanamh ar rátaí dearfacha i measc daoine aonair a bhfuil ionfhabhtuithe ili nó conaire riospráide géarmhíochaine orthu chuig cúram príomhúil. \\ \\ \hline 
\end{tabular}
\caption[Extracts from the {\em gaHealth} corpus]{Extracts from the {\em gaHealth} corpus are illustrated in this table. The sentences, EN-1 / GA-1, are drawn from strategy statements, EN-2 / GA-2 are taken from annual reports whereas EN-3 / GA-3 are from Covid sources.}
\label{tab:corpus-samples}
\end{table}

This amalgamated corpus, {\em gaHealth}, consists of 16,201 lines of parallel text files. The combined English and Irish vocabulary size is 19,269 unique words. The constituent elements of the dataset, prior to applying the toolchain, are outlined in Table \ref{tab:elements}.

\begin{table}
\centering
\begin{tabular}{llllll}
\hline
\textbf{Documents} &
 \textbf{Source} &
 \textbf{Lines} & \\ \hline
Strategy Statement 2020 & HSE & 3k &  \\
Strategy Statement 2017 & HSE & 2.5k & \\
Strategy Statement 2015 & HSE & 3k &  \\
Annual Report 2020 & HSE & 2k &  \\
Annual Report 2019 & HSE & 2k &  \\
Annual Report 2017 & HSE & 2k &  \\
Website (Covid) & Citizen's Advice & 4k & \\
Publications (Covid) & HSE & 4k \\ \hline
\end{tabular}

\caption{Sources used in corpus development}
\label{tab:elements}
\end{table}

\subsection{Toolchain used for {\em gaHealth} Development}
\par

\label{subsec:creatCorp}

The HSE PDF and Word documents were pre-processed using a toolchain currently under development as part of the Irish Language Resource Infrastructure project (ILRI), funded by the Department of the Gaeltacht. This toolchain has been written to accept primarily data that originates in public administration organisations, i.e. relatively formal text for which the translation quality is assumed to be high, the structure and formatting to be reasonably consistent, and the potential for noise to be low.

The source material consisted of a combination of twelve input files: six in English (all PDF) and six in Irish (five PDF, one Word); all PDFs had a text layer. They ranged from relatively short (30 to 40 A4 pages) to more substantial (ca. 200 pages). PDFs in particular can be problematic for creating high-quality corpora for a variety of reasons. In other words, while the quality of the input content in this case can be said to be high, the quality of the input medium is low.

\begin{figure*}
\includegraphics [width=15cm]{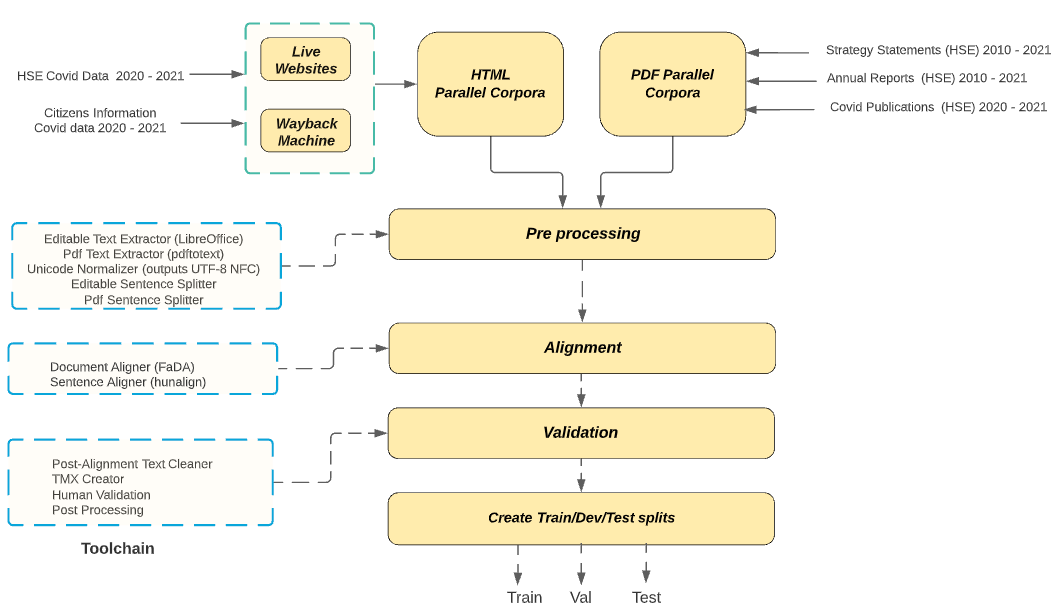}
\caption[Corpus development process]{Corpus development process.  In developing the corpus, the key steps of data collection, pre-processing, alignment and validation were followed. The role of the toolchain at various stages is highlighted.}
\label{fig:corpus_dev}
\end{figure*}

The process used for developing the corpus is illustrated in Figure \ref{fig:corpus_dev}. The toolchain consists of a set of components run in sequence over a set of input documents, to convert it from raw content to a sentence-aligned corpus. Several of the components listed below have different implementations depending on the source type and intended output.

\paragraph{Text extractors}These are wrappers around external components that extract text based on input type; for the {\em gaHealth} file types, wrappers were written for LibreOffice and pdftotext.

\paragraph{Unicode normalizer}This is used to achieve Unicode equivalence, and optionally to substitute certain (e.g. corrupted) characters.

\paragraph{Language detector}In this toolchain, the language detector is a wrapper around langdetect, itself a port from language-detection by Nakatani Shuyo~\parencite{nakatani2010langdetect}. The wrapper was written to allow this to be run conveniently either on a string or on an entire file.

\paragraph{Sentence splitters}These are custom-written components to reconstruct sentence boundary information. For editable file types like plain text and Word, this process is relatively straightforward. However, PDFs present particular challenges in this regard. Along with ordering issues, the absence of sentence boundary information is one of the biggest reasons why it is so difficult to construct a high-quality corpus from PDFs. A custom sentence splitter was used to determine sentence boundaries from text extracted from PDFs specifically, primarily using capitalisation and language-specific lists of abbreviations to determine where sentences should be broken.

\paragraph{Document aligner}This aligns sets of files whose languages have been identified. A wrapper was written around an external component called FaDA~\parencite{lohar2016fada} to adapt it to the toolchain. As FaDA always names an alignment for each input file, sometimes even mapping two different files to the same one, it was necessary to put some selection logic in here, as well as a mechanism for determining when there is no appropriate mapping. Constraints may be put on the relative size of the files to accept an alignment, and the process may be re-run multiple times, with previously rejected files being run again.

\paragraph{Sentence aligner}This aligns pairs of files at the sentence level. This was a wrapper around the external component hunalign~\parencite{varga2005hunalign}. No special parameters were used, as default settings produced results of high quality.

\paragraph{Text cleaners}These remove sentence pairs that are believed to be incorrect alignments, such as empty segments and those with obviously mismatched content.

\subsection{Guidelines}
\label{subsubsec:guide}

With the above considerations in mind, the following set of rules was decided upon when processing the {\em gaHealth} dataset. Many of these could be specified as parameters to the toolchain, while others were hard-coded into the system.

\begin{enumerate}

\item Unicode standard: normalise all characters to Unicode UTF-8 NFC. Remove any byte order marks.

\item Whitespacing and capitalisation: merge sequences of whitespace characters into a single space. Do not perform tokenization or truecasing.

\item File language detection: scan the first 50 lines, and then every 100th line.

\item Document alignment: assume that specific patterns like a line beginning with a single letter in parentheses or a number followed by a full stop indicate a sentence break from the previous line. Ensure each document is 0.75-1.33 times the size of the document it is being aligned with. Run for a maximum of three iterations.

\item Sentence alignment: allow one-to-many alignments.

\item Cleaning: remove any pairs where source or target:
\begin{itemize}
\item is empty
\item contains no non-alphabetical characters
\item is of an incorrect language. This will remove most untranslated segments. The language is only to be detected for segments that have at least 40 characters
\end{itemize}

\end{enumerate}

\subsection{Transformer Architecture}

All  EN$\leftrightarrow$GA  models, trained with the {\em gaHealth} corpus, were developed using a Transformer architecture. A reduction in the number of hidden layer neurons and increasing dropout significantly improved performance. Furthermore, within the context of low-resource EN$\rightarrow$GA translation, using a 16k BPE submodel resulted in the highest-performing models. Optimal hyperparameters were selected in line with these findings from our previous and are outlined in Table \ref{tab:hpo-table-gaHealth}.
\begin{center}
\begin{table}
\center
\begin{tabular}{ll}
\hline
\textbf{Hyperparameter} & \textbf{Values}                \\ \hline
Learning rate            & 0.1, 0.01, 0.001, \textbf{2}            \\ \hline
Batch size               & 1024, \textbf{2048},  4096, 8192       \\ \hline
Attention heads          & \textbf{2}, 4, \textbf{8}                     \\ \hline
Number of layers         & 5, \textbf{6}                           \\ \hline
Feed-forward dimension   & \textbf{2048}                           \\ \hline
Embedding dimension      & 128, \textbf{256}, 512                  \\ \hline
Label smoothing          & \textbf{0.1}, 0.3                       \\ \hline
Dropout                  & 0.1, \textbf{0.3}                       \\ \hline
Attention dropout        & \textbf{0.1}                            \\ \hline
Average Decay            & 0, \textbf{0.0001}                      \\ \hline
\end{tabular}
\caption[Hyperparameter optimisation for Transformer models]{Hyperparameter optimisation for Transformer models. Optimal parameters are highlighted in bold \parencite{lankford2021transformer}.}
\label{tab:hpo-table-gaHealth}
\end{table}
\end{center}

\section{Empirical Evaluation}
\label{sec:exp_gaHealth}

In addition to developing the {\em gaHealth} corpus, the effectiveness of the dataset was evaluated by training models for EN$\leftrightarrow$GA translation in the Health domain. All experiments involved concatenating source and target corpora to create a shared vocabulary and a shared SentencePiece~\parencite{kudo2018sentencepiece} subword model.  The impact of using separate source and target subword models was not explored.

To benchmark the performance of our models, the EN$\rightarrow$GA and GA$\rightarrow$EN test datasets from the LoResMT2021 Shared Task ~\parencite{ojha2021findings} were used. These test datasets enabled the evaluation of the {\em gaHealth} corpus, and associated models since this shared task focused on an application of the health domain i.e. the translation of Covid-related data. Furthermore, using a shared task test dataset enables the comparison of gaHealth models' performance with models entered by other teams. 

The results from the IIITT~\parencite{puranik2021attentive} and UCF~\parencite{chen2021ucf} teams are included in Tables \ref{tab:en2gahealth} and \ref{tab:ga2en-gahealth} so the performance of the {\em gaHealth} models can be easily compared with the findings of LoResMT2021. IIITT fine-tuned an Opus MT model from Helsinki NLP on the training dataset. UCF \parencite{chen2021ucf} used transfer learning, unigram and subword segmentation methods for EN$\leftrightarrow$GA translation.

\subsection{Infrastructure}
Rapid prototype development was enabled through a Google Colab Pro subscription using NVIDIA Tesla P100 PCIe 16GB graphic cards and up to 27GB of memory when available \parencite{Bisong2019}. Our MT models were trained using the Pytorch implementation of OpenNMT 2.0, an open-source toolkit for NMT~\parencite{klein2017opennmt}. 

\subsection{Metrics}

Automated metrics were used to determine the translation quality. All models were trained and evaluated using the BLEU \parencite{papineni-etal-2002-bleu}, TER \parencite{snover2006study} and ChrF \parencite{popovic2015chrf} evaluation metrics. Case-insensitive BLEU scores, at the corpus level, are reported. Model training was stopped after 40k training steps or once an early stopping criteria of no improvement in validation accuracy for four consecutive iterations was recorded.

\subsection{Results: Automatic Evaluation}

The hyperparameters used for developing the models are outlined in Table \ref{tab:hpo-table-gaHealth}. The details of the training, validation and test sets used by our NMT models are outlined in Table \ref{tab:en2ga-stats} and Table \ref{tab:ga2en-stats}. In all cases, 502 lines were used from the LoResMT2021 validation dataset whereas the test dataset used 502 lines for EN$\rightarrow$GA translation and 250 lines for GA$\rightarrow$EN translation. Both were independent health-specific Covid test sets which were provided by LoResMT2021. There was one exception, due to a data overlap between test and training data, a reduced test set was used when testing the {\em gaHealth} en2ga* system.  

\begin{table} [ht]
\centering
\begin{tabular}{lcccccc}
\hline
\textbf{Team} &
 \textbf{System} &
  \textbf{Train}  &
  \textbf{Dev}  &
  \textbf{Test}  \\ \hline
adapt & covid\_extended & 13k & 502 & 500 \\
adapt & combined\_domains & 65k & 502 & 500 \\
IIITT  & en2ga-b & 8k & 502 & 500 \\
UCF     & en2ga-a & 8k & 502 & 500   \\ 
{\em gaHealth} & en2ga & 24k & 502 & 500 \\
{\em gaHealth} & en2ga* & 24k & 502 & 338 \\
\hline
\end{tabular}
\caption[EN$\rightarrow$GA training, validation and test datasets]{EN$\rightarrow$GA training, validation and test dataset distributions. The baseline {\em gaHealth} system was augmented with an 8k Covid dataset provided by LoResMT2021. A smaller test set was used when evaluating {\em gaHealth} en2ga* due to an overlap with the training data. An alternative approach of removing the overlap from the {\em gaHealth} corpus, before training, was also carried out to produce the {\em gaHealth} en2ga system.} 
\label{tab:en2ga-stats}
\end{table}

\begin{table} [ht]
\centering
\begin{tabular}{lcccccc}
\hline
\textbf{Team} &
 \textbf{System} &
  \textbf{Train}  &
  \textbf{Dev}  &
  \textbf{Test}  \\ \hline
IIITT  & ga2en-b & 8k & 502 & 250 \\
UCF     & ga2en-b & 8k & 502 & 250   \\ 
{\em gaHealth} & ga2en & 24k & 502 & 250 \\
\hline
\end{tabular} 
\caption[GA$\rightarrow$EN training, validation and test datasets]{GA$\rightarrow$EN training, validation and test dataset distributions. The baseline {\em gaHealth} system was augmented with an 8k Covid dataset provided by LoResMT2021. All overlaps were removed from the {\em gaHealth} corpus prior to training the {\em gaHealth} ga2en model.}
\label{tab:ga2en-stats}
\end{table}

\begin{figure}[h]
    \centering
    {\includegraphics[width=7.25cm]{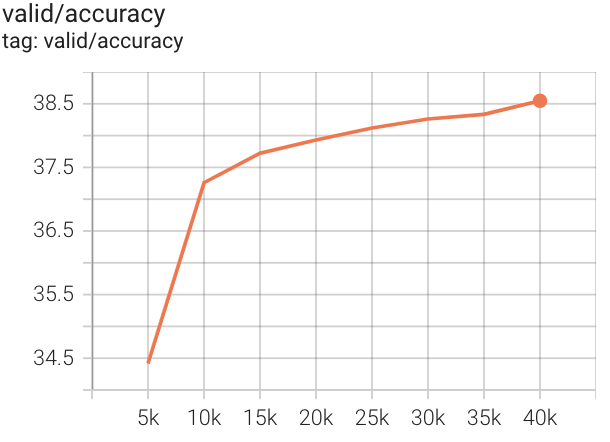}}
    {\includegraphics[width=7.25cm]{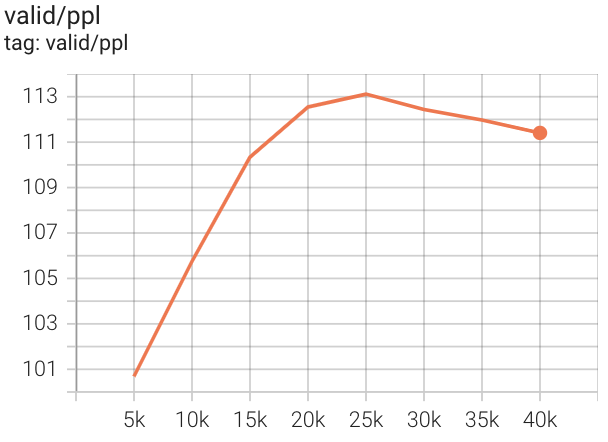}}
    \caption[{\em gaHealth} en2ga* system: training EN$\rightarrow$GA model]{{\em gaHealth} en2ga* system: training EN$\rightarrow$GA model with combined 16k gaHealth corpus and 8k LoResMT2021 covid corpus achieving a max validation accuracy of 38.5\% and perplexity of 111 after 40k steps. BLEU score: \textbf{37.6}. }
    \label{fig:en-ga-gaHealth-ch4}
\end{figure}

\begin{figure}[h]
    \centering
    {\includegraphics[width=7.25cm]{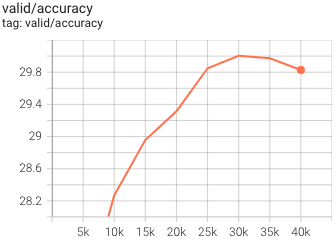}}
    {\includegraphics[width=7.25cm]{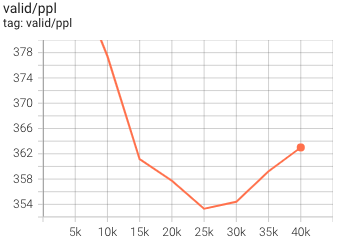}}
    \caption[adapt covid\_extended system:training EN$\rightarrow$GA model]{ adapt covid\_extended system: training EN$\rightarrow$GA model with 8k LoResMT2021 covid corpus achieving a max validation accuracy of 30.0\% and perplexity of 354 after 30k steps. BLEU score: \textbf{36.0}.}
    \label{fig:en-ga-covid-ch4}
\end{figure}

\begin{figure}[h]
    \centering
    {\includegraphics[width=7.25cm]{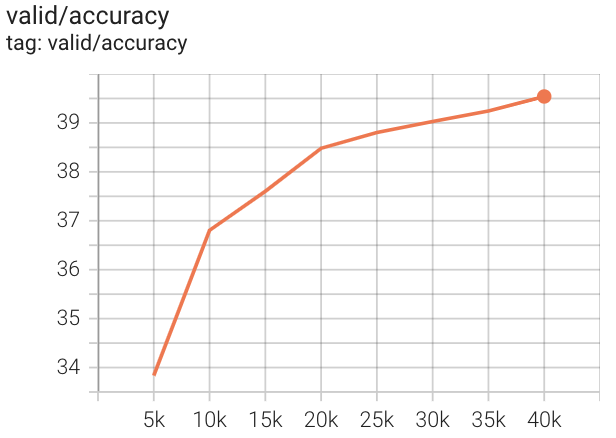}}
    {\includegraphics[width=7.25cm]{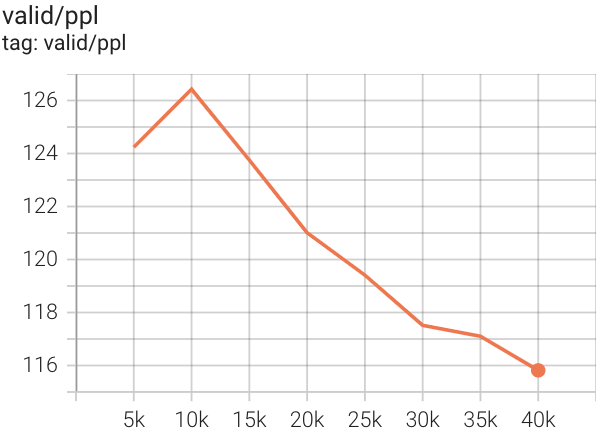}}
    \caption[{\em gaHealth} ga2en system: training GA$\rightarrow$EN model]{{\em gaHealth} ga2en system: training GA$\rightarrow$EN model with combined 16k gaHealth corpus and 8k LoResMT2021 Covid corpus achieving a max validation accuracy of 39.5\% and perplexity of 116 after 40k steps. BLEU score: \textbf{57.6.}}
    \label{fig:ga-en-gaHealth-ch4}
\end{figure}

\par 
Experimental results achieved using a Transformer architecture, are summarised in Table \ref{tab:en2gahealth} and Table \ref{tab:ga2en-gahealth}. In the LoResMT2021 Shared Task, the highest-performing EN$\rightarrow$GA system was submitted by the ADAPT team. The system uses an extended Covid dataset (13k, which is a combination of the MT summit Covid baseline and a custom DCU Covid dataset. This Transformer model, with 2 heads, performs well across all key translation metrics (BLEU: 36.0, TER: 0.531 and ChrF3: 0.6). 
\begin{table}[ht!]
\centering
\begin{tabular}{lcccccc}
\hline
\textbf{Team} &
 \textbf{System} &
  \textbf{BLEU} $\uparrow$ &
  \textbf{TER} $\downarrow$ &
  \textbf{ChrF3} $\uparrow$ \\ \hline
UCF     & en2ga-b & 13.5 & 0.756 & 0.37   \\
IIITT  & en2ga-b & 25.8 & 0.629 & 0.53 \\
adapt & combined & 32.8 & 0.590 & 0.57 \\
{\em gaHealth} & en2ga & 33.3 & 0.604 & 0.56 \\
adapt & covid\_extended & 36.0 & 0.531 & 0.60 \\
{\em gaHealth} & en2ga* & \textbf{37.6} & 0.577 & 0.57 \\  
\hline
\end{tabular}
\caption{EN$\rightarrow$GA {\em gaHealth} system compared with LoResMT2021 systems.}
\label{tab:en2gahealth}
\end{table}

\par
Validation accuracy, and model perplexity, in developing the {\em gaHealth} models are illustrated in Figure \ref{fig:en-ga-gaHealth-ch4} and Figure \ref{fig:ga-en-gaHealth-ch4} whereas Figure \ref{fig:en-ga-covid-ch4} illustrates model training on just the covid\_extended dataset. Rapid convergence was observed while training the {\em gaHealth} models such that little accuracy improvement occurs after 30k steps. Only marginal gains were achieved after this point and it declined in the case of the system trained using the covid\_extended dataset. 

Perplexity (PPL) shows how many different, equally probable words can be produced during translation. As a metric for translation performance, it is important to keep low scores so the number of alternative translations is reduced. 

\par
Of the models developed by the ADAPT team, the worst-performing model uses a larger 65k dataset. This is not surprising given the dataset is from a generic domain of which only 20\% is health-related. The performance of this higher-resourced 65k line model lags the augmented {\em gaHealth} model which was developed using just 24k lines. 

\begin{table}[ht!]
\centering
\begin{tabular}{lcccccc}
\hline
\textbf{Team} &
 \textbf{System} &
  \textbf{BLEU} $\uparrow$ &
  \textbf{TER} $\downarrow$ &
  \textbf{ChrF3} $\uparrow$ \\ \hline
UCF & ga2en-b & 21.3 & 0.711 & 0.45\\
IIITT  & ga2en-b & 34.6 & 0.586 & 0.61\\
{\em gaHealth} & ga2en & \textbf{57.6} & 0.385 & 0.71\\
\hline
\end{tabular} 
\caption{GA$\rightarrow$EN {\em gaHealth} system compared with LoResMT2021 systems.}
\label{tab:ga2en-gahealth}
\end{table}

\par
For GA$\rightarrow$EN translation, the best-performing model for the LoResMT2021 Shared Task was developed by IIITT with a BLEU of 34.6, a TER of 0.586 and ChrF3: 0.6. This effectively serves as the baseline by which our GA$\rightarrow$EN model, developed using the {\em gaHealth} corpus, can be benchmarked. The performance of the {\em gaHealth} model offers an improvement across all metrics with a BLEU score of 57.6, a TER of 0.385 and a CHrF3 result of 0.71. In particular, the 40\% improvement in BLEU score is very significant.

\section{Discussion}
\label{sec:res}
Although the main objective of this work is to develop the first bilingual corpus of EN$\leftrightarrow$GA data, we conduct initial experiments on the effectiveness of such datasets in training MT models. We have used our {\em gaHealth} dataset to train an MT model on test data from the LoResMT2021 Shared task to evaluate how the system performs translating EN$\leftrightarrow$GA health data. Our systems, developed using the {\em gaHealth} corpus achieved significantly higher scores. 

\section{Conclusion and Future Work}
\label{sec:concFut}
The main contribution of this work is to present an ongoing translation project that aims at building the first ever parallel corpus of health data for the Irish language -- {\em gaHealth} -- by fully utilising freely available parallel documents. 

Due to the issues encountered during the conversion of PDF documents, we developed guidelines to aid in the conversion process. In addition to developing the {\em gaHealth} corpus, we trained and evaluated translation models for in-domain health data.  

In our experiments, the models achieved a BLEU score of 37.6 (Table \ref{tab:en2gahealth}) for translating EN$\rightarrow$GA test data and 57.6 (Table \ref{tab:ga2en-gahealth}) for GA$\rightarrow$EN translation, which is encouraging performance given this is the beginning of our work on {\em gaHealth}. There is no such corpus available according to the best of our knowledge, so {\em gaHealth} will become a useful resource in the NLP community, especially for those working with the Irish language domain. 

For future work, we intend to extend the corpus, as more Irish language documents become available. Upon extension, we will refine our models. One important aspect which needs further investigation is to understand why the EN$\rightarrow$GA model ({\em gaHealth} en2ga), tested with the full test set, performed worse than the model ({\em gaHealth} en2ga*) which was tested with the reduced test set. A deep linguistic investigation involving a sentence-level BLEU analysis will be conducted as part of a future study. 

In addition, we aim to build in-domain datasets for other key domains such as Education and Finance. We will also apply deep learning techniques to further refine our in-domain models. We have released the {\em gaHealth} corpus online\footnote{\url{https://github.com/seamusl/gaHealth}} to facilitate further research on this data set.

\chapter{Human Evaluation of EN$\leftrightarrow$GA Transformer-Based NMT}
\section{Context}

In addressing RQ3, we employed a rigorous quantitative human evaluation of the translation outputs of RNN and Transformer models for a specific low-resource language pair. These methods and results are detailed in the paper \textit{``Human evaluation of English–Irish Transformer-based NMT''}. Even though a different, human-centric approach was adopted, there is continuity with our previous research efforts. In particular, we sought a human validation of Transformer HPO for low-resource languages whereas in our previous work, \textit{``Transformers for low-resource languages - is féidir linn!''}, the focus was on automatic metrics. 

For our assessment, we employed an MQM error taxonomy complemented by an SQM to pinpoint the error types in both RNN and Transformer systems. Collaborating with native Irish speakers, our study devised a method for human evaluation under limited resources. Engaging native speakers to assess reference translations, we verified the outputs of both models. Our data reveals the Transformer system substantially reduces errors in accuracy and fluency. A pivotal aspect of our research is the linguistic feedback from our annotators.

\clearpage

   \begin{center}
       \vspace*{1cm}

       \textbf{Human Evaluation of EN$\leftrightarrow$GA Transformer-Based NMT}
            
       \vspace{1.5cm}

       \textbf{Séamus Lankford \\ Haithem Afli \\ Andy Way}

       \vfill
            
       Information 13.7, MDPI \\
        January 10th, 2023

       \vspace{0.8cm}
                 
       ADAPT Centre\\
       Dublin City University\\
       Ireland\\

                    \vspace{0.5cm}
 \url{https://www.mdpi.com/2078-2489/13/7/309}   
   \end{center}

\clearpage

\section{Abstract}

In this study, a human evaluation is carried out on how hyperparameter settings impact the quality of Transformer-based neural machine translation for the low-resource English-Irish language pair. SentencePiece models using both byte pair encoding (BPE) and unigram approaches were appraised. Variations in model architectures included modifying the number of layers, evaluating the optimal number of heads for attention and testing various regularisation techniques. The greatest performance improvement was recorded for a Transformer-optimised model with a 16k BPE subword model. Compared with a baseline recurrent neural network (RNN) model, a Transformer-optimised model demonstrated a BLEU score improvement of 7.8 points. When benchmarked against Google Translate, our translation engines demonstrated significant improvements. Furthermore, a quantitative fine-grained manual evaluation was conducted which compared the performance of machine translation systems. Using the multidimensional quality metrics error taxonomy, a human evaluation of the error types generated by an RNN-based system and a Transformer-based system was explored. Our findings show the best-performing Transformer system significantly reduces both accuracy and fluency errors when compared with an RNN-based model.

\section{Introduction}
A new era of high-quality translations has been heralded with the advent of neural machine translation (NMT). Given that large datasets are a prerequisite for high-quality NMT, these improvements are not always evident in the translation of low-resource languages. In the context of such languages, which suffer from a sparsity of data, alternative approaches must be adopted. 

Developing applications and models to address the challenges of low-resource language technology is an important part of this research. This technology incorporates new methods, which reduce the impact that data scarcity has on the digital engagement of low-resource languages. One approach is to use a mechanism that helps NMT systems to learn from unlabelled data using dual-learning~\parencite{he2016dual, ahmadnia2019augmenting}.
  
Out-of-the-box NMT systems, trained on English$\leftrightarrow$Irish (EN$\leftrightarrow$GA) data, have been shown to achieve a lower translation quality compared with using a tailored SMT system~\parencite{dowling2018smt}. It is in this context that further research is required in the development of NMT for low-resource languages, and the Irish language in particular.

Most research on the choice of subword models has focused on high-resource languages~\parencite{ding2019call, gowda2020finding}. Translation, by its nature, requires an open vocabulary and the use of subword models aims to address the fixed-vocabulary problem associated with NMT. Rare and unknown words are encoded as sequences of subword units. By adapting the original BPE algorithm \parencite{gage1994new}, the use of BPE subword models can improve translation performance \parencite{sennrich2015neural,kudo2018subword}. In the context of developing models for EN$\leftrightarrow$GA translation, there were no clear recommendations on the choice of subword model types. Character-based models were briefly explored due to their simplicity and reduced memory requirements. However, they were not considered suitable given that most single characters do not carry meaning in the English and Irish languages. Therefore, one of the objectives of our research is to identify which type of subword model performs best in this low-resource scenario.

An important goal of this study is to extend our previous work \parencite{lankford2021transformer} by providing a human evaluation and comparison of English to Irish (EN$\rightarrow$GA) machine translation (MT) on systems that use either a baseline RNN architecture or a subword-model optimised Transformer~model. 

This paper describes the context in which our research was conducted and provides a background of the types of available architecture in Section \ref{s2}. A detailed overview of our approach is outlined in Section \ref{s3}, where we provide details of the data and parameters used in our NMT systems. The empirical results, using both automatic metrics and a human evaluation, are presented in Section \ref{s4}. Finally, our findings are discussed in Section \ref{s6} and the possibilities for future work are outlined in Section \ref{s7}.

\section{Background\label{s2}}
   
Native speakers of low-resource languages are often excluded from useful content since, more often than not, online content is not available to them in their language of choice. This digital divide experienced by second-language speakers has been well-documented in the research literature \parencite{macfarlane2008responses, alam2015digital}.

Research on MT in low-resource scenarios seeks to directly address this challenge of exclusion via pivot languages~\parencite{liu2018pivot}, and indirectly, via domain adaptation of models ~\parencite{ghifary2016deep}. Consequently, research efforts focusing on NMT~\parencite{bahdanau2014neural, cho2014properties} have resulted in a state-of-the-art (SOTA) performance being attained for multiple language pairs~\parencite{bojar-etal-2017-findings, bojar-etal-2018-findings}. The Irish language is a primary example of a low-resource language that will benefit from this research. NMT involving Transformer model development will improve performance in specific domains of low-resource languages. 

\subsection{Hyperparameter Optimisation}

Hyperparameters are employed to customise machine learning models such as translation models. It has been shown that machine learning performance may be improved through hyperparameter optimisation (HPO) rather than just using default settings~\parencite{sanders2017informing}. The principal methods of HPO are grid search~\parencite{montgomery2001design} and random search \parencite{JMLR:v13:bergstra12a}. 

\subsubsection{RNN}

The tasks of natural language processing (NLP), speech recognition and MT are often performed by RNNs. This architecture enables previous outputs to be used as inputs while having hidden states. In the context of MT, such neural networks were ideal due to their ability to process inputs of any length. Furthermore, the model sizes do not necessarily increase with the input size. Commonly used variants of RNN include bidirectional (BRNN) and deep (DRNN) architectures. However, the problem of vanishing gradients coupled with the development of attention-based algorithms often leads to Transformer models performing better than RNNs.

\subsubsection{Transformer}
The greatest improvements have been demonstrated when either the RNN or the CNN architecture is abandoned completely and replaced with an attention mechanism creating a much simpler and faster architecture known as Transformer. Experiments in MT tasks show such models are better in quality due to greater parallelisation while requiring significantly less time to train~\parencite{vaswani2017attention}. 

Transformer models use attention to focus on previously generated tokens. The approach allows for models to develop a long memory, which is particularly useful in the domain of language translation. Performance improvements to both RNN and CNN approaches may be achieved through the introduction of such attention layers in the translation architecture.

\subsection{SentencePiece}

Designed for NMT, SentencePiece is a language-independent subword tokenizer that provides an open-source C++ and a Python implementation for subword units. An attractive feature of the tokenizer is that SentencePiece directly trains subword models from raw sentences~\parencite{kudo2018sentencepiece}.

\subsection{Human Evaluation}

Human evaluation, within NLP and MT, is a topic of growing importance, which often has a dedicated research track or workshop at major conferences \parencite{humeval-2021-human}. This focus has resulted in many publications in the area of human evaluation that relate to MT \parencite{toral-etal-2018-attaining, castilho2017neural} and it has particularly benefited the evaluation of low-resource languages \parencite{bayon-sanchez-gijon-2019-evaluating, imankulova2019exploiting}.

The best practice for the human evaluation of MT has been published in the form of a series of recommendations \parencite{laubli2020set}. As part of our research, we adopted these recommendations, which are in line with similar human evaluation studies of EN$\leftrightarrow$GA MT at the ADAPT centre \parencite{dowling2020human}. Specifically, these recommendations encourage the use of professional translators, evaluation at the document level and assessments of both fluency and accuracy. Original source texts were also used in the training and test data.

These recommendations have been complemented by a fine-grained human analysis, which uses both a Scalar Quality Metric (SQM) and Multidimensional Quality Metrics (MQM).

\section{Proposed Approach\label{s3}}

Considerable performance improvements have been achieved using the HPO of RNN models in low-resource settings. One of the key research questions, evaluated as part of this study, is to identify the extent to which such optimisation techniques may be applied to low-resource Transformer models. Evaluations included modifying the number of attention heads, changing the number of layers and experimenting with regularisation techniques such as dropout and label smoothing. Most importantly, the choice of subword model type and vocabulary size is evaluated. Furthermore, previous research focuses on using an automatic evaluation of performance, whereas we propose combining a human evaluation approach with automatic metrics.

\begin{figure}[h]

\centering 

\includegraphics[width=16cm]{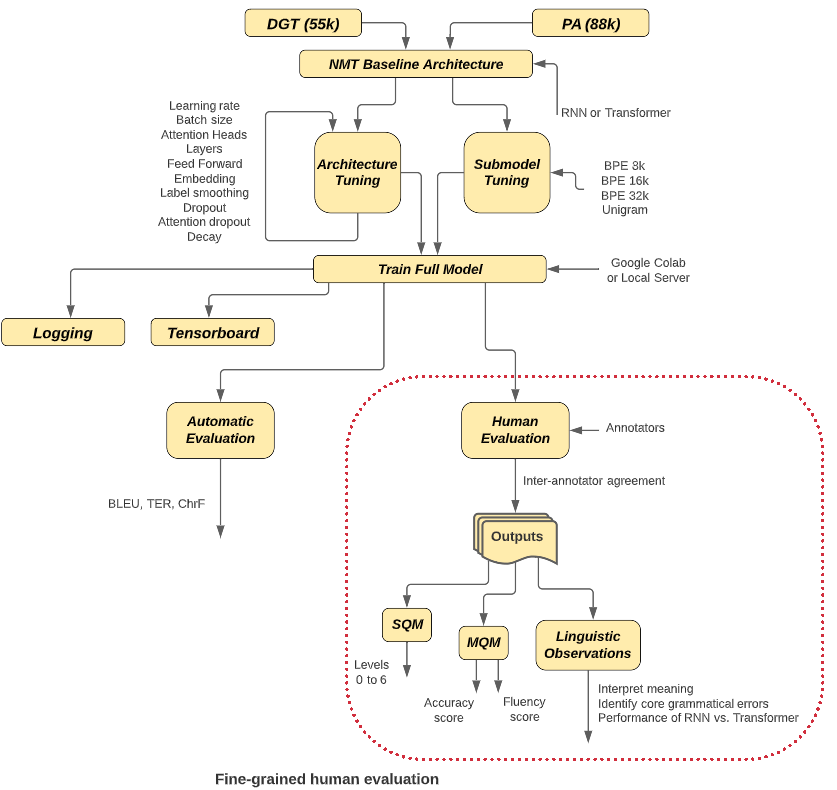}

\caption[Proposed approach to evaluate RNN and Transformer architectures]{Proposed approach to evaluate the baseline architectures of RNN and Transformer models. Using a random search approach, the values outlined in Table~\ref{tab:hpo-table-nmt-he} were tested to determine the optimal hyperparameters. Short cycles of 5k training steps were applied to test a range of values for each parameter. Once an optimal value was identified within the sampled range, it was locked in for tests on subsequent parameters. A fine-grained human evaluation was conducted on the output from the DGT dataset and its results were compared with an automatic evaluation.}
\label{fig:approach_human evaluation}
\end{figure}

To test the effectiveness of our approach, optimisation was carried out on an EN$\leftrightarrow$GA parallel dataset: a general corpus of 52k lines from the Directorate General for Translation (DGT). With DGT, the test set used 1.3k lines and the development set comprised 2.6k lines. All experiments involved concatenating source and target corpora to create a shared vocabulary and a shared SentencePiece subword model. The adopted approach is illustrated in Figure \ref{fig:approach_human evaluation}.

\subsection{Architecture Tuning}

It is difficult and costly to tune systems using a conventional grid search approach given the long training times associated with NMT. Therefore, we adopted a random search approach in the HPO of our Transformer models. 

Using smaller and fewer layers with low-resource datasets has previously been shown to improve performance~\parencite{araabi2020optimizing}. Furthermore, the use of shallow Transformer models has been demonstrated to improve the translation performance of low-resource NMT \parencite{van2020optimal}. Guided by these findings, configurations were tested, which varied the number of neurons in each layer and modified the number of layers used in the Transformer architecture.

Varying degrees of dropout were applied to Transformer models to evaluate the impact of regularisation. Configurations using smaller (0.1) and larger values (0.3) were applied to the output of each feed-forward layer.

\subsection{Subword Models}

Incorporating a word segmentation approach, such as BPE, is now standard practice when developing NMT models. Subword models are particularly beneficial for low-resource languages since rare words are often a problem. In the context of EN$\rightarrow$GA translation, there is no clear agreement as to what constitutes the best approach. Consequently, subword regularisation techniques involving BPE and unigram models were evaluated as part of this study to determine the optimal parameters for maximising translation performance. BPE models with varying vocabulary sizes of 4k, 8k, 16k and 32k were evaluated.

\begin{table}[h]
\centering
\small

\caption[Transformer HPO using a random search approach]{Transformer HPO using a random search approach. The optimal hyperparameters are highlighted in bold. The best-performing model used two attention heads and was trained on a 55k DGT corpus.}
\begin{tabular}{ll}
\hline
\textbf{Hyperparameter} & \textbf{Values}                \\ \hline
Learning rate            & 0.1, 0.01, 0.001, \textbf{{2}}            \\ 
Batch size               & 1024, \textbf{{2048}},  4096, 8192       \\
Attention heads          & \textbf{{2}}, 4, \textbf{{8}}              \\ 
Number of layers         & 5, \textbf{{6}}                           \\ 
Feed-forward dimension   & \textbf{{2048}}                           \\ 
Embedding dimension      & 128, \textbf{{256}}, 512                  \\ 
Label smoothing          & \textbf{{0.1}}, 0.3                       \\ 
Dropout                  & 0.1, \textbf{{0.3}}                       \\ 
Attention dropout        & \textbf{{0.1}}                            \\ 
Average Decay            & 0, \textbf{{0.0001}}                      \\ \hline
\end{tabular}
\label{tab:hpo-table-nmt-he}
\end{table}

\subsection{Human Evaluation of NMT}

Morphological-rich languages, such as Irish, have a high degree of inflection and free word order that gives rise to specific translation issues when translating from English. Grammatical categories such as gender or case inflections in nouns are often difficult to reliably generate in an Irish translation. 
\par
One of the goals of this research is to explore how an NMT system handles these issues compared with an RNN approach. Existing research suggests NMT systems should improve these linguistic aspects. NMT, with its use of subword models, implicitly addresses the problem in an unsupervised manner, without understanding the actual formal rules of grammatical categories.
\par
Previous human evaluation studies that evaluate EN$\leftrightarrow$GA MT performance have focused on the differences between an SMT and an NMT approach~\parencite{dowling2018smt}. In the context of our research, human evaluation was conducted on purely NMT methods, which included RNN and Transformer approaches. Furthermore, our study is differentiated by using both SQM and MQM as our human evaluation metrics.

\par
It is clear from our earlier experimental findings, based solely on automatic evaluation metrics, that a Transformer approach leads to significant improvements compared to traditional RNN systems. However, as with most automatic scoring methods, these simply provide an overall score for each system but do not indicate the exact nature of the linguistic problems that may be encountered in translation. Therefore, it can be said that automatic evaluation does not address the question of the linguistic or grammatical quality of the target output. Nuances, such as how gender or cases are handled, are not covered by this approach.    
\par
To achieve a deeper understanding of the linguistic errors created by our RNN and Transformer systems, a fine-grained human evaluation was conducted. The outputs from these systems were systematically analysed and compared in a manual error analysis. This approach captures the nature of the translation errors for each of the evaluated systems. The output from this study forms the basis of future work, which will help to improve the translation quality of our models. The annotation framework, the overall annotation process and the inter-annotator agreement are discussed below and broadly follow the approach adopted by other fine-grained human evaluation studies \parencite{klubivcka2018quantitative}.

\subsubsection{Scalar Quality Metrics}

SQM \parencite{freitag-etal-2021-experts} adapts the WMT shared-task settings to collect segment-level scalar ratings with a document context. SQM uses a scale from 0 to 6 for translation quality assessment. This is a modification of the WMT approach \parencite{ma-etal-2017-blend}, which uses a range from 0 to 100. 
 
With this evaluation approach, annotators must select a rating from 0 through 6 when presented with the source and target sentences. The SQM quality levels for 0, 2, 4 and 6 are outlined in Table \ref{tab:sqm-nmt}. Annotators may also choose intermediate levels of 1, 3 and 5 in cases where the translations do not exactly match the core SQM levels.

\begin{table}[h]
\caption[SQM levels explained]{SQM levels explained \parencite{freitag-etal-2021-experts}.}

\centering 
\begin{tabular}{cp{12cm}}
\hline
\multicolumn{1}{c}{\textbf{SQM Level}} & \textbf{Details of Quality} \\ \hline
\multirow{2}{*}{ 6} &
  Perfect Meaning and Grammar: The meaning of the translation is completely consistent with the source and the surrounding context (if applicable). The grammar is also correct. \\ \hline
\multirow{2}{*}{ 4} &
  Most Meaning Preserved and Few Grammar Mistakes: The translation retains most of the meaning of the source. This may contain some grammar mistakes or minor contextual inconsistencies. \\  \hline
\multirow{2}{*}{ 2} & Some Meaning Preserved: The translation preserves some of the meaning of the source but misses significant parts. The narrative is hard to follow due to fundamental errors. Grammar may be poor. \\  \hline
 \multirow{2}{*}{ 0}&
  Nonsense/No meaning preserved: Nearly all information is lost between the translation and source. Grammar is irrelevant. \\ \hline
\end{tabular}
\label{tab:sqm-nmt}

\end{table}

\subsubsection{Multidimensional Quality Metrics}

As part of QTLaunchpad project\footnote{\url{https://www.qt21.eu/}} the MQM framework\footnote{\url{https://themqm.org/the-mqm-full-typology/}} was developed to provide a framework of how manual evaluation could be performed via a detailed error analysis. A single metric for all uses is not imposed. Instead, a comprehensive catalogue of quality issue types, with standardised names and definitions, is provided. This catalogue may be customised for specific tasks. In addition to forming a reliable methodology for quality assessment, it also allows for us to specify which error tags were relevant to our task.
\par
To adapt the generic MQM framework for our context, we followed the official guidelines for scientific research \parencite{Lommel2018}. The details of our customisation of MQM are discussed below. 
\par
\begin{figure}[h]
 \centering
\includegraphics[scale=0.75]{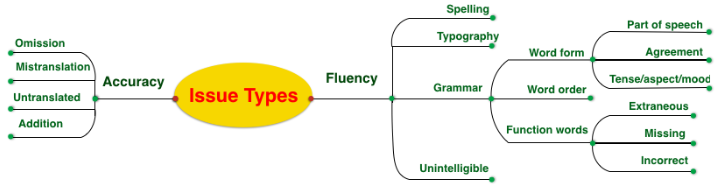}
    \caption{The core set of error categories proposed by the MQM guidelines.}
    \label{fig:mqm_core}
\end{figure}

A large variety of tags, on several annotation layers, are proposed within the original MQM guidelines. However, this full MQM tagset is too detailed for a specific annotation task. Therefore, when evaluating our MT output, the smaller default set of evaluation categories, specified in the core tagset, was used. These standard top-level categories of accuracy and fluency, which are proposed by the MQM guidelines, are illustrated in Figure \ref{fig:mqm_core}. A special non-translation error was used to tag an entire sentence, which was too badly translated to allow for the identification of individual errors.

Error severities are specified as either major or minor errors and are assigned independently of category. These correspond to actual translation and grammatical errors or smaller imperfections, respectively.  
The recommended default weights \parencite{Lommel2018} were used, which allocate a weight of 1 to minor errors whereas major errors are assigned a weight of 10. Furthermore, the non-translation category was allocated a weight of 25, an approach which is in line with the best practice established in previous studies \parencite{freitag-etal-2021-experts}. 

The annotators were instructed to identify all errors within each sentence of the translated output for both systems. The error categories used by the annotators are outlined in Table~\ref{tab:mqmcat-human}.

\begin{table}[h]
\small
\caption[Description of error categories within the core MQM framework]{Description of error categories within the core MQM framework \parencite{freitag-etal-2021-experts}.}

\centering 
\begin{tabular}{p{2.7cm}p{3.3cm}p{7.75cm}}
\hline
\textbf{Category} & \textbf{Sub-Category} & \textbf{Description}                                 \\ \hline
\textbf{Non-translation} &                & Impossible to reliably characterise the 5 most severe errors.       \\ \hline
\textbf{Accuracy}        & Addition       & Translation includes information not present in the source.         \\
         & Omission              & Translation is missing content from the source.      \\
                & Mistranslation & Translation does not accurately represent the source.               \\
         & Untranslated text     & Source text has been left untranslated.              \\
        \hline
\textbf{Fluency}  & Punctuation           & Incorrect punctuation                                \\
         & Spelling              & Incorrect spelling or capitalisation.                \\
         & Grammar               & Problems with grammar, other than orthography.       \\
                         & Register       & Wrong grammatical register (e.g., inappropriately informal pronouns). \\
                  & Inconsistency         & Internal inconsistency (not related to terminology). \\
                  & Character encoding & Characters are garbled due to incorrect encoding.   \\ \hline
\end{tabular}

\label{tab:mqmcat-human}
\end{table}

\subsubsection{Annotation Setup}

Annotations were carried out using the simpler SQM approach and a more detailed, fine-grained MQM approach. The hierarchical taxonomy of our MQM implementation is illustrated in Figure \ref{fig:mqm_core}, whereas the SQM categories are summarised in Table~\ref{tab:sqm-nmt}. 
\par
Two annotators with similar backgrounds were used for the annotation of outputs from an RNN system and a Transformer system. Both annotators are native speakers of Irish and neither had prior experience with MQM. Prior to annotation, they were thoroughly familiarised with the process and the official MQM annotation guidelines. These guidelines offer detailed instructions for annotation within the MQM framework.

Both annotators have been very involved in the education sector for decades. One of the annotators has edited numerous English-language and Irish-language books during her career as a university lecturer. The second annotator has a PhD in Irish-language place names. In addition, he has written numerous books in both English and Irish. Given their experience and strong language backgrounds, they were well-equipped to handle the task at hand.

Using a test set of 20 randomly selected sentences, the annotators were presented with the English source text, an Irish reference translation and the two unannotated system outputs: one generated using an RNN model and the other created using a Transformer model. Potential bias was removed by using blind annotation such that annotators did not know which model the translation output came from. The annotators worked independently of each other but were occasionally in contact to discuss the process and how to approach difficult sentences.

Translations from the RNN and the Transformer system were annotated by both annotators, meaning that each system translated the same 20 sentences and each annotator annotated the resulting 40 translated sentences (20 source sentences for 2 MT systems), producing a total of 80 annotated sentences. The annotated dataset is publicly available on GitHub.\footnote{\url{https://github.com/seamusl/isfeidirlinn}}  

Once the annotation data were extracted, each annotator analysed the output to determine the performance of each system for each error category. 

\subsubsection{Inter-Annotator Agreement}
Low inter-annotator agreement (IAA) scores are a common problem experienced when using manual MT evaluation approaches such as MQM \parencite{lommel-etal-2014-using,callison2007meta}.To determine the validity of the findings of our research, it is important to check the level of agreement between our annotators~\parencite{artstein2017inter}. 

Cohen’s $kappa$ ($k$) \parencite{cohen1960coefficient} was used to determine inter-annotator agreement. The agreement was calculated based on the annotations of each system, with the agreement being observed at the sentence level. With this approach, the differences in agreement across systems were explored and we also gained a high-level view of overall agreement between the annotators. Furthermore, Cohen’s $kappa$ was calculated separately for every error type and the findings are outlined in Table \ref{tab:my-cohen-nmt}.

\begin{table}[h]
\centering
\caption{Inter-annotator agreement using Cohen values}
\begin{tabular}{@{}clll@{}}
\hline
\textbf{Error   type} & \multicolumn{1}{c}{\textbf{RNN}} & \multicolumn{1}{c}{\textbf{NMT}}  \\ \hline
\multicolumn{1}{l}{Non-translation}               & 1.0 & 1.0    \\
\multicolumn{1}{l}{Accuracy}                      & 1.0 & 1.0   \\
Addition                                          & 1.0 & 1.0   \\
Omission                                          & 1.0 & 1.0   \\
Mistranslation                                    & -0.071 & 1.0   \\
Untranslated text                                 & 0.0 & 1.0   \\
\multicolumn{1}{l}{Fluency}                       &  &  &  \\
Punctuation                                       & 0.651 & 1.0    \\
Spelling                                          & 0.0 & 0.0    \\
Grammar                                           & 0.867 & 0.895   \\
Register                                          & 1.0 & 1.0   \\
Inconsistency                                     & 1.0 & 1.0   \\
Character Encoding                                & 1.0 & 1.0    \\ 
 \hline
\end{tabular}

\label{tab:my-cohen-nmt}
\end{table}  
   \vspace{4.5cm}
   
\section{Empirical Evaluation\label{s4}}

\subsection{Experimental Setup}
\subsubsection{Datasets}
The performance of the Transformer and RNN approaches is evaluated on a publicly available EN$\leftrightarrow$GA parallel dataset from the Directorate General for Translation (DGT).\footnote{\url{https://ec.europa.eu/info/departments/translation}} The Joint Research Centre of the DGT has made all its translation memory (i.e. sentences and their professionally produced translations) available, which covers the official European Union languages~\parencite{steinberger2013dgt}. Included in the training data are parallel texts from the Digital Corpus of the European Parliament (DCEP) and the DGT. Crawled data, from sites of a similar domain, are also incorporated. This dataset is broadly categorised as generic and is publicly available.  

\subsubsection{Infrastructure}
Model development was conducted using local workstations, each of which was built with an AMD Ryzen 7 2700X processor, 16GB memory, a 256SSD and an NVIDIA GeForce GTX 1080 Ti. 

In addition, a Google Colab Pro subscription enabled rapid prototype development and created zero-emission models. The available computing power of the Google Cloud was much higher than our local infrastructure and provided servers with 16GB graphic cards (NVIDIA Tesla P100 PCIe) and up to 27GB of memory~\parencite{Bisong2019}. Larger Transformer models were built on local infrastructure since long builds timed out on Colab due to Google restrictions. The Pytorch implementation of OpenNMT 2.0, an open-source toolkit for NMT \parencite{klein2017opennmt}, was used to train all MT models.

\subsubsection{Metrics}

The performance of all models was evaluated using the automated metrics of BLEU~\parencite{papineni-etal-2002-bleu}, TER \parencite{snover2006study} and ChrF \parencite{popovic2015chrf}. Case-insensitive BLEU scores are reported at the corpus level.  

\subsection{Automatic Evaluation Results}

\subsubsection{Performance of Subword Models}
The impact that choice of subword model has on translation is highlighted in Tables \ref{tab:dgtvanilla-table} and  \ref{tab:trans-table}. 
Incorporating any subword model type led to improvements in model accuracy when training both RNN and Transformer architectures.  

\begin{table}[h]
\small
\caption[RNN performance on DGT dataset of 52k Lines]{RNN performance on DGT dataset of 52k Lines. There were zero carbon emissions in building these models since smaller RNN models were trained on Google Colab servers, which are carbon-neutral.}

\centering
\begin{tabular}{lcccccccc}
\hline
\textbf{Architecture} &
  \textbf{BLEU} \boldmath{$\uparrow$} &
  \textbf{TER} \boldmath{$\downarrow$} &
  \textbf{ChrF3} \boldmath{$\uparrow$} &
  \textbf{Steps} &
  \textbf{Runtime  (h)} &
  \textbf{kgCO\textsubscript2} \\ \hline
dgt-rnn-base    & 52.7       & 0.42  & 0.71 & 75k  & 4.47 & 0 \\
dgt-rnn-bpe8k   & 54.6       & 0.40 & 0.73 & 85k  & 5.07 & 0 \\
dgt-rnn-bpe16k  & 55.6 & 0.39 & 0.74 & 100k & 5.58 & 0 \\
dgt-rnn-bpe32k  & 55.3       & 0.39 & 0.74 & 95k  & 4.67 & 0 \\ 
dgt-rnn-unigram & 55.6 & 0.39 & 0.74 & 105k & 5.07 & 0 \\ \hline
\end{tabular}

\label{tab:dgtvanilla-table}
\end{table}

\begin{table}[h]
\small
\caption[Transformer performance on 52k DGT dataset]{Transformer performance on 52k DGT dataset. The highest-performing model uses 2 attention heads. All other models use 8 attention heads. Transformer models were long-running builds, which had to be carried out on local servers.}

\centering 

\begin{tabular}{lcccccccc}
\hline
\textbf{Architecture} &
  \textbf{BLEU} \boldmath{$\uparrow$} &
  \textbf{TER} \boldmath{$\downarrow$} &
  \textbf{ChrF3} \boldmath{$\uparrow$} &
  \textbf{Steps} &
  \textbf{Runtime (h)} &
  \textbf{kgCO\textsubscript2} \\ \hline
dgt-trans-base      & 53.4 & 0.41 & 0.72 & 55k  & 14.43 & 0.81 \\
dgt-trans-bpe8k     & 59.5 & 0.34 & 0.77 & 200k & 24.48 & 1.38 \\
dgt-trans-bpe16k    & 60.5 & 0.33 & 0.78 & 180k & 26.90 & 1.52 \\
dgt-trans-bpe32k    & 59.3 & 0.35 & 0.77 & 100k & 18.03 & 1.02 \\ 
dgt-trans-unigram   & 59.3 & 0.35  & 0.77 & 125k & 21.95 & 1.24 \\ \hline
\end{tabular}

\label{tab:trans-table}
\end{table}

A baseline RNN model, illustrated in Table \ref{tab:dgtvanilla-table}, achieved a BLEU score of 52.7, whereas the highest-performing BPE variant,  using a 16k vocabulary, recorded an improvement of nearly three points, with a score of 55.6. 

In the context of Transformer architectures, highlighted in Table \ref{tab:trans-table}, the use of subword models delivers significant performance improvements. The performance gains for Transformer models are much higher than the improvements recorded by the RNN models. A baseline Transformer model achieves a BLEU score of 53.4, whereas a Transformer model, with a 16k BPE subword model, has a score of 60.5, representing a BLEU score improvement of 13\% at 7.1 BLEU points. 

For translating into a morphological-rich language, such as Irish, the ChrF metric has proven successful in showing a strong correlation with human translation \parencite{stanojevic2015results}. In the context of our experiments, this worked well in highlighting the performance differences between RNN and Transformer architectures. 

\subsubsection{Transformer Performance Compared with RNN}

\begin{figure}[h]
    \includegraphics[width=15cm]{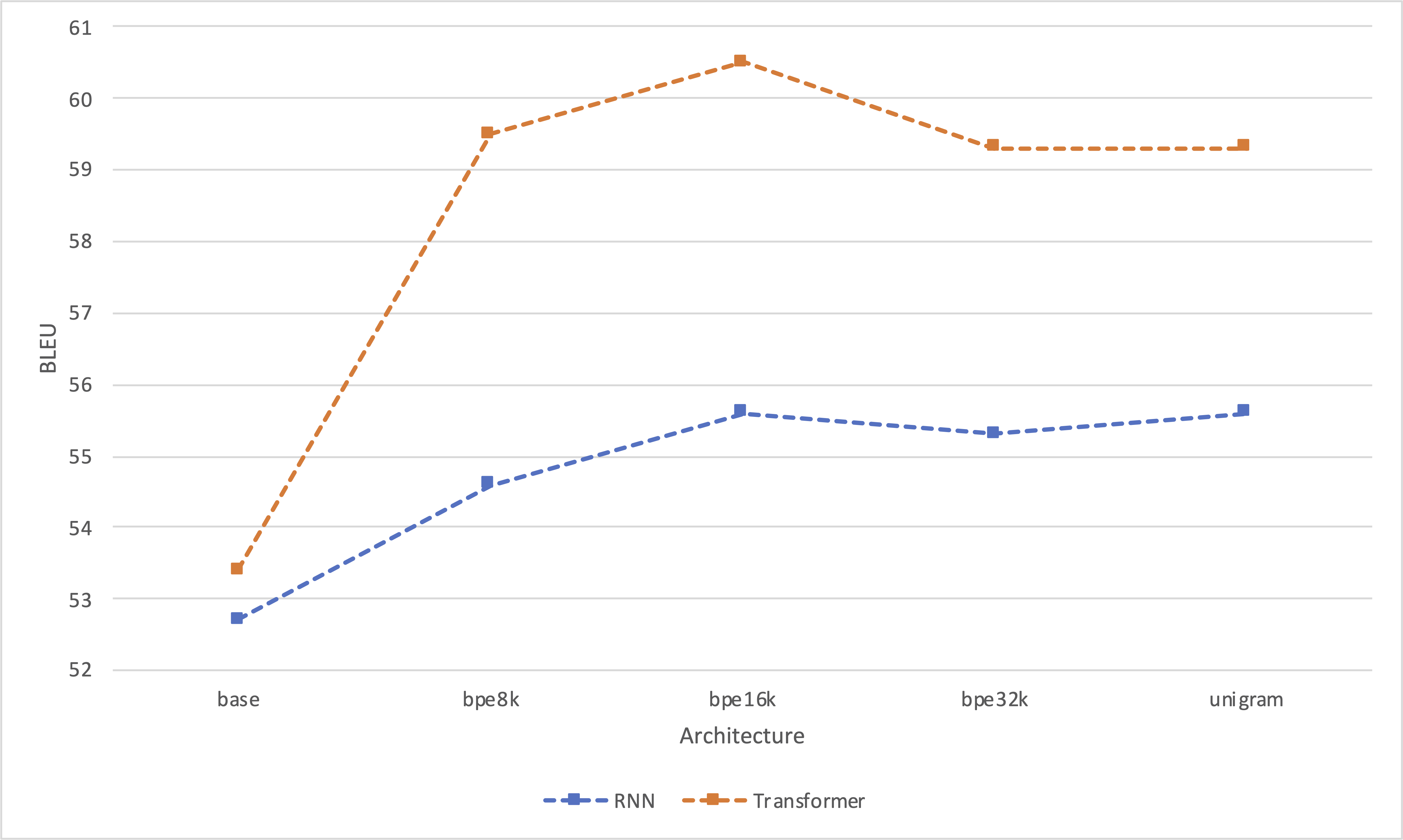}
    \caption[BLEU performance for all model architectures]{BLEU performance for all model architectures. The use of a BPE subword model improved translation performance in all cases. The best-performing model was built using a 16k BPE subword model on a Transformer architecture. }
    \label{fig:dgt}
\end{figure}

\begin{figure}[h]
    \includegraphics[width=15cm]{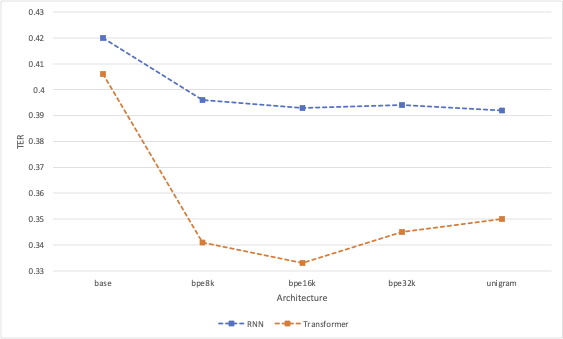}
    \caption[TER performance for all model architectures]{TER performance for all model architectures. The highest-performing model uses a 16k BPE subword model on a Transformer architecture. In all instances, incorporating a subword model improves TER.}
    \label{fig:pacompare}
\end{figure}

The performance of RNN models is contrasted with the Transformer approach in \linebreak  Figures \ref{fig:dgt} and \ref{fig:pacompare}. Transformer models, as anticipated, outperformed all their RNN counterparts. It is interesting to note the impact of choosing the optimal vocabulary size for BPE subword models. Choosing a BPE vocabulary of 16k words yields the highest performance. 

Furthermore, the TER scores highlighted in Figure \ref{fig:pacompare} reinforce the findings that using 16k BPE subword models on Transformer architectures leads to a better translation performance. The TER score for the 16k BPE Transformer model is significantly better (0.33) when compared with the baseline performance (0.41).

\subsection{Human Evaluation Results}

The aggregate total of errors found by annotators for each system is highlighted in Table~\ref{tab:mqmtotals}. Looking at the aggregate data alone, it is evident that both annotators have judged that the RNN system contains more errors and that the NMT system contains fewer errors.

\begin{table}[h]
\centering
\caption{Total errors found by each annotator using the MQM metric.}
\begin{tabular}{@{}lcccc@{}}
\hline
                      & \multicolumn{2}{c}{\textbf{Annotator 1}} & \multicolumn{2}{c}{\textbf{Annotator 2}} \\ \hline
\textbf{System}       & RNN                 & Transformer                & RNN                 & Transformer                \\ \hline
\textbf{Total Errors} & 41                     & 23                    & 43                     & 23                   \\ \hline
\end{tabular}

\label{tab:mqmtotals}
\end{table}

\par
While such a high-level view is instructive in determining which system is better, it lacks the granularity required to pinpoint the linguistic aspects of how these translations can be improved. To achieve a deeper insight, a fine-grained analysis of the error types was conducted, the results of which are displayed in Table~\ref{tab:mqm_combined}.  Categorised by error type, the sum of error tags by each annotator for each system is outlined.

\begin{table}[h]
\centering
\caption[Transformer and RNN approach compared]{Transformer and RNN approach compared using concatenated annotation data across both annotators. In all MQM error categories, the Transformer architecture performs better, apart from a tie in the omission category.}
\begin{tabular}{@{}lcccc@{}}
\hline
\multicolumn{1}{l}{} &
  \multicolumn{1}{c}{\textbf{RNN}} &
  \multicolumn{1}{c}{\textbf{NMT}} \\ \hline
\textbf{Error Type} &
  \multicolumn{1}{c}{\textbf{Error}} &
  \multicolumn{1}{c}{\textbf{Error}} \\ \hline
Non-translation       & 0 & 0    \\
\textbf{Accuracy}              &  &     \\
Addition                                  & 10 & 4   \\
Omission                                  & 12 & 12   \\
Mistranslation                            & 26  & 14    \\
Untranslated text                         & 4 & 1   \\
\textbf{Fluency}               &  &    \\
Punctuation                               & 5 & 4   \\
Spelling                                  & 1 & 0   \\
Grammar                                   & 20 & 11   \\
Register                                  & 2 &  0  \\
Inconsistency                             & 2 & 0   \\
Character Encoding                        & 0 & 0   \\ \hline
\multicolumn{1}{l}{\textbf{Total errors}} & 82 & 46    \\ \hline
\end{tabular}

\label{tab:mqm_combined}
\end{table}

\section{Environmental Impact\label{s5}}
The environmental impact of all aspects of computing has received increased research interest in recent times. Much of this effort has concentrated on NMT's carbon footprint~\parencite{info13020088, bender2021dangers}. To assess the environmental impact of our NMT models, we tracked energy consumption during their development. 

Prototype model development was carried out using Google Colab which is a carbon-neutral platform~\parencite{lacoste2019quantifying}. However, longer running Transformer experiments were conducted on local servers using 324 gCO\textsubscript2 per kWh\footnote{\url{https://www.seai.ie/publications/Energy-in-Ireland-2020.pdf}} \parencite{sei2020}. The net result was just under 10 kgCO\textsubscript2, created for a full run of model development. Models developed during this study will be reused for ensemble experiments in future work.

The environmental costs of our model development were tracked to serve as a benchmark for future work. Awareness of such costs will impose a discipline on our work, such that we opt for carbon-neutral cloud providers. In cases where models are developed on local infrastructure, this will encourage the use of more efficient GPUs and the utilisation of techniques that result in faster builds.

\section{Discussion\label{s6}}
 
\begin{figure}[h]
\centering
    \includegraphics[width=7cm]{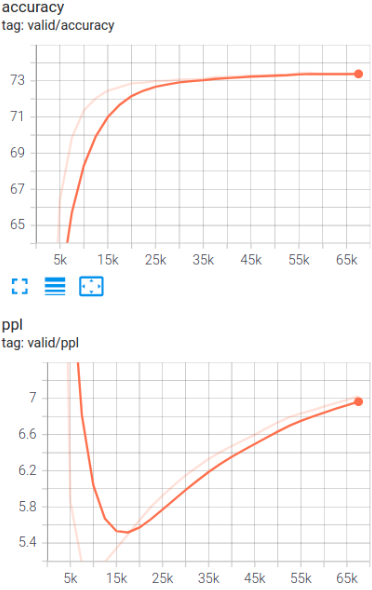}
    \caption{Transformer baseline.}
    \label{fig:train-base}
\end{figure}

\begin{figure}[h] 
\centering
    \includegraphics[width=7cm]{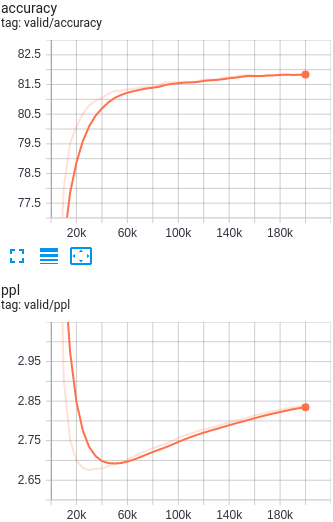}
    \caption{Transformer 16k BPE subword model}
    \label{fig:train-bpe}
\end{figure}

Validation accuracy and model perplexity (PPL) in developing the baseline and optimal Transformer models are illustrated in Figures \ref{fig:train-base} and  \ref{fig:train-bpe}. Training a Transformer model with a 16k BPE subword model boosted the validation accuracy by over 8\% compared to its baseline.

Rapid convergence was observed while training the baseline model such that little accuracy improvement occurred after 20k steps. Including a subword model led to slower converging models, with only marginal gains recorded after 60k steps. Examining Figures \ref{fig:train-base} and  \ref{fig:train-bpe}, we can see that PPL achieves a lower global minimum when the Transformer approach is used with a 16k BPE subword model. The PPL global minimum (2.7) is over 50\% lower than the corresponding PPL for the Transformer base model (5.5). This finding illustrates that choosing an optimal subword model delivers significant performance gains.

Translation engine performance, at the corpus level, was benchmarked against Google Translate's\footnote{\url{https://translate.google.com/}} EN$\leftrightarrow$GA translation service, which is freely available on the internet. Four random samples were selected from the English source test file and are presented in Table~\ref{tab:translations}. Translation of these samples was carried out on the optimal Transformer model and using Google Translate. Case-insensitive, sentence-level BLEU scores were recorded and are presented in Table~\ref{tab:trans-google}. It must be acknowledged that this comparison is not entirely valid given that Google does not have access to our training data, nor do we have unlimited access to the Google Cloud infrastructure. Nonetheless, the results are encouraging and indicate a good performance by our translation models on the DGT dataset.

\begin{table} [h]
\small
\caption{Random samples of human reference translations from the test dataset.}
\begin{small} 
\centering 

\begin{tabular}{p{7.2cm}p{7.2cm}}\hline
\textbf{Source Language (English)} & \textbf{Reference Human Translation (Irish)} \\  \hline
A clear harmonised procedure, including the necessary criteria for disease–free status, should be established for that purpose. & Ba cheart nós imeachta comhchuibhithe soiléir, lena n-áirítear na critéir is gá do stádas saor ó ghalar, a bhunú chun na críche sin. \\ \hline
the mark is applied anew, as appropriate. & déanfar an mharcáil arís, mar is iomchuí. \\ \hline
If the court decides that a review is justified on any of the grounds set out in paragraph 1, the judgment given in the European Small Claims Procedure shall be null and void. & Má chinneann an chúirt go bhfuil bonn cirt le hathbhreithniú de bharr aon cheann de na forais a leagtar amach i mír 1, beidh an breithiúnas a tugadh sa Nós Imeachta Eorpach um Éilimh Bheaga ar neamhní go hiomlán. \\ \hline
households where pet animals are kept; & teaghlaigh ina gcoimeádtar peataí; \\ \hline

\end{tabular}

\label{tab:translations}
\end{small}

\end{table}

\begin{table}[h]
\caption[Transformer model compared with Google Translate]{Transformer model compared with Google Translate using random samples from the DGT corpus. Full evaluation of Google Translate's engines on the DGT test set, with 1.3k lines, generated a BLEU score of 46.3 and a TER score of 0.44. Comparative scores on the test set using our Transformer model, with 2 attention heads and 16k BPE subword model realised 60.5 for BLEU and 0.33 for TER.}
\begin{small} 
\centering 
\begin{tabular}{p{5cm}p{1.75cm}p{5cm}p{1.75cm}}\hline

\textbf{Transformer (16k BPE)} & \textbf{BLEU} \boldmath{$\uparrow$} & \textbf{Google Translate} & \textbf{BLEU} \boldmath{$\uparrow$}\\ \hline

Ba cheart nós imeachta soiléir comhchuibhithe, lena n-áirítear na critéir is gá maidir le stádas saor ó ghalair, a bhunú chun na críche sin. & 61.6 & Ba cheart nós imeachta comhchuibhithe soiléir, lena n-áirítear na critéir riachtanacha maidir le stádas saor ó ghalair, a bhunú chun na críche sin. & 70.2 \\ \hline
go gcuirtear an marc i bhfeidhme, de réir mar is iomchuí. & 21.4 & cuirtear an marc i bhfeidhm as an nua, de réir mar is cuí. & 6.6 \\ \hline

Má chinneann an chúirt go bhfuil bonn cirt le hathbhreithniú ar aon cheann de na forais a leagtar amach i mír 1, beidh an breithiúnas a thugtar sa Nós Imeachta Eorpach um Éilimh Bheaga ar neamhní. & 77.3 &  Má chinneann an chúirt go bhfuil údar le hathbhreithniú ar aon cheann de na forais atá leagtha amach i mír 1, beidh an breithiúnas a thugtar  sa Nós Imeachta Eorpach um Éilimh Bheaga ar neamhní & 59.1 \\ \hline
teaghlaigh ina gcoimeádtar peataí; & 100 & teaghlaigh ina gcoinnítear peataí; & 30.2 \\ \hline
\end{tabular}

\label{tab:trans-google}
\end{small}

\end{table}

The optimal hyperparameters selected in this discovery process are identified in bold in Table~\ref{tab:hpo-table-nmt-he}. A higher initial learning rate of 2 coupled with an average decay of 0.0001 led to longer training times but more accurate models. Despite setting an early stopping parameter, many of the Transformer builds continued for the full cycle of 200k steps over periods of 20+ hours. 

Training Transformer models with a reduced number of attention heads led to a marginal improvement in translation accuracy with a smaller corpus. Our best-performing model achieved a BLEU score of 60.5 and a TER score of 0.33 with 2 heads and a 16k BPE subword model. By comparison, using 8 heads with the same architecture and dataset yielded 60.3 for BLEU and 0.34 in terms of TER. 

Transformer models developed, using state-of-the-art techniques, were evaluated as part of the LoResMT2021 Shared Task \parencite{ojha2021findings}. Models developed using our approach, as outlined above, were entered into the competition, and the highest-performing EN$\rightarrow$GA direction system was submitted by our team (ADAPT) \parencite{lankford2021machine}.

\subsection{Inter-Annotator Reliability}

In Cohen's original article \parencite{cohen1960coefficient}, the interpretation of specific $k$ scores is clearly outlined. There is no agreement with values $\le$ 0, none to slight agreement when scores are in the range of 0.01--0.20, fair agreement is represented by 0.21--0.40, 0.41--0.60 is moderate agreement, 0.61--0.80 is substantial agreement, and 0.81--1.00 is almost perfect agreement. 

The literature \parencite{mchugh2012interrater} recommends a minimum of 80\% agreement for good inter-annotator agreement. As illustrated in Table \ref{tab:my-cohen-nmt}, there is almost perfect agreement between the annotators when evaluating output from the NMT models. In the case of the RNN outputs, there is disagreement in the mistranslation category but agreement in all other categories. Given these scores, we have a high degree of confidence in our human evaluation of both the RNN and NMT outputs.

\subsection{Performance of Models Relative to Google}

Using standard Transformer parameters, such as a  batch size of 2048 and setting the number of encoder and decoder layers to 6, were observed to perform well. Increasing the regularisation dropout to 0.3 and reducing hidden neurons to 256 improved translation performance. Consequently, these values were selected when building all Transformer models.

\subsection{Linguistic Observations}\label{lo}

A linguistic analysis of the outputs from the Transformer-optimised model is illustrated in Table \ref{tab:ling-anal-lrev}. The English language source sentences and their Irish language translations are presented. Sentences have been selected from the fine-grained human evaluation since they highlight some of the key error types that are encountered. The analysis focuses on the shortcomings of our model outputs, which fall into the following categories: interpretative meaning, core grammatical errors and commonly used irregular verbs. Finally, using the human evaluation metrics of SQM and MQM, the performance of an RNN approach is contrasted with that of the Transformer approach.  

\subsubsection{Interpreting Meaning}

The generic Irish verb ``déan'' (to do or to make) is used to express more precise concepts such as ``to conduct'', ``to put into effect'' or ``to carry out''. Both the RNN and Transformer systems make use of ``déan'' in a generic way, but they fail to capture the refinement of the concept expressed in each of these meanings. An example of this problem is illustrated in GA-1 in Table~\ref{tab:ling-anal-lrev}. In this context, a more natural and intuitive translation to capture the expression ``to conduct'' would be to substitute ``a dhéanamh'' with ``a sheoladh''.

\clearpage
A similar lack of refinement from both systems is also found in the usage of other words. For example, ``cuid'' (part) is used to translate ``operative part'' in GA-2. However, a more precise interpretation would be the usage of ``gné'', leading to the correct translation ``gné oibríochtúil'' i.e., ``operative part''.

Another example where the translation models failed to correctly interpret the true sense of an English source word into a corresponding Irish translation can be seen in GA-3. The Irish verb ``Mainnigh'' meaning ``to default'' would not be used in the context of the source text in EN-3. Using the Irish verb ``teip'', meaning ``to fail'', is the correct translation of the idea ``fails to meet the performance requirements'': ``má theipeann an t-oibreoir na ceanglais feidhmíochta a chomhlíonadh.'' This error was observed in both the RNN and Transformer model outputs.

\begin{table}[h!]
\small
\setlength{\tabcolsep}{4.85mm}
\caption[Linguistic analysis of system outputs]{Linguistic analysis of system outputs. Sources of errors are flagged in blue and in red.}

\centering \begin{tabular}{p{2cm}p{10.5cm}}
\hline
\textbf{Type} &
  \textbf{Sentence} \\ \hline 
\textbf{EN-1} &
  The lead supervisory authority may request at any time other supervisory authorities concerned to provide mutual assistance pursuant to Article 61 and \textcolor{blue}{may conduct} joint operations pursuant to Article 62, in particular for carrying out investigations or for monitoring the implementation of a measure concerning a controller or processor established in another Member State. \\ \hline 
\textbf{GA-1} &
  Féadfaidh an príomhúdarás  maoirseachta iarraidh, tráth ar bith, ar bith eile lena mbaineann cúnamh   frithpháirteach a chur ar fáil de bhun Airteagal 61 agus féadfaidh sé oibríochtaí comhpháirteacha a dhéanamh de bhun Airteagal 62, go háirithe   maidir le himscrúduithe a dhéanamh nó maidir le faireachán \textcolor{red}{a dh\'{e}anamh} ar chur chun feidhme beart i ndáil le rialaitheoir nó próiseálaí atá bunaithe i mBallstát eile. \\\hline
\textbf{EN-2} &
  The Office shall mention the judgment in the Register and shall take the necessary measures to comply with its operative \textcolor{blue}{part}. \\ \hline
\textbf{GA-2} &
 Luafaidh an Oifig an breithiúnas   sa Chlár agus glacfaidh sí na bearta is gá chun cloí lena \textcolor{red}{chuid} oibríochtúil. \\ \hline
\textbf{EN-3} &
The competent authority may at any time wholly or partially suspend or terminate the contract awarded under this provision if the operator \textcolor{blue}{fails} to meet the performance requirements. \\\hline
\textbf{GA-3} &
Féadfaidh an t-údarás inniúil an conradh a dámhadh faoin bhforáil seo a chur ar fionraí nó a fhoirceannadh go hiomlán nó go páirteach \textcolor{red}{m\'{a} mhainn\'{i}onn} an t-oibreoir na ceanglais feidhmíochta a chomhlíonadh. \\ \hline
\textbf{EN-4} &
This Directive shall enter into force on the day following that of its \textcolor{blue}{publication} in the Official Journal of the European Union. \\ \hline
\textbf{GA-4} &
Tiocfaidh an Treoir seo i bhfeidhm an lá tar éis lá \textcolor{red}{a fhoilsithe} in Iris Oifigiúil an Aontais Eorpaigh. \\ \hline
\textbf{EN-5} &
Such special measures are interim in nature, and \textcolor{blue}{shall not be} subject to the conditions set out in Article 7(1) and (2). \\ \hline
\textbf{GA-5} &
Tá bearta speisialta den sórt sin eatramhach, agus \textcolor{red}{ní bheidh said} faoi réir na gcoinníollacha a leagtar amach in Airteagal 7(1) agus (2) iad. \\\hline
\end{tabular}

\label{tab:ling-anal-lrev}
\end{table}

\subsubsection{Core Grammatical Errors}

Grammatical mistakes in the form of the misuse of lenitions (e.g., GA-4), incorrect pronouns (e.g., GA-5) and register errors (e.g., GA-5) were observed in both translation architectures. However, as is evident from both the automatic and MQM evaluations, there were far fewer errors with the Transformer model. Evidence of this can be seen in Table~\ref{tab:ling-anal-lrev}. In the case of GA-4, the RNN model included the lenition in ``a foilsithe'', whereas the Transformer model correctly removed ``h''. The correct use of the feminine noun ``treoir'' requires the removal of ``h'' in ``fhoilsithe''. 

The misuse of pronouns was observed in the RNN translation model and, to a lesser degree, in the Transformer model. In the case of GA-5, the RNN's incorrect use of the pronoun ``ní bheidh siad'' (they will not) is illustrated, whereas the Transformer approach used the correct form ``ní bheidh sé'' (he will not).  

Within the same sentence, GA-5, there is also evidence of a register error. In the English source text EN-5, the use of ``shall not be subject to'' expresses a stipulation. This is not registered in the Irish translation of ``ní bheidh said'', which simply, and less forcefully, means ``they will not''. This incorrect use of register was observed with both the RNN and the Transformer approaches. A more formal and closer interpretation of the English source would be the use of the imperative mode: ``ná bídís'' (let it not be).

\subsubsection{Commonly-Used Irregular Verbs}

One of the main inadequacies observed in both the RNN and Transformer systems is a lack of refinement of verbal usage, particularly when using the verbs ``déan'' (to do or to make ) and ``bí'' (to be). As in many languages, the fact that these are possibly the two most universally used verbs in Irish further exacerbates the problem. An illustration of this problem can be seen in the output GA-1, which highlights the incorrect usage of  ``déan''. Similarly, GA-5 demonstrates how the system misinterprets the usage of the verb ``bí'', e.g., ``ní bheidh said''.

\subsubsection{Performance of RNN Approach Relative to Transformer Approach}

There is a strong correlation between automatic and human evaluation of the translation systems that we developed. The automatic BLEU scores are contrasted with the human evaluation scores for both the RNN and Transformer models in Table~\ref{tab:all-metrics}. 

\begin{table}[h]
\small
\setlength{\tabcolsep}{10.85mm}

\caption[Transformer approach compared to the RNN approach]{Transformer approach compared to the RNN approach across all metrics for the DGT dataset. The results from our human evaluation, using SQM and MQM metrics, validate the BLEU automatic evaluation results.}
\begin{tabular}{@{}llccc@{}}
\hline
\textbf{Approach}    &  & \textbf{BLEU} \boldmath{$\uparrow$} & \textbf{SQM} \boldmath{$\uparrow$} & \textbf{MQM} \boldmath{$\uparrow$} \\ \hline
\textbf{Transformer} &  & 60.5          & 4.53         & 77.92        \\
\textbf{RNN}         &  & 52.7          & 3.30         & 43.05        \\ \hline
\end{tabular}

\label{tab:all-metrics}
\end{table}

\subsection{Limitations of the Study}
Certain aspects of this study could be further developed, given more time and resources. Although there is a high inter-annotator agreement, it would help to have more annotators. In addition, the human evaluation of a greater number of lines, coupled with a more detailed MQM taxonomy, may provide greater insight into the MT outputs. This would help in uncovering other aspects, such as how gender is handled by the MT models.

\section{Conclusions and Future Work\label{s7}}

With this research, we have presented the first human evaluation study that compares the output of EN$\rightarrow$GA RNN systems with that of Transformer-based EN$\rightarrow$GA systems. Automatic metrics were shown to differentiate the systems and highlighted that Transformer models are superior to RNN models. In our paper, we demonstrated that a random search approach to HPO enabled the development of high-performing translation models. We have shown there is a high level of correlation between a human evaluation and an automatic approach. Both the automatic metrics and our human evaluation demonstrated that the Transformer-based system is the most accurate. 

The importance of selecting hyperparameters when training low-resource Transformer models was also demonstrated. By increasing dropout and reducing the number of hidden-layer neurons, our models performed significantly better than Google Translate and our baseline models.

We have demonstrated that choosing the correct subword models is an important performance driver for low-resource MT. Within the context of low-resource EN$\rightarrow$GA translations, we achieved optimal performance on a 55k generic corpus when a Transformer architecture with a 16k BPE subword model was used. Improvements in the performance of our optimised Transformer models were observed across all key indicators, namely, PPL was achieved at a lower global minimum, with a lower post-editing effort and a higher translation accuracy.

As part of future work, steps can be taken to deal with the inadequacies highlighted in our linguistic analysis. The issue of misusing common irregular verbs could be addressed by fine-tuning our models with a dataset specifically tailored for that purpose. Similarly, fine-tuning after the careful selection of training data would also reduce the register errors encountered in our linguistic analysis. As it is difficult to train systems for all eventualities, using post-editing tools would be the best approach to correcting core grammatical errors involving pronouns, lenitions and lemmatization.

\authorcontributions{All authors have read and agreed to the published version of the manuscript.}

\funding{This work was supported by ADAPT, which is funded under the SFI Research Centres Programme (Grant 13/RC/2016) and is co-funded by the European Regional Development Fund. This research was also funded by the Munster Technological University.}

\informedconsent{Informed consent was obtained from all subjects involved in the study.}

\dataavailability{The data presented in this study are openly available at \url{https://github.com/seamusl/isfeidirlinn} }

\acknowledgments{We would like to thank the annotators, Dr Éamon Lankford and Ms. Máirín Lankford for their meticulous work in annotating the system outputs. }

\conflictsofinterest{ The authors declare no conflict of interest. The funders had no role in the design of the study; in the collection, analyses, or interpretation of data; in the writing of the manuscript, or in the decision to publish the results.} 

\abbreviations{Glossary}{Irish terms referenced and used in this manuscript:\\
\noindent 
\begin{tabular}{@{}ll}
Déan & To do or to make\\
Bí & To be \\ 
Ná bídís & Let it not be\\
Ní bheidh siad & They will not\\
Ní bheidh sé & He will not \\
\end{tabular}}

\chapter{Design of an Open-Source Architecture for NMT}
\section{Context}

With RQ4, the process of how NMT development, evaluation, and deployment could be streamlined for both developers and translators was considered. The purpose of this short paper for the inaugural CrowdMT workshop at Tampere was to showcase the adaptNMT application \parencite{lankford-etal-2023-design}. As a workshop paper, this chapter is effectively a compressed version of the longer and more comprehensive journal paper outlined in Chapter 7. Readers who wish for an in-depth discussion of the adaptNMT application are encouraged to go directly to Chapter 7, adaptNMT: Open-Source Neural Machine Translation. 

By simplifying NMT model development, building and deploying NMT models can be streamlined. This approach recognises that newcomers to the field face challenges in setting up the development environment and creating the necessary data splits for training, validation, and testing. The community is invited to contribute new ideas and improvements, and a spirit of collaboration will be fostered among researchers. The research contribution is to provide an open-source approach which can lead to the continuous improvement and evolution of the adaptNMT project.

\clearpage

   \begin{center}
       \vspace*{1cm}

       \textbf{Design of an Open-Source Architecture for Neural Machine Translation}
            
       \vspace{1.5cm}

       \textbf{Séamus Lankford \\ Haithem Afli \\ Andy Way}

       \vfill

crowdMT23: Workshop on Open Community-Driven Machine Translation \\ 
June 15, 2023\\ 
Tampere, Finland.
       \vspace{0.8cm}
                 
       ADAPT Centre\\
       Dublin City University\\
       Ireland\\

        \vspace{0.5cm}
    \url{https://aclanthology.org/2023.crowdmt-1.2.pdf}            
   \end{center}

\clearpage

\section{Abstract}
 
adaptNMT is an open-source application that offers a streamlined approach to the development and deployment of recurrent neural networks and Transformer models. This application is built upon the widely-adopted OpenNMT ecosystem, and is particularly useful for new entrants to the field, as it simplifies the setup of the development environment and creation of training, validation, and test splits. The application offers a graphing feature that illustrates the progress of model training and employs SentencePiece for creating subword segmentation models. Furthermore, the application provides an intuitive user interface that facilitates hyperparameter customisation. Notably, a single-click model development approach has been implemented, and models developed by adaptNMT can be evaluated using a range of metrics. To encourage eco-friendly research, adaptNMT incorporates a green report that flags the power consumption and kgCO\textsubscript2 emissions generated during model development. The application is freely available.\footnote{\url{http://github.com/adaptNMT}}

\section{Credits}
This research is supported by Science Foundation Ireland through the ADAPT Centre (Grant 13/RC/2106) (www.adaptcentre.ie) at Dublin City University. This research was also funded by the Munster Technological University.

\section{Introduction}\label{sec1_he}

Explainable Artificial Intelligence (XAI) \parencite{arrieta2020explainable} aims to ensure the outcomes of AI solutions are easily comprehensible to humans. In light of this goal, adaptNMT has been developed to provide users with a form of explainable neural machine translation (XNMT). The typical neural machine translation (NMT) process comprises several independent stages, including setting up the environment, preparing the dataset, training subword models, setting the parameters and training the main models, evaluating and deploying them. By adopting a modular approach, this framework has established an effective NMT model development process that caters to both technical and non-technical practitioners in the field. To address the environmental impact of building and running large AI models \parencite{henderson2020towards,jooste-etal-2022-knowledge}, we have also produced a ``green report'' that calculates carbon emissions. While primarily intended as an information aid, this report will hopefully encourage the development of reusable and sustainable models.

This research endeavours to create models and applications that address the challenges of language technology, which will be particularly beneficial for those new to the field of Machine Translation (MT) and those seeking to learn more about NMT.

The application is built on OpenNMT\footnote{\url{https://opennmt.net}}~\parencite{klein2017opennmt} and thus inherits all of its features. Unlike many NMT toolkits, a command line interface is not used, and the interface is designed and fully implemented in Google Colab.\footnote{\url{https://colab.research.google.com}} For both educational and research purposes, a cloud-hosted solution like Colab is often more user-friendly. Additionally, the training of models can be monitored and controlled on a mobile phone using Google Colab's responsive design, which is useful for long-run builds. The adaptNMT framework also includes GUI controls that allow for the customisation of all crucial parameters needed for NMT model training.

The application can be run in local mode to utilise existing infrastructure or hosted mode for rapid infrastructure scaling. A deploy function is also included to allow for the immediate deployment of trained models.

This paper begins by presenting background information on NMT and NMT tools in Section \ref{related-finland}, followed by a detailed description of the adaptNMT architecture and its key features in Section \ref{arch-finland}. The system is discussed in Section \ref{disc-finland} before concluding with a discussion of future work in Section \ref{concl-finland}. A more in-depth system description, coupled with an empirical evaluation
of models developed using the application, is outlined in a separate paper \parencite{lankfordlrev}. 

\section{Related Work}\label{related-finland}

\subsection{NMT}\label{subsec2_he}
In addition to the ongoing research dedicated to developing state-of-the-art (SOTA) NMT models, comprehensive descriptions of this technology are readily available in the literature, making it accessible to individuals who are new to the field or have limited technical expertise \parencite{WayBlooms}.

NMT has benefited from the availability of large parallel corpora, leading to the development of high-performing MT models. The field of MT has experienced significant advancements through the application of NMT, particularly after the introduction of the Transformer \parencite{vaswani2017attention} architecture, which has resulted in SOTA performance across multiple language pairs \parencite{bojar-etal-2017-findings,bojar-etal-2018-findings,lankford2021transformer, lankford2021machine,lankford2022lrec,lankford2022human}.

\subsection{NMT Tools}\label{tools}

In essence, adaptNMT is an IPython wrapper built on OpenNMT, enabling it to benefit from OpenNMT's extensive feature set and continuous code maintenance. However, adaptNMT takes abstraction to a higher level than OpenNMT, with a greater focus on usability, particularly for newcomers. As a result, adaptNMT facilitates easy and fast deployment, offering features such as more pre-processing, as well as GUI control over model creation. Moreover, it incorporates green features in line with current research efforts towards smaller models with reduced carbon footprints, making it suitable for educational and research environments alike.

Other commonly used frameworks for developing NMT systems include FAIRSEQ\footnote{\url{https://github.com/facebookresearch/fairseq}} \parencite{ott2019fairseq}, an open-source sequence modelling toolkit based on PyTorch that allows for training models for translation, summarisation, and language modelling. Marian\footnote{\url{https://marian-nmt.github.io}}~\parencite{junczys2018marian}, on the other hand, is an NMT framework based on dynamic computation graphs and developed using C++. OpenNMT is an open-source NMT framework that has been widely adopted in the research community and covers the entire MT workflow from data preparation to live inference.

\section{Architecture of adaptNMT}\label{arch-finland}

\begin{figure*}[ht]
    \centering
    \includegraphics[width=15cm]{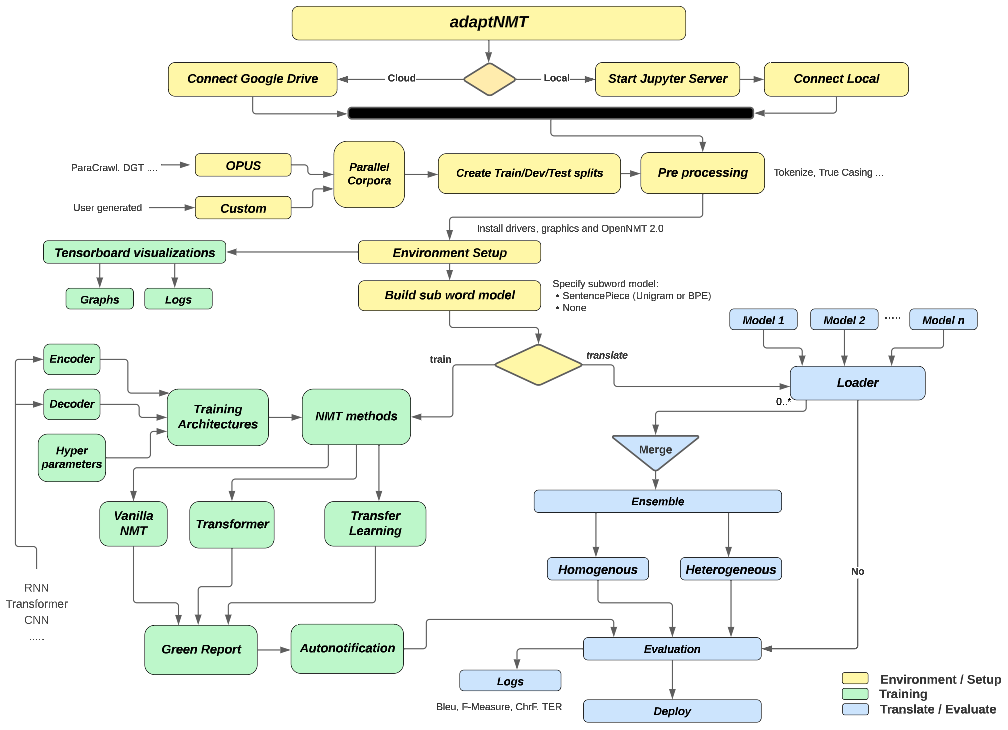}
    \caption[Proposed architecture for adaptNMT]{Proposed architecture for adaptNMT: a language-agnostic NMT development environment. The system is designed to run either in the cloud or using local infrastructure. Models are trained using parallel corpora. Visualisation and extensive logging enable real-time monitoring. Models are developed using vanilla RNN-based NMT, Transformer-based approaches or transfer learning using a fine-tuning approach. Translation and evaluation can be carried out using either single models or ensembles.}
    \label{fig:approach}
\end{figure*}

After providing a general overview of NMT and NMT development systems, we introduce the adaptNMT tool, which enables users to configure the components of the NMT development process. The platform's system architecture is depicted in Figure \ref{fig:approach}. The tool is built as an IPython notebook and leverages the Pytorch implementation of OpenNMT for training models. Additionally, SentencePiece is used to train subword models. Using a Jupyter notebook facilitates sharing the application with other members of the MT community, and the application's setup is simplified since all necessary packages are downloaded dynamically as the application runs.

The system has two deployment options: running it locally or as a Colab instance via Google Cloud. To build translation models, the system requires parallel text corpora for both the source and target languages. A Tensorboard visualization allows for real-time monitoring of the model training process. At run time, users can select to use the system for either model building or translation services, or both. Additionally, as depicted in Figure \ref{fig:approach}, the system enables the generation of an ensemble output during translation. Finally, trained models can be easily deployed to a pre-configured location.

\subsection{adaptNMT}\label{aNMT_he} 
The application may be run as an IPython Jupyter notebook or as a Google Colab application. Given the ease of integrating large Google Drive storage into Colab, the application has been used exclusively as a Google Colab application for our experiments. 

\subsubsection{Initialisation and Logging}

Initialisation enables connection to Google Drive to run experiments, automatic installation of Python, OpenNMT,\footnote{\url{https://opennmt.net}} SentencePiece,\footnote{\url{ https://github.com/google/sentencepiece}} Pytorch and other applications. The visualisation section enables real-time graphing of model development. All log files are stored and can be viewed to inspect training convergence, the model’s training and validation accuracy and changes in learning rates.

\subsubsection{Modes of Operation}
There are two modes of operation: local and cloud. In local mode, the application is run so that models are built using the user's local GPU resources. The option to use cloud mode enables users to develop models using Google's GPU clusters. For shorter training times, the unpaid Colab option is adequate. However, for a small monthly subscription, the Google Colab Pro option is worthwhile since users have access to improved GPU and compute resources. Furthermore, using Google Cloud may be considered as the ``green option'' since its platform uses 100\% renewables \parencite{lacoste2019quantifying}. 

\subsubsection{Customisation of Models}
The system has been developed to allow users to select variations to the underlying model architecture. A vanilla RNN or Transformer approach may be selected to develop the NMT model. The customisation mode enables users to specify the exact parameters required for the chosen approach. One of the features, AutoBuild, enables a user to build an NMT model in three simple steps: (i) upload source and target files, (ii) select  RNN or Transformer, and (iii) click AutoBuild.

\subsubsection{Use of Subword Segmentation}
In the NMT development process, users can specify the type of optimiser for learning and choose from different subword models. The subword model functionality allows for the selection of a subword model type and the choice of vocabulary size, currently offering either a SentencePiece unigram or a SentencePiece BPE model.

A user may upload a dataset which includes the train, validation and test splits for both source and target languages. In cases where a user has not already created the required splits for model training, single source and target files may be uploaded. Automated splitting of the uploaded dataset into the train, validation, and test files is then performed based on the user's chosen split ratio. 

Given that building NMT models typically demands long training times, an automatic notification feature is incorporated that informs the user by email when model training has been completed.

\subsubsection{Translation and Evaluation}

The application supports not only the training of models but also the translation and evaluation of model performance. For translation using pre-built models, users can specify the model name as a hyperparameter which is subsequently used to translate and evaluate the test files. The option for creating an ensemble output is also available, with users simply naming the models to be used in generating the ensemble output.

Once the system has been built, the user can select the model to be used for translating the test set. While human evaluation is often considered the most insightful approach for evaluating translation quality, it can be limited by factors such as availability, cost, and subjectivity. Thus, automatic evaluation metrics are frequently employed, particularly by developers monitoring the incremental progress of systems. A further discussion on the advantages and disadvantages of human and automatic evaluation is available in the literature \parencite{Way2018}.

Several automatic evaluation metrics provided by SacreBleu\footnote{\url{https://github.com/mjpost/sacrebleu}} \parencite{post2018call} are used: BLEU \parencite{papineni-etal-2002-bleu}, TER \parencite{snover2006study} and ChrF \parencite{popovic2015chrf}. Translation quality can also be evaluated using Meteor~\parencite{denkowski2014meteor} and F1 score~\parencite{melamed-etal-2003-precision}. Note that BLEU, ChrF, Meteor and F1 are precision-based metrics, so higher scores are better, whereas TER is an error-based metric and lower scores indicate better translation quality. Evaluation options available include standard (truecase) and lowercase BLEU scores, a sentence-level BLEU score option, ChrF1 and ChrF3.  

There are three levels of logging: model development logs for graphing, training console output and experimental results. A references section outlines resources which are relevant to developing, using and understanding adaptNMT. Validation during training is currently conducted using model accuracy and perplexity (PPL). 

\subsection{Infrastructure}
Rapid prototype development is possible through a Google Colab Pro subscription using NVIDIA Tesla P100 PCIe 16GB graphic cards and up to 27GB of memory when available. 

\section{Discussion}\label{disc-finland}

Numerous tools have been developed to assess the carbon footprint of NLP \parencite{bannour-etal-2021-evaluating}. The notion of sustainable NLP has also gained momentum as an independent research track, with high-profile conferences such as the \textit{EACL 2021 Green and Sustainable NLP} track dedicating resources to this area.\footnote{\url{https://2021.eacl.org/news/green-and-sustainable-nlp}} 

Given these developments, we have incorporated a ``green report'' into adaptNMT that logs the kgCO\textsubscript2 generated during model development. This aligns with the industry's increasing focus on quantifying the environmental impact of NLP. In fact, it has been demonstrated that high-performing MT systems can be developed with much lower carbon footprints, leading to significant energy cost savings for a real translation company \parencite{info13020088}.

The risks associated with relying on large language models (LLMs) have been well-documented in the literature. The discussion surrounding these models emphasises not only their environmental impact but also the inherent biases and dangers they pose for low-resource languages \parencite{bender2021dangers}. It is important to note that smaller, in-domain datasets can yield high-performing NMT models, and the adaptNMT framework makes this approach easily accessible and understandable.

\section{Conclusion and Future Work}\label{concl-finland}

We have introduced adaptNMT, an application that manages the entire NMT model development, evaluation, and deployment workflow.

As for future work, our development efforts will be directed towards incorporating new transfer learning methods and improving our ability to track environmental costs. We will integrate modern zero-shot and few-shot approaches, as seen in the GPT3 \parencite{brown2020language} and Facebook LASER \parencite{artetxe2019massively} frameworks. While the existing adaptNMT application is focused on customising NMT models, we will also develop a separate application, adaptMLLM \parencite{lankford2023adaptLLM}, for fine-tuning multilingual language models (MLLMs) and LLMs. In particular, adaptMLLM will cater for low-resource language pairs covered by Meta's NLLB \parencite{costa2022no} models.

The green report integrated into the application represents our first implementation of a sustainable NLP feature within adaptNMT. We plan to enhance this feature by improving the user interface and providing recommendations on how to develop greener models. As an open-source project, we invite the community to contribute new ideas and improvements to the development of this feature.

\chapter{adaptNMT: Open-Source Neural Machine Translation}
\section{Context}

The motivation for the research on adaptNMT is to address several key challenges in the field of NMT and NLP in general. Overall, the research behind adaptNMT addresses critical issues in NMT development and deployment, while also promoting sustainable and collaborative practices within the NLP community.

The primary motivations for this research were initially raised in RQ4 and are outlined in more detail below:

\begin{itemize}
  
\item \textbf{Streamlining Development and Deployment Processes}: The primary goal of adaptNMT is to simplify and streamline all aspects of developing and deploying RNN and Transformer neural translation models. This includes processes like setting up the development environment, creating data splits, and training the models.

\item  \textbf{Accessibility for Both Developers and Translators}: The application is designed to be user-friendly for individuals with varying levels of technical expertise. This inclusivity aims to make MT more accessible to a wider audience, including those who may not have extensive technical backgrounds.

\item \textbf{Support for New Entrants in the Field}: adaptNMT is particularly valuable for newcomers to the field of MT. By simplifying the setup and training process, it lowers the barrier to entry, allowing more individuals to participate in NLP research and development.

\item \textbf{Built Upon the OpenNMT Ecosystem}: By leveraging the widely adopted OpenNMT ecosystem, adaptNMT benefits from a strong foundation of established tools and resources in the NMT community.

\item \textbf{Visualisation and Monitoring of Model Training}: The application provides visualisations to track the progress of model training. This feature aids researchers and developers in understanding how their models are evolving during the training process.

\item \textbf{Utilisation of SentencePiece for Subword Segmentation Models}: SentencePiece is employed to create subword segmentation models. This helps in handling morphologically rich languages and improving the overall performance of the translation models.

\item \textbf{Intuitive Hyperparameter Customisation}: The application offers an intuitive user interface (UI) for customising hyperparameters. This facilitates experimentation with different settings to optimise model performance.

\item \textbf{Efficient Model Development with Single-Click Approach}: The one-click model development approach simplifies and accelerates the process of creating and fine-tuning translation models.

\item \textbf{Comprehensive Model Evaluation and Deployment}: Models developed with adaptNMT can be rigorously evaluated using a range of metrics. Additionally, the application allows for easy deployment of these models as a translation service.

\item \textbf{Environmental Considerations}: A strong emphasis has been placed on eco-friendly practices in NLP research. A ``green report'' feature that tracks the power consumption and carbon emissions generated has been implemented which promotes sustainable research practices.

\item \textbf{Demonstrated Success in Shared Tasks}: The performance of adaptNMT was validated by generating an EN$\rightarrow$GA translation model, which achieved 1st place in the LoResMT2021 shared task. This success demonstrates the practical effectiveness of the application.

\item \textbf{Expanding to Fine-Tuning Large Language Models}: While adaptNMT focuses on customising NMT models, a separate application, adaptMLLM, has been developed to fine-tune MLLMs and LLMs. This is particularly important for low-resource language pairs.

\end{itemize}

\clearpage

   \begin{center}
       \vspace*{1cm}

       \textbf{adaptNMT: an open-source,
language-agnostic development environment for Neural MT }
            
       \vspace{1.5cm}

       \textbf{Séamus Lankford \\ Haithem Afli \\ Andy Way}

       \vfill
            
    Journal of Language Resources Evaluation, Springer \\
             March 1st 2023

       \vspace{0.8cm}
                 
       ADAPT Centre \\
       Dublin City University \\
       Ireland \\

 \vspace{0.5cm}
    \url{https://link.springer.com/article/10.1007/s10579-023-09671-2}   
   \end{center}

\clearpage

\section{Abstract}
adaptNMT streamlines all processes involved in the development and deployment of RNN and Transformer neural translation models. As an open-source application, it is designed for both technical and non-technical users who work in the field of MT. Built upon the widely-adopted OpenNMT ecosystem, the application is particularly useful for new entrants to the field since the setup of the development environment and creation of training, validation and test splits is greatly simplified. Graphing, embedded within the application, illustrates the progress of model training, and SentencePiece is used for creating subword segmentation models. Hyperparameter customisation is facilitated through an intuitive user interface, and a single-click model development approach has been implemented. Models developed by adaptNMT can be evaluated using a range of metrics and deployed as a translation service within the application. To support eco-friendly research in the NLP space, a green report also flags the power consumption and kgCO\textsubscript2 emissions generated during model development. The application is freely available.\footnote{\url{http://github.com/adaptNMT}}

\clearpage

\section{Graphical abstract}

\begin{figure}[htp]
  \includegraphics[width=15.5cm]{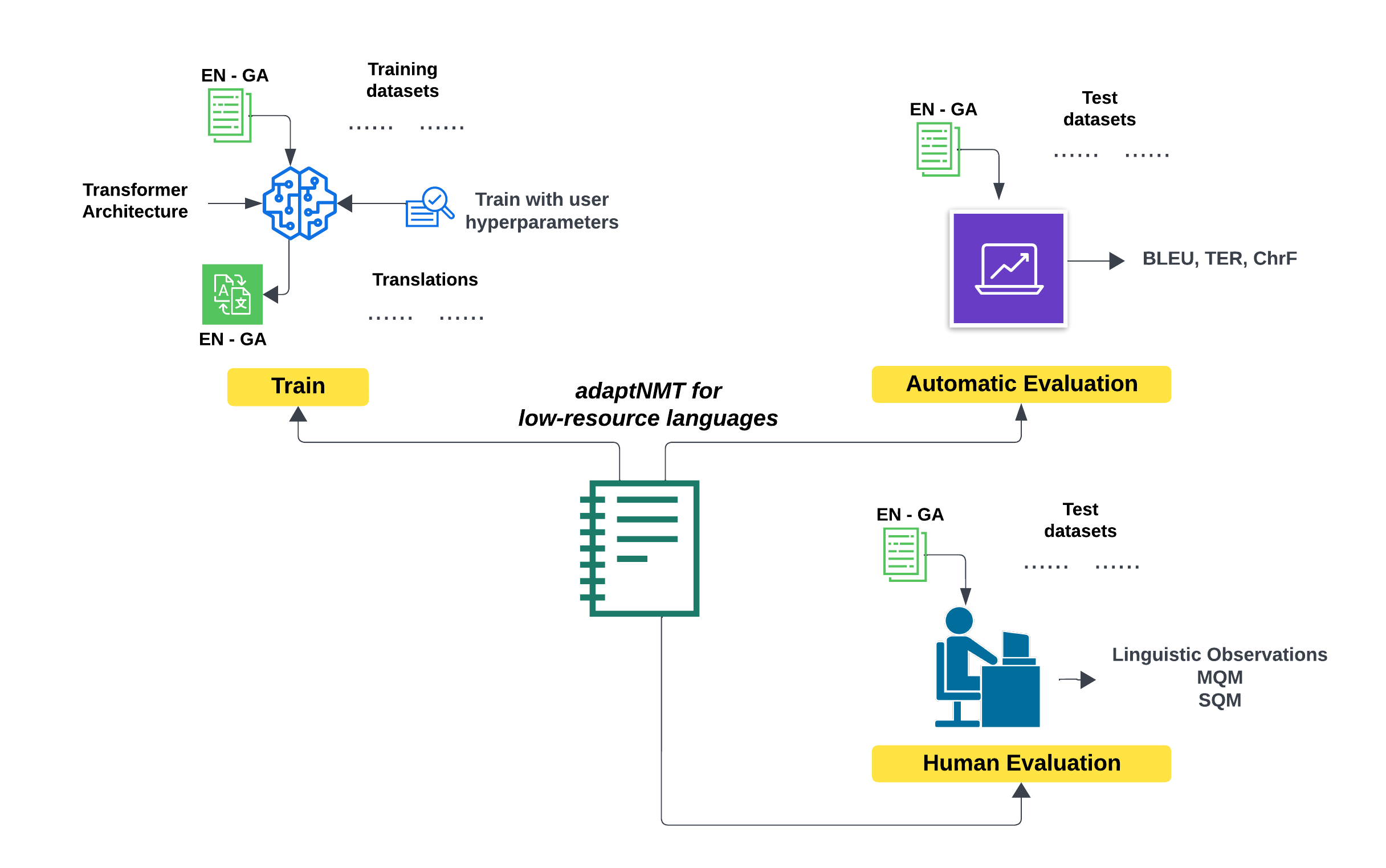}
  \caption{Graphical abstract summarising the adaptNMT system}
 \label{fig:graph_arch_nmt}
\end{figure}

\section{Introduction}\label{sec1_nmt_mdpi}

Explainable artificial intelligence (XAI) \parencite{gunning2019xai, arrieta2020explainable} seeks to ensure that the results of AI solutions are easily understood by humans. It is against this backdrop that adaptNMT has been developed to afford users a form of {\em explainable neural machine translation (XNMT)}. The stages involved in a typical neural machine translation (NMT) process are broken down into a series of independent steps including environment setup, dataset preparation, training of subword models, parameterising and training of main models, evaluation and deployment. This modular approach has created an effective NMT model development process for both technical and less technical practitioners in the field. Given the environmental impact of building and running large AI models \parencite{strubell-etal-2019-energy,henderson2020towards,jooste-etal-2022-knowledge}, we also compute carbon emissions in a ``green report'', primarily as an information aid, but hopefully as a way to encourage reusable and sustainable model development.

An important part of this research involves developing applications and models to address the challenges of language technology. It is hoped that such work will be of particular benefit to newcomers to the field of machine translation (MT) and in particular to those who wish to learn more about NMT.

To have a thorough understanding of how NMT models are trained, the individual components and the mathematical concepts underpinning both RNN- and Transformer-based models are explained and illustrated in this paper. The application is built upon OpenNMT \parencite{klein2017opennmt} and subsequently inherits all of its features. Unlike many NMT toolkits, a command line interface approach is not used. The interface is designed and fully implemented in Google Colab.\footnote{\url{https://colab.research.google.com}} For an educational setting, and indeed for research practitioners, a Colab cloud-hosted\footnote{\url{https://cloud.google.com}} solution is often more intuitive to use. Furthermore, the training of models can be viewed and controlled using the Google Colab responsive website which is ideal for builds with long run times. GUI controls, also implemented within adaptNMT, enable the customisation of all key parameters required when training NMT models.

The application can be run in local mode enabling existing infrastructure to be utilised, or in hosted mode which allows for rapid scaling of the infrastructure. A deploy function allows for the immediate deployment of trained models. 

This paper is organised by initially presenting background information on NMT and related work on system-building environments in Section \ref{nmt-detail}. This is followed by a detailed description of the adaptNMT architecture and its key features in Section \ref{arch}. An empirical evaluation of models is carried out in Section \ref{sec:exp-nmt}. The system is discussed in Section \ref{disc-nmt} before drawing conclusions and describing future work in Section \ref{concl-nmt}. For newcomers to the field, we suggest going straight to Section \ref{arch} to examine the platform's capabilities, and then discovering more about the various components and their statistical underpinning in Section \ref{nmt-detail}. This can be followed by the remaining sections in their logical sequence. 

\section{Neural Networks for MT}\label{nmt-detail}

\subsection{Recurrent Neural Network Architecture}
 
Recurrent neural networks (RNNs)~\parencite{sennrich2016edinburgh, sennrich2019revisiting, araabi2020optimizing} are often used for the tasks of natural language processing (NLP), speech recognition and MT. RNNs, such as long short-term memory (LSTM) \parencite{hochreiter1997long}, were designed to support sequences of input data. LSTM models use an encoder-decoder architecture which enables variable length input sequences to predict variable length output sequences. This architecture is the cornerstone of many complex sequence prediction problems such as speech recognition and MT.

RNN models enable previous outputs to be used as inputs through the use of hidden states. In the context of MT, such neural networks were ideal due to their ability to process inputs of any length. In the initial stages of NMT, the RNN encoder-decoder framework was adopted and variable-length source sentences were encoded as fixed-length vectors~\parencite{cho2014properties, sutskever2014sequence}. An improvement upon the basic RNN approach was proposed in \parencite{bahdanau2014neural} which enhanced translation performance of the basic encoder-decoder architecture by replacing fixed-length vectors with variable-length vectors. A bidirectional RNN was now employed to read input sentences in the forward direction to produce forward hidden states while also producing backward hidden states by reading input sentences in the reverse direction. This development enabled neural networks to more accurately process long sentences, which previously had served as bottlenecks to performance, given their tendency to `forget' words in long input sequences which are `too far away' from the current word being processed. 

More importantly, \parencite{bahdanau2014neural} introduced the concept of `attention' to the basic RNN architecture, similar in spirit and intention to `alignments' in the forerunner to NMT, statistical MT \parencite{och-ney-2003-systematic}. In attention-augmented NMT, the system could now pay special heed to the most relevant other source-sentence words and use them as contextual clues when considering how best to select the most appropriate target words for translationally ambiguous words in the same string.

\subsection{Transformer Architecture}

Following the introduction of the attention mechanism, a natural line of investigation was to see whether attention could do most of the heavy lifting of translation by itself. Accordingly, \parencite{vaswani2017attention} proposed that  “attention is all you need” in their `Transformer architecture', which has achieved state-of-the-art (SOTA) performance on many NLP benchmarks by relying solely on an attention mechanism,  removing recurrence and convolution while allowing the use of much simpler feed-forward neural networks.

\begin{figure}[htb]
    \centering
    \includegraphics[width=12cm]{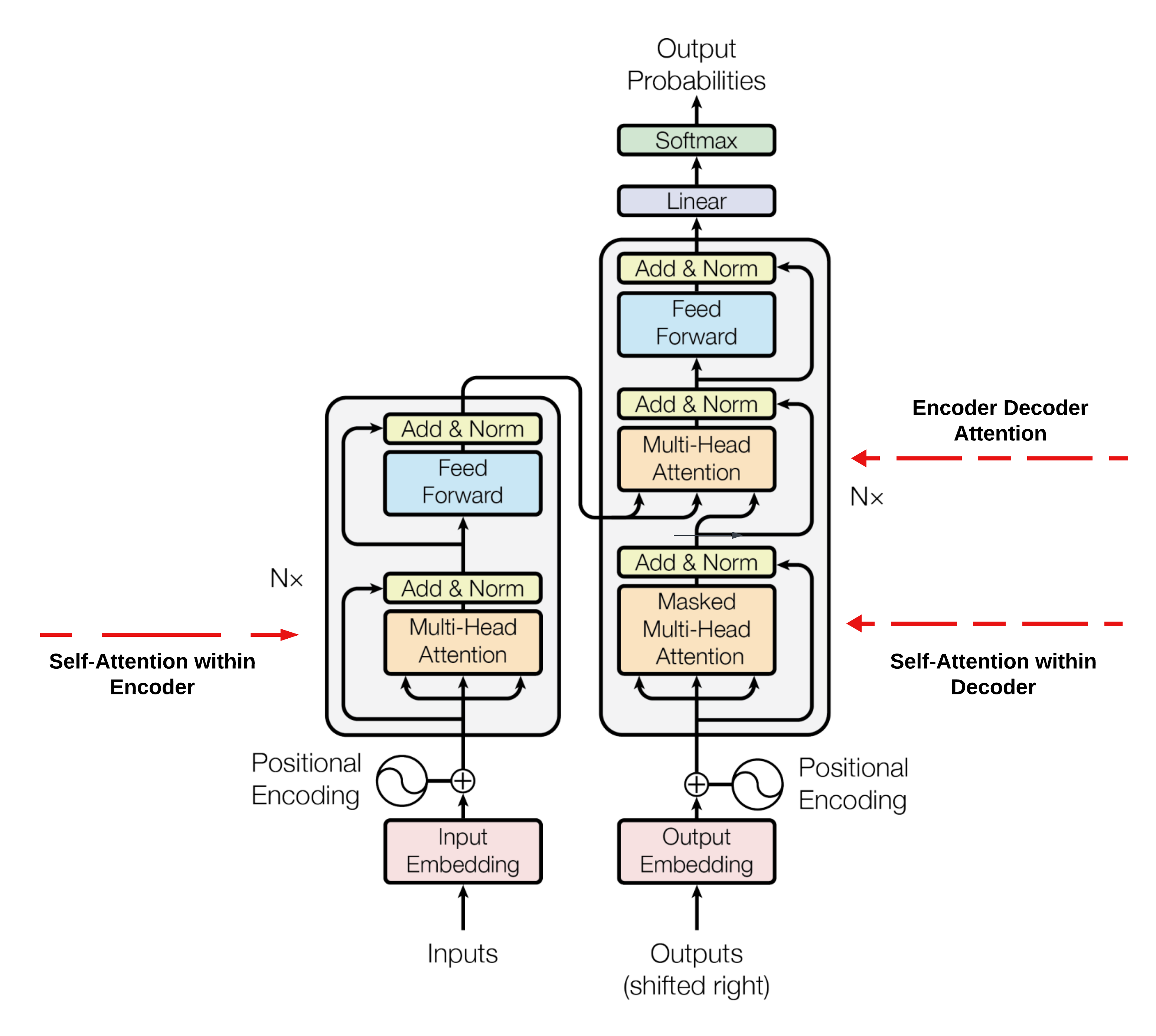}
    \caption[Transformer architecture using an encoder-decoder]{The Transformer architecture using an encoder-decoder~\parencite{vaswani2017attention}. The encoder maps an input sequence to the decoder. The decoder generates a new output by combining the encoder output with the decoder output from the previous step.}
    \label{fig:translayers}
\end{figure}

This approach follows an encoder-decoder structure and allows models to develop a long memory which is particularly useful in the area of language translation. The task of the encoder is to map an input sequence to a sequence of continuous representations, which is then passed to a decoder to generate an output sequence by using the output of the encoder together with the decoder output from the previous time step. Both the encoder and decoder each consist of a stack of 6 identical layers, whose structure is illustrated in Figure \ref{fig:translayers}. In the encoder, each layer is composed of two sub-layers: a multi-head self-attention mechanism and a fully connected feed-forward network. In the case of the decoder, there are three sub-layers: one which takes the previous output of the decoder stack, another which implements a multi-head self-attention mechanism, and the final layer which implements a fully connected feed-forward network.

\subsection{Attention}

\begin{figure}[h!]
    \centering
    \includegraphics[width=2.5cm]{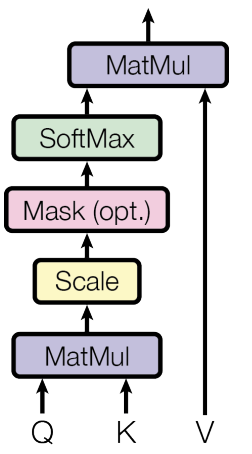}
    \caption[Multi-head attention in the decoder]{Multi-head attention in the decoder~\parencite{vaswani2017attention}. In the decoder, a multi-head layer receives queries from the previous decoder sublayer, and the keys and values from the encoder output. The decoder can now attend to all words in the input sequence.}
    \label{fig:attention}
\end{figure}

As illustrated in Figure \ref{fig:attention}, the attention function can be described as mapping a query and a set of key-value pairs to an output, where the query, keys, values, and output are all vectors. The output is computed as a weighted sum of the values, where the weight assigned to each value is computed by a compatibility function of the query with the corresponding key, as shown in Equation \ref{eqn8}.

The query, keys and values used as inputs to the attention mechanism are different projections of the same input sentence (`self-attention') and capture the relationships between the different words of the same sentence.

Both a scaled dot-product attention and a multi-head attention are used in the Transformer architecture. With scaled dot-product attention, a dot product is initially computed for each query $q$ with all of the keys $k$. The input consists of queries and keys of dimension $d_k$. Subsequently, each result is divided by $\sqrt{d_k}$ and a Softmax function is applied. The process leads to the weights which are used to scale the values, $v$.

The Softmax function allows us to perform multiclass classification which makes it a good choice in the final layer of neural network-based classifiers. The function forces the outputs of the neural network to a total sum to 1, which can be viewed as a probability distribution across multiple classes. Therefore, Softmax is the ideal choice as the output activation function, given that NMT is essentially a multiclass classification problem where the output classes represent the words within the vocabulary.

Computations performed by scaled dot-product attention can be efficiently applied to the entire set of queries simultaneously. To achieve this, the matrices, $Q$, $K$ and $V$, are supplied as inputs to the attention function:

\begin{equation}\label{eqn8}
attention(Q,K,V)=softmax(QK^T/\sqrt{d_k})V
\end{equation}

\subsection{NMT}\label{subsec2_nmt_mdpi}
While much research effort concentrates on creating new SOTA NMT models, excellent descriptions of the technology are also available within the literature for those starting out in the field, or for those with a less technical background \parencite{forcada2017making,WayBlooms}.

The availability of large parallel corpora has enabled NMT to develop high-performing MT models. Breakthrough performance improvements in the area of MT have been achieved through research efforts focusing on NMT~\parencite{bahdanau2014neural} but the advent of the Transformer architecture has greatly improved MT performance. Consequently, SOTA performance has been attained on multiple language pairs~\parencite{bojar-etal-2017-findings, bojar-etal-2018-findings, lankford2021transformer, lankford2022human, lankford2022lrec}. 

Similar to many deep-learning approaches, NMT development is underpinned by the mathematics of probability. At a fundamental level, the goal is to predict the probabilistic distribution $P (y\vert x)$ given a dataset $D$, where $x$ represents the source input sentence and $y$ represents the target output sentence. 

Supervised training of an NMT model develops the model weights by comparing the predicted $P (y\vert x)$ with the correct $y$ sentences of the training dataset, $D\textsubscript{Train}$. In evaluating the performance of an NMT model, automatic evaluation results are determined when the predicted $P (y\vert x)$ sentences are compared with the correct $y$ sentences of the test dataset, $D\textsubscript{Test}$. 

In adopting a deep learning paradigm, MT inherits the mathematical first principles which are inherent to this approach. To understand these principles, the manner in which neural networks model a conditional distribution is outlined. Furthermore, the encoder-decoder mechanism used for training NMT models is presented in the modelling subsection, and model optimisation using training objectives is outlined in the learning subsection. Finally, the mathematics of how translated sentences are generated is explored in the inference subsection.

\subsubsection{Modelling}
In NMT, sentence-level translation is modelled using input and output sentences as sequences. Using this approach, an NMT model implements a sequence-to-sequence model with a given source sentence,  $x = (x_{1},...,x_{s})$ generating a target sentence $y = (y_{1},...,y_{t})$.

In effect, such a sequence-to-sequence NMT model acts as a conditional language model. The decoder within the model predicts the next word of the target sentence $y$, while such predictions are conditioned on the source sentence $x$.

By applying the chain rule, a model’s prediction (i.e. translation $y$ of length $T$) maximises the probability $P (y\vert x)$ identified in Equations \ref{eqn1} and \ref{eqn2}:

\begin{equation} \label{eqn1}
P(y\vert x)=P(y_1\vert x)P(y_2\vert y_1,x)P(y_3\vert y_1, y_2,x)P(y_T\vert y_1,...,y_{T-1},x)
\end{equation}

\begin{equation} \label{eqn2}
P(y\vert x)=\prod_{t=1}^{T}P(y_t\vert y_1,...,y_{T-1},x)  
\end{equation}

\begin{figure}[htb]
    \centering
    \includegraphics[width=8cm]{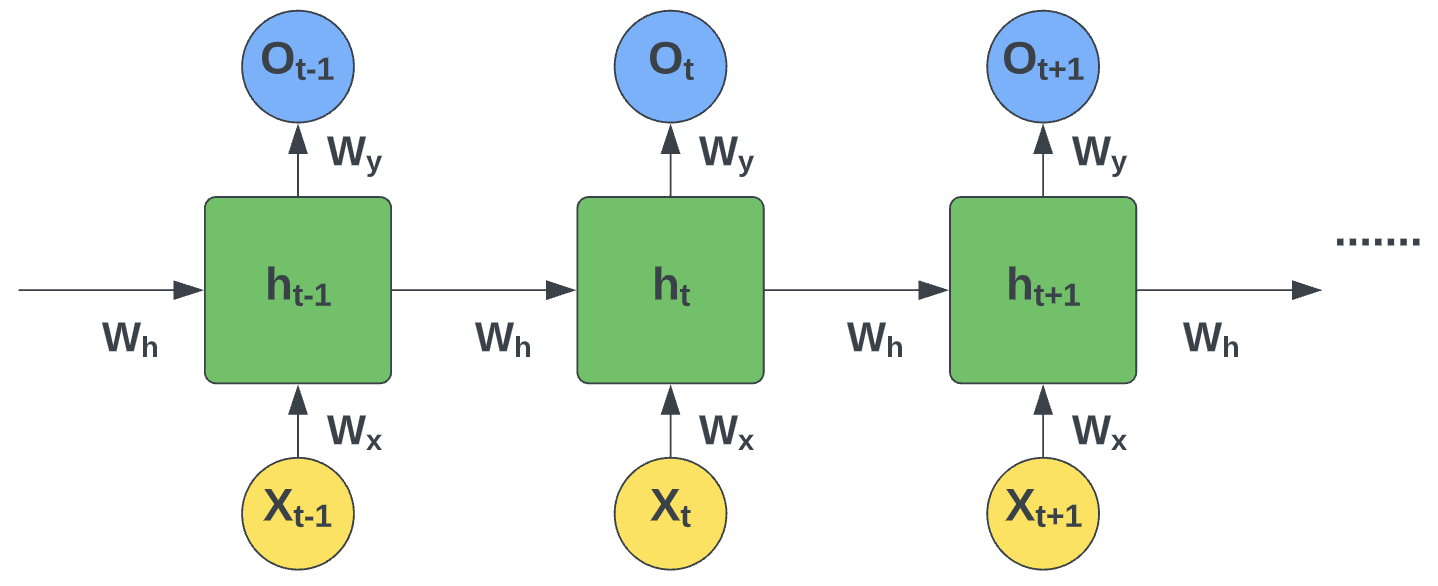}
    \caption[Neurons within an RNN]{Neurons within an RNN. At the input side, the neuron's input at time $t$ is a function of the encoded word (i.e. input vector $x_t$) and a hidden state vector $h_{t-1}$ which contains the previous sequence. The output generated by the neuron is represented by the vector $O_t$.}
    \label{fig:neurons}
\end{figure}

Prior to Transformer, encoder-decoder models that incorporate RNNs were the most common method of representing text sequences in NMT. RNNs are networks which accumulate information composed of similar units repeated over time. In NMT, a primary function of the RNN encoder is that it encodes text, i.e. it turns text into a numeric representation. Neurons within an RNN are illustrated in Figure \ref{fig:neurons}.

\begin{figure}[htb]
    \centering
    \includegraphics[width=12cm]{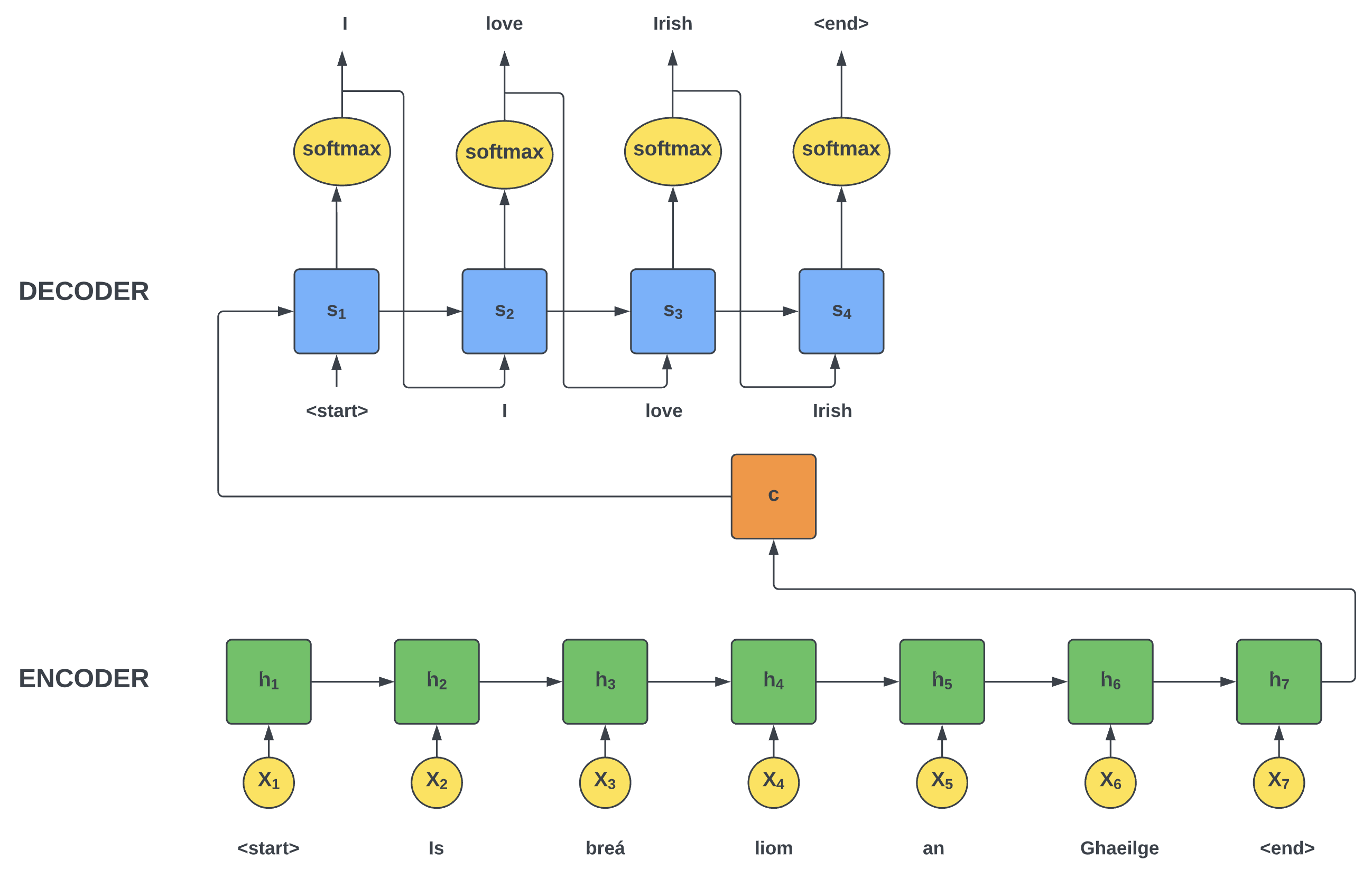}
    \caption[Encoder-decoder architecture]{Encoder-decoder architecture. The encoder encodes the entire input sequence into a fixed-length \textit{context vector}, $c$, by processing input time steps. The function of the decoder is to read this \textit{context vector} while stepping through output time steps.}
    \label{fig:encdec}
\end{figure}

Decoders unfold the vector representing the sequence state and return text. An important distinction between an encoder and a decoder is illustrated in Figure \ref{fig:encdec}, where it can be seen that both the encoder hidden state and the output from the previous decoding state are required by the decoder.

To kick-start the processing of the decoder, a special token $<start>$ is used since there is no previous output. The calculations carried out by the encoder are summarised in Equation \ref{eqn3}:

\begin{equation} \label{eqn3}
h_t=RNN_{ENC}(x_t,h_{t-1})  
\end{equation}
\\
The $RNN_{ENC}$ function is iteratively applied over the input sequence to generate the final encoder state, $h_s$ which is fed to the decoder. The complete source sentence is effectively represented by $h_s$. The decoder within the model predicts the next word of the target sentence y, while such predictions are conditional on the source sentence $x$. 


The RNN decoder, $RNN_{DEC}$, creates a state vector $s_t$ by compressing the decoding history ${y_0,…,y_{t-1}}$ which is described in Equation \ref{eqn4}. The distribution of target tokens is predicted by a classification layer which typically uses the Softmax activation function. 

\begin{equation} \label{eqn4}
s_t=RNN_{DEC}(y_{t-1},s_{t-1})  
\end{equation}

\subsubsection{Learning}

It is possible to optimise models using different types of training objectives, although maximum log-likelihood (MLE) is the most commonly used method. Given a set of training examples $D =\{(x^s,y^s)\}_{s=1}^{S}$, the MLE is maximised according to Equations \ref{eqn5} and \ref{eqn6}.

\begin{equation} \label{eqn5}
\boldsymbol{\hat{\theta}}_{MLE}= \arg \max_{\theta} \{\mathcal{L}(\theta)\}
\end{equation}

\begin{equation} \label{eqn6}
\mathcal{L}(\theta)\ = \sum_{s=1}^{S}logP(y^{s}\vert x^{s});\theta)
\end{equation}

The gradient of $\mathcal{L}$ with respect to $\theta$ is calculated using back-propagation \parencite{rumelhart1986learning} as an automatic differentiation algorithm for calculating gradients of the neural network weights, where $\theta$ is the set of model parameters.  

Many NMT approaches implement Stochastic Gradient Descent (SGD) as the optimisation algorithm for minimising the loss of the predictive model with regard to the training data. For reasons of computational efficiency, SGD typically computes the loss function and gradients on a minibatch of the training set. The standard SGD optimiser updates parameters of an NMT model according to Equation \ref{eqn7}, where the learning rate is specified by $\alpha$: 

\begin{equation} \label{eqn7}
\theta \leftarrow \theta - \alpha \bigtriangledown \mathcal{L}(\theta)
\end{equation}
\\
There are several alternatives to using SGD for optimisation, among which the ADAM optimiser has proven popular due to a reduction in training times \parencite{kingma2014adam}.

\begin{figure}[htp]
    \centering
    \includegraphics[width=10cm]{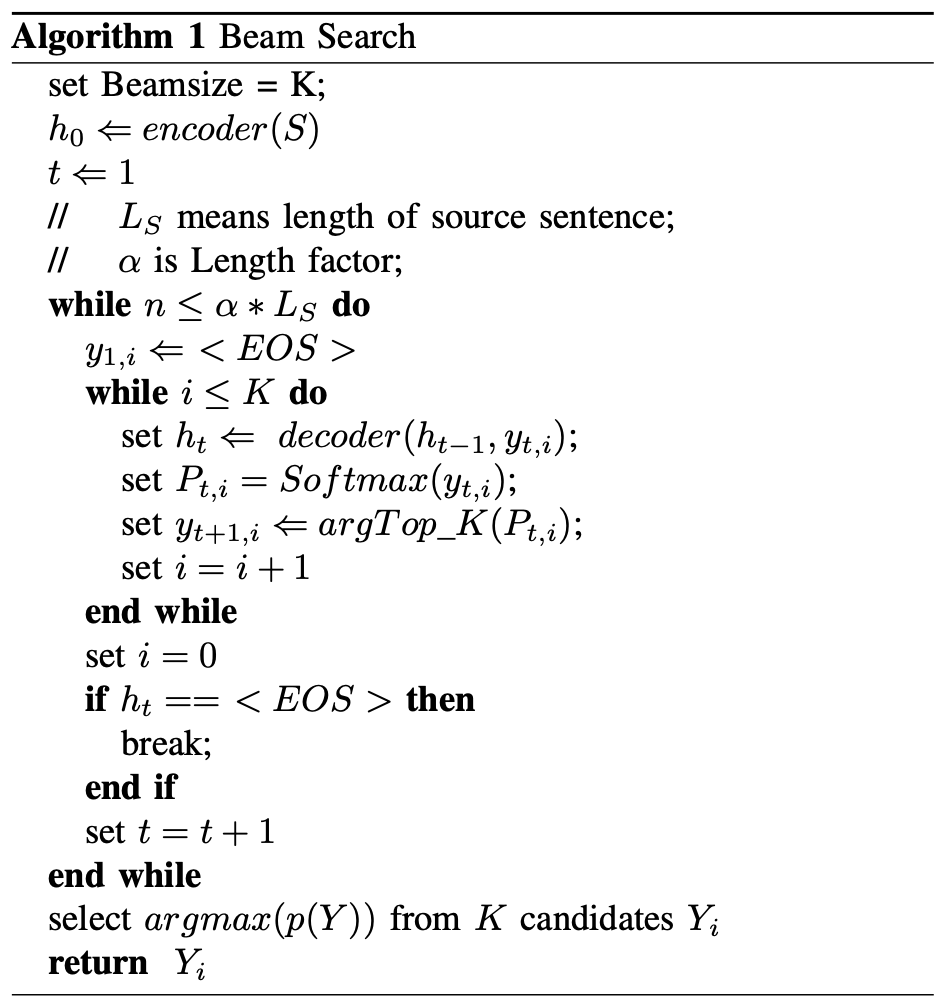}
    \caption[Beam Search Algorithm]{Beam Search Algorithm} \parencite{yang2020survey}
    \label{fig:beam}
\end{figure}

\subsubsection{Inference}
In the context of NMT, inference should ideally find the target translated sentence $y$ from the source $x$ which maximises the model prediction $P(y \vert x;\theta)$. However, in practice, it is often difficult to find the translation with the highest probability due to the impractically large search space. Accordingly, to find a good but not necessarily the very `best' (i.e. that with the highest probability given the model) translation, NMT usually relies instead on local search algorithms such as greedy search or beam search (cf. Figure \ref{fig:beam}). Translations are carried out by default using beam search, although the option exists to switch to greedy search if needed. This approach is consistent with many other NMT tools since beam search is a classic local search algorithm. Using a pre-defined beam width parameter K, the beam search algorithm keeps only the top-K possible translations as potential candidates.  With each iteration, a new potential translation is formed by combining each candidate word with a new word. New candidate translations compete with each other using log probability values to obtain the new top-K most probable results. This process is continued until the end of the translation process, and the 1-best translation is output.

\subsection{Subword Models}
\label{section:subword}
Translation by its very nature requires an open vocabulary, but restricted (e.g. 30k, 50k, or 70k) vocabularies are typically used for reasons of computational efficiency.  However, the use of subword models aims to address this fixed vocabulary problem associated with NMT. The problem manifests itself in how previously unseen `out-of-vocabulary' (OOV) words are handled. In such cases, a single `UNK' (for `unknown') token is used to `recognise' the OOV word. Encoding rare and unknown words into sequences of subword units significantly reduces the problem and has thus given rise to a number of subword algorithms.

Optimally, this will be performed via morphological processing \parencite{passban-etal-2018-tailoring}, but good quality wide-coverage morphological analysers are not always available. Therefore it is common practice to use methods such as Byte Pair Encoding (BPE)~\parencite{gage1994new} to break down rare and previously unseen words into subword models in order to significantly improve translation performance \parencite{sennrich2015neural,kudo2018subword}. 

Designed for NMT, SentencePiece \parencite{kudo2018sentencepiece}, is a language-independent subword tokenizer that provides an open-source C++ and a Python implementation for subword units. An attractive feature of the tokenizer is that SentencePiece trains subword models directly from raw sentences.

\subsection{NMT Tools}
\parencite{kreutzer2019joey} describe their Joey NMT platform\footnote{\url{https://github.com/joeynmt/joeynmt}} as a minimalist NMT toolkit, based on PyTorch, which is designed especially for newcomers to the field. Joey NMT provides many popular NMT features in a simple code base enabling novice users to easily adapt the system to their particular requirements. The toolkit supports both RNN and Transformer architectures.

Given that adaptNMT is essentially an IPython wrapper layered on top of OpenNMT, it inherits all of OpenNMT’s features and continues to benefit from the work which goes into developing and maintaining its code base. adaptNMT offers a higher level of abstraction over OpenNMT where the focus is much more on usability, especially for newcomers to the field. Accordingly, it provides for easy and rapid deployment by enabling new features such as greater pre-processing, as well as GUI control over model building. It also contains green features in line with the current research drive towards smaller models with lower carbon footprints (cf. Sections~\ref{sec:envimp-nmt} and \ref{disc-nmt}). Such features make adaptNMT suitable for both educational and research environments. The key features differentiating adaptNMT from Joey NMT are outlined in Table \ref{tab:joey}.

\begin{table}[]
\begin{tabular}{@{}l@{}}
\hline
adaptNMT is built upon OpenNMT and subsequently inherits all of its features. \\ \hline
\begin{tabular}[c]{@{}l@{}}The interface is designed and fully implemented in Google Colab.\end{tabular} \\ \hline
\begin{tabular}[c]{@{}l@{}}Colab is easier to follow for practitioners since each step can be executed individually.\\ The approach is ideal in education since the progression of the pipeline is demonstrated.\end{tabular} \\ \hline
Training of models can be viewed and controlled using Colab Android or Apple apps. \\ \hline
\begin{tabular}[c]{@{}l@{}}adaptNMT can be run in local mode enabling existing infrastructure to be utilised or in \\ hosted mode which allows rapid scaling of the infrastructure.\end{tabular} \\ \hline
\begin{tabular}[c]{@{}l@{}}Colab Pro+ provides individual researchers, or even small teams, the capacity to build \\ large models on an excellent infrastructure with very little resources.\end{tabular} \\ \hline
\begin{tabular}[c]{@{}l@{}}GUI controls can split a corpus into training, validation and test datasets. \end{tabular} \\ \hline
GUI controls are available for hyperparameter customisation in NMT training. \\ \hline
A green report outlines the country-specific kgCO\textsubscript2 generated when training a model.  \\ \hline
Autonotification notifies the user on the completion of training. \\ \hline
A deploy function enables the immediate deployment of trained models. \\ \hline
The functionality of serverNMT is not available within Joey NMT. \\ \hline
\end{tabular}
\caption{Key features differentiating adaptNMT from Joey NMT}
\label{tab:joey}
\end{table} 
Other popular frameworks for NMT system-building include FAIRSEQ\footnote{\url{https://github.com/facebookresearch/fairseq}}~\parencite{ott2019fairseq}, an open-source sequence modelling toolkit based on PyTorch, that enables researchers to train models for translation, summarisation and language modelling. Marian\footnote{\url{https://marian-nmt.github.io}}~\parencite{junczys2018marian}, developed using C++, is an NMT framework based on dynamic computation graphs. OpenNMT\footnote{\url{https://opennmt.net}}~\parencite{klein2017opennmt} is an open-source NMT framework that has been widely adopted in the research community. The toolkit covers the entire MT workflow from the preparation of data to live inference. 

\subsection{Hyperparameter Optimisation}

Hyperparameters are employed to customise machine learning models such as translation models. It has been shown that machine learning performance may be improved through hyperparameter optimisation (HPO) rather than just using default settings~\parencite{sanders2017informing}.

The principal methods of HPO are grid search~\parencite{montgomery2001design} and random search~\parencite{JMLR:v13:bergstra12a}. Grid search is an exhaustive technique which evaluates all parameter permutations. However, as the number of features grows, the amount of data permutations grows exponentially making optimisation expensive in the context of developing translation models which require long build times. Accordingly, an effective, less computationally intensive alternative is to use random search which samples random configurations.

\begin{figure}[ht]
    \centering
    \includegraphics[width=15cm]{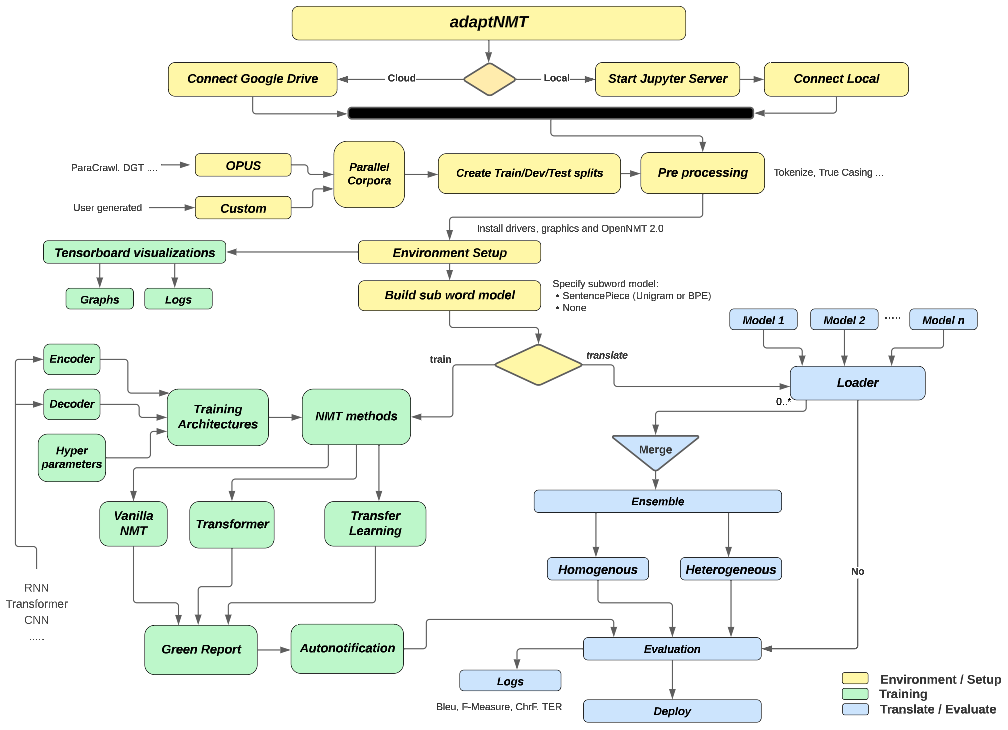}
    \caption[Proposed architecture for adaptNMT]{Proposed architecture for adaptNMT: a language-agnostic NMT development environment. The system is designed to run either in the cloud or using local infrastructure. Models are trained using parallel corpora. Visualisation and extensive logging enable real-time monitoring. Models are developed using vanilla RNN-based NMT, Transformer-based approaches or transfer learning using a fine-tuning approach. Translation and evaluation can be carried out using either single models or ensembles.}
    \label{fig:approach_lrev}
\end{figure}

\section{Architecture of adaptNMT}\label{arch}

Having described the individual components of RNN- and Transformer-based NMT systems, we now present the adaptNMT tool itself, in which these components can be configured by the user. A high-level view of the system architecture of the platform is presented in Figure \ref{fig:approach_lrev}. Developed as an IPython notebook, the application uses the Pytorch implementation of \textit{OpenNMT} for training models with SentencePiece used for training subword models. By using a Jupyter notebook, the application may be easily shared with others in the MT community. Furthermore, the difficulties involved in setting up the correct development environment have largely been removed since all required packages are downloaded on the fly as the application runs. 

\begin{figure} [htp!]
\centering
\begin{subfigure}{.45\textwidth}
  \includegraphics[width=.9\linewidth]{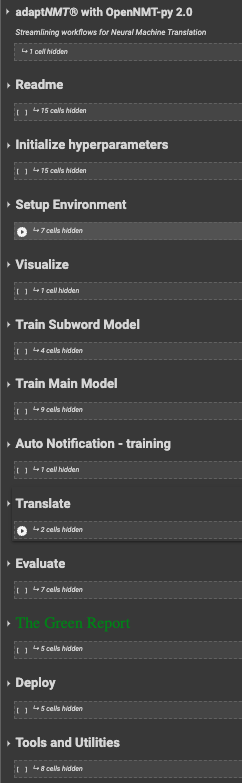}
  \caption{\footnotesize {Overview of adaptNMT. Key areas include initialisation, pre-processing, environment setup, visualisation, auto and custom NMT, training of subword model, training of main model, evaluation and deployment (cf. Section \ref{archNMT}).} }
  \label{fig:sub1}
\end{subfigure}\hspace{5mm}%
\begin{subfigure}{.45\textwidth}
  \includegraphics[width=.9\linewidth]{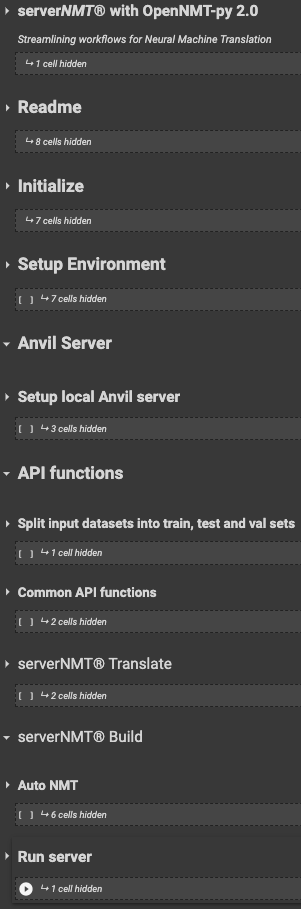}
  \caption{\footnotesize {Overview of serverNMT. Highlighted cells include initialisation, environment setup, Anvil server, API functions, translation, model building, adaptNMT and running the server (cf. Section \ref{sNMT}).}}
  \label{fig:sub2}
\end{subfigure}
\caption{adaptNMT and serverNMT}
\label{fig:notebookcells}
\end{figure}

There are options to run the system on local infrastructure or to run it as a Colab instance using Google Cloud. Translation models are developed using parallel text corpora of the source and target languages. A Tensorboard visualisation provides a real-time graphical view of model training. The primary use cases for the system are model building and a translation service, one or both of which can be selected at run-time. As illustrated in the system diagram in Figure \ref{fig:approach_lrev}, generating an ensemble output while translating has also been facilitated. Models may also be deployed to a pre-configured location. 

\subsection{adaptNMT}\label{archNMT} 
The application may be run as an IPython Jupyter notebook or as a Google Colab application. Given the ease of integrating large Google Drive storage into Colab, the application has been used exclusively as a Google Colab application for our experiments, some of which are described in Section~\ref{sec:exp-nmt}. The key features of the notebook are illustrated in Figure \ref{fig:notebookcells}.

\subsubsection{Initialisation and logging}

Initialisation enables connection to Google Drive to run experiments, automatic installation of Python, OpenNMT, SentencePiece, Pytorch and other applications. The visualisation section enables real-time graphing of model development.  All log files are stored and can be viewed to inspect training convergence, the model’s training and validation accuracy, changes in learning rates and cross-entropy.

\subsubsection{Modes of operation}
There are two modes of operation: local and cloud. In local mode, the application is run so that models are built using the user's local GPU resources. The option to use cloud mode enables users to develop models using Google's GPU clusters. For shorter training times, the unpaid Colab option is adequate. However, for a small monthly subscription, the Google Colab Pro option is worthwhile since users have access to improved GPU and compute resources. Nevertheless, there are also environmental and running costs to consider (cf. Sections~\ref{sec:envimp-nmt} and \ref{disc-nmt}), although the Google Cloud is run on a platform which uses 100\% renewables \parencite{lacoste2019quantifying}. It is also a very cost-effective option for those working in the domain of low-resource languages since developing smaller models requires shorter training times. However, users requiring long training times and very high compute resources will need to use their own hardware and run the application in local mode unless they have access to large budgets.

\subsubsection{Customisation of models}
The system has been developed to allow users to select variations to the underlying model architecture. A vanilla RNN or Transformer approach may be selected to develop the NMT model. The customisation mode enables users to specify the exact parameters required for the chosen approach. One of the features, AutoBuild, enables a user to build an NMT model in three simple steps: (i) upload source and target files, (ii) select  RNN or Transformer, and (iii) click AutoBuild.

\subsubsection{Use of subword segmentation}
The type of optimiser to be used for learning can be specified, and users may also choose to employ different types of subword models when building the system. The subword model functionality allows the user to choose whether or not to use a subword model. Currently, the user specifies the vocabulary size and chooses either a SentencePiece unigram or a SentencePiece BPE subword model (cf. Section \ref{section:subword}). 

A user may upload a dataset which includes the training, validation and test splits for both source and target languages. In cases where a user has not already created the required splits for model training, single source and target files may be uploaded. The splits needed to create the training, validation and test files are then automatically generated according to the user-specified split ratio. Given that building NMT models typically demands long training times, an automatic notification feature is incorporated that informs the user by email when model training has been completed.

\subsubsection{Translation and evaluation}
In addition to supporting the training of models, the application also allows for translation and evaluation of model performance. Translation using pre-built models is also parameterised. Users specify the name of the model as a hyperparameter which is then subsequently used to translate and evaluate the test files. The option for creating an ensemble output is also catered for, and users simply name the models which are to be used in generating the ensemble output.

Once the system has been built, the model to be used for translating the test set may be selected. To evaluate the quality of translation, humans usually provide the best insight, but they may not always be available, do not always agree, and are expensive to recruit for experiments. Accordingly, automatic evaluation metrics are typically used, especially by developers monitoring the incremental progress of systems (cf. \cite{Way2018} for more on the pros and cons of human and automatic evaluation).

Several automatic evaluation metrics provided by SacreBleu\footnote{\url{https://github.com/mjpost/sacrebleu}} \parencite{post2018call} are used: BLEU \parencite{papineni-etal-2002-bleu}, TER \parencite{snover2006study} and ChrF \parencite{popovic2015chrf}. Translation quality can also be evaluated using Meteor~\parencite{denkowski2014meteor} and F1 score~\parencite{melamed-etal-2003-precision}. Note that BLEU, ChrF, Meteor and F1 are precision-based metrics, so higher scores are better, whereas TER is an error-based metric and lower scores indicate better translation quality. Evaluation options available include standard (truecase) and lowercase BLEU scores, a sentence-level BLEU score option, ChrF1 and ChrF3.  

There are three levels of logging for model development, training and experimental results. A references section outlines resources which are relevant to developing, using and understanding adaptNMT. Validation during training is currently conducted using model accuracy and perplexity (PPL). 

\subsection{serverNMT}\label{sNMT}

A  server application, serverNMT, was also developed and implemented as an IPython notebook. It can be configured to run either as a translation server or as a build server. A secure connection, implemented from serverNMT, can be made to websites hosting embedded web apps. At the core of serverNMT, there are two embedded Python web apps, one for translation services and another for developing models, both of which use the anvil.works platform.\footnote{\url{https://anvil.works}} 

As a build server, serverNMT enables a window to the underlying cloud infrastructure in which NMT models can be trained. A web app hosted on another system may connect to this infrastructure made available by serverNMT.

Using an Anvil server embedded within serverNMT, the application continuously waits for communication to web apps and effectively enables a cloud infrastructure for NMT. Written as a REST server, it acts as an API for serving previously built models and facilitates the integration of translation models with other systems.

\section{Empirical Evaluation}
\label{sec:exp-nmt}
 
Having described the theoretical background and the tool itself, we now evaluate the effectiveness of the adaptNMT approach by training models for English to Irish (EN$\rightarrow$GA) and Irish to English (GA$\rightarrow$EN) translation in the health domain using the {\em gaHealth} \parencite{lankford2022lrec} corpus.\footnote{\url{https://github.com/seamusl/gaHealth}} All experiments involved concatenating source and target corpora to create a shared vocabulary and a shared SentencePiece subword model. To benchmark the performance of our models, the EN$\leftrightarrow$GA test datasets from the LoResMT2021 Shared Task\footnote{\url{https://github.com/loresmt/loresmt-2021}}~\parencite{ojha2021findings} were used. These test datasets enabled the evaluation of the {\em gaHealth} models since the shared task focused on an application of the health domain, namely the translation of Covid-related data. Furthermore, using an official test dataset from a shared task enables the direct comparison of our models' performance with models entered by other teams, as well as future implementations. 

The hyperparameters used for developing the models are outlined in Table \ref{tab:hpo-table-nmt}. The details of the training, validation and test sets used by our NMT models are outlined in Tables \ref{tab:en2ga-stats-nmt} and  \ref{tab:ga2en-stats-nmt}. In all cases, 502 lines were used from the LoResMT2021 validation dataset whereas the test dataset used 502 lines for EN$\rightarrow$GA translation and 250 lines for GA$\rightarrow$EN translation. Both were independent health-specific Covid test sets which were provided by LoResMT2021. There was one exception; due to a data overlap between the test and train datasets, a reduced test set was used when testing the {\em gaHealth} en2ga* system.  

The results from the IIITT~\parencite{puranik2021attentive} and UCF~\parencite{chen2021ucf} teams are included in Tables \ref{tab:en2ga-nmt} and \ref{tab:ga2en-nmt} so the performance of the {\em gaHealth} models can be easily compared with the findings of the participating LoResMT2021 systems. IIITT fine-tuned an Opus MT model\footnote{\url{https://github.com/Helsinki-NLP/Opus-MT}}~\parencite{tiedemann-thottingal-2020-opus} on the training dataset. UCF used transfer learning \parencite{zoph-etal-2016-transfer}, unigram and subword segmentation methods for EN$\leftrightarrow$GA translation.

\begin{center}
\begin{table}
\center
\begin{tabular}{ll}
\hline
\textbf{Hyperparameter} & \textbf{Values}                \\ \hline
Learning rate            & 0.1, 0.01, 0.001, \textbf{2}            \\ \hline
Batch size               & 1024, \textbf{2048},  4096, 8192       \\ \hline
Attention heads          & \textbf{2}, 4, \textbf{8}                     \\ \hline
Number of layers         & 5, \textbf{6}                           \\ \hline
Feed-forward dimension   & \textbf{2048}                           \\ \hline
Embedding dimension      & 128, \textbf{256}, 512                  \\ \hline
Label smoothing          & \textbf{0.1}, 0.3                       \\ \hline
Dropout                  & 0.1, \textbf{0.3}                       \\ \hline
Attention dropout        & \textbf{0.1}                            \\ \hline
Average Decay            & 0, \textbf{0.0001}                      \\ \hline
\end{tabular}
\caption[Hyperparameter optimisation for Transformer models]{Hyperparameter optimisation for Transformer models. Optimal parameters are highlighted in bold. \parencite{lankford2021transformer}.}
\label{tab:hpo-table-nmt}
\end{table}
\end{center}

\begin{table} [ht]
\centering
\begin{tabular}{lcccccc}
\hline
\textbf{Team} &
 \textbf{System} &
  \textbf{Train}  &
  \textbf{Validation}  &
  \textbf{Test}  \\ \hline
adapt & covid\_extended & 13k & 502 & 500 \\
adapt & combined\_domains & 65k & 502 & 500 \\
IIITT  & en2ga-b & 8k & 502 & 500 \\
UCF     & en2ga-a & 8k & 502 & 500   \\ 
{\em gaHealth} & en2ga & 24k & 502 & 500 \\
{\em gaHealth} & en2ga* & 24k & 502 & 338 \\
\hline
\end{tabular}
\caption[EN$\rightarrow$GA training, validation and test dataset distributions]{EN$\rightarrow$GA training, validation and test dataset distributions. The baseline {\em gaHealth} system was augmented with an 8k Covid dataset provided by LoResMT2021.} 
\label{tab:en2ga-stats-nmt}
\end{table}

\begin{table} [ht]
\centering
\begin{tabular}{lcccccc}
\hline
\textbf{Team} &
 \textbf{System} &
  \textbf{Train}  &
  \textbf{Validation}  &
  \textbf{Test}  \\ \hline
IIITT & ga2en-b & 8k & 502 & 250 \\
UCF & ga2en-b & 8k & 502 & 250   \\ 
{\em gaHealth} & ga2en & 24k & 502 & 250 \\
\hline
\end{tabular} 
\caption[GA$\rightarrow$EN training, validation and test dataset distributions]{GA$\rightarrow$EN training, validation and test dataset distributions. The baseline {\em gaHealth} system was augmented with an 8k Covid dataset provided by LoResMT2021. All overlaps were removed from the {\em gaHealth} corpus prior to training the {\em gaHealth} ga2en model.}
\label{tab:ga2en-stats-nmt}
\end{table}

\subsection{Infrastructure}
Rapid prototype development was enabled through a Google Colab Pro subscription using NVIDIA Tesla P100 PCIe 16GB graphic cards and up to 27GB of memory when available~\parencite{Bisong2019}. All {\em gaHealth} MT models were trained using adaptNMT. 

\subsection{Metrics}

Automated metrics were used to determine the translation quality. To compare against our previous work, the performance of models is measured using three evaluation metrics, namely BLEU, TER and ChrF. These metrics indicate the accuracy of the translations derived from our NMT systems.

Case-insensitive BLEU scores at the corpus level are reported. Model training was stopped after 40k training steps or once an early stopping criterion of no improvement in validation accuracy for four consecutive iterations was recorded.

PPL is often used to evaluate language models within NLP. It measures the effectiveness of a probability model in predicting a sample. As a metric for translation performance, it is important to keep low scores so that the number of alternative translations is reduced.

\subsection{Results: Automatic Evaluation}

\begin{figure}[htbp]
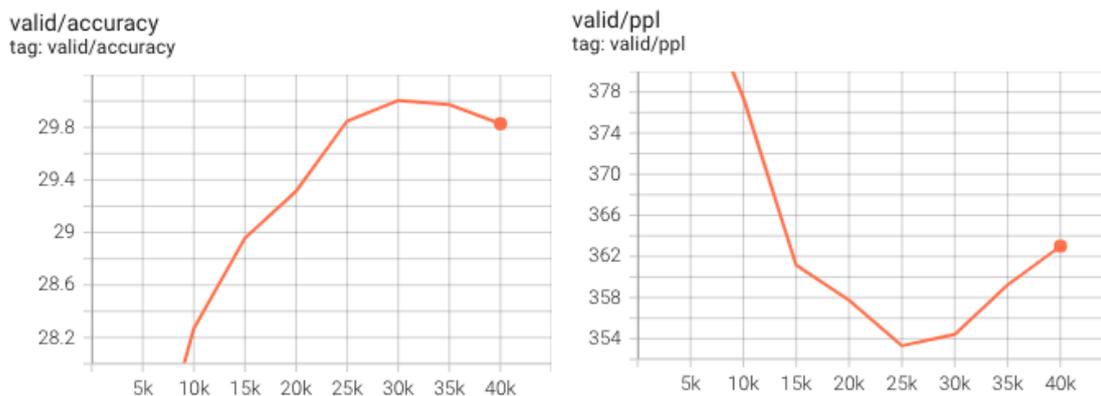

    \centering
    {\includegraphics[width=7.25cm]{en-ga_acc_covid.png}}
    {\includegraphics[width=7.25cm]{en-ga_ppl_covid.png}}
    \caption[adapt covid\_extended system]{adapt covid\_extended system: training EN$\rightarrow$GA model with 13k lines consisting of the ADAPT 5k corpus and an 8k LoResMT2021 Covid corpus. The graph on the left illustrates accuracy and the graph on the right demonstrates perplexity.}
    \label{fig:en-ga-covid}
\end{figure}

The experimental results from LoResMT 2021 are summarised in Tables \ref{tab:en2ga-nmt} and \ref{tab:ga2en-nmt}. In the LoResMT2021 Shared Task, the highest-performing EN$\rightarrow$GA system was submitted by the ADAPT team \parencite{lankford2021machine}. The system uses an extended Covid dataset, which is a combination of the 2021 MT Summit Covid baseline and a custom ADAPT Covid dataset. The model, developed within adaptNMT, uses a Transformer architecture with 2 heads. It performs well across all key translation metrics (BLEU: 36.0, TER: 0.531 and ChrF3: 0.6). The training of this EN$\rightarrow$GA model is illustrated in Figure \ref{fig:en-ga-covid}. The model achieved a maximum validation accuracy of 30.0\% and perplexity of 354 after 30k steps. 

\begin{table}[ht!]
\centering
\begin{tabular}{lcccccc}
\hline
\textbf{Team} &
 \textbf{System} &
  \textbf{BLEU} $\uparrow$ &
  \textbf{TER} $\downarrow$ &
  \textbf{ChrF3} $\uparrow$ \\ \hline
UCF & en2ga-b & 13.5 & 0.756 & 0.37   \\
IIITT & en2ga-b & 25.8 & 0.629 & 0.53 \\
adapt & combined & 32.8 & 0.590 & 0.57 \\
{\em gaHealth} & en2ga & 33.3 & 0.604 & 0.56 \\
adapt & covid\_extended & 36.0 & 0.531 & 0.60 \\
{\em gaHealth} & en2ga* & \textbf{37.6} & 0.577 & 0.57 \\  
\hline
\end{tabular}
\caption[EN$\rightarrow$GA {\em gaHealth} system compared with LoResMT 2021]{EN$\rightarrow$GA system compared with LoResMT 2021 EN$\rightarrow$GA systems.}
\label{tab:en2ga-nmt}
\end{table}

\begin{figure}[ht!]
    \centering
    {\includegraphics[width=7.25cm]{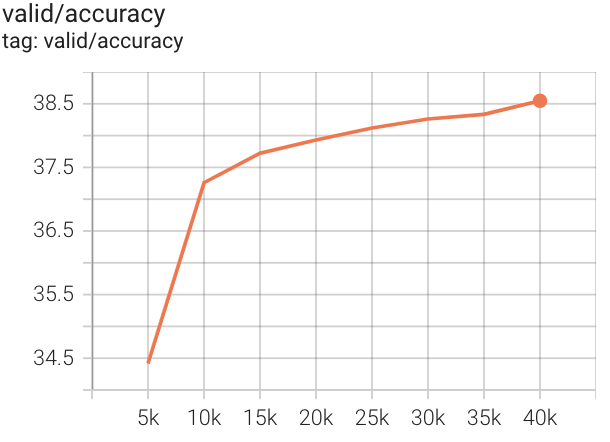}}
    {\includegraphics[width=7.25cm]{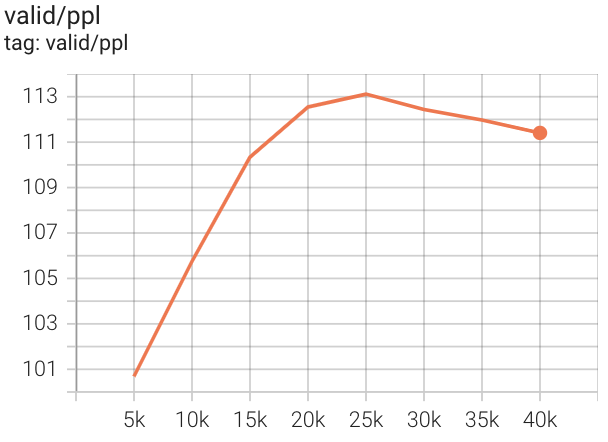}}
    \caption[{\em gaHealth} en2ga* system: training EN$\rightarrow$GA model]{{\em gaHealth} en2ga* system: training EN$\rightarrow$GA model with combined 16k gaHealth corpus and 8k LoResMT2021 Covid corpus. The graph on the left illustrates OpenNMT accuracy and the graph on the right demonstrates perplexity.}
    \label{fig:en-ga-gaHealth}
\end{figure}

\begin{figure}[h]
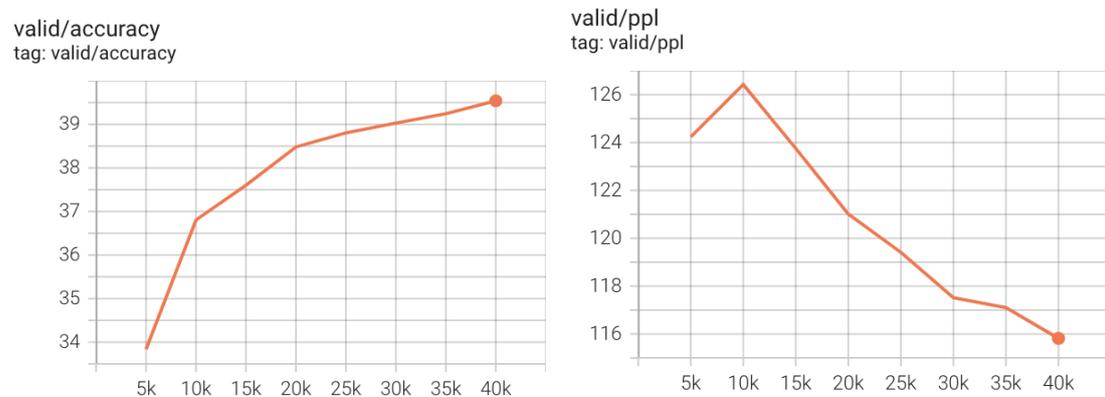

    \centering
    {\includegraphics[width=7.25cm]{ga-en_acc.png}}
    {\includegraphics[width=7.25cm]{ga-en_ppl.png}}
    \caption[{\em gaHealth} ga2en system: training GA$\rightarrow$EN model]{{\em gaHealth} ga2en system: training GA$\rightarrow$EN model with combined 16k gaHealth corpus and 8k LoResMT2021 Covid corpus. The graph on the left illustrates OpenNMT accuracy and the graph on the right demonstrates perplexity.}
    \label{fig:ga-en-gaHealth}
\end{figure}

The results from the LoResMT2021 Shared Task were further improved by developing models using a bespoke health dataset, {\em gaHealth}. Table \ref{tab:en2ga-nmt} shows an improvement of 1.6 BLEU points, a relative improvement of almost 4.5\%, although TER and ChrF3 scores are a little worse. Validation accuracy and PPL in training the {\em gaHealth} models with adaptNMT are illustrated in Figures \ref{fig:en-ga-gaHealth} and \ref{fig:ga-en-gaHealth}. Figure \ref{fig:en-ga-covid} illustrates model training using the covid\_extended dataset, also developed using adaptNMT. In training the {\em gaHealth} en2ga* system, as highlighted in Figure \ref{fig:en-ga-gaHealth}, the EN$\rightarrow$GA model was trained with the combined 16k {\em gaHealth} and 8k LoResMT2021 corpora. The model's validation accuracy of 38.5\% and perplexity of 113 achieved a BLEU score of 37.6 when evaluated with the test data.

The training of the GA$\rightarrow$EN {\em gaHealth} ga2en system with the combined 16k gaHealth corpus and 8k LoResMT2021 Covid corpus is shown in Figure \ref{fig:ga-en-gaHealth}. This model achieves a validation accuracy of 39.5\% and perplexity of 116 which results in a BLEU score of 57.6. This is significantly better (by 20 BLEU points) than for the reverse direction, as it is well-known that translating into a morphological-rich language like Irish is always more difficult compared to when the same language acts as the source. This is confirmed by comparing the results for the UCF (13.5 vs. 21.3 BLEU) and IIITT (25.8 vs. 34.6) systems in Tables \ref{tab:en2ga-nmt} and \ref{tab:ga2en-nmt}.

Rapid convergence was observed while training the {\em gaHealth} models such that little accuracy improvement occurs after 30k steps, 10K fewer than for the reverse direction. Only marginal gains were achieved after this point and it declined in the case of the system trained using the covid\_extended dataset, as the left-hand graph in Figure~\ref{fig:en-ga-covid} shows. 

\par
Of the models developed by the ADAPT team, the worst-performing model uses a larger 65k dataset. This is not surprising given that the dataset is from a generic domain of which only 20\% is health-related. The performance of this higher-resourced 65k line model lags behind the augmented {\em gaHealth} model which was developed using just 24k lines. 

\begin{table}[ht!]
\centering
\begin{tabular}{lcccccc}
\hline
\textbf{Team} &
 \textbf{System} &
  \textbf{BLEU} $\uparrow$ &
  \textbf{TER} $\downarrow$ &
  \textbf{ChrF3} $\uparrow$ \\ \hline
UCF & ga2en-b & 21.3 & 0.711 & 0.45\\
IIITT  & ga2en-b & 34.6 & 0.586 & 0.61\\
{\em gaHealth} & ga2en & \textbf{57.6} & 0.385 & 0.71\\
\hline
\end{tabular} 
\caption[GA$\rightarrow$EN {\em gaHealth} systems compared with LoResMT2021]{GA$\rightarrow$EN {\em gaHealth} systems compared with LoResMT2021 GA$\rightarrow$EN systems}
\label{tab:ga2en-nmt}
\end{table}

\par
For GA$\rightarrow$EN translation, the best-performing model for the LoResMT2021 Shared Task was developed by IIITT with a BLEU of 34.6, a TER of 0.586 and ChrF3 of 0.6. Accordingly, this serves as the baseline score by which our GA$\rightarrow$EN model, developed using the {\em gaHealth} corpus, can be benchmarked. The performance of the {\em gaHealth} model offers an improvement across all metrics with a BLEU score of 57.6, a TER of 0.385 and a ChrF3 result of 0.71. In particular, the 66\% relative improvement in BLEU score against the IIITT system is very significant.

\subsection{Environmental Impact}
\label{sec:envimp-nmt}
We were motivated by the findings of \parencite{strubell-etal-2019-energy} and \parencite{bender2021dangers} to track the energy consumption required to train our models. Prototype model development used Colab Pro, which as part of Google Cloud is carbon neutral \parencite{lacoste2019quantifying}. However, longer running Transformer experiments were conducted on local servers using 324 gCO\textsubscript2 per kWh\footnote{\url{https://www.seai.ie/publications/Energy-in-Ireland-2020.pdf}} \parencite{sei2020}. The net result was just under 10 kgCO\textsubscript2 created for a full run of model development. Models developed during this study will be reused for ensemble experiments in the future so that work will have a life beyond this paper.

\subsection{Stochastic Nuances}
\label{sec:stoc}

To evaluate the translation performance of an IPython-based application such as adaptNMT, a comparison with a Python script version of the same application, myNMT.py, was conducted.  We built EN$\leftrightarrow$GA translation models using this script. The models developed with adaptNMT were trained on Google Colab using a 12GB Tesla K80 GPU, whereas the myNMT models were trained on a local machine using a 12GB Gigabyte 3060 graphics card. The results from evaluating these models are presented in Tables~\ref{tab:engastoc} and \ref{tab:gaenstoc}. 

Despite setting the same random seed, it is clear from Tables~\ref{tab:engastoc} and \ref{tab:gaenstoc} that the translation performance of the adaptNMT models is better by 1.2 BLEU points (3.3\% relative improvement) in EN$\rightarrow$GA translation and 1.0 BLEU point (1.8\% relative improvement) in GA$\rightarrow$EN translation. 

Given the stochastic nature of machine learning, training models on different systems can yield different results even with the same training, validation and test data. The performance differences can be attributed to the stochastic nature of the learning algorithm and evaluation procedure. Furthermore, the platforms had different underlying system architectures which is another source of stochastic error.

\begin{table}[ht!]
\centering
\begin{tabular}{lcccccc}
\hline
 \textbf{System} &
  \textbf{BLEU} $\uparrow$ &
  \textbf{TER} $\downarrow$ &
  \textbf{ChrF3} $\uparrow$ \\ \hline
adaptNMT & 37.6 & 0.577 & 0.570 \\
myNMT  & 36.4 & 0.622 & 0.56 \\
\hline
\end{tabular} 
\caption{Stochastic differences between EN$\rightarrow$GA systems}
\label{tab:engastoc}
\end{table}

\begin{table}[ht!]
\centering
\begin{tabular}{lcccccc}
\hline
 \textbf{System} &
  \textbf{BLEU} $\uparrow$ &
  \textbf{TER} $\downarrow$ &
  \textbf{ChrF3} $\uparrow$ \\ \hline
adaptNMT & 57.6 & 0.385 & 0.71 \\
myNMT  & 56.6 & 0.399 & 0.703 \\
\hline
\end{tabular} 
\caption{Stochastic differences between GA$\rightarrow$EN systems}
\label{tab:gaenstoc}
\end{table}

\section{Discussion}\label{disc-nmt}
The mathematical first principles governing NMT development were presented to demonstrate the mechanics of what happens during model training. Several parameters in Equations \ref{eqn1} - \ref{eqn7} are configurable within the adaptNMT application.
 
The environmental impact of technology, and the measurement of its effects, has gained a lot of prominence in recent years \parencite{henderson2020towards}. Indeed, this may be viewed as a natural response to truly massive NLP models which have been developed by large multinational corporations.  In particular, HPO of NMT models can be particularly demanding if hyperparameter fine-tuning is conducted across a broad search space. As part of their work on NMT architectures, the Google Brain team required more than 250,000 GPU hours for NMT HPO \parencite{britz2017massive}. Training of these models was conducted using Tesla K40m and Tesla K80 GPUs with maximum power consumption between 235 W and 300 W, giving rise to potentially in excess of 60 MWh of energy usage. Even though the Google Cloud is carbon neutral, one must consider the opportunity cost of this energy usage. 
 
A plethora of tools to evaluate the carbon footprint of NLP \parencite{bannour-etal-2021-evaluating} has subsequently been developed and the concept of sustainable NLP has become an important research track in its own right at many high-profile conferences such as the EACL 2021 \textit{Green and Sustainable NLP} track.\footnote{\url{https://2021.eacl.org/news/green-and-sustainable-nlp}} 
In light of such developments, a `green report' was incorporated into adaptNMT whereby the kgCO\textsubscript2 generated during model development is logged. This is very much in line with the industry trend of quantifying the impact of NLP on the environment; indeed, \parencite{info13020088} have demonstrated that high-performing MT systems can be built with much lower footprints, which not only reduce emissions but also in the post-deployment phase deliver savings of almost 50\% in energy costs for a real translation company.

To evaluate system performance in translating health data in the EN$\rightarrow$GA direction, we used the adaptNMT application to develop an MT model for the LoResMT2021 Shared Task. The application was subsequently used to develop an MT model for translating in the GA$\rightarrow$EN direction. In both cases, high-performing models achieving SOTA scores were achieved by using adaptNMT to develop  Transformer models capable of generating high-quality output.    

The danger of relying on increasingly large language models (LLMs) has been well-documented in the literature. Such discussion focuses not just on the environmental impact but also highlights the impact of in-built bias and the inherent risks that large models pose for low-resource languages \parencite{bender2021dangers}. Using an easily-understood framework such as adaptNMT, the benefits of developing high-performing NMT models with smaller in-domain datasets should not be overlooked. 

\section{Conclusion and Future Work}\label{concl-nmt}

We introduced adaptNMT, an application for NMT which manages the complete workflow of model development, evaluation and deployment. The performance of the application was demonstrated in the context of generating an EN$\rightarrow$GA translation model which ranked 1st in the LoResMT2021 shared task and validated against a standalone reimplementation of EN$\leftrightarrow$GA systems outside the tool, where no drop-off in performance was seen. 

With regard to future work, development will focus more on tracking environmental costs and integrating new transfer learning methods. Modern zero-shot and few-shot approaches, adopted by GPT3 \parencite{brown2020language} and Facebook LASER \parencite{artetxe2019massively} frameworks, will be integrated. Whereas the existing adaptNMT application focuses on customising NMT models, a separate application adaptMLLM will be developed to fine-tune multilingual language models and LLMs, in particular those that focus on low-resource language pairs such as NLLB \parencite{costa2022no}.

The green report embedded within the application is our first implementation of a sustainable NLP feature within adaptNMT. It is planned to develop this feature further to include an improved user interface and user recommendations about how to develop greener models.  As an open-source project, we hope the community will add to its development by contributing new ideas and improvements.

\chapter{adaptMLLM: Fine-Tuning Multilingual Language Models}
\section{Context}

\begin{figure}[ht!] 
  \includegraphics{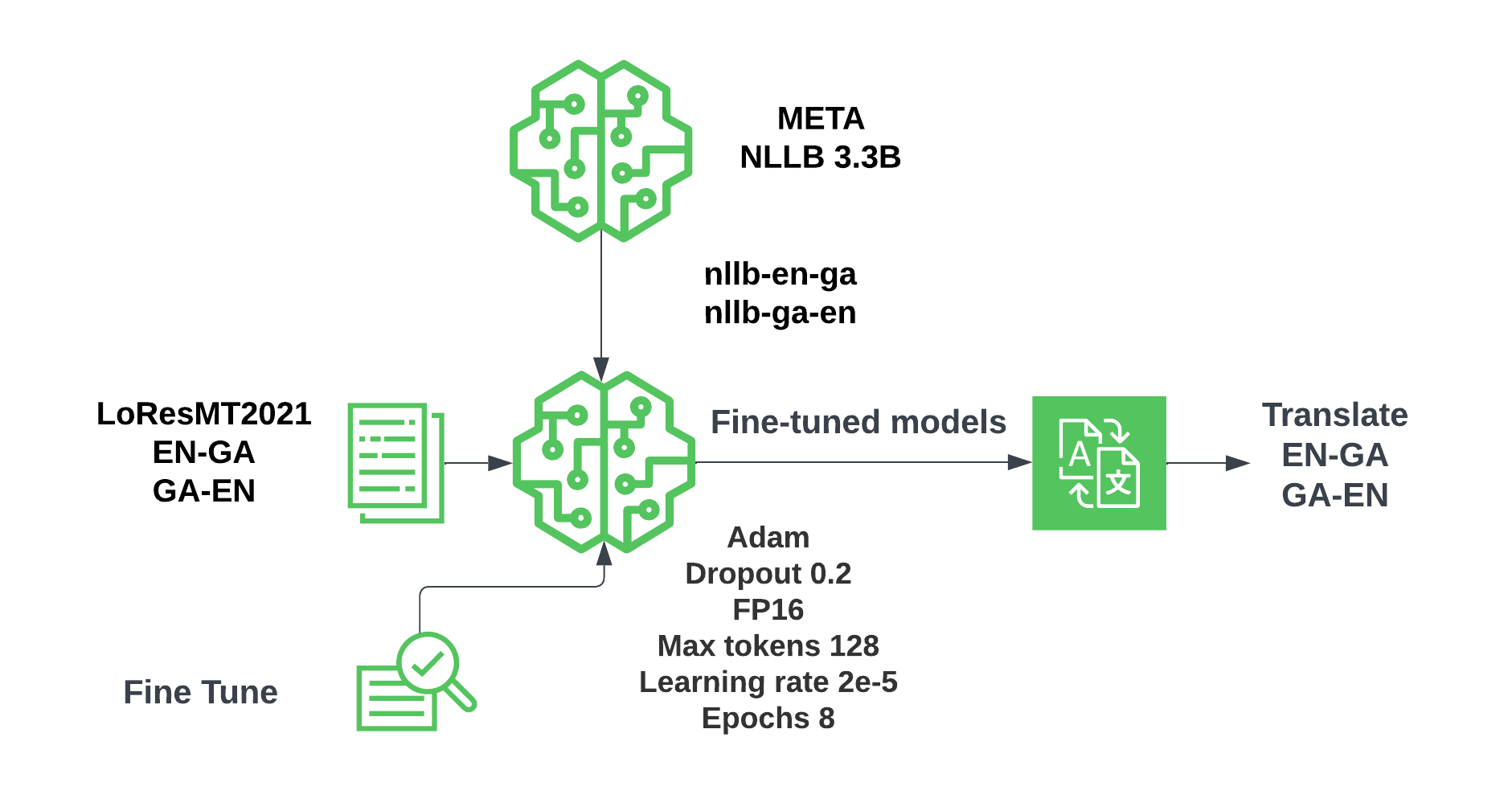}
  \caption{Fine-tuned MLLM approach of adaptMLLM for EN$\leftrightarrow$GA translation}
 \label{fig:adaptMLLMenga}
\end{figure}

As part of the research effort required to answer RQ4, two low-resource language pairs (EN$\leftrightarrow$GA and EN$\leftrightarrow$MR) were chosen for running fine-tuning experiments. The research contribution of this work resulted in the development of the adaptMLLM application which was used to carry out the fine-tuning experiments. The fine-tuned EN$\leftrightarrow$GA and EN$\leftrightarrow$MR MLLM models demonstrated superior translation quality when compared to baselines from the LoResMT2021 Shared Task.

The first suite of tests was conducted on the EN$\leftrightarrow$GA pair and our approach in fine-tuning a pre-trained 3.3B parameter No Language Left Behind (NLLB) model is illustrated in Figure \ref{fig:adaptMLLMenga}. With EN$\rightarrow$GA translation, the following scores were attained: 41.2 BLEU, 0.531 TER and 0.6 ChrF. In the GA$\rightarrow$EN direction, the system achieved 75.1 BLEU, 0.385 TER and 0.71 ChrF which surpassed by a wide margin the winning scores of the shared task. An overview of the NLLB project is outlined in Section \ref{nllb_overview} and a more in-depth discussion of the NLLB architecture is covered in Appendix A.

\begin{figure}[ht!]
  \includegraphics{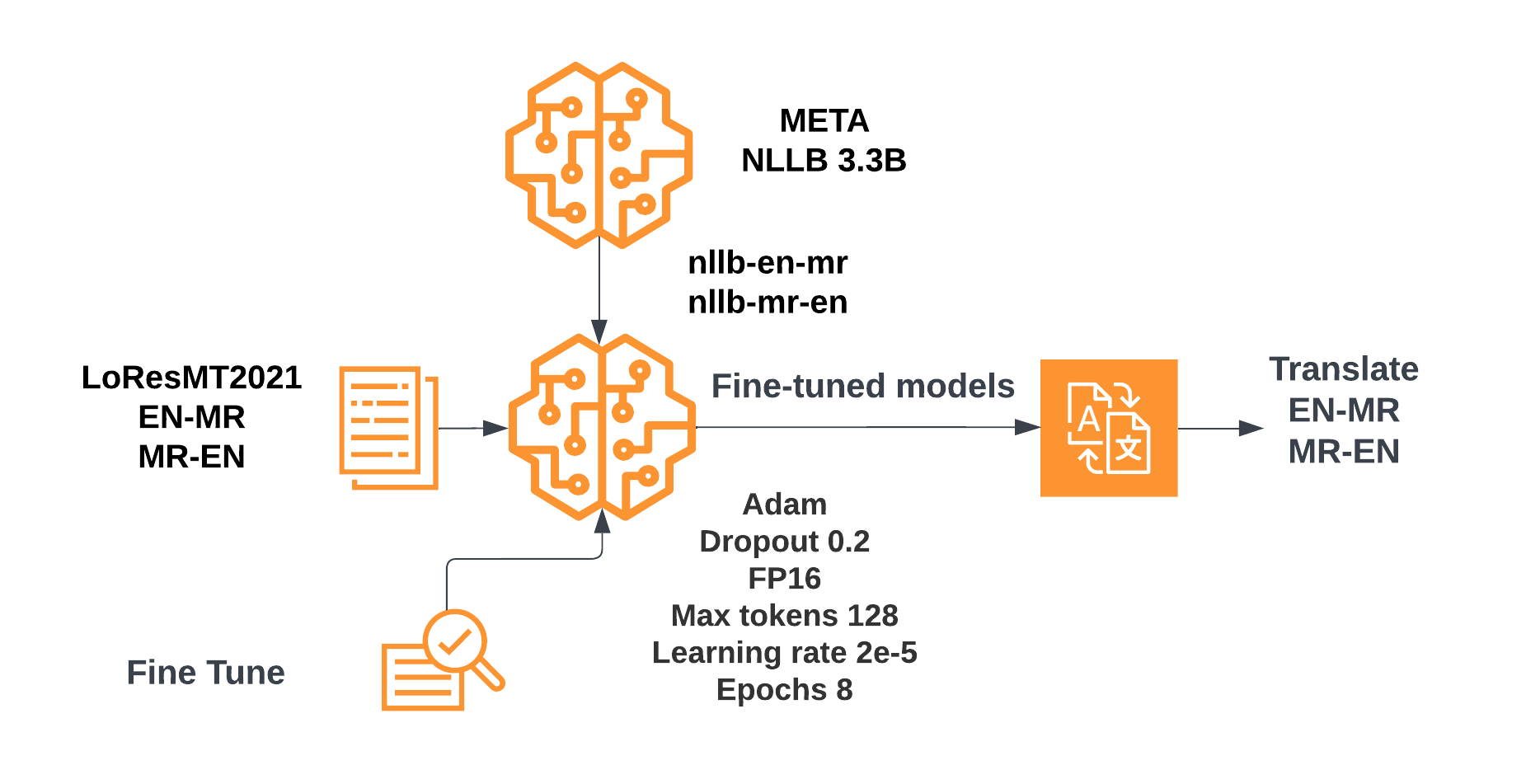}
  \caption{
 Fine-tuned MLLM approach of adaptMLLM for EN$\leftrightarrow$MR translation}
 \label{fig:adaptMLLM-enmrboth}
\end{figure}

The approach taken in fine-tuning a pre-trained 3.3B parameter NLLB model on the EN$\leftrightarrow$MR pair is illustrated in Figure \ref{fig:adaptMLLM-enmrboth}. Again these scores improved upon all previous scores which were achieved in LoResMT2021. In the EN$\rightarrow$MR direction, the following scores were attained: 26.4 BLEU, 0.56 TER and 0.608 ChrF. In the MR$\rightarrow$EN direction 52.6 BLEU, 0.409 TER and 0.704 ChrF were achieved.

\clearpage

   \begin{center}
       \vspace*{1cm}

       \textbf{adaptMLLM: Fine-Tuning Multilingual Language Models on Low-Resource Languages with integrated LLM playgrounds}
            
       \vspace{1.5cm}

       \textbf{Séamus Lankford \\ Haithem Afli \\ Andy Way}

       \vfill
            
       Information, MDPI \\
       November, 2023
       \vspace{0.8cm}
                 
       ADAPT Centre\\
       Dublin City University\\
       Ireland\\

              \vspace{0.5cm}
    \url{https://doi.org/10.3390/info14120638}
            
   \end{center}

\clearpage

\section{Abstract}

The advent of multilingual language models (MLLMs) and large language models (LLMs) has spawned innovation in many areas of natural language processing. Despite the exciting potential of this technology, its impact on developing high-quality machine translation (MT) outputs for low-resource languages remains relatively under-explored. Furthermore, an open-source application, dedicated to both fine-tuning MLLMs and managing the complete MT workflow for low-resources languages, remains unavailable. We aim to address these imbalances through the development of adaptMLLM which streamlines all processes involved in the fine-tuning of MLLMs for MT. This open-source application is tailored for developers, translators, and users who are engaged in MT. It is particularly useful for newcomers to the field, as it significantly streamlines the configuration of the development environment. An intuitive interface allows for easy customisation of hyperparameters, and the application offers a range of metrics for model evaluation and the capability to deploy models as a translation service directly within the application. As a multilingual tool, we used adaptMLLM to fine-tune models for two low-resource language pairs: English to Irish (EN$\leftrightarrow$GA) and English to Marathi (EN$\leftrightarrow$MR). Compared with baselines from the LoResMT2021 Shared Task, the adaptMLLM system demonstrated significant improvements. In the EN$\rightarrow$GA direction, an improvement of 5.2 BLEU points was observed and an increase of 40.5 BLEU points was recorded in the GA$\rightarrow$EN direction representing relative improvements of 14\% and 117\% respectively. Significant improvements in the translation performance of the EN$\leftrightarrow$MR pair were also observed notably in the MR$\rightarrow$EN direction with an increase of 21.3 BLEU points which corresponds to a relative improvement of 68\%. Finally, a fine-grained human evaluation of the MLLM output on the EN$\rightarrow$GA pair was conducted using the Multidimensional Quality Metrics and Scalar Quality Metrics error taxonomies. The application and models are freely available.\footnote{\url{http://github.com/adaptNMT/adaptMLLM}}

\section{Graphical abstract}

\begin{figure}[h]
  \includegraphics[width=15.5cm]{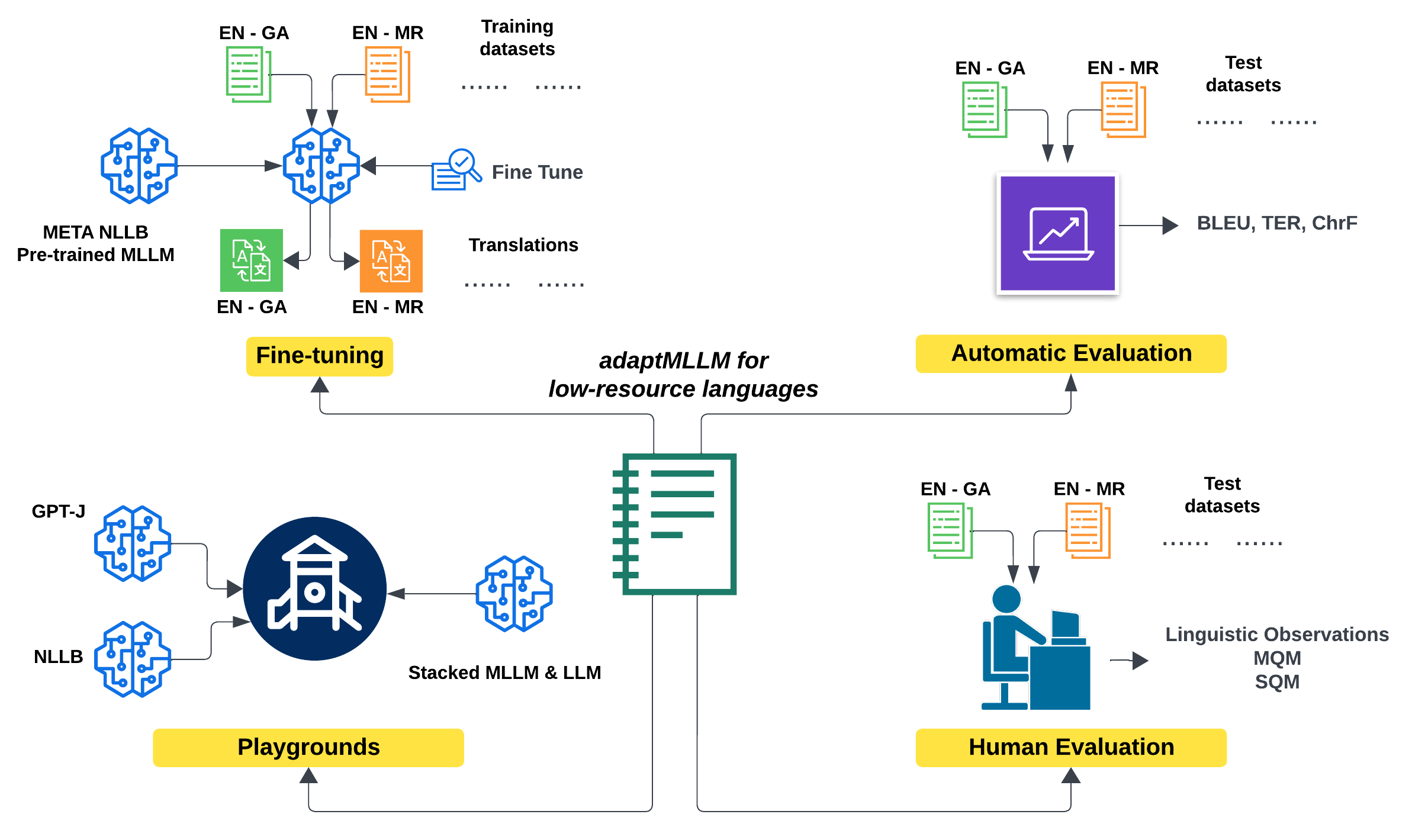}
  \caption{Graphical abstract summarising the adaptMLLM system}
 \label{fig:graphical_arch_mllm}
\end{figure}

\section{Introduction}
\label{Introduction}

Large language models (LLMs), are AI models that use deep learning techniques to generate human-like text. These models are trained on vast amounts of text data, often using unsupervised learning, to learn the patterns and relationships within language. This results in models that can generate text which is often indistinguishable from text written by a human.
   
The excitement surrounding LLMs stems from their potential to revolutionise many fields, from language translation \parencite{costa2022no} and content generation \parencite{brown2020language} to chatbots\footnote{\url{https://openai.com/blog/chatgpt}} and virtual assistants.\footnote{\url{https://genie.stanford.edu/}} With their ability to understand natural language and generate complex responses, LLMs have the potential to enhance human communication and productivity in ways that were previously unimaginable. LLMs can also be used in creative applications, such as generating music\footnote{\url{https://soundraw.io/}} or art.\footnote{\url{https://labs.openai.com/}}

No Language Left Behind (NLLB) \parencite{costa2022no} represents a groundbreaking AI project in the area of multilingual language models (MLLMs). The project has released open-source models proficient in delivering high-quality translations across 200 languages and has enhanced translations for low-resource languages on platforms like Facebook and Instagram. The NLLB-200 model, integrated into the Wikimedia Foundation's Content Translation tool, aids Wikipedia editors in translating content into their preferred languages. These editors can now more effectively translate articles from lesser-known languages, such as Luganda and Icelandic, enriching Wikipedia's language diversity. The open-sourced nature of the NLLB-200 model also empowers the research community and Wikipedia editor groups to expand upon their findings.

When building MLLMs and LLMs, the focus is on designing and training the model architecture. This involves selecting the appropriate neural network architecture and hyperparameters, as well as deciding on the training data and optimisation techniques to use.

Tuning an MLLM or LLM, on the other hand, involves adjusting the parameters of the model to improve its performance on a specific task. In neural networks such as MLLMs and LLMs, the weights and biases are parameters that the network adjusts through training to minimise a cost function. This is done by training the model on a task-specific dataset and adjusting the model's hyperparameters to optimise its performance. Tuning an MLLM can be a challenging task, as the model is often very complex and the training process can take a long time. Our paper concentrates on fine-tuning pre-built MLLMs to enhance machine translation (MT) with a particular focus on low-resource language pairs. 

The process of fine-tuning an MLLM involves several distinct stages which are broken down into individual steps. These steps include setting up the environment, preparing the dataset, parameterising and fine-tuning the chosen MLLM, and evaluating and deploying the model. This modular approach has proven to be effective in fine-tuning MLLMs and we have structured our adaptMLLM application to cater for both developers and translators. In light of the environmental impact of developing and running large AI models \parencite{strubell-etal-2019-energy,henderson2020towards,info13020088}, we also calculate carbon emissions in a ``green report''. It is envisaged that such a report will incentivise more responsible and sustainable model development.

A significant aspect of our research involves creating applications and models to address language technology challenges. Similar to our previous work which focused on developing NMT models \parencite{lankfordlrev}, we hope this paper will be particularly helpful for those new to MT wishing to learn more about fine-tuning MLLMs.

Unlike many translation toolkits, our application does not use a command line interface. Instead, we have designed and fully implemented the interface in Google Colab,\footnote{\url{https://colab.research.google.com}} a cloud-hosted solution\footnote{\url{https://cloud.google.com}} that is more intuitive for both educational and research settings. Furthermore, our application provides graphical user interface (GUI) controls within adaptMLLM, enabling users to customise all key hyperparameters required for MLLMs.

Our application is designed to operate as a platform as a service (PaaS) cloud computing application, allowing for quick and efficient scaling of the infrastructure. Additionally, the deploy function allows for immediate deployment of trained models.

This paper is organised by initially presenting related work and background information on MLLMs and LLMs in Section \ref{related}. This is followed by a description of our datasets in Section \ref{approach}. The key features of the adaptMLLM architecture are discussed in Section \ref{aLLM} and an empirical evaluation of our trained models, including a human evaluation is carried out in Section \ref{sec:exp_mllm}. The system is discussed in Section \ref{sec:discussion} before drawing conclusions and describing future work in Section \ref{concl}.

\section{Related Work}
\label{related}

\subsection{Transformer Architecture}

Following the introduction of the attention mechanism, a natural line of investigation was to see whether attention could do most of the heavy lifting of translation by itself. Accordingly, \parencite{vaswani2017attention} proposed that  ``attention is all you need'' in their Transformer architecture, which has achieved state-of-the-art (SOTA) performance on many natural language processing (NLP) benchmarks by relying solely on an attention mechanism,  removing recurrence and convolution while allowing the use of much simpler feed-forward neural networks. In the context of our research, we have previously demonstrated that Transformer-based models deliver high-functioning models for the low-resource EN$\rightarrow$GA language pair \parencite{lankford2021transformer}.

The default Transformer architecture follows an encoder-decoder structure generating its output without relying on recurrence and convolutions. The task of the encoder is to map an input sequence to a sequence of continuous representations, which is then passed to a decoder. The decoder generates an output sequence by using the output of the encoder together with the decoder output from the previous time step.

\subsection{Multilingual Language Models - NLLB}\label{nllb_overview}

MT has become a significant area of research in AI to eliminate language barriers worldwide. However, the current focus is limited to a small number of languages, neglecting the vast majority of low-resource languages. To address this issue, the NLLB initiative was launched. This project aims to overcome the challenges of using MT for low-resource language translation by developing datasets and models that bridge the performance gap between low- and high-resource languages. The NLLB team has also created architectural and training enhancements tailored to support MT for low-resource languages. Their work is open-source,\footnote{\url{https://github.com/facebookresearch/fairseq/tree/nllb}} and many of their models serve as baselines for fine-tuning with adaptMLLM.

\subsection{Large Language Models}\label{llm_background}

The increasing availability of large datasets provides the raw material for LLM training \parencite{radford2019language,conneau-etal-2020-unsupervised,winata-etal-2021-language}
enabling performance improvement on NLP tasks which can learn from a wide variety of sources. 

Another key factor in driving the ubiquity of LLMs has been the growth in computational power dedicated to the domain. As a consequence, more powerful computers now have the capability to train LLMs on massive datasets, which in turn has led to SOTA results on many common NLP tasks \parencite{devlin-etal-2019-bert}. New training algorithms developed through advancement in AI research have further boosted LLM performance \parencite{lepikhin2020gshard}. 

LLMs have the potential to improve the use of technology across a wide range of domains, among which include medicine, education and computational linguistics. In education, LLMs may be used for personalised student learning experiences \parencite{KASNECI2023102274} while in the medical domain, analysing large amounts of medical files can assist doctors in treating patients \parencite{iftikhar2023docgpt}. 
Of particular interest to our research is how LLMs can be used within the realm of computational linguistics, more specifically in the field of MT. 

\subsubsection{GPT-J}

Transformers are increasingly the architecture of choice for NLP problems, replacing recurrent neural networks (RNNs) such as long short-term memory (LSTM) \parencite{hochreiter1997long}. 

GPT-J is an open-source implementation of a particular class of LLMs known as generative pre-trained transformer (GPT) models \parencite{radford2018improving}. GPT-J is a Transformer model trained using Wang's Mesh Transformer JAX.\footnote{\url{https://github.com/kingoflolz/mesh-transformer-jax}} GPT-J-6B\footnote{\url{https://6b.eleuther.ai}} is an autoregressive language model, created by EleutherAI,\footnote{\url{https://www.eleuther.ai}} with 6 billion trainable parameters. As an advanced alternative to OpenAI's GPT-3, it performs very well on a wide array of NLP tasks such as chat, summarisation, and question answering. 

\subsubsection{GPT-4}

The primary distinction between GPT-3.5 and GPT-4\footnote{\url{https://openai.com/product/gpt-4}} is that while the former is a text-to-text model, the latter is more of a data-to-text model, exhibiting the ability to perform tasks that its predecessor could not. For example, GPT-4 is capable of processing visual input as part of a prompt, such as images or web pages, and can even generate text that explains the humour in memes. Consequently, GPT-4 can be classified as a ``multimodal model''. Furthermore, GPT-4 has a longer memory than its previous versions, with a short-term memory closer to 64,000 words, enabling it to maintain coherence during extended interactions. GPT-4 also enables users to select different personalities for the model's responses.

The number of parameters utilised in the training of GPT-4 has not been disclosed by OpenAI; however, other sources, such as AX Semantics,\footnote{\url{https://en.ax-semantics.com/}} have estimated the number to be around 100 trillion. AX Semantics maintains that such a number makes the language model more akin to the functioning of the human brain for language and logic.\footnote{\url{https://en.ax-semantics.com/blog/gpt-4-and-whats-different-from-gpt-3/}}

Additionally, GPT-4 outperformed GPT-3.5 in various standardised tests, such as the LSAT, SAT, Uniform Bar Exam, and GRE, and was shown to be 82\% less likely to respond when prompted inappropriately and 60\% less likely to generate false information \parencite{OpenAI2023GPT4TR}.
 
\subsubsection{BARD}

BARD\footnote{\url{https://bard.google.com/}} utilises a lightweight version of the Language Model for Dialogue Applications (LaMDA) \parencite{thoppilan2022lamda}, which is an AI engine developed by Google. BARD has two primary objectives: to ensure the accuracy of its responses and to integrate the benefits of AI into its everyday products. Google has a rich history of employing AI to improve the search experience for billions of users. Its earlier Transformer model, BERT,\footnote{\url{https://github.com/google-research/bert}} was a breakthrough in comprehending the intricacies of human language. The company has since introduced MUM,\footnote{\url{https://blog.google/products/search/introducing-mum/}} which is a thousand times more potent than BERT. Recent AI technologies like LaMDA, PaLM, Imagen, and MusicLM are building on these developments, creating new ways to interact with information from language and images to video and audio. Furthermore, in 2018, Google was one of the pioneering companies to release a set of AI principles.\footnote{\url{https://ai.google/principles/}}

Apart from its products, Google aims to assist developers in innovating with AI by simplifying and scaling the benefits of these advances. In the future, the company intends to create a suite of tools and APIs that will make it easier to build innovative applications with BARD and more generally with its AI.

\subsection{DeepSpeed}\label{deepspeed}

The advent of DeepSpeed \parencite{10.1145/3394486.3406703}, a free software library from Microsoft, was a significant breakthrough for researchers looking to implement and fine-tune MLLMs and LLMs with limited resources. Large model training, in terms of scale, speed, and cost, is now achievable for most people. Additionally, DeepSpeed's most recent Transformer kernel improvements enabled the DeepSpeed team to achieve SOTA performance, setting a new record for the fastest BERT \parencite{devlin-etal-2019-bert} pre-training.

For small teams, DeepSpeed's Zero Redundancy Optimizer (ZeRO) is particularly advantageous, providing fresh memory optimisation for large-scale distributed deep learning. With minor changes to a PyTorch model, DeepSpeed can improve the speed and scale of model training.

\subsection{HuggingFace}\label{hf}

The Hugging Face Transformers library\footnote{\url{https://github.com/huggingface/transformers}} \parencite{wolf-etal-2020-transformers} is an open-source software library that provides a wide range of pre-trained SOTA NLP models, including models for language modelling, question answering, text classification, and MT, among others.
 
The library is built on top of popular deep learning frameworks such as PyTorch\footnote{\url{https://github.com/pytorch/pytorch}} and TensorFlow,\footnote{\url{https://github.com/tensorflow/tensorflow}} and it provides a simple and consistent API for accessing pre-trained models and fine-tuning them for downstream tasks. The library also includes a set of tools for data pre-processing, model evaluation, and visualisation, which make it easier for researchers and developers to experiment with different NLP models and tasks.

The Hugging Face Transformers library has become one of the most popular and widely used NLP libraries in the industry and the research community, and it has been adopted by many companies and organisations to build NLP applications and systems.

\subsection{Human Evaluation}

Within the fields of NLP and MT, human evaluation is increasingly recognised as critical, often meriting its own specialised research track or workshop at leading conferences \parencite{humeval-2021-human}. This emphasis has spurred a wealth of studies focusing on human evaluation related to MT, proving especially valuable in assessing low-resource languages \parencite{bayon-sanchez-gijon-2019-evaluating,imankulova2019exploiting}.

A set of best practices for human evaluation in MT has emerged, detailed in a collection of suggested guidelines \parencite{laubli2020set}. Our study incorporates these guidelines, aligning with comparable EN${\leftrightarrow}$GA studies at the ADAPT centre. To enhance these guidelines, a detailed human analysis was conducted, employing both the Scalar Quality Metric (SQM) \parencite{freitag-etal-2021-experts} and the Multidimensional Quality Metric (MQM) \parencite{lommel2014multidimensional} for a nuanced assessment. SQM and MQM, are both widely used in industry and academia, to evaluate the quality of machine-generated text.

SQM is a simple, single-number metric that is used to measure the overall MT quality. It is often used when a quick evaluation of the quality of the text is required.

MQM, on the other hand, is a more complex metric that measures the quality of the text across multiple dimensions such as fluency, adequacy, and coherence, to name a few. It provides a more comprehensive evaluation of MT by measuring the quality of the text across different aspects.

\section{Datasets}\label{approach}

\subsection{Language Pairs}

To evaluate the translation performance of adaptMLLM in fine-tuning MLLMs for low-resource languages, we had to choose suitable language pairs. Furthermore, appropriate datasets upon which we could benchmark our performance also had to be sourced. The EN${\leftrightarrow}$GA and EN${\leftrightarrow}$MR language pairs were selected since they fulfilled the criteria of low-resource languages.

The Irish language, also known as Irish Gaelic, is the first official language of the Republic of Ireland and is also recognised as a minority language in Northern Ireland. According to the 2022 Irish census,\footnote{\url{https://www.cso.ie/en/releasesandpublications/ep/p-cpsr/censusofpopulation2022-summaryresults/educationandirishlanguage/}} 1.87 million people in the Republic of Ireland reported being able to speak Irish to some degree, which represents 40.4\% of the population. Irish is also spoken by a small number of people in other countries, particularly in the United States, Canada, and Australia, as well as in Irish-speaking communities in other parts of the world. It is also one of the official languages of the European Union and a recognised minority language in Northern Ireland with an ISO code of ``GA''.\footnote{\url{https://www.iso.org/}}

The dominant language spoken in India’s Maharashtra state is Marathi with an ISO code of ``MR''. It has over 83 million speakers and is a member of the Indo-Aryan language family. Despite being spoken by a significant number of people, Marathi is considered to be relatively under-resourced when compared to other languages used in the region.

\subsection{Shared Task Datasets}

To benchmark the performance of our EN${\leftrightarrow}$GA models, trained using adaptMLLM, datasets from the LoResMT2021 Shared Task\footnote{\url{https://github.com/loresmt/loresmt-2021}} \parencite{ojha2021findings} were used. These datasets enabled the evaluation of adaptMLLM models since the shared task focused on low-resource languages which included both the EN${\leftrightarrow}$GA pair and the EN${\leftrightarrow}$MR pair. Furthermore, using official datasets from a shared task enables the direct comparison of our models' performance with models entered by other teams. 

Both datasets focused on the specific domain of translation of Covid-related data. A parallel corpus of EN${\leftrightarrow}$GA sentences concentrating on the Covid domain were mainly drawn from the Government of Ireland\footnote{\url{https://www.gov.ie/}} and the Health Service Executive\footnote{\url{https://www.hse.ie/}} websites. EN${\leftrightarrow}$MR parallel Covid sentences were extracted from the Government of India\footnote{\url{https://www.mygov.in/}} website, BBC Marathi\footnote{\url{https://www.bbc.com/marathi}} and online newspapers. A detailed breakdown of all sources is available in \cite{ojha2021findings}.

The datasets from the shared task provided 502 Irish and 500 Marathi validation sentences whereas 250 GA${\rightarrow}$EN, 500 EN${\rightarrow}$GA, 500 EN${\leftrightarrow}$MR sentences were made available in the test datasets i.e. exactly the same as our other experiments to allow direct comparison with previous work. Training data consisted of 20,933 lines of parallel data for the EN${\leftrightarrow}$MR language pair and 13,171 lines of parallel data were used to train the EN${\leftrightarrow}$GA models.

\section{Approach}\label{aLLM}
\label{disc}

\begin{figure}[h]
  \includegraphics[width=15.5cm]{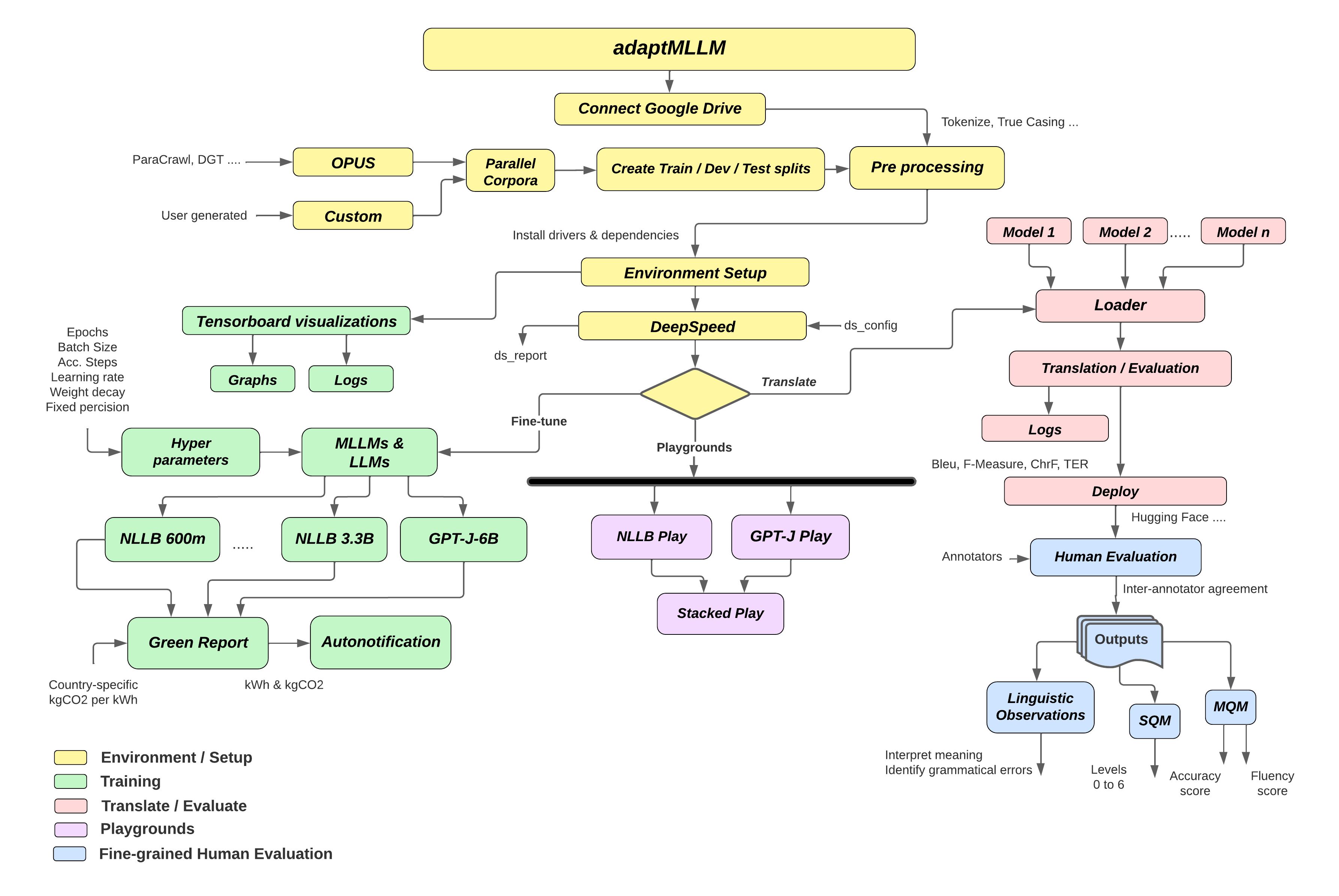}
  \caption{Proposed architecture for adaptMLLM: a system for fine-tuning MLLMs}
 \label{fig:arch_mllm}
\end{figure}

After outlining the background that gave rise to the creation of MLLMs and LLMs, we now introduce the adaptMLLM tool. This tool allows users to customise these components to their liking. Figure \ref{fig:arch_mllm} offers a high-level overview of the platform's system architecture. 

The application is designed as an IPython notebook and employs Pytorch for model training. The utilisation of a Jupyter notebook format facilitates easy sharing within the AI community. Additionally, the challenge of configuring the proper development environment is substantially reduced, as all necessary packages are automatically downloaded while the application is running. 

\begin{figure} [htp]
\begin{center}

\includegraphics[height=22cm]{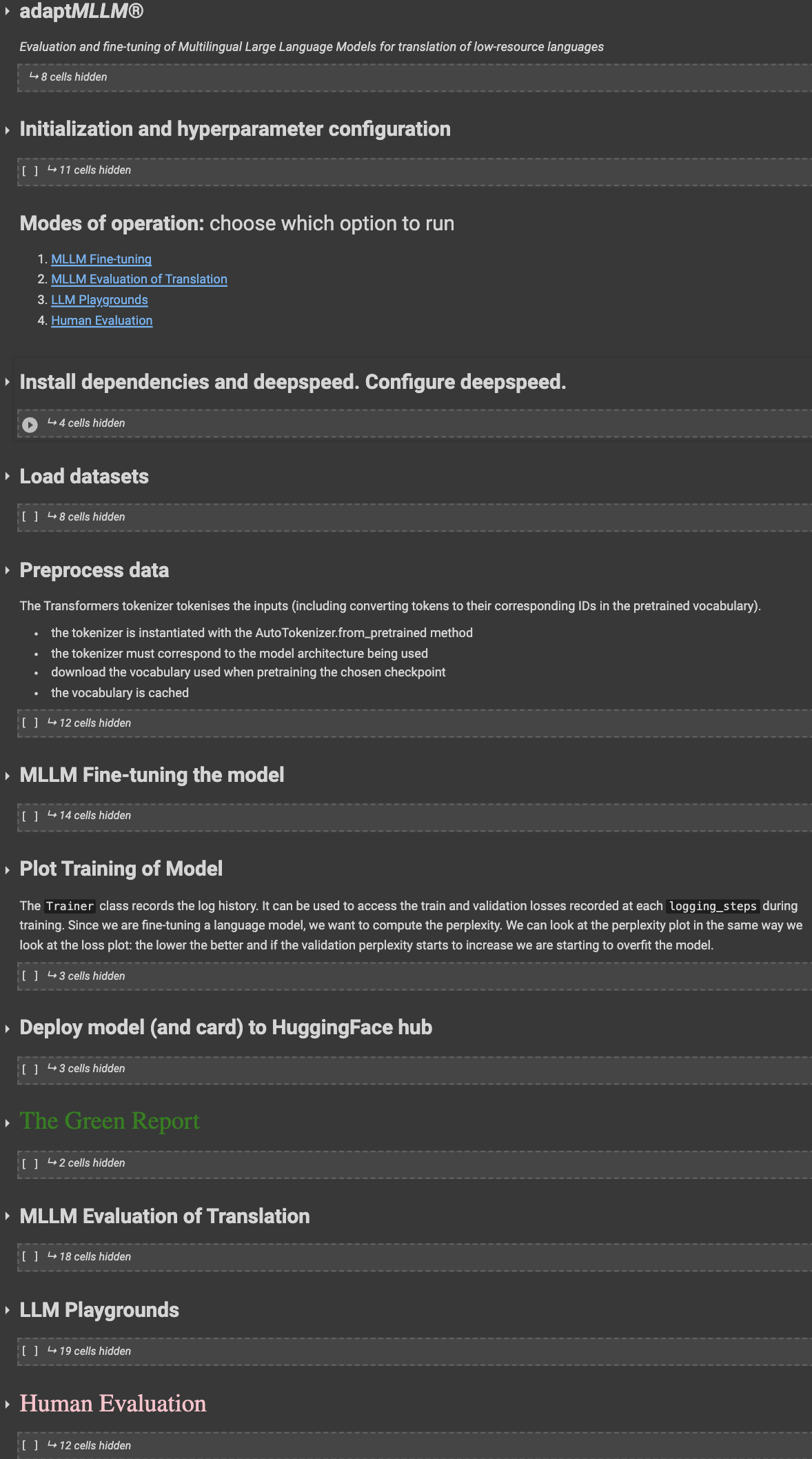}
  \caption [Overview of adaptMLLM] {Overview of adaptMLLM. Key areas include initialisation, a menu of operation modes, loading and pre-processing, MLLM fine-tuning, visualisation, deployment, a green report, MLLM translation and evaluation, LLM playgrounds and a human evaluation (cf. Section \ref{aLLM}).}
  \label{fig:features}

\end{center}
\end{figure}

There are options to run the system for fine-tuning MLLMs, evaluating MLLM translation performance, testing LLM playgrounds and conducting a human evaluation of the translation performance. The application is run as a Colab instance on the Google Cloud.  Translation models are developed using aligned text corpora from both the original and the target languages. Tensorboard offers a live graphical representation of the training process of the model. The system is primarily employed for training models and functioning as a translation service, either of which can be chosen at run-time.

The application is primarily run as a Google Colab application but may also be run as a Jupyter notebook. Given the ease of integrating Google Drive storage into Colab, we have used adaptMLLM exclusively as a Google Colab application for our experiments, some of which are described in Section \ref{sec:exp_mllm}. Key features of the notebook are highlighted in Figure \ref{fig:features}.

\subsection{Initialisation and Pre-processing}

Initialisation enables connection to Google Drive to run experiments, automatic installation of Python, SentencePiece\footnote{\url{ https://github.com/google/sentencepiece}} \parencite{kudo2018sentencepiece}, Pytorch, HuggingFace Transformer's library (cf. Section \ref{hf}) and other libraries. 

The training, validation and test splits for both source and target languages may be uploaded by the users. In cases where a user has not already created the required splits for model training, single source and target files may be uploaded. The necessary splits to form the training, validation, and test files will be automatically created based on the split ratio specified by the user. 

\subsection{Modes of Operation}
There are several modes of operation, namely MLLM fine-tuning, evaluation of MLLM translation performance, experimentation with LLM playgrounds and a human evaluation of the translation output. 

With MLLM fine-tuning, the application develops models using Google's GPU-based cloud platform. For a monthly subscription, the Google Colab Pro+ is a prerequisite since fine-tuning demands access to high-end GPU and compute resources.  

Apart from low-cost access to a high-specification infrastructure, model development on the Google Cloud is also recommended given the platform uses 100\% renewables \parencite{lacoste2019quantifying}. This has emerged as an economical choice for practitioners in the field of low-resource languages, as the creation of smaller models involves reduced training times.

\subsection{Fine-tuning and Visualisation}

The system has been designed to enable users to choose variations of the base MLLM architecture. In the current release, users can choose to fine-tune the following baselines: i) NLLB-200-600M, ii) NLLB-200-1.3M, iii) NLLB-200-3.3B or iv) a user-specified baseline. The fine-tuning mode allows users to specify, using GUI controls, the exact hyperparameters required for the chosen approach.

The visualisation segment provides live graphing of model progression, allowing for the monitoring of model convergence. All log files are preserved and accessible for review to examine the training convergence, as well as to evaluate the model's accuracy during the training and validation phases.

\subsection{Deployment}

Gradio\footnote{\url{https://gradio.app/}} \parencite{abid2019gradio} is an open-source Python library that enables the development of easy-to-use web applications for ML models. The library integrates with the most popular Python libraries, including Scikit-learn and PyTorch. 

A key advantage is that it allows interaction with a web app developed for a Jupyter or Colab notebook. Consequently, it was selected as the library used for the deployment of our custom fine-tuned models. 

\subsection{Green Report}

In recent years, the ecological footprint of technology, along with the assessment of its impacts, has become increasingly prominent \parencite{henderson2020towards}. Indeed, this may be viewed as a natural response to truly massive NLP models which have been developed by large multinational corporations with little apparent regard for their environmental impact.

Specifically, HPO for finely-tuned MLLMs can be especially demanding when the fine-tuning of hyperparameters spans a wide search space.

Consequently, a wide array of tools for assessing NLP's carbon footprint has been created \parencite{bannour-etal-2021-evaluating}, and the idea of sustainable NLP has emerged as a significant area of research. This has been recognised at numerous prestigious conferences, for instance, the Green and Sustainable NLP track at EACL 2021.\footnote{\url{https://2021.eacl.org/news/green-and-sustainable-nlp}}

Reflecting these advancements, adaptMLLM has integrated a ``green report'' feature that records the kgCO\textsubscript2 emitted during the development of the model. This aligns closely with the current industry movement towards measuring the environmental impact of NLP activities; indeed, \parencite{info13020088} have demonstrated that high-performing MT systems can be built with much lower footprints, which not only reduce emissions but also in the post-deployment phase deliver savings of almost 50\% in energy costs for a real translation company.

\subsection{MLLM Translation and Evaluation}

Besides facilitating model fine-tuning, the application also provides functionality for translation and assessing model performance. The use of pre-trained models for translation is also parameterised; users specify the model's name as a hyperparameter, which is then used to perform translation and evaluation on the test files.

After building the system, users can select the model they wish to use for translation of the test set. While human judgement is often the most reliable for assessing translation quality, human evaluators are not always accessible, may have differing opinions, and can be costly to engage for experimental purposes. As a result, automatic evaluation metrics are commonly employed, particularly by developers who are tracking the step-by-step advancement of their systems (cf. \cite{Way2018} for more on the pros and cons of human and automatic evaluation).

Several automatic evaluation metrics provided by SacreBleu\footnote{\url{https://github.com/mjpost/sacrebleu}} \parencite{post2018call} are used: BLEU \parencite{papineni-etal-2002-bleu}, TER \parencite{snover2006study} and ChrF \parencite{popovic2015chrf}. Translation quality can also be evaluated using Meteor \parencite{denkowski2014meteor} and F1 score \parencite{melamed-etal-2003-precision}. 

It's important to recognise that BLEU, ChrF, Meteor, and F1 are metrics based on precision, thus higher values signify better performance. On the other hand, TER is a metric based on errors, with lower values denoting superior translation quality. The available evaluation options include standard (truecase) and lowercase BLEU scores, along with sentence-level BLEU scoring, as well as ChrF1 and ChrF3.

Logging occurs at three tiers: a model development log for charting progress, an output log from the training console, and a log of the evaluation outcomes. Additionally, there is a references section that provides materials pertinent to the development, utilisation, and comprehension of adaptMLLM. Presently, validation throughout the training process is performed based on model loss.

\subsection{LLM Playgrounds}

When OpenAI\footnote{\url{https://openai.com/}} released a playground for its GPT-3 model, the community was quick to create demos. Given that OpenAI's GPT-3 is proprietary, generating text using its API would incorporate a cost and involve sending data to the site. Ideally, we sought to host an open-source text generation model, and associated playground app in our environment. 

In 2021, Eleuther AI created GPT-J, an open-source text generation model to rival GPT-3 and the model is freely available on the Hugging Face Model Hub allowing us to download variations of this model. In this spirit, we have developed our own fully customisable text generation playground using GPT-J. Using Gradio, a web interface that can interact with these GPT models was developed. 

\section{Empirical Evaluation}
\label{sec:exp_mllm}
After outlining the theoretical framework and the tool itself, we proceed to assess the efficacy of the adaptMLLM methodology by training models for the EN${\leftrightarrow}$GA and the EN${\leftrightarrow}$MR language pairs.   

\subsection{Infrastructure and Hyperparameters}

A Google Colab Pro+ subscription facilitated the rapid development of prototypes using NVIDIA 40GB GPU graphics cards (A100-SXM4-40GB) and compute resources of up to 89GB of system memory when available \parencite{Bisong2019}. All MT models were trained using the adaptMLLM application.

The DeepSpeed library (cf. Section \ref{deepspeed}) is a critical component in making the adaptMLLM system work since it enables our models to be loaded across both GPU and system memory. Without such a library, very significant compute resources would be required which would be prohibitively costly for our team to hire. The hyperparameters used for developing models for both language pairs are outlined in Table \ref{tab:hpo-table-mllm}.

\begin{table} [hbt!]
\centering
\begin{tabular}{ll}
\hline
\textbf{Hyperparameter} & \textbf{Values}      \\ \hline
Epochs  & 1, 3, \textbf{5} \\ \hline
Batch size        & 8, 12, \textbf{16}       \\ \hline
Gradient accumulation steps  & 2,  4, \textbf{8}   \\ \hline
Learning rate         & 1e-5, \textbf{3e-5}, 9e-5  \\ \hline
Weight decay   & 0.01, \textbf{0.1}, 1, 2 \\ \hline
Mixed precision   &   False, \textbf{True}       \\ \hline
\end{tabular}
\caption[Hyperparameter Optimisation]{HPO with optimal hyperparameters, within the search space, are highlighted in bold.}

\label{tab:hpo-table-mllm}
\end{table}

\subsection{Results: Automatic Evaluation}

To determine the quality of our translations, automated metrics were employed. For comparison with our prior studies, the performance of models was gauged using three evaluative metrics: BLEU, TER, and ChrF. These metrics reflect the precision of translations produced by our fine-tuned MLLM systems. We report case-insensitive BLEU scores at the corpus level.

\subsubsection{Translation in the EN$\leftrightarrow$GA directions}
The experimental results from the LoResMT2021 Shared Task in the EN$\leftrightarrow$GA directions are summarised in Tables \ref{tab:en2ga}--\ref{tab:ga2en} and are compared with our experimental findings which adaptMLLM achieved by fine-tuning a 3.3B parameter NLLB MLLM. 

The highest-performing EN$\rightarrow$GA system in the LoResMT2021 Shared Task was submitted by the ADAPT team \parencite{lankford2021machine}. The model was developed with an in-house application, adaptNMT \parencite{lankfordlrev} using a Transformer architecture. It performed well across all key translation metrics (BLEU: 36.0, TER: 0.531 and ChrF3: 0.6). 

Subsequently, these results were improved upon (BLEU: 37.6, TER: 0.577 and ChrF3: 0.57) by training a Transformer model on a bespoke health dataset, gaHealth \parencite{lankford2022lrec}.

\begin{table}[h]
\centering
\begin{tabular}{lcccccc}
\hline
\textbf{Team} &
 \textbf{System} &
  \textbf{BLEU} $\uparrow$ &
  \textbf{TER} $\downarrow$ &
  \textbf{ChrF3} $\uparrow$ \\ \hline
adaptMLLM & en2ga-tuned & \textbf{41.2} & \textbf{0.51} & \textbf{0.48} \\
adapt & covid\_extended & 36.0 & 0.531 & 0.60 \\
adapt & combined & 32.8 & 0.590 & 0.57 \\
adaptMLLM & en2ga-baseline & 29.7 & 0.595 & 0.559 \\
IIITT  & en2ga-b & 25.8 & 0.629 & 0.53 \\
UCF     & en2ga-b & 13.5 & 0.756 & 0.37   \\
\hline
\end{tabular}
\caption[EN${\rightarrow}$GA: adaptMLLM systems compared with LoResMT2021]{EN${\rightarrow}$GA: adaptMLLM systems compared with LoResMT2021. The impact of fine-tuning the baseline NLLB model is evident with the BLEU score rising from 29.7 to 41.2 representing a 39\% relative improvement. Models developed using adaptMLLM were trained using the optimal hyperparameters set out in Table \ref{tab:hpo-table-mllm}.}
\label{tab:en2ga}
\end{table}

By fine-tuning the NLLB MLLM, using the hyperparameters outlined in Table \ref{tab:hpo-table-mllm}, a significant improvement in translation performance was achieved. The adaptMLLM EN$\rightarrow$GA en2ga system, shown in Table \ref{tab:en2ga}, achieves a BLEU score of 41.2 which is 5.2 BLEU points higher than our previous score which won the shared task in 2021. This represents a relative improvement of 14\%.
 
\par
For translation in the GA$\rightarrow$EN direction, illustrated in Table \ref{tab:ga2en}, the best-performing model for the LoResMT2021 Shared Task was developed by IIITT with a BLEU of 34.6, a TER of 0.586 and ChrF3 of 0.6. Accordingly, this serves as the baseline score by which we can benchmark our GA$\rightarrow$EN model, developed by fine-tuning a 3.3B parameter NLLB using adaptMLLM. Similar to the results achieved in the EN$\rightarrow$GA direction, significant translation performance was observed using this new method. The performance of the adaptMLLM model offers an improvement across all metrics with a BLEU score of 75.1, a TER of 0.385 and a ChrF3 result of 0.71. In particular, the 117\% relative improvement in BLEU score against the IIITT system is very significant. The adaptMLLM model is a fine-tuned pre-trained NLLB 3.3B parameter MLLM whereas the IIITT model fine-tuned a smaller Opus MT model from Helsinki NLP. MLLMs and LLMs have already learned to represent natural language patterns and structures from large amounts of data, which can be adapted to specific tasks or domains by updating the model's parameters with a smaller amount of annotated data. The effect of this approach is demonstrated in the substantially higher BLEU achieved by the adaptMLLM model relative to the IIITT model which was trained on a much smaller Opus model.

The improvement in translation performance is real and not just a BLEU score anomaly given that large improvements were simultaneously observed across the BLEU, TER and CHRF metrics. More specifically, Meta's nllb-200-3.3B model has a memory footprint of 17.58GB enabling 3.3 billion parameters to be trained compared to the Helsinki-NLP model, opus-mt-ga-en, which is just 295MB and has a correspondingly much smaller set of trainable parameters. Another aspect differentiating the adaptMLLM approach is the relatively broad hyperparameter search space compared to other teams outlined in Tables \ref{tab:en2ga}--\ref{tab:ga2en}. We experimented with the number of epochs, the batch size, the gradient accumulation steps, the learning rate, the weight decay and the type of precision used. The exact hyperparameters used are illustrated in Table \ref{tab:hpo-table-mllm}.

\begin{table*}[ht!]
\centering
\begin{tabular}{lcccccc}
\hline
\textbf{Team} &
 \textbf{System} &
  \textbf{BLEU} $\uparrow$ &
  \textbf{TER} $\downarrow$ &
  \textbf{ChrF3} $\uparrow$ \\ \hline
adaptMLLM & ga2en-tuned & \textbf{75.1} & \textbf{0.385} & \textbf{0.71}\\
adaptMLLM & ga2en-baseline & 47.8 & 0.442 & 0.692 \\
IIITT  & ga2en-b & 34.6 & 0.586 & 0.61\\
UCF & ga2en-b & 21.3 & 0.711 & 0.45\\
\hline
\end{tabular} 
\caption[GA${\rightarrow}$EN: adaptMLLM systems compared with LoResMT2021]{GA${\rightarrow}$EN: adaptMLLM systems compared with LoResMT2021. The impact of fine-tuning the baseline NLLB model is evident with the BLEU score rising from 47.8 to 75.1 representing a 57\% relative improvement. Models developed using adaptMLLM were trained using the optimal hyperparameters set out in Table \ref{tab:hpo-table-mllm}.}
\label{tab:ga2en}
\end{table*}

\subsubsection{Translation in the EN$\leftrightarrow$MR directions}

The experimental results from the LoResMT2021 Shared Task in the EN$\leftrightarrow$MR directions are summarised in Tables \ref{tab:en2mr}--\ref{tab:mr2en} and are compared with our experimental findings in developing adaptMLLM. For the shared task, the highest-performing EN$\rightarrow$MR system was submitted by the IIITT team. Their model used a Transformer architecture and achieved a BLEU score of 34.6, a TER of 0.586 and ChrF3 of 0.61. 

\begin{table*}[ht!]
\centering
\begin{tabular}{lcccccc}
\hline
\textbf{Team} &
 \textbf{System} &
  \textbf{BLEU} $\uparrow$ &
  \textbf{TER} $\downarrow$ &
  \textbf{ChrF3} $\uparrow$ \\ \hline
adaptMLLM & en2mr-tuned & \textbf{26.4} & \textbf{0.56} & \textbf{0.608} \\  
IIITT & en2mr-IndicTrans-b & 24.2 & 0.59 & 0.597   \\
oneNLP-IIITH & en2mr-Method2-c & 22.2 & 0.56 & 0.746 \\
oneNLP-IIITH & en2mr-Method3-c & 22.0 & 0.56 & 0.753 \\
oneNLP-IIITH & en2mr-Method1-c & 21.5 & 0.56 & 0.746 \\
adaptMLLM & en2mr-baseline & 19.8 & 0.656 & 0.57 \\
adaptNMT & en2mr & 13.7 & 0.778 & 0.393\\
\hline
\end{tabular}
\caption[EN${\rightarrow}$MR: adaptMLLM systems compared with LoResMT2021]{EN${\rightarrow}$MR: adaptMLLM systems compared with LoResMT2021. The impact of fine-tuning the baseline NLLB model is evident with the BLEU score rising from 19.8 to 26.4 representing a 33\% relative improvement. Models developed using adaptMLLM were trained using the optimal hyperparameters set out in Table \ref{tab:hpo-table-mllm}.}
\label{tab:en2mr}
\end{table*}

\begin{table*}[ht!]
\centering
\begin{tabular}{lcccccc}
\hline
\textbf{Team} &
 \textbf{System} &
  \textbf{BLEU} $\uparrow$ &
  \textbf{TER} $\downarrow$ &
  \textbf{ChrF3} $\uparrow$ \\ \hline
adaptMLLM & mr2en-tuned & \textbf{52.6} & \textbf{0.409} & \textbf{0.704}\\  
adaptMLLM & mr2en-baseline & 42.7 & 0.506 & 0.639 \\
oneNLP-IIITH & mr2en-Method3-c & 31.3 & 0.58 & 0.646 \\
oneNLP-IIITH & mr2en-Method2-c & 30.6 & 0.57 & 0.659 \\
oneNLP-IIITH & mr2en-Method1-c & 20.7 & 0.48 & 0.735 \\
adaptNMT & mr2en & 19.9 & 0.758 & 0.429\\
UCF & mr2en-UnigramSegmentation-b & 7.7 & 0.24 & 0.833 \\
IIITT & mr2en-IndicTrans-b & 5.1 & 0.22 & 1.002 \\
\hline
\end{tabular} 
\caption[MR${\rightarrow}$EN: adaptMLLM systems compared with LoResMT2021]{MR${\rightarrow}$EN: adaptMLLM systems compared with LoResMT2021. The impact of fine-tuning the baseline NLLB model is evident with the BLEU score rising from 42.7 to 52.6 representing a 23\% relative improvement. Models developed using adaptMLLM were trained using the optimal hyperparameters set out in Table \ref{tab:hpo-table-mllm}.}
\label{tab:mr2en}
\end{table*}

Again the approach taken by adaptMLLM in fine-tuning a 3.3.B parameter NLLB MLLM yielded the best performance compared with other systems entered for the shared task. The EN$\rightarrow$MR adaptMLLM en2mr system achieves the highest BLEU score of 26.4 compared with IIITT, the winning team in the EN$\rightarrow$MR Shared task. IIITT had a BLEU score of 24.2 which represents a relative improvement of 9\% for the adaptMLLM system. The other key translation metrics of TER and ChrF3 were also improved upon indicating that the adaptMLLM system is the best approach in the EN$\rightarrow$MR direction.  

\par
For translation in the MR${\rightarrow}$EN direction, the best-performing model for the LoResMT2021 Shared Task was developed by oneNLP-IIITT with a BLEU score of 31.3, a TER of 0.58 and ChrF3 of 0.646. This serves as the baseline score by which our MR${\rightarrow}$EN model, developed using adaptMLLM, can be benchmarked. The performance of the adaptMLLM model offers a significant improvement across all metrics with a BLEU score of 52.6, a TER of 0.409 and a ChrF3 of 0.704. Again this represents a very strong relative improvement of 68\% in BLEU compared with the winning team from the shared task.

\subsection{Human Evaluation Results}

Irish, characterised by its complex morphology, flexible sentence structure, and extensive inflection, presents unique challenges in translation from English. As a result, accurately producing grammatical aspects like gender or case inflections in nouns within Irish translations often proves to be a difficult task.

\par
This research aims to investigate how a neural machine translation (NMT) system, like a fine-tuned NLLB model, manages these linguistic complexities. Current studies imply that fine-tuned MLLMs are likely to enhance these language features \parencite{costa2022no}. MLLMs and LLMs tackle the issue indirectly through subword models in an unsupervised fashion, without grasping the explicit formal principles of grammatical categories.

\par
Past human evaluation studies examining EN$\rightarrow$GA MT performance have centred on outputs from NMT systems that did not use pre-trained models \parencite{lankford2022human}. In the context of this research, we now conduct a human evaluation of the output from our MLLM models. The work is further differentiated in that it examines the output in both the EN$\rightarrow$GA and GA$\rightarrow$EN directions. The approach taken in the previous study and our current work are similar in that we use SQM and MQM as our human evaluation metrics.
\par
While automatic evaluation metrics show that a fine-tuned MLLM approach leads to significant improvements compared to building a Transformer model from scratch, it fails to address the issue of grammatical or linguistic quality in the translated output. Such an approach does not account for the subtleties of handling gender or cases in the target language. To gain a more comprehensive understanding of the linguistic errors produced by MLLM systems, a fine-grained human evaluation was conducted through a manual error analysis. This approach allowed for the identification and categorisation of specific translation errors associated with each of the evaluated systems, providing a foundation for future work aimed at improving the translation quality of the models.

\par 
We also describe the annotation framework, the overall annotation process, and the level of agreement among annotators, which broadly follows the approach taken by other fine-grained human evaluation studies \parencite{klubivcka2018quantitative,lankford2022human}.

\subsubsection{Scalar Quality Metrics}

The SQM framework modifies the WMT shared-task settings to acquire segment-level scalar ratings with document context. SQM assesses the quality of translations using a scale that ranges from 0 to 6, which is different from the WMT approach \parencite{ma-etal-2017-blend}, which employs a range of 0 to 100.

When using this evaluation method, annotators are required to choose a rating ranging from 0 to 6 after being presented with the source and target sentences. Table \ref{tab:sqm} provides the SQM quality levels for ratings 0, 2, 4, and 6. In situations where the translations do not precisely align with the core SQM levels, annotators may select intermediate ratings of 1, 3, or 5.

\begin{table}[h]
\small
\setlength{\tabcolsep}{8.16mm}
\caption{SQM levels explained \parencite{freitag-etal-2021-experts}.}
											
\centering 

\begin{tabular}{cp{10cm}}
\hline
\multicolumn{1}{c}{\textbf{SQM Level}} &
  \textbf{Details of Quality} \\ \hline
6 &
  Perfect Meaning and Grammar: The meaning of the translation is completely consistent with the source and the surrounding context (if applicable). The grammar is also correct. \\ \hline
4 &
  Most Meaning Preserved and Few Grammar Mistakes: The translation retains most of the meaning of the source. This may contain some grammar mistakes or minor contextual inconsistencies. \\  \hline
2 & Some Meaning Preserved: The translation preserves some of the meaning of the source but misses significant parts. The narrative is hard to follow due to fundamental errors. Grammar may be poor. \\  \hline
0 &
  Nonsense/No meaning preserved: Nearly all information is lost between the translation and source. Grammar is irrelevant. \\ \hline
\end{tabular}
\label{tab:sqm}

\end{table}

\par
The average annotator SQM scores arising from our human evaluation were compared with automatic metric scores recorded by adpatMLLM when evaluating the EN${\leftrightarrow}$GA systems. These results, illustrated in Table \ref{tab:sqm_results} indicate a high level of correlation between the automatic metrics and the SQM outputs of the human evaluation. Clearly, the system translating in the GA${\rightarrow}$EN direction performs better, when evaluated using both automatic and human evaluation, than its counterpart when translating in the opposite direction. These results are consistent with our previous work which also shows better GA${\rightarrow}$EN translation performance \parencite{lankfordlrev}. This performance difference is attributed to the morphological-rich nature of the Irish language which relies heavily on inflection, derivation and its case system. 
\begin{table}[h]
\centering
\begin{tabular}{lllllllll}
\hline
\textbf{System} &
  \textbf{BLEU} $\uparrow$ &
  \textbf{TER} $\downarrow$ &
  \textbf{ChrF3} $\uparrow$ &
  \textbf{SQM} $\uparrow$ \\ \hline
adaptMLLM en2ga    & 41.2  & 0.51  & 0.48 & 4.38  \\
adaptMLLM ga2en   & 75.1  & 0.385 & 0.71 & 5.63 \\
\hline
\end{tabular}
\caption[Annotator SQM scores for adaptMLLM systems]{Average SQM scores for adaptMLLM systems compared with automatic metrics.}
\label{tab:sqm_results}
\end{table}

\subsubsection{Multidimensional Quality Metrics}

Within the QTLaunchpad project,\footnote{\url{https://www.qt21.eu}} the development of the MQM framework\footnote{\url{https://themqm.org/the-mqm-full-typology/}} aimed to offer a structured approach to conducting manual evaluations through meticulous error analysis. This framework does not mandate a uniform metric for all applications; rather, it supplies an extensive list of potential quality issues, each with standardised names and definitions, which can be tailored to particular tasks. Beyond establishing a dependable method for quality evaluation, the MQM framework also enables us to identify and select error tags pertinent to our specific task.

\par
We customised the MQM framework to suit our context by following the official scientific research guidelines \parencite{Lommel2018}. Our modifications to MQM are explained below.

\par
The original MQM guidelines propose a wide range of tags on different annotation layers. However, for our specific annotation task, this comprehensive tagset is too detailed. Hence, we evaluated our MT output using the smaller default set of evaluation categories outlined in the core tagset. These standard top-level categories, which include accuracy and fluency, are recommended by the MQM guidelines and are presented in Table \ref{tab:mqmcat}.

We used a special non-translation error tag to label entire sentences that were so poorly translated that individual errors could not be identified. Error severities were designated as major or minor errors, and they were assigned independently of the category. These corresponded to actual translation or grammatical errors and minor imperfections, respectively. We used the default recommended weights \parencite{Lommel2018} which assign a weight of 1 to minor errors, while major errors are given a weight of 10. Additionally, the non-translation category was assigned a weight of 25, which is consistent with best practices established in previous studies \parencite{freitag-etal-2021-experts}.

Our annotators were instructed to identify all errors in each sentence of the translated output using the error categories provided in Table \ref{tab:mqmcat}.

\begin{table}[h]
\small

\caption[Description of error categories within the core MQM framework]{Description of error categories within the core MQM framework \parencite{freitag-etal-2021-experts}.}
				
\centering 
\begin{tabular}{p{2.4cm}p{3.5cm}p{7.8cm}}
\hline
\textbf{Category} & \textbf{Sub-Category} & \textbf{Description}                                 \\ \hline
\textbf{Non-translation} &                & Impossible to reliably characterise the 5 most severe errors.       \\ 
       \hline
\textbf{Accuracy}        & Addition       & Translation includes information not present in the source.         \\
         & Omission              & Translation is missing content from the source.      \\
                & Mistranslation & Translation does not accurately represent the source.               \\
         & Untranslated text     & Source text has been left untranslated.              \\
        \hline
\textbf{Fluency}  & Punctuation           & Incorrect punctuation                                \\
         & Spelling              & Incorrect spelling or capitalisation.                \\
         & Grammar               & Problems with grammar, other than orthography.       \\
                         & Register       & Wrong grammatical register (e.g., inappropriately informal pronouns). \\
                  & Inconsistency         & Internal inconsistency (not related to terminology). \\
                  & Character encoding    & Characters are garbled due to incorrect encoding.   \\ \hline
\end{tabular}

\label{tab:mqmcat}
\end{table}

\subsubsection{Annotation Setup}

Annotations were carried out using a detailed, fine-grained MQM approach and a simpler SQM approach. The SQM categories are summarised in Table \ref{tab:sqm} whereas the hierarchical taxonomy of our MQM implementation is outlined in Table \ref{tab:mqmcat}.

\par
Working independently of one another, two annotators with similar backgrounds were selected for the annotation of fine-tuned EN$\leftrightarrow$GA systems. Both annotators are fluent speakers of Irish and neither had prior experience with MQM. The annotators are postgraduate students of the Máistir Gairmiúil san Oideas (Postgraduate Masters in Education) at the University of Galway.\footnote{\url{https://universityofgalway.ie}}

Before starting the annotation process, they were extensively briefed on the process and the MQM annotation guidelines. These guidelines provide in-depth directions for carrying out annotation activities under the MQM framework.

\par
In conducting the EN$\rightarrow$GA human evaluation of the translation output, we presented our annotators with a test set of 25 randomly selected sentences, which consisted of the English source text, an Irish reference translation and the unannotated fine-tuned MLLM EN$\rightarrow$GA system output. 

A similar approach was adopted for the GA$\rightarrow$EN human evaluation where the annotator test set consisted of 25 randomly selected sentences, which consisted of the Irish source text, an English reference translation and the unannotated fine-tuned MLLM GA$\rightarrow$EN system output.

\par
After extracting the annotation data, the annotators individually examined the output to assess the performance of each system across the different error categories. 
 
\subsubsection{Inter-Annotator Agreement}

To ensure the validity of our research findings, it is essential to assess the degree of consensus among our annotators \parencite{artstein2017inter}. Manual evaluation methods for MT, such as MQM, often result in low inter-annotator agreement (IAA) \parencite{lommel-etal-2014-using,callison2007meta}. We computed inter-annotator agreement using Cohen's $kappa$ ($k$) coefficient \parencite{cohen1960coefficient}, a widely recognised metric in the field. The evaluation was performed at the sentence level for each system, and the agreement discrepancies across systems were examined. This approach also allowed us to obtain an overall view of the level of agreement between annotators.

Table \ref{tab:mqmtotals-human} highlights the cumulative number of errors identified by the annotators for each system. Looking at the aggregate data alone, it is evident that both annotators have judged the EN$\rightarrow$GA system to contain significantly more errors which supports the findings of the automatic evaluation. 

\begin{table}[h]
\centering
\small
\caption{System errors found by each annotator using the MQM metric}
\begin{tabular}{@{}lcccc@{}}
\hline 
\textbf{Num Errors}       & \textbf{EN$\rightarrow$GA}                 & \textbf{GA$\rightarrow$EN}                \\ \hline
Annotator 1  & 53                     & 7       \\ \hline
Annotator 2  & 82                     & 11       \\ \hline
\end{tabular}

\label{tab:mqmtotals-human}
\end{table}

\par
Table \ref{tab:mqmtotals-human} provides a useful overview for evaluating which system performs better overall, but it does not offer the detailed analysis necessary to identify specific linguistic areas for improvement in the translations. For a more comprehensive understanding, we delved into a detailed examination of the types of errors present, with the findings presented in Table \ref{tab:combined}. This table breaks down the total number of error tags noted by each annotator for each system, categorised by the type of error. The detailed analysis underscores how the GA$\rightarrow$EN system outperforms the EN$\rightarrow$GA system. Notably, the GA$\rightarrow$EN system's translations display significantly greater fluency, as evidenced by just two errors recorded in this category.

\par 
One way to measure inter-rater reliability is to use Cohen's $kappa$, which is a rigorous method. It determines the percentage of items that raters agree on while also taking into account the possibility of them agreeing on some items by chance. Cohen’s $kappa$ was calculated separately for every error type and the findings are outlined in Table \ref{tab:my-cohen} and discussed in further detail later in Section \ref{obs}. To calculate Cohen’s $kappa$ the following formula is used: 
\begin{equation} \label{e1}
k = (p\textsubscript{o} - p\textsubscript{e}) / (1 - p\textsubscript{e}) 
\end{equation}

$p\textsubscript{o}$: Relative observed agreement among raters

$p\textsubscript{e}$: Hypothetical probability of chance agreement

\begin{table}[h]
\centering
\small

\caption{Fine-grained analysis with concatenated errors across both annotators}
\begin{tabular}{@{}lcccc@{}}
\hline
\textbf{Error Type} &
  \multicolumn{1}{c}{\textbf{EN$\rightarrow$GA errors}} & \textbf{GA$\rightarrow$EN errors} \\ \hline
Non-translation       & 0 & 0  \\
\textbf{Accuracy}              &  \\
Addition                                  & 12 & 5 \\
Omission                                  & 14 & 3 \\
Mistranslation                            & 41 & 6 \\
Untranslated text                         & 9 & 2 \\
\textbf{Fluency}               &  \\
Punctuation                               & 10 & 0 \\
Spelling                                  & 6 & 0 \\
Grammar                                   & 27 & 0 \\
Register                                  & 19 & 2 \\
Inconsistency                             & 6 & 0 \\
Character Encoding                        & 0 & 0\\ \hline
\multicolumn{1}{l}{\textbf{Total errors}} & 135 & 18 \\ \hline
\end{tabular}

\label{tab:combined}
\end{table}

\begin{table}[h]
\centering
\caption[Inter-annotator agreement using Cohen values]{Inter-annotator agreement using Cohen values. Perfect observed agreement is indicated by $p\textsubscript{o}$ = 1.}
\begin{tabular}{@{}cccl@{}}
\hline
\textbf{Error type} & \multicolumn{1}{c}{\textbf{EN$\rightarrow$GA}} & \textbf{GA$\rightarrow$EN}  \\ \hline
\multicolumn{1}{l}{Non-translation}               & $p\textsubscript{a}$=1 & $p\textsubscript{a}$=1 \\
\multicolumn{1}{l}{Accuracy}                      &  &  \\
Addition                                          & 0.24 & 0 \\
Omission                                          & 0.31  & 0 \\
Mistranslation                                    & 0.32 & -0.11 \\
Untranslated text                                 & 0.07 & 0 \\
\multicolumn{1}{l}{Fluency}                       &  &  \\
Punctuation                                       & 1 & $p\textsubscript{o}$=1 \\
Spelling                                          & 0.24 & $p\textsubscript{o}$=1 \\
Grammar                                           & 0.59 & $p\textsubscript{o}$=1 \\
Register                                          & -0.07 & 0 \\
Inconsistency                                     & 0.34 & $p\textsubscript{o}$=1 \\
Character Encoding                                & $p\textsubscript{o}$=1 & 1.0\\ 
 \hline
\end{tabular}

\label{tab:my-cohen}
\end{table}  

\subsubsection{Inter-Annotator Reliability}\label{sec:rater}

In Cohen's seminal paper \parencite{cohen1960coefficient}, he precisely defines the interpretation of various $k$ scores. Scores $\le$ 0 indicate no agreement, scores from 0.01 to 0.20 suggest none to slight agreement, scores from 0.21 to 0.40 denote fair agreement, scores from 0.41 to 0.60 reflect moderate agreement, scores from 0.61 to 0.80 correspond to substantial agreement, and scores from 0.81 to 1.00 represent almost perfect agreement. The $kappa$ values of each error type are displayed in Table \ref{tab:my-cohen}.

Many chance-adjusted indices of inter-rater reliability estimate agreement using a distribution-based approach. A problem arises when there is only one observed response category, resulting in a score of NaN (Not a Number). This occurs when the observed agreement, $p\textsubscript{o}$ and the chance agreement, $p\textsubscript{e}$ are both 1, which cannot be computed as seen in Equation \ref{e1}. In such cases, it is better to report $p\textsubscript{o}$ instead of $kappa$ since there is perfect observed agreement i.e. $p\textsubscript{o}$ = 1.

Illustrated in Table \ref{tab:my-cohen}, we observe a high level of agreement overall. There is either fair agreement, or perfect observed agreement, in 16 out of 22 sub-categories. Given these scores, we have a high degree of confidence in the human evaluation of the fine-tuned MLLM outputs.
 
\subsection{Environmental Impact}
\label{sec:envimp}

Motivated by research which examines the environmental impact of NLP \parencite{strubell-etal-2019-energy,bender2021dangers}, we monitored the energy and carbon emissions required to train our models.

Model development was carried out using Colab Pro+, which as part of Google Cloud is carbon neutral \parencite{lacoste2019quantifying}. All fine-tuning experiments of MLLMs were conducted on Google Cloud servers and consequently were emission-free.\footnote{\url{https://cloud.google.com/sustainability/region-carbon}}

In terms of energy consumption, the total power draw for each experimental run is outlined in Table \ref{tab:environ}. As part of our Google Colab subscription, NVIDIA a100-sxm4-40gb graphics cards were used which have a max power consumption of 400W. The calculations are based on the graphics card running at 80\% max power during model training.  

\begin{table}[h]
\centering
\begin{tabular}{lllllllll}
\hline
\textbf{System} &
  \textbf{BLEU} $\uparrow$ &
  \textbf{TER} $\downarrow$ &
  \textbf{ChrF3} $\uparrow$ &
  \textbf{Lines} &
  \textbf{\begin{tabular}[c]{@{}l@{}}Runtime \\ (hours)\end{tabular}} &
  \textbf{kWh} \\ \hline
adaptMLLM en2ga    & 41.2  & 0.51  & 0.48 & 13k  & 3.51 & 1.1  \\
adaptMLLM ga2en   & 75.1  & 0.385 & 0.71 & 13k  & 3.41 & 1.1  \\
adaptMLLM en2mr  & 26.4 & 0.56 & 0.608 & 21k & 5.49 & 1.8  \\
adaptMLLM mr2en  & 52.6 & 0.409 & 0.74 & 21k & 5.43 & 1.7 \\ 
\hline
\end{tabular}
\caption[Energy consumption during MLLM fine-tuning experiments]{Energy consumption during MLLM fine-tuning experiments. All experiments were carried out on Google Cloud with 0 kgCO\textsubscript2 emissions.}
\label{tab:environ}
\end{table}

\section{Discussion} \label{sec:discussion}
 
We used the adaptMLLM application to create MT models with datasets from the LoResMT2021 Shared Task to assess system efficiency when translating in the EN$\leftrightarrow$GA directions.

High-performing models achieving SOTA scores were developed by fine-tuning the NLLB MLLM pre-trained models with adaptMLLM. Using an easily-understood framework such as adaptMLLM, the benefits of developing high-performing fine-tuned models with small in-domain datasets are thus clear.

\subsection{Performance of adaptMLLM Relative to Google Translate}

Translation engine performance, at the corpus level, was benchmarked against Google Translate's \footnote{\url{https://translate.google.com}} EN${\leftrightarrow}$GA translation service, which is freely available on the internet. 

A full evaluation of Google Translate's engines on the EN$\rightarrow$GA test set generated a BLEU score of 38.7, a TER score of 0.493 and a ChrF3 of 0.633. The comparative scores on the test set using our fine-tuned MLLM realised 41.2 for BLEU, 0.489 for TER and 0.653 for ChrF3. Therefore, in the EN$\rightarrow$GA direction, the adaptMLLM system demonstrates a relative BLEU score improvement of 6.5\% compared to Google Translate.

The translation output from our fine-tuned MLLMs was also compared with Google Translate using random samples from the LoResMT2021 EN$\rightarrow$GA corpus. Table \ref{tab:translations_en2ga} highlights random samples which were picked from the English source test file. A perfect match, with a BLEU of 100, was recorded in one instance which is unusual. However, this may occur on occasion with the translation of short sentences. Any duplicates between training and test data were removed before fine-tuning but the possibility exists of the test sentence forming part of the original training of the NLLB model.

Translation of these samples was independently carried out on the optimal fine-tuned MLLM model and also using Google Translate. Case-insensitive, sentence-level BLEU scores were recorded and are presented in Table \ref{tab:en2ga_gt}.

\begin{table}[h]
\caption{EN${\rightarrow}$GA test dataset of LoResMT2021: samples of human reference translations.}

\begin{tabular}{p{7.2cm}p{7.2cm}}\hline
\textbf{Source Language (English)} & \textbf{Human Translation (Irish)} \\ \hline
Temporary Covid-19 Wage Subsidy Scheme & Scéim Fóirdheontais Shealadaigh Pá Covid-19\\\hline
how Covid-19 spreads and its symptoms & conas a scaipeann Covid-19 agus na siomptóim a bhaineann leis\\\hline
\end{tabular}
\label{tab:translations_en2ga}
\end{table}

\begin{table}[h]
\caption{EN${\rightarrow}$GA fine-tuned MLLM model compared with Google Translate. }

\begin{tabular}{p{5cm}p{1.75cm}p{5cm}p{1.75cm}}\hline

\textbf{fine-tuned MLLM} & \textbf{BLEU} \boldmath{$\uparrow$} & \textbf{Google Translate} & \textbf{BLEU} \boldmath{$\uparrow$}\\ \hline
Scéim Fóirdheontais Pá Sealadach Covid-19 & 25.4 & Scéim Fóirdheontais Pá Shealadach Covid-19 & 25.4 \\\hline
conas a scaipeann Covid-19 agus na siomptóim a bhaineann leis & 100 & conas a scaipeann Covid-19 agus na hairíonna a bhaineann leis & 65.8 \\\hline
\end{tabular}
\label{tab:en2ga_gt}
\end{table}

The translation output from our fine-tuned MLLMs was also compared with Google Translate using random samples from the LoResMT2021 EN$\rightarrow$MR corpus. A full evaluation of Google Translate's engines on the EN$\rightarrow$MR test set, with 500 lines, generated a BLEU score of 25.9, a TER score of 0.566 and a ChrF3 of 0.601. The comparative scores on the test set using our fine-tuned MLLM realised 26.4 for BLEU, 0.565 for TER and 0.608 for ChrF3. Therefore, in the EN$\rightarrow$MR direction, the adaptMLLM system demonstrates a relative BLEU score improvement of 1.9\% compared to Google Translate.

Samples from the EN$\rightarrow$MR test set, along with the corresponding human translation are illustrated in Table \ref{tab:translations_en2mr}. The performance of these individual samples from the MLLM output and the Google Translation output is compared in Table \ref{tab:en2mr_gt}.
The results are encouraging and indicate a good performance by our translation models on the datasets from LoResMT2021.

\begin{table}[h]
\caption{Samples of human reference translations from EN${\rightarrow}$MR LoResMT2021}
\includegraphics[width=1.0\textwidth]{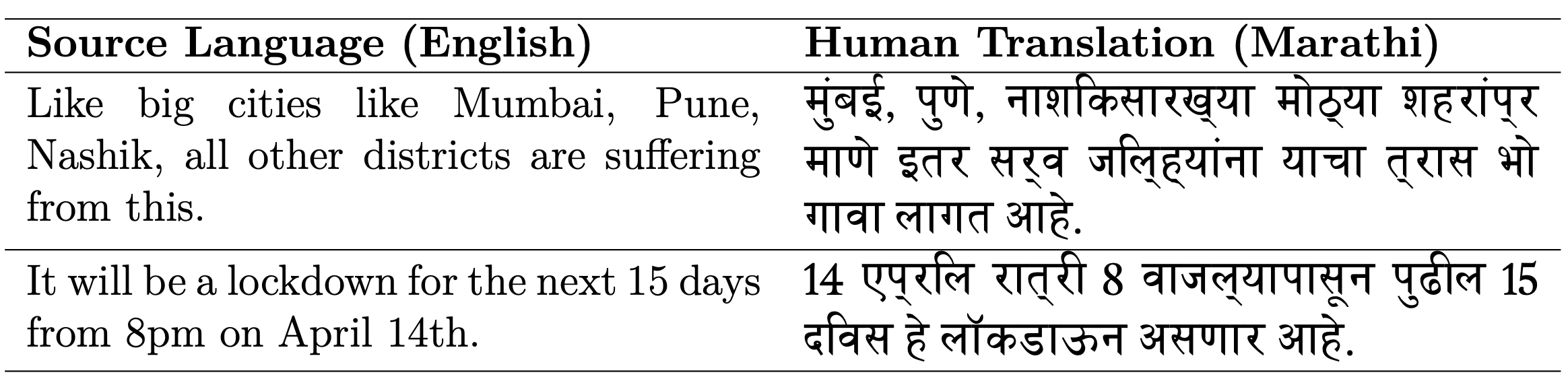}
\label{tab:translations_en2mr}
\end{table}

\begin{table}[h]
\caption[EN${\rightarrow}$MR fine-tuned MLLM model compared with Google Translate]{EN${\rightarrow}$MR fine-tuned MLLM model compared with Google Translate. MR phrases are back-translated to EN and highlighted immediately below each MR sentence pair.}
\includegraphics[width=1.0\textwidth]{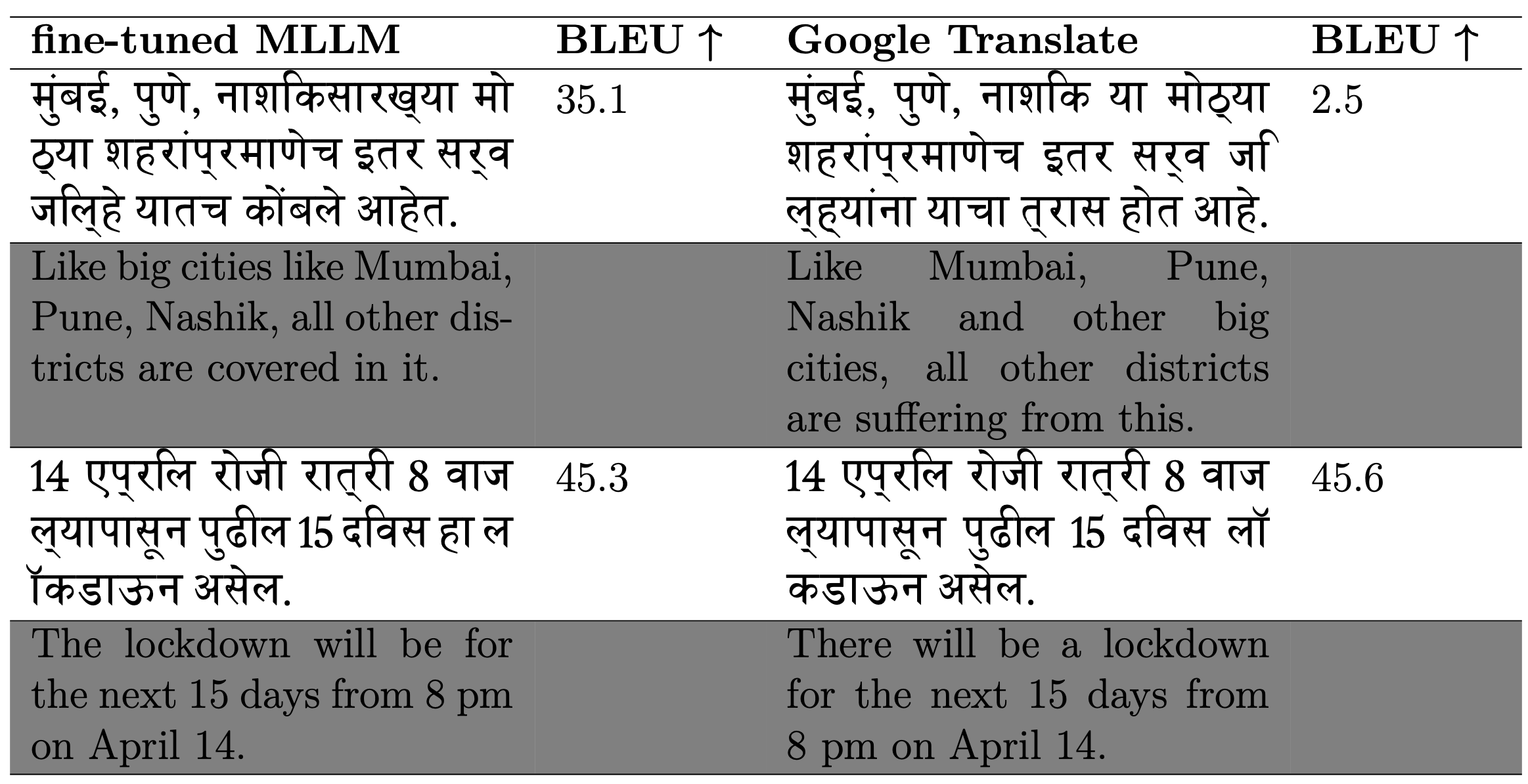}
\label{tab:en2mr_gt}
\end{table}

\subsection{Linguistic Observations}\label{obs}

Table \ref{tab:ling-anal} provides a linguistic analysis of the EN$\rightarrow$GA MLLM outputs, showcasing the source sentences alongside their corresponding translations. These sentences were chosen specifically for this detailed human evaluation since they underscore the principal types of errors observed. The approach adopted is similar to the analysis taken in our previous human evaluation of EN$\rightarrow$GA translation \parencite{lankford2022human} in that it focuses on model output errors which fall into the categories of `interpreting meaning' and `core grammatical errors'.  

\begin{table}[h]

\small
\setlength{\tabcolsep}{4.85mm}
\caption[Linguistic analysis of EN$\rightarrow$GA system output]{Linguistic analysis of EN$\rightarrow$GA system output. Errors in the target translation are flagged in red with the corresponding source highlighted in blue.}

\centering 
\begin{tabular}{p{2cm}p{10cm}}
\hline
\textbf{Type} &
  \textbf{Sentence} \\ \hline
\textbf{EN-1} & 
\textcolor{blue}{Covid-19 information} and advice for taxpayers and agents \\ \hline
\textbf{GA-1} &
Eolas agus \textcolor{red}{comhairle Covid-19} díocóirí cánach agus dionadaithe \\ \hline
\textbf{EN-2} &
 \textcolor{blue}{We understand} the unprecedented situation facing taxpayers as a result of the Covid-19 pandemic.    \\ \hline
\textbf{GA-2} &
\textcolor{red}{Tuigeann muid} an cás gan fasach atá roimh cháiníocóirí mar thoradh ar an bpaindéim Covid-19. \\ \hline
\textbf{EN-3} &
Further information on the Employment Wage Subsidy Scheme (EWSS) is available from the \textcolor{blue}{Employing people section} on this website. \\\hline
\textbf{GA-3} &
Tá tuilleadh faisnéise ar Scéim Fóirdheontais Pá Fostaíochta (EWSS) ar fáil ón \textcolor{red}{gcuid Fostaithe} ar an láithreán gréasáin seo.\\ \hline
\textbf{EN-4} &
Information for \textcolor{blue}{employers} on the Temporary Covid-19 Wage Subsidy Scheme is available from the Employing people section on this website. \\ \hline
\textbf{GA-4} &
Tá faisnéis \textcolor{red}{dfhostóirí} ar an Scéim Fóirdheontais Pá Sealadach Covid-19 ar fáil ón gcuid Fostaithe ar an láithreán gréasáin seo.\\ \hline

\end{tabular}

\label{tab:ling-anal}
\end{table}

\subsubsection{Interpreting Meaning}

When examining the relationship of one noun to another noun, it should not necessarily be directly translated from English to Irish. This is illustrated in EN-1 where “COVID-19 information and advice” refers to the information and advice that is related to COVID. However, the EN$\rightarrow$GA system translates this to “Comhairle COVID-19” which effectively means “COVID-19's information and advice”, i.e. COVID-19 is treated as a possessive noun which is incorrect.

At times the translated output does not reflect the context in which particular words should be used. An example of this can be seen in the translation of the words “Employing people section” in EN-3 which was interpreted by the EN$\rightarrow$GA system as “gcuid Fostaithe”. In this English source sentence, the meaning focuses on a section related to a website and the correct translation would be “rannán Daoine a Fhostú”. This is outlined in more detail on the reference website, Fóclóir.\footnote{\url{https://www.focloir.ie}} It is interesting to note that Google Translate correctly interprets this meaning in its translation of the sentence.

Given the nature of the source text, one word frequently encountered was ``Information''. The word was accurately translated to ``faisnéis'' over the text, but it is important to note this word is not widely used in the Irish language. We recommend using the word ``eolas'' (knowledge) since it is a more natural and intuitive translation.\footnote{\url{https://www.teanglann.ie/en/fgb/eolas}}

\subsubsection{Core Grammatical Errors}

Common mistakes which were encountered throughout the texts involved the use of the apostrophe. Most of these mistakes were flagged as minor errors but in some cases, a missing apostrophe conveyed an entirely different meaning. An example of this can be seen in EN-4 and GA-4 where “information for employers” has been translated to  “faisnéis dfhostóirí” which means “employers’ information”. By simply correcting this to “faisnéis d'fhostóirí”, the correct meaning would have been preserved.

\section{Conclusion and Future Work}\label{concl}

We presented adaptMLLM, a comprehensive application designed for fine-tuning MLLMs and which handles the entire process of model development, evaluation, and deployment. The performance of the application was showcased through the creation of EN$\leftrightarrow$GA translation models, which exhibited substantial improvements over the top-ranked models from the EN$\leftrightarrow$GA LoResMT2021 Shared Tasks.

To further validate this work, a fine-grained human evaluation was conducted by annotators on the translation output in the EN$\leftrightarrow$GA directions and the findings are outlined in our linguistic observations (cf. Section \ref{obs}).

As a multilingual tool, systems derived from adaptMLLM were also compared with the winning entries from the EN$\leftrightarrow$MR LoResMT2021 Shared Tasks. Fine-tuning 3.3B parameter NLLB models, using adaptMLLM demonstrated that our models for the EN$\leftrightarrow$MR language pair performed significantly better across all translation metrics when compared with the winning entries in the EN$\leftrightarrow$MR LoResMT2021 Shared Tasks.

The performance of our translation models developed for this study was compared with the output generated by Google Translate on both the EN$\leftrightarrow$GA and EN$\leftrightarrow$MR language pairs. In all language directions, the performance of the adaptMLLM models was better than that of Google Translate demonstrating a new SOTA in low-resource MT of the EN$\leftrightarrow$GA and EN$\leftrightarrow$MR language pairs. 

In terms of future work, there is much which can be done to extend our research. There are several avenues which we plan on exploring further. Firstly we would like to establish the effects of fine-tuning larger MLLMs, such as the 54B parameter NLLB network, on our existing datasets. It is anticipated this will most likely improve our results and will also establish the trend in which increasingly larger MLLMs drive MT performance. The availability of the MTU and DCU GPU clusters within the ADAPT centre, coupled with the DeepSpeed library, provides the platform upon which this can be achieved.

At this juncture, we have just scratched the surface of the MT performance enhancements which are possible through hyperparameter optimisation. Using a random search approach \parencite{JMLR:v13:bergstra12a}, we will extend our search space by examining a greater number of hyperparameters and a larger range of associated values.   
Against this backdrop, it will be possible to apply adaptMLLM to new shared tasks and WMT competitions. This will also address another goal of our future work which is to apply our approach to other low-resource language pairs.

Furthermore, integration of GPT-3, GPT-4 and BARD (cf. Section \ref{llm_background}) playgrounds into adaptMLLM, in addition to fine-tuning these LLMs will be explored as part of future work.

Once the preserve of large research teams with very significant compute infrastructure, our approach has shown much smaller research teams can fine-tune MLLMs on modest budgets. In doing so, we have succeeded in developing SOTA results for two low-resource language pairs. As an open-source initiative, we look forward to the community contributing to its advancement with the addition of fresh concepts and feature enhancements.

We have shown in the context of our low-resource EN$\leftrightarrow$MR and EN$\leftrightarrow$GA pairs that fine-tuning a pre-trained MLLM such as NLLB is a more efficient and effective approach than training a bespoke Transformer model from scratch. 

In addition to improved performance, fine-tuning MLLMs saves both time and computational resources. Consequently, given the right infrastructure, we recommend using such an approach when developing MT systems for low-resource pairs in the future.

\section{Limitations of the Study}
With additional resources, some elements of this research could be expanded upon. While there is a satisfactory level of agreement between annotators, the inclusion of a larger pool of annotators would be beneficial. Moreover, evaluating a more extensive selection of lines with a finer classification of the MQM taxonomy could yield a deeper understanding of the MT outputs.

Whereas fine-tuning the baseline NLLB models highlighted a demonstrable improvement in translation quality using automatic metrics, a corresponding human evaluation of the baseline NLLB outputs was not conducted. As part of our future work, it is planned to conduct such an evaluation.

The focus of the study primarily centred on fine-tuning the NLLB base model since it was the most likely candidate for success in producing high-quality MT output for low-resource languages. Other LLMs, such as GPT-J, should also be investigated for fine-tuning experiments.

With more hardware resources, and a larger research team, the impact of even larger models such as NLLB-54B would have been explored. It is planned to address these limitations in our future work (cf. Section \ref{concl}).

\authorcontributions{Writing—original draft, S.L.; Writing—review \& editing, H.A. and A.W. All authors have read and agreed to the published version of the manuscript.}

\funding{This research is supported by Science Foundation Ireland through ADAPT Centre (Grant 13/RC/2106) at Dublin City University. This research was also funded by the Staff Doctorate Scheme at the Munster Technological University. }

\institutionalreview{In the “Related Work” section of this paper, we discuss academic papers published at conferences and in academic journals. We ensure that all data used in our analysis were obtained legally and ethically. With regard to licensing for our application, adaptMLLM, it is covered by the Creative Commons Attribution 4.0 International License. We recognise the importance of responsible and ethical conduct in AI research and will continue to prioritise these values in our work.}

\dataavailability{The data presented in this study are openly available and can be found at \url{https://github.com/adaptNMT/adaptMLLM/}}

\acknowledgments{ We also thank our anonymous reviewers for their comments, and our annotators Darragh Lankford and Muireann Ní Chorcora for their meticulous work in annotating the system outputs.}

\conflictsofinterest{ The authors declare no conflict of interest. The funders had no role in the design of the study; in the collection, analyses, or interpretation of data; in the writing of the manuscript, or in the decision to publish the results.}

\chapter{Conclusion}
The main research contributions are summarised and the initial research questions laid out in Chapter 1 are reflected upon in this conclusion chapter. The lessons learnt are subsequently discussed and finally, roadmaps for future work are outlined.

This research has contributed to the field of NMT by thoroughly evaluating hyperparameter settings, proposing an innovative framework for the human evaluation of low-resource languages, and developing the adaptable and user-friendly tools of adaptNMT and adaptMLLM. The research findings provide valuable insights into improving translation performance for low-resource languages and lay the groundwork for further advancements in MT.

\section{Research contributions}

\subsection{HPO for Low-resource Languages}
A comprehensive evaluation of hyperparameter settings and model variations in the context of Transformer-based NMT for low-resource language pairs was conducted as a means of addressing RQ1. The study explored the impact of modifications in the number of layers, regularisation techniques, attention heads, subword model types, and vocabulary size, aiming to identify optimal settings for improving translation performance. The research demonstrated that HPO significantly enhances the translation quality of Transformer models, and recommendations for optimal hyperparameter settings for our low-resource models were proposed. The beneficial impact of subword models, such as BPE and unigram models, on translation performance for low-resource languages was also highlighted. However, it must be highlighted that HPO of the corresponding RNN models was not conducted as part of these experiments. Future work in this area should ensure the HPO of RNN models if such models are to be used as a baseline for comparison purposes.

\subsection{Corpus Development and Guidelines} 
The research question identified in RQ2 addresses a significant gap in NMT: while translation technology is advanced for high-resource languages, it lags in some domains for low-resource languages due to the scarcity of parallel datasets. Most efforts for such languages are concentrated on constructing large, generic datasets without focusing on the potential advantages of specialised, in-domain datasets. The impact of domain-specific data on MT quality was examined by formulating a health-focused dataset for the EN$\leftrightarrow$GA language pair, a notably low-resource combination. This work resulted in the development of gaHealth, a bilingual health data corpus tailored for the Irish language and also created linguistic guidelines instrumental for its development. Another contribution arising from RQ2 was the experimental findings from using the gaHealth dataset. By making these results, and the gaHealth corpus, openly accessible online, the study furnishes empirical evidence underscoring the utility of domain-centric datasets while laying a foundation for future explorations of in-domain corpora.

\subsection{Human Evaluation of Low-resource Languages}
With RQ3, we systematically compared the outputs of RNN and Transformer systems, analysing and categorising the nature of translation errors produced by each system. A human evaluation based on the MQM error taxonomy demonstrated that Transformer-based NMT systems substantially reduce both accuracy and fluency errors compared to RNN-based models. A key research contribution was the development of a modified MQM framework which was subsequently incorporated into the tools developed in response to answering RQ4.  This analysis serves as a foundation for future work to enhance translation quality in NMT models for low-resource language pairs. 

\subsection{Explainable AI Architectures}

Explainable AI (XAI) refers to AI systems that are designed to be transparent and understandable by humans. Furthermore, XAI should also make the decision-making processes of AI systems clear so that users can understand how and why a particular decision or prediction was made. 

In the context of my research, one of the core objectives of XAI has been achieved, namely the delivery of transparent and easily understandable applications. Further work is required to address how particular translations are arrived at. As part of our future work, it should be relatively easy to expose the final SoftMax layer within the neural models so that probabilities associated with each potential translation are displayed. Of course, for a truly explainable architecture, it is important to go deeper into the network where gradient or attention-based methods could be used to find the most salient words in the input when determining system outputs. 

The key research contribution arising from answering RQ4 was to develop a user-friendly and flexible tool (adaptNMT) for training and deploying NMT models. The adaptNMT application simplifies the entire development and deployment process, offering a streamlined approach for developers, translators and end users. Notable features include simplified setup and customisation with evaluation and deployment capabilities. A significant contribution was the introduction of a green report feature which promotes eco-friendly research by monitoring power consumption and kgCO\textsubscript2 emissions. 

In particular, adaptNMT offers a real-time graphical view of model training via TensorBoard visualisation, enabling users to easily monitor their model's training progress. It also incorporates SentencePiece to establish subword segmentation models, though users have the flexibility to select other subword model types during system development if they choose. Moreover, adaptNMT is versatile in its deployment; users can run it on local setups or utilise it as a Colab instance on Google Cloud, with the latter offering enhanced GPU and computing resources through the Colab Pro subscription.

The research contributions of RQ4 were further refined with the development of a separate application to concentrate solely on MLLMs and LLMs including NLLB and GPT-J. Open-sourced as a standalone tool, adaptMLLM was evaluated for the EN$\leftrightarrow$GA and EN$\leftrightarrow$MR language pairs. The evaluation showcased significant improvements in translation performance through fine-tuning a 3.3B parameter NLLB MLLM using the adaptMLLM approach. Relative improvements ranging from 9\% to 117\% were achieved compared to baseline models submitted in the LoResMT2021 Shared Task.

The tools developed through research on RQ4 are both open-source applications that streamline all processes involved in the development and deployment of NMT models.  Furthermore, models can be easily evaluated with various metrics and seamlessly deployed as an in-app translation service. The applications provide an intuitive interface for hyperparameter customisation, allowing users to fine-tune parameters specific to their chosen methodology.

The study also emphasised the vital role played by the DeepSpeed library in enabling efficient model training and prototype development to be conducted on a moderate infrastructure. Fine-tuning of MLLMs which had traditionally been the preserve of large tech corporations has effectively been democratised by this library in that large MLLMs and LLMs can be stored and trained on combined GPU and system RAM.

\section{Research Impact}

The motivation for this research was primarily driven by a desire to use AI technology to improve MT for low-resource languages, particularly the EN$\leftrightarrow$GA language pair. The skills developed as part of an MSc. in AI were used to run fine-tuning experiments and make recommendations for optimal Transformer settings for low-resource languages. To facilitate such experimental runs, architectures were developed to provide standardised approaches which were repeatable. The tools developed as part of this work were used to train MT systems for EN$\leftrightarrow$GA and EN$\leftrightarrow$MR translations. These architectures have subsequently been made available to the community as open-source tools. We hope that such tools will be extended by others in ways which we have not foreseen, and also in ways which we have anticipated through the roadmaps set out below.

The impact on MT quality of fine-tuning with a small, high-quality in-domain corpus was clearly demonstrated. These results should provide the motivation for developing similar corpora in other domains using the guidelines laid out as part of our work. 

While automatic evaluation has an important role to play in validating MT systems, humans are the ultimate arbiters who validate the quality of such systems. In recognising this, a framework has been put in place to assist practitioners in carrying out human validation experiments. This approach has been used to successfully evaluate the quality of the output from both XAI architectures which were developed. It is anticipated that our approach will be useful for human evaluation of other low-resource models. 

\section{Lessons Learnt}

NMT models often struggle when dealing with language pairs that have limited training data. However, HPO of the Transformer model can lead to significant performance enhancements, especially in these low-resource scenarios. The choice of the correct subword model emerges as a pivotal factor driving translation performance. 

Further improvements can be achieved by optimising certain aspects of the model, such as the number of attention heads, the number of layers, and the employment of various regularisation techniques. Interestingly, using smaller and fewer layers within the Transformer model was found to be beneficial for performance when working with low-resource datasets. 

For the HPO of Transformer models in such settings, a random search approach is favoured. This preference helps reduce the costs and extended training durations typically associated with the grid search method. The research underscored the potential of Transformers, proving they can be effectively harnessed for low-resource translation tasks, such as EN$\leftrightarrow$GA and EN$\leftrightarrow$MR translations.

The custom development of an in-domain corpus is a tedious and time-consuming process. However, we have learnt that even a small in-domain dataset can have a very significant impact on the translation quality of MT systems. This was clearly illustrated in the development of a Covid corpus for the LoResMT2021 Shared Task and the subsequent development of gaHealth. Adhering to a pre-defined approach as laid out in our paper for LREC 2022 helped to accelerate both the development time and quality of the corpus.

The linguistic quality of NMT systems was assessed by comparing the performance of RNN and Transformer systems using SQM and MQM as human evaluation metrics. The proposed approach in this study combines human evaluation with automatic metrics. From our human evaluation research, several key insights emerged regarding MT, particularly concerning the comparison between RNN- and Transformer-based EN$\leftrightarrow$GA systems. This study presents a human evaluation that compares the output from EN$\leftrightarrow$GA RNN systems against Transformer-based EN$\leftrightarrow$GA models. The findings show a substantial correlation between human evaluation and automated methods. Both these evaluation methods converged on the finding that Transformer-based models displayed significantly higher accuracy. As part of the human evaluation, the misuse of common irregular verbs was flagged as a common problem but that could perhaps be rectified by refining our models with datasets meticulously curated for this particular challenge. Likewise, the strategic selection of training data for model fine-tuning could help reduce the register errors highlighted in our linguistic analysis. 

Several lessons were learnt from the empirical evaluation of the adaptMLLM approach, which was used to train models for two language pairs namely EN$\leftrightarrow$GA and EN$\leftrightarrow$MR. 

Infrastructure and hyperparameters play a foundational role in the adaptMLLM system. The availability of the Google Colab Pro+ subscription, supplied with NVIDIA 40GB GPU graphics cards and access to up to 89GB of system memory, proved invaluable. This resource allowed for rapid prototype development, which was further amplified by the DeepSpeed library. This library is integral to the adaptMLLM system, facilitating the loading of models across both GPU and system memory, effectively keeping computational costs in check.

Fine-tuning results were particularly promising. When using adaptMLLM to refine the NLLB MLLM, there was a marked improvement in translation performance. Specifically, for EN$\rightarrow$GA, adaptMLLM demonstrated a relative enhancement of 14.4\% over the top-performing system in the LoResMT2021 Shared Task. The GA$\rightarrow$EN counterpart displayed a more significant improvement still, with a relative improvement of 117\% in BLEU score compared to the task's best model. Likewise, for translation in the EN$\rightarrow$MR direction, the adaptMLLM system surpassed the winning team's results in the shared task by 9\%. In the reverse direction, MR$\rightarrow$EN, the gain was a relative improvement of 68\% in the BLEU score.

\section{Future Work}

\subsection{Hyperparameter Tuning}

An initial evaluation of the impact of varying the number of attention heads, the number of layers, and experimentation with other regularisation techniques, such as label smoothing, has been conducted. Widening the hyperparameter search space should lead to marginal improvements in the performance of Transformer models in low-resource scenarios. In addition, the performance of the optimised Transformer model should be evaluated on other low-resource language pairs. 

The effects of using different training data, such as news articles or social media posts, on the performance of Transformer models in low-resource settings could be evaluated. This exploration would provide insights into the effectiveness of using different types of data for training.

The performance of the optimised Transformer model could be compared with other SOTA models, such as convolutional neural networks. This comparison will identify the strengths and weaknesses of the Transformer model in low-resource scenarios.

While some transfer learning techniques have been incorporated into the adaptNMT and adaptMLLM applications, others will be explored to improve the performance of the Transformer model in low-resource scenarios. 

\subsection{Corpus Development}

The substantial impact that a small in-domain health dataset can have on MT quality has been observed, and empirical evidence has been provided through RQ2. This underscores the importance of further development of in-domain datasets. As part of the background work for gaHealth, the availability of parallel data from other domains such as Finance, Education and Agriculture was explored. The departments of the Irish government publish much of their work, including strategy statements and annual reports, bilingually in English and Irish. Although such reports are not organised as parallel text corpora, the application of the guidelines developed by RQ2 could be applied in the creation of a new set of concise in-domain corpora. The availability of an array of in-domain corpora could be a platform for developing a high-performing MT ensemble for EN$\leftrightarrow$GA systems.

\subsection{Human Evaluation}

Further human evaluation studies should be conducted to gain a deeper understanding of the linguistic errors generated by NMT models and to develop strategies to address these errors. Techniques to incorporate external linguistic resources, such as morphological analysers or dictionaries \parencite{passban-etal-2018-tailoring}, into NMT models could be developed which would potentially improve MT performance. The proposed approach could be applied to other low-resource language pairs to test how well it generalises. 

The outputs from the human evaluation aspect of the study have helped in understanding the quality and the failings of our translation of our models. In our human evaluation, we propose combining a human evaluation approach with automatic metrics to test the effectiveness of the models in creating high-quality MT outputs. The research aims to explore how NMT systems handle translation issues compared to RNN approaches in morphologically rich languages. 

\subsection{Explainable AI Architectures}

\subsubsection{Roadmap for adaptNMT}

The application could be expanded to incorporate other NMT architectures, such as the recently introduced non-autoregressive models. Other subword segmentation models, beyond the currently implemented SentencePiece models, should be integrated into the adaptNMT system.

The following enhancements to the adaptNMT architecture should be incorporated as part of a future research roadmap:
\begin{itemize}
     
\item	\textbf{Enhancement of the green report feature} to provide more detailed information on the environmental impact of NMT model development and deployment, and suggestions for eco-friendly practices. Currently, the green report flags the power consumption and kgCO\textsubscript2 emissions generated during model development, but additional green features such as a recommender function that suggests energy-efficient model architectures or automatic selection of eco-friendly compute resources could be explored.
\item	\textbf{Development of a user community} for sharing models, datasets, and best practices, which could be facilitated through the application's Colab notebook format.
\item	\textbf{Improvement of the automatic notification} feature to include more detailed information on model training convergence and accuracy, and suggestions for model refinement.
\item	\textbf{Integration with additional datasets and languages}: adaptNMT could be expanded to include a wider range of language pairs and training datasets, enabling the tool to be used for a broader range of applications.
\item	\textbf{Development of a mobile version}: The mobile web version of adaptNMT works similarly to other Colab notebooks running on mobile platforms, and users can build and deploy NMT models on the go. However, a dedicated application would improve its usability for mobile users. 
\item	\textbf{Integration with other NLP tools}: Integrating adaptNMT with other NLP tools such as part-of-speech taggers, named entity recognition tools, and sentiment analysers could enhance the capabilities of the tool and enable more sophisticated NMT models to be built. Such integration would also facilitate more advanced pre-processing and post-processing of NMT outputs.
\item	\textbf{Implementation of additional evaluation metrics}: While adaptNMT currently supports a range of evaluation metrics, additional metrics such as BLEU+ could be implemented to provide a more comprehensive evaluation of NMT models.
\item	\textbf{Integration with cloud-based ML platforms}: Integrating adaptNMT with cloud-based ML platforms such as Amazon SageMaker or Microsoft Azure could enable more powerful model development and deployment, as well as easier collaboration between researchers and teams.
\end{itemize}

\subsubsection{Roadmap for adaptMLLM}

To achieve better translation performance, further optimisation of the infrastructure and hyperparameters for adaptMLLM models using multiple language pairs should be considered. Additionally, the adaptMLLM approach can be evaluated on more language pairs to see if it generalises well to other languages. Overall, the feature roadmap for adaptMLLM should focus on making the application more user-friendly, versatile, and accessible to a wider audience. In particular, the following roadmap should be considered:
\begin{itemize}
\item \textbf{Greater multilingual support}: adaptMLLM is currently available for two language pairs, EN$\leftrightarrow$GA and EN$\leftrightarrow$MR. To make the application more versatile, additional language pairs could be added in the future.
\item 	\textbf{Fine-tuning with different pre-trained models}: Currently, adaptMLLM fine-tunes MLLMs using pre-trained models like NLLB. This could be extended to include the ability to fine-tune with other pre-trained models.
\item 	\textbf{Interactive visualisation of model training}: While TensorBoard provides real-time graphical views of model training, an interactive visualisation could provide a more intuitive way to track the training progress.
\item 	\textbf{Easy integration with other NLP libraries}: adaptMLLM could be made more user-friendly by providing easy integration with other popular NLP libraries like spaCy and NLTK.
\item 	\textbf{Integration of different pre-processing techniques}: adaptMLLM could be enhanced by integrating different pre-processing techniques like data cleaning, normalisation, and augmentation to improve the quality of training data.
\item 	\textbf{Automatic tuning of hyperparameters}: While adaptMLLM provides an intuitive user interface for hyperparameter customisation, it could be further enhanced by incorporating automatic tuning of hyperparameters.
\item 	\textbf{Support for transfer learning}: adaptMLLM could be enhanced by incorporating support for transfer learning, which could enable users to transfer knowledge from one language pair to another.
\item 	\textbf{Generalising the adaptMLLM approach}: the adaptMLLM approach was developed specifically for low-resource MT. Future work could include exploring ways to generalise the approach for other NLP tasks.
\item 	\textbf{Scaling the adaptMLLM approach}: the adaptMLLM approach was developed using a 3.3B parameter NLLB MLLM. Future work could involve scaling the adaptMLLM approach to bigger MLLMs.
\item 	\textbf{Optimising the infrastructure and hyperparameters}: the infrastructure and hyperparameters used for developing models for the language pairs could be significantly enhanced. Future work will involve enhancing the training process by optimising both the infrastructure and hyperparameters used for developing models for the language pairs. 
\end{itemize}

\section{Final Remarks}

The applications developed to improve the XAI of MT address a common problem, namely the streamlining of NMT development for developers and translators. However, each application's approach and infrastructure requirements differ significantly. With adaptNMT, Transformer models can be built from scratch using a modest architecture with limited system memory and less powerful, cheaper GPUs. 

In the case of adaptMLLM, the approach taken is to fine-tune large pre-trained MLLM models which is demanding both in terms of GPU and system memory requirements. In our research, we have concentrated on pre-trained models with 3.3 billion parameters but with the right infrastructure, there is the capacity to scale to 54 billion parameters. 

Furthermore, both applications are built on very different MT engines. The OpenNMT framework is the underlying MT engine for adaptNMT whereas the HuggingFace platform is at the core of adaptMLLM. This has important implications for how both applications can be developed by the MT community and possibly integrated into other tools. Therefore, it is envisaged that both projects will coexist with each other.

The research contributions arising from this work will not solve all the problems encountered in developing MT systems for low-resource languages. However, the approach adopted may serve as a handbook for how low-resource MT systems can be developed. Through the use of HPO and XAI architectures, coupled with small custom-developed corpora and a straightforward human evaluation framework, SOTA in the field of low-resource MT can be achieved.

\appendix
\chapter{No Language Left Behind}

\begin{figure}[h]
  \includegraphics[width=15.5cm]{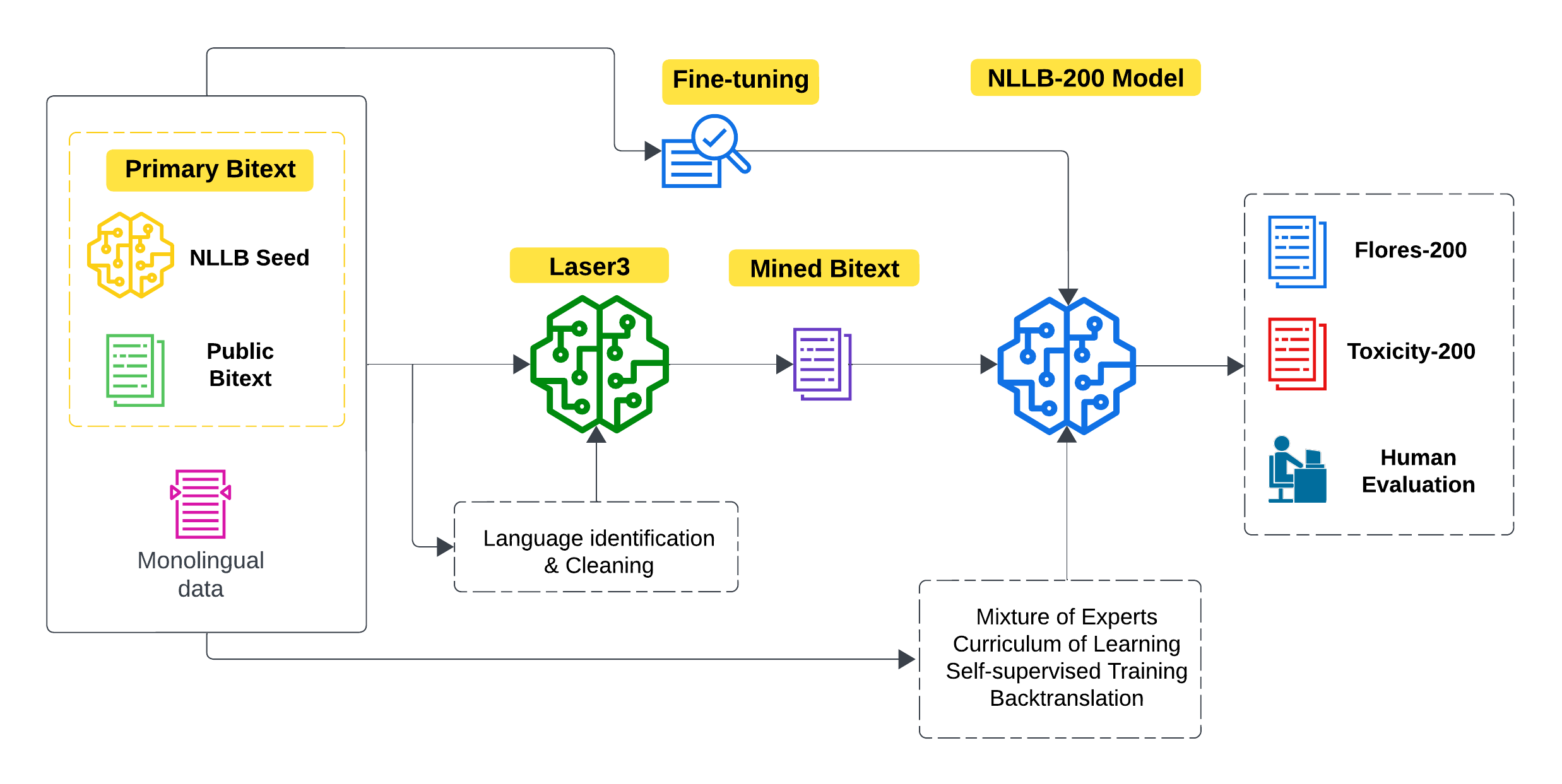}
  \caption{Overview of the NLLB approach \parencite{costa2022no}}
 \label{fig:nllb_overview}
\end{figure}

Meta employs several advanced techniques in its No Language Left Behind (NLLB) architecture including a mixture of experts (MoE), curriculum learning, self-supervised training, backtranslation and an NLLB Seed. The MoE approach efficiently scales NNs using `experts' (sub-networks) which are specialised in different types of tasks or data. MoE enables the model to handle the diversity of languages more effectively. Each `expert' can focus on specific language features or translation nuances, making the model powerful and efficient.

Curriculum learning involves training a model on tasks of increasing complexity. This step-by-step approach can help in better generalisation and handling of complex language translations. Furthermore, backtranslation is used to generate synthetic parallel sentences, which are particularly useful for training translation models in language pairs with scarce data.
   
An NLLB-seed is a foundational dataset or pre-trained model used as the starting point for the NLLB project.  The NLLB-seed facilitates transfer learning by providing a robust base model that can be fine-tuned or adjusted for specific languages or language pairs, especially those with limited training data. Using NLLB-Seed means that the model already has a significant understanding of linguistic patterns and structures, which it can then build upon.

The NLLB project uses a diverse training dataset to support its goal of providing high-quality translations for a wide range of languages, including many that are underrepresented low-resource languages. The principal components of the training dataset are illustrated in Figure \ref{fig:nllb_overview}.  

Publicly available datasets that are commonly found in MT and language research are used. These include datasets from WMT\footnote{https://machinetranslate.org/wmt} (Workshop on Machine Translation), OPUS\footnote{https://github.com/Helsinki-NLP/Opus-MT} (from the University of Helsinki), and other similar resources. Furthermore, specially curated datasets for low-resource languages have been created. This involves collecting texts from diverse sources, including books and websites which are then translated and included in the training dataset.

To further boost the quality of the datasets, the NLLB project has collaborated with language experts and native speakers, especially for low-resource languages. These collaborations help in verifying the quality of translations and in collecting authentic language data. Finally, the dataset also includes data from social media platforms and other online sources which is useful for capturing contemporary usage and evolving language trends.

\printbibliography

\end{document}